\documentclass[10pt,twocolumn,letterpaper]{article}

\usepackage[pagenumbers]{cvpr} %

\usepackage{url}            %
\usepackage{amsfonts}       %
\usepackage{nicefrac}       %
\usepackage{microtype}      %
\usepackage{makecell}       

\usepackage{enumitem}
\usepackage{colortbl}
\usepackage{placeins}
\definecolor{mildgreen}{rgb}{0.88, 1, 0.88}
\definecolor{mildred}{rgb}{1, 0.8, 0.8}

\PassOptionsToPackage{numbers, sort&compress}{natbib}
\newcommand{\appref}[1]{\hyperref[#1]{Appendix~\ref*{#1}}}

\usepackage{graphicx}
\usepackage{booktabs}
\usepackage{multirow}
\usepackage{multicol}
\usepackage{indentfirst}
\usepackage{subcaption}
\usepackage{makecell}
\usepackage{mdframed}
\usepackage{placeins}
\usepackage{CJKutf8}
\usepackage{listings}
\usepackage[ruled,vlined]{algorithm2e}
\usepackage[dvipsnames]{xcolor}
\definecolor{mygreen}{RGB}{58, 130, 51}
\definecolor{myblue}{RGB}{40, 84, 156}
\definecolor{mygray}{RGB}{142, 142, 142}
\definecolor{commentcolor}{RGB}{59,116,116}   %

\newcolumntype{M}[1]{>{\centering\raggedright\arraybackslash}m{#1}}
\usepackage{pifont}   %
\newcommand{\cmark}{\textcolor[rgb]{0.004, 0.663, 0}{\ding{51}}} 
\newcommand{\xmark}{\textcolor{red}{\ding{55}}}
\newcommand{\qmark}{{\color{orange}\textbf{?}}}

\usepackage{nicematrix}
\definecolor{mygray}{gray}{0.85}
\definecolor{softgray}{rgb}{0.9, 0.9, 0.9}
\definecolor{softblue}{rgb}{0.88, 0.92, 1.0}
\definecolor{softgreen}{rgb}{0.88, 1.0, 0.88}
\definecolor{softyellow}{rgb}{1.0, 1.0, 0.88}
\definecolor{softred}{rgb}{1.0, 0.88, 0.88}
\definecolor{softpink}{rgb}{1.0, 0.88, 0.94}

\usepackage{color-edits}

\addauthor{gn}{magenta}
\addauthor{zq}{magenta}

\usepackage[most]{tcolorbox}
\usepackage{enumitem}
\usepackage{fontawesome5}
\usepackage{xcolor}

\setlist[enumerate]{label=(\arabic*), leftmargin=*, itemsep=1pt, topsep=2pt, parsep=0pt}

\newtcolorbox{pointsbox}[1][]{
  enhanced,
  colback=gray!2,
  colframe=gray!40,
  boxrule=0.4pt,
  arc=2pt,
  left=10pt, right=8pt, top=12pt, bottom=8pt, %
  fontupper=\small,                            %
  before skip=6pt, after skip=6pt,
  overlay={
    \node[anchor=north west, inner sep=2pt]
      at ([xshift=4pt,yshift=-4pt]frame.north west)
      {\textcolor{yellow!65!orange}{\faLightbulb[regular]}};
  },
  #1
}

\usepackage{listings}
\usepackage[T1]{fontenc}
\usepackage[scaled=0.85]{beramono} %

\definecolor{cvprblue}{rgb}{0.21,0.49,0.74}
\usepackage[pagebackref,breaklinks,colorlinks,allcolors=cvprblue]{hyperref}

\def\paperID{1719} %
\def\confName{CVPR}
\def\confYear{2026}

\title{Building a Precise Video Language with Human–AI Oversight}

\author{
Zhiqiu Lin$^{1}$ \quad
Chancharik Mitra$^{1}$ \quad
Siyuan Cen$^{1}$ \quad
Isaac Li$^{1}$ \quad
Yuhan Huang$^{1}$ \quad 
Yu Tong Tiffany Ling$^{1}$ \\
Hewei Wang$^{1}$ \quad
Irene Pi$^{1}$ \quad
Shihang Zhu$^{1}$ \quad
Ryan Rao$^{1}$ \quad
George Liu$^{1}$ \quad
Jiaxi Li$^{1}$ \quad
Ruojin Li$^{1}$ \\
Yili Han$^{1}$ \quad
Yilun Du$^{2}$ \quad
Deva Ramanan$^{1}$  \\
{\small $^1$Carnegie Mellon University} \quad
{\small $^2$Harvard University} 
}

\begin{document}
\maketitle
\begin{abstract}
Video–language models (VLMs) learn to reason about the dynamic visual world through natural language. We introduce a suite of open datasets, benchmarks, and recipes for scalable oversight that enable precise video captioning. First, we define a structured specification for describing subjects, scenes, motion, spatial, and camera dynamics, grounded by hundreds of carefully defined visual primitives developed with professional video creators such as filmmakers. Next, to curate high-quality captions, we introduce CHAI (Critique-based Human–AI Oversight), a framework where trained experts critique and revise model-generated pre-captions into improved post-captions. This division of labor improves annotation accuracy and efficiency by offloading text generation to models, allowing humans to better focus on verification. Additionally, these critiques and preferences between pre- and post-captions provide rich supervision for improving open-source models (Qwen3-VL) on caption generation, reward modeling, and critique generation through SFT, DPO, and inference-time scaling.
Our ablations show that critique quality in precision, recall, and constructiveness, ensured by our oversight framework, directly governs downstream performance. With modest expert supervision, the resulting model outperforms closed-source models such as Gemini-3.1-Pro. Finally, we apply our approach to re-caption large-scale professional videos (e.g., films, commercials, games) and fine-tune video generation models such as Wan to better follow detailed prompts of up to 400 words, achieving finer control over cinematography including camera motion, angle, lens, focus, point of view, and framing. Our results show that precise specification and human–AI oversight are key to professional-level video understanding and generation. Data and code on \href{https://linzhiqiu.github.io/papers/chai/}{our site}.
\end{abstract}

\begin{figure*}[t!]
\centering
    \includegraphics[width=\textwidth]{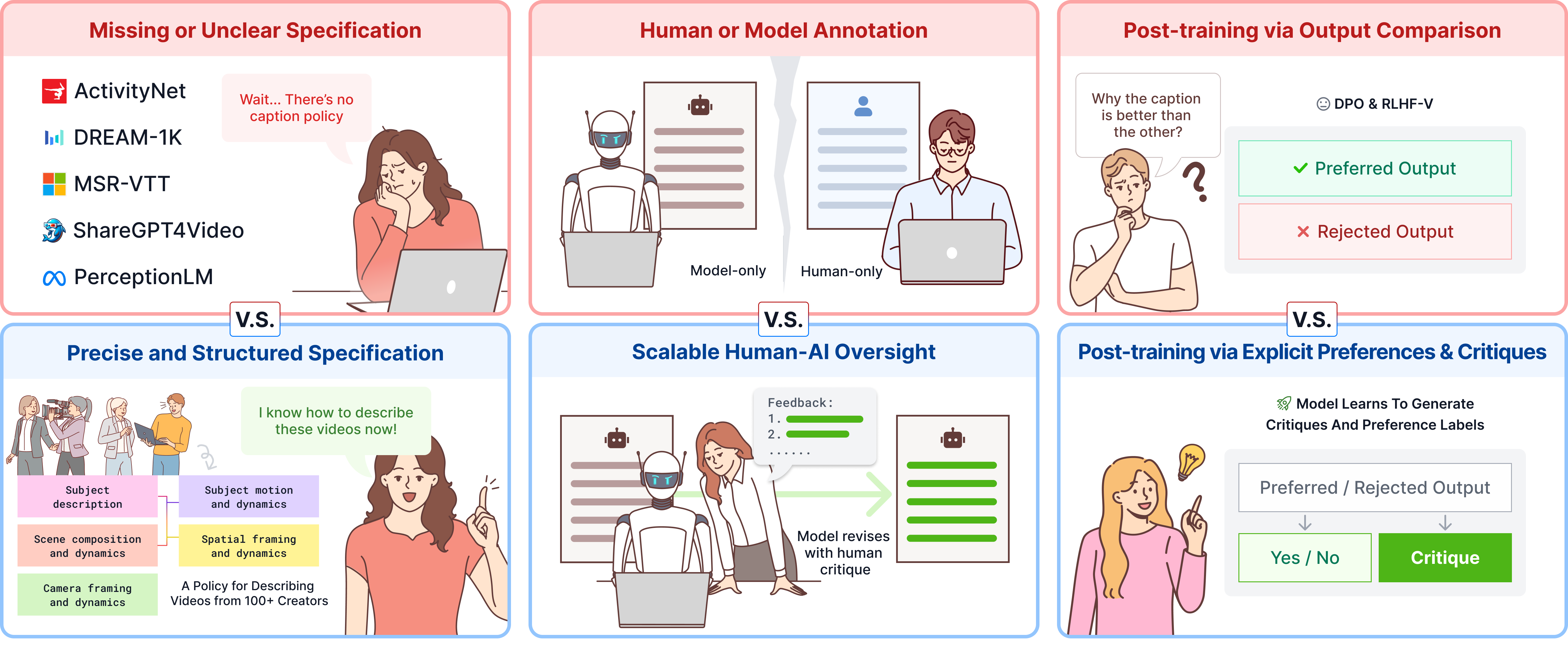}
    \caption{\small {\bf Recipe for precise video language.} We present a recipe for producing high-quality video captions (\textcolor{blue}{blue, bottom row}) and compare it with prior work (\textcolor{red}{red, top row}). (1) While prior video–text datasets lack a clear policy to teach annotators and models {\em what} and {\em how} to describe, we develop a {\bf structured specification} with professional video creators to collect descriptions that support comprehensive video understanding. (2) Instead of relying solely on human or model annotators, we introduce a scalable {\bf human–AI oversight} framework where humans critique model-generated captions with LLM-assisted writing to ensure annotation quality. (3) Our critique-based human-AI (CHAI) oversight framework naturally produces preference pairs and natural-language critiques for {\bf post-training} models, leading to better performance via SFT, DPO, and inference-time scaling. \autoref{fig:videogeneration} shows that video generators fine-tuned on our captions follow detailed prompts with professional-level control over camera dynamics, shot composition, and cinematography. We release our recipe, data, and models to support future research in fine-grained video understanding and generation.
    }\label{fig:teaser}
    \vspace{-10pt}
\end{figure*}

\vspace{-5pt}
\section{Introduction}
\label{sec:intro}
\begin{quote}
\raggedright %
{\em The limits of my language mean the limits of my world.}\\
\hfill-- {\scriptsize Ludwig Wittgenstein, \emph{Tractatus Logico-Philosophicus}~\cite{wittgenstein1922tractatus}} %
\end{quote}

Video–language models (VLMs) learn a world model of visual dynamics through language supervision that teaches them {\em what} is present, {\em where} it is, and {\em how} it unfolds over time. Prior work~\cite{dalle3, bahng2025cycle, fan2023improving, onoe2024docci} shows that high-quality language supervision improves alignment and control in image-text models, yet research on curating precise video annotations remains scarce. Existing video–text datasets~\cite{activitynet, xu2016msr, liu2025good, tarsier, meng2025videocap, chen2024panda, chen2025sharegpt4video} often lack a clear specification of what to describe, producing inconsistent captions that cause models to hallucinate or miss key visual details. To differentiate our goal from that of prior captioning work, we treat precise video language as something to be {\em built} rather than merely collected, requiring three key ingredients (\autoref{fig:teaser}): (1) a clear {\em specification} of what to describe and (2) a scalable {\em oversight framework} for high-quality annotation, and (3) {\em post-training strategies} for scaling model capability with modest expert supervision.

\begin{figure*}[t!]
\centering
    \includegraphics[width=\textwidth]{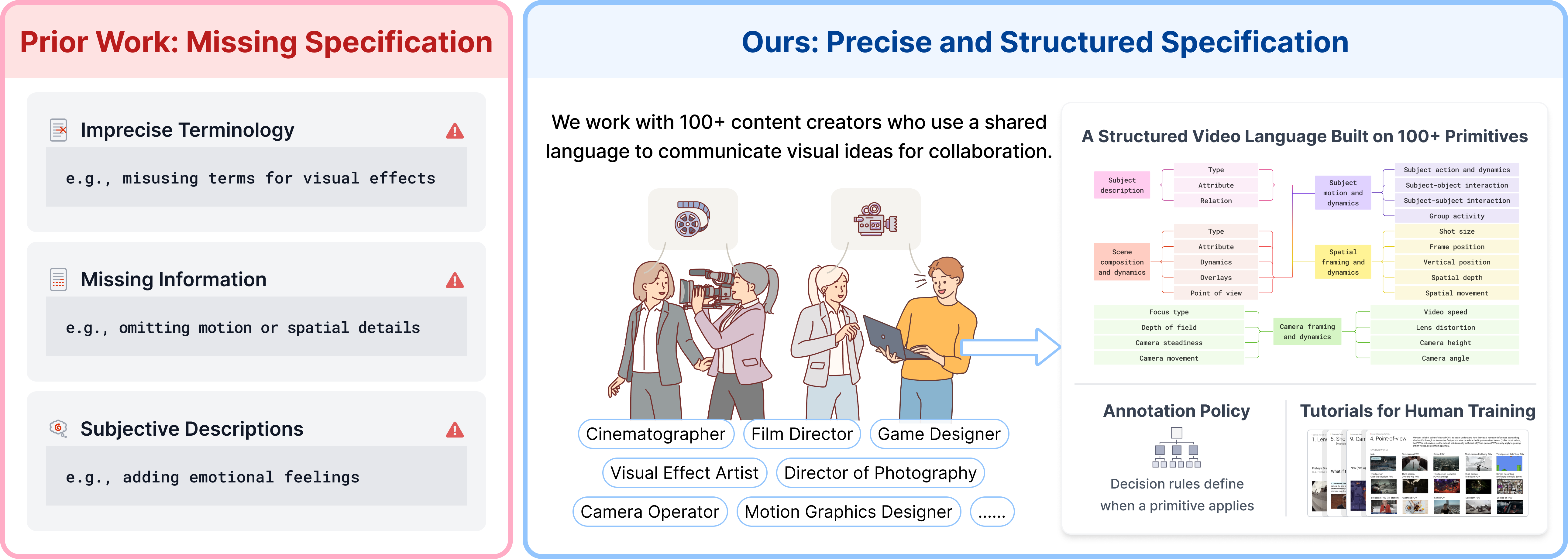}
    \caption{\small {\bf Specification is crucial in video language.} Prior datasets (\textcolor{red}{red, left column}) that lack a clear specification often misuse visual and motion terms, omit key information, and include subjective or emotional descriptions that distract from the observable video content. Over the course of a year, we worked (\textcolor{blue}{blue, right column}) with more than 100 professional video creators, including filmmakers, cinematographers, visual effects artists, and game designers, who rely on a shared language to communicate visual intent precisely in their work. Together, we define a structured specification of video language that spans subjects, scenes, motion, spatial, and camera dynamics, grounded by hundreds of primitives such as POVs, overlays, playback speed, subject actions and transitions, shot size and angle, and camera lens and movement. Each primitive is paired with clear decision rules and training materials to help both professionals and non-professionals apply them consistently. We refer readers to CameraBench-Pro in the Moodio technical report~\cite{rao2026moodio} for full descriptions of these primitives and guidelines. We find this specification effective in professional workflows and will release it publicly to support future research in precise video understanding and generation. 
    }\label{fig:specification}
\end{figure*}

{\bf (1) Precise specification} (\autoref{fig:specification}).
Describing a video is inherently ambiguous without explicit guidelines, as one may focus on what the subject does, how the camera moves, or how the framing changes over time. Prior cognitive studies show that language affects visual perception~\cite{winawer2007russian, maier2018native, navarrete2020language}, making a precise specification essential for consistent video description. To build this foundation, we collaborate with professional content creators such as filmmakers who rely on a shared vocabulary~\cite{filmgrammar, camerabench} to coordinate complex creative workflows that often require teamwork. Together, we develop a {\em comprehensive specification} that formalizes this vocabulary from first principles, defining a structured framework that spans {\it subjects}, {\it scenes}, {\it motion}, {\it spatial framing}, and {\it camera dynamics}. Unlike popular datasets~\cite{activitynet, chen2025sharegpt4video, tarsier} that lack structured annotation policies, our specification defines hundreds of visual and motion primitives (introduced as CameraBench-Pro in the Moodio technical report~\cite{rao2026moodio}) and training guidelines that together improve caption consistency and coverage in human evaluations. %

{\bf (2) Oversight framework} (\autoref{fig:oversight}). Writing detailed video descriptions is cognitively demanding even with clear guidelines. Even a short 5-second clip can contain multiple subjects entering and exiting the frame, each performing distinct actions. Describing such video dynamics can take hundreds of words and more than ten minutes to complete. As a result, many recent video–text datasets~\cite{chen2025sharegpt4video, chen2024panda, xue2025ultravideo, wang2025spatialvid, ju2024miradata} instead rely on video–language models~\cite{qwen25vl, chatgpt} to write captions following human-provided instructions. While these models are highly fluent writers, they often hallucinate visual details, such as confusing left and right, inventing non-existent objects or motion, or misidentifying effects like fisheye lens or dolly zoom. On the other hand, in language tasks such as summarization and code review, models have begun to outperform humans in fluency and coverage, motivating research into {\em scalable oversight} methods~\cite{saunders2022self, bowman2022measuring, chatterji2025people} that let humans supervise models that may exceed their own capabilities in certain task-relevant skills. We extend this idea to video captioning by implementing an oversight framework where models first generate high-recall {\bf pre-captions}, and trained human experts focus on {\bf critiquing} rather than writing from scratch, guiding models to produce improved {\bf post-captions}. 
Our user study in \autoref{sec:prior_work_errors} shows that this approach improves annotation accuracy, writing quality, and efficiency over prior work~\cite{yu2024rlhf, cho2025perceptionlm} that rely on manual caption editing, presumably because shifting limited human attention from text {\em generation} to {\em verification} allows for more effective use of cognitive resources.

{\bf (3) Post-training strategies (\autoref{fig:critique_quality}).}
Popular post-training methods such as DPO~\cite{rafailov2023direct} and GRPO~\cite{guo2025deepseek} rely on preference-based supervision, comparing candidate outputs ranked by humans or reward models. Our data engine not only provides such preference signals but also adds explicit language feedback that explains how to improve weaker outputs into stronger ones through triplets of {\bf (pre-caption, critique, post-caption)}. We find that explicitly training on these preferences and critiques greatly improves standard supervised fine-tuning and offline RL methods such as DPO and RLHF-V~\cite{yu2024rlhf}. Crucially, our ablations show that successful post-training depends on the high-quality critiques produced by our framework, which are more accurate, complete, and constructive than those collected in prior work such as in OpenAI's GDC~\cite{saunders2022self, zhang2025mm}.

{\bf Benchmarks and findings.}
Using our collected triplets on $\sim$4k videos spanning films, games, commercials, and user-generated clips, we release the first unified benchmark for precise video–language understanding that jointly evaluates (1) {\em caption generation}, (2) {\em reward modeling}, and (3) {\em critique generation}. Key findings are:
\begin{tcolorbox}[
  title={\textbf{Key Takeaways}},
  colback=yellow!5,
  colframe=black!60,
  coltitle=white,
  colbacktitle=black!60,
  fonttitle=\sffamily,
  boxrule=0.8pt,
  arc=3pt,
  left=6pt, right=6pt, top=4pt, bottom=4pt,
  breakable
]
\begin{enumerate}[leftmargin=*, itemsep=6pt]
\item Off-the-shelf video-language models capture subject appearance and scene context well but struggle with fine-grained aspects such as subject motion and camera dynamics.
\item Explicit preference and critique supervision improves standard SFT and RL methods, enabling open-source models (Qwen3-VL) to outperform closed-source Gemini-3.1 with modest expert supervision.
\item Critique quality (precision, recall, and constructiveness) is crucial for post-training success. Our oversight framework enforces these properties by design, and we share additional guidelines on quality control and workflow.
\end{enumerate}
\end{tcolorbox}

{\bf Application: improving text-to-video generation.}
We apply our models to re-caption~\cite{dalle3} large-scale professional videos, sourcing diverse, high-quality footage from films, commercials, music videos, and games. This re-captioning helps fine-tuned video generation models such as Wan~\cite{wan2025} to better follow detailed prompts of up to 400 words, achieving finer control over motion, camera, visual composition, and cinematography. We hope this serves as a promising first step toward professional-grade text-to-video generation with precise language control. See \autoref{fig:videogeneration} for examples.

{\bf Summary.}
We make three contributions toward building a precise video language: 
(1) a {\em comprehensive specification} that defines what and how to describe in videos developed with professional video creators; 
(2) a {\em scalable oversight framework} that shifts human effort from writing to verification via critique-based correction; and 
(3) {\em post-training recipes} that explicitly leverage these preference and feedback signals to further improve the model.
We hope our released data, framework, and benchmarks pave the way for scalable human–AI data curation for professional video understanding and generation.

\section{Related Work}

{\bf Video–text datasets.}
Video captioning has evolved from short, loosely aligned descriptions~\cite{xu2016msr, activitynet} to longer, detailed ones~\cite{kong2025tuna, cho2025perceptionlm, camerabench, meng2025videocap, zhang2025critic, dang2025synpo, lian2025describe, wang2025sota, zhang2024direct, wu2025ugc}. However, current datasets curated by humans or models~\cite{chen2024panda, chen2025sharegpt4video, ju2024miradata, xue2025ultravideo, yang2024vript, tang2025video, zhang2025vcapsbench, ma2025omni, chai2024auroracap} often suffer from incomplete coverage, visual hallucinations, and writing problems. Our human evaluations show that these issues largely stem from the lack of clear specification and effective oversight. In addition, many foundation models are trained on closed-source datasets~\cite{chen2025skyreels, qwen25vl, yang2025kwai} that cannot be verified for quality.

{\bf Scalable oversight.}
As AI systems begin to match or surpass humans in complex tasks such as coding, mathematics, and writing, a key question arises: how can people continue to supervise models that are stronger than themselves in many of the skills required for these tasks? Scalable oversight~\cite{saunders2022self, bowman2022measuring} in natural language processing studies this challenge through the lens of human–AI collaboration; rather than human-only supervision, models assist humans in providing high-quality oversight through error detection in tasks such as question answering, summarization, and code review~\cite{mcaleese2024llm}. However, most video–text datasets still rely on either human-written~\cite{activitynet, xu2016msr, tarsier} or model-generated captions~\cite{chen2025sharegpt4video, xue2025ultravideo} without cooperation. We adapt the principle of scalable oversight to curate video captions, allowing models to generate fluent captions while humans focus on visual details, each specializing in what they do best.

{\bf Post-training strategies.}
Post-training aligns model behavior with human preferences or task-specific goals through supervised fine-tuning, reinforcement learning, or inference-time scaling~\cite{rafailov2023direct, Schulman2017ProximalPO, guo2025deepseek, parashar2024neglected, lin2024revisiting, lin2023crossmodal, liu2023language, chai2025activation, mitra2025enhancing, chen2025minimax, yu2025dapo}. In multimodal learning, recent work extends these methods to vision–language models using self-critique or reflection~\cite{ankner2024critique, yu2024self, liu2025inference}, yet most rely on synthetic or image-only feedback~\cite{chen2024dress, li2024fire, wei2025perception, yu2025diverse, liu2023language, li2024evaluating, madaan2023self, wang2025critique, yasunaga2025multimodal, zhao2025omnialign, wu2025visco, gordon2025unblocking}.
Ours is the first to show that high-quality feedback provides an effective supervision signal for video understanding.

\section{Specification For Video Captioning}
How the captioning task is framed directly affects what annotators perceive and choose to describe~\cite{stangl2021going, gurari2020captioning}. We first identify common issues in prior datasets that lack consistent annotation policies and then present our structured specification built with professional video creators. See \autoref{fig:specification} for an overview.

{\bf Issue: lack of specification.} Without clear instruction, annotators may not know what to describe or how much detail to include. To examine this, we manually evaluate eight widely used video–text datasets: MSR-VTT~\cite{xu2016msr}, ActivityNet~\cite{activitynet}, ShareGPT4Video~\cite{chen2025sharegpt4video}, UltraVideo~\cite{xue2025ultravideo}, VDC~\cite{chai2024auroracap}, Dream1K~\cite{tarsier}, PerceptionLM (PE-Video)~\cite{cho2025perceptionlm}, and TUNA-Bench~\cite{kong2025tuna}. We find that most datasets do not provide a detailed policy for annotators (the only exception being \cite{kong2025tuna}, whose guideline is not public). We observe three major issues caused by the lack of specification and provide detailed error examples in \autoref{sec:prior_work_errors}:
\begin{itemize}
    \item {\bf (1) Imprecise terminology}. Without clear guidelines, annotators may lack the proper vocabulary to describe visual or motion effects. For example, they often confuse {\it camera translation} (camera physically moves forward or sideways) with {\it zoom} (focal length change without moving the camera) or {\it rotation} (camera pivots in place). They also misuse cinematography terms, calling a {\it full shot} (entire subject in view) a {\it close-up} (small part of the subject), an {\it aerial shot} (high altitude) a {\it bird’s-eye view} (top-down angle), or describing {\it fisheye distortion} (straight lines bending outward from a wide lens) as a ``circular'' scene. See \autoref{fig:crowdsourced} for such mistakes made by untrained annotators.
    \item (2) {\bf Missing information.} Without clear rules on what to include, annotators often miss key elements essential for understanding a shot. Some captions focus only on the subject and its actions, while others describe only the scene or camera movement, leading to incomplete descriptions. For instance, although many datasets attempt to capture camera work~\cite{kong2025tuna, cho2025perceptionlm}, their captions frequently omit crucial details such as camera shake, focus changes, and tracking movement.
    \item (3) {\bf Subjective descriptions}. Without an objective annotation policy, captions often include personal opinions or emotional language, such as calling a scene ``inspiring'' or ``informative,'' which other annotators may not agree with. Such subjective phrasing varies across annotators and distracts from describing the actual visual content.
\end{itemize}

{\bf Building specification with content creators.} The lack of clear specification among prior datasets motivates us to learn from professionals who already use one. We view video captioning as a form of {\it visual storytelling}: the best caption should let someone who has never seen the video picture it vividly~\cite{gurari2020captioning}. This idea naturally aligns with how content creators work. In filmmaking, for example, directors must tell camera operators exactly how the camera should move, what angle to use, and how the frame should look to achieve the intended shot~\cite{filmgrammar, camerabench}. Such teamwork depends on a precise shared language to communicate visual intent before the shot is made. We adopt this practice to build a structured, teachable video language with over
100 creators from the US and China working in film, game design, and motion graphics, with 2–5+ years of professional experience. Through a year-long collaboration, we take a bottom-up approach to formalize this language. We start by asking professional creators to describe diverse videos—from films and games to user-generated clips—and collect the aspects they naturally mention. These are then grouped into a structured framework of five major aspects with clear subcategories: {\it subject} (types, attributes, relations), {\it scene} (point of view, overlays, setting, time of day), {\it motion} (actions, interactions, group activities), {\it spatial} (shot size, frame position, spatial depth, movement), and {\it camera} (playback speed, camera height, angle, lens, focal plane, steadiness, movement). Each aspect captures both what appears and how it changes over time. We note that our framework focuses only on {\em objectively observable} visual aspects, rather than subjective impressions or emotional interpretations.

{\bf Primitives for precise specification.}
Common cinematic terms are often used inconsistently across practitioners and datasets.
To address this, our CameraBench-Pro~\cite{rao2026moodio} defines hundreds of fundamental visual and motion {\bf primitives} through bottom-up collaboration with professional video creators, establishing rigorous definitions, video examples, corner cases, and decision rules for each primitive.
For example, CameraBench-Pro clarifies that {\it bird's-eye view} refers to a strict top-down angle rather than any high vantage point, and extends playback speed beyond {\it slow motion} and {\it time-lapse} to include {\it speed ramp} and {\it stop motion}.
The full taxonomy includes over 200 primitives spanning camera motion ($\sim$50, overlapping with CameraBench~\cite{camerabench}), camera setup ($\sim$100), and video cinematography ($\sim$70).
Each static property (e.g., focal plane, shot size) is labeled at both the start and end of a shot to capture temporal change, with decision rules resolving common ambiguities such as subject identification and dynamic framing.
We refer readers to the CameraBench-Pro technical report~\cite{rao2026moodio} for the complete taxonomy, training guidelines, and labeling platform.

{\bf Structured captioning policy.} While primitives define the building blocks of our specification, they alone cannot describe how these elements unfold within a video. A complete caption must link multiple primitives into a coherent description; for example, a transition from a {\it bird’s-eye view} to a {\it level angle} through camera {\it tilting upward}, or a {\it shallow depth of field} shot where the focus shifts from {\it foreground} to {\it midground} through a {\it rack focus}. To capture such structure, we develop detailed guidelines that extend primitive-based labeling into full captioning, which specifies how to describe each aspect under different conditions. For instance, when a video lacks a clear or consistent {\it subject}, annotators note the absence (e.g., an establishing shot) or describe how the subject changes over time (e.g., a revealing shot). \autoref{sec:captioning_policy} provides the full guidelines in detail.

\begin{figure*}[t!]
\centering
    \includegraphics[width=\textwidth]{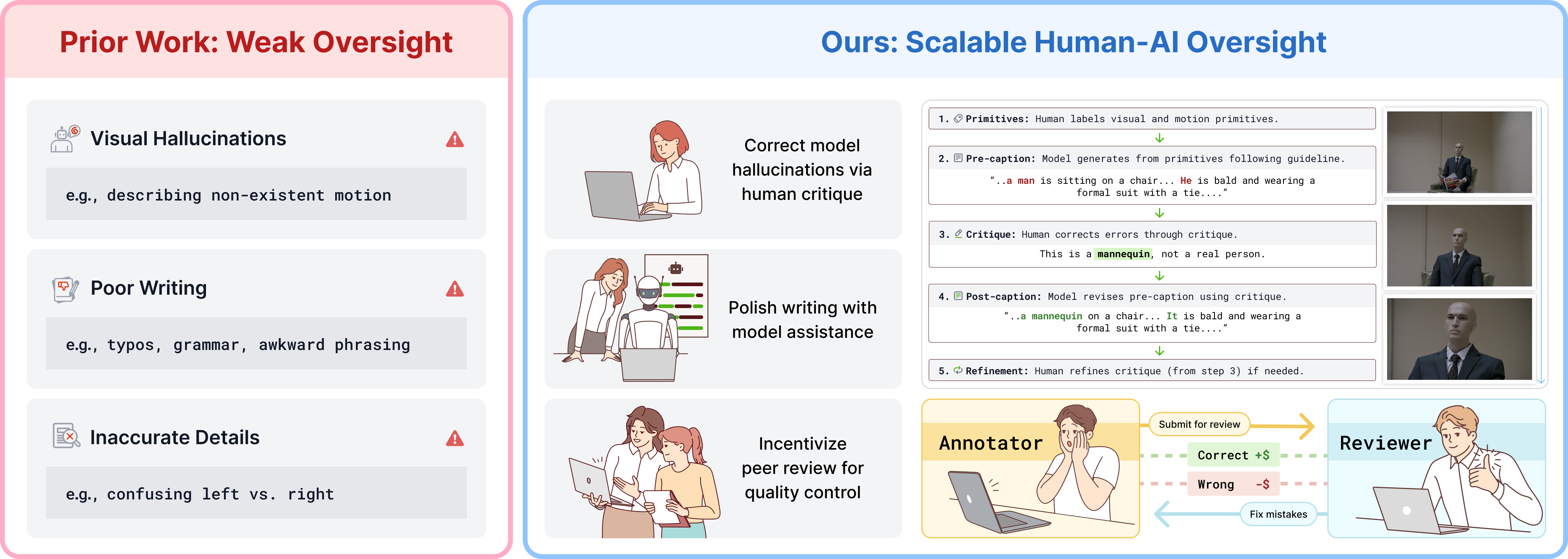}
    \caption{\small {\bf Oversight is crucial for quality annotations.} Even with clear task guidelines, writing detailed video descriptions remains error-prone for both humans and models. Our human evaluations show that prior datasets (\textcolor{red}{red, left column}) suffer from recurring issues: human-written captions often contain typos, grammatical errors, awkward phrasing, and events described out of order; while model-written captions frequently exhibit visual hallucinations, describing non-existent subjects or motion. To address these issues, we introduce a {\bf C}ritique-based {\bf H}uman–{\bf AI} oversight ({\bf CHAI}) framework (\textcolor{blue}{blue, right column}) that combines the strengths of both. The model first generates a comprehensive {\bf pre-caption} following task guidelines, and the human writes a {\bf critique} that corrects hallucinations and guides the model toward a refined {\bf post-caption}. This division of labor (1) eliminates {\it writing problems} since the language model handles text generation and editing, and (2) allows humans to focus on {\it verification} rather than writing a full caption from scratch or worrying about phrasing. Lastly, since both humans and models still struggle with fine details such as spatial relations (e.g., left vs. right), we implement (3) a {\bf quality control system} with a peer-review bonus: annotators earn bonuses when reviewers find no errors, and reviewers earn bonuses for valid corrections. This incentive keeps both sides motivated for accuracy. \autoref{sec:oversight_details} provides implementation details including annotator screening, training, and interface design. \autoref{sec:more_examples} shows complete (pre-caption, critique, post-caption) examples from our dataset (\autoref{fig:subject_captions} to \autoref{fig:camera_captions}).
    }\label{fig:oversight}
\end{figure*}

\section{Human-AI Oversight Framework}
With or without a clear specification, writing detailed video descriptions is difficult and prone to error. Even a short five-second clip can include several subjects and camera or spatial movements that must be described precisely and in order. Oversight is therefore key to keeping annotations accurate and consistent. To study its effect on data quality, we first analyze datasets with limited or no oversight and the issues that result, then present our {\bf CHAI} framework ({\bf C}ritique-based {\bf H}uman–{\bf AI} oversight). \autoref{fig:oversight} provides a summary.

{\bf Issue: lack of oversight.} Most existing video–text datasets rely on either human-written or model-generated captions without systematic quality control. We provide a detailed report of them in \autoref{sec:prior_work_errors} and summarize the major issues caused by insufficient oversight below:

\begin{itemize}
    \item (4) {\bf Poor writing.} Human captions often contain typos, grammatical mistakes, or awkward phrasing. Beyond these surface-level issues, untrained annotators may describe events out of temporal order or use vague references when multiple subjects appear, such as saying ``the first person hugs the second person'' instead of “the person on the left hugs the one on the right.” 
    \item (5) {\bf Visual hallucinations.} Video captioning models frequently make confident but incorrect statements about the video, e.g., describing actions or objects that do not exist, or claiming the camera is static when it is clearly moving. 
    \item (6) {\bf Inaccurate details.} Both human and AI annotators struggle with subtle visual and motion details, such as mistaking a hand on the right side of the frame for the subject’s right hand (which is actually the left), or missing a small handheld motion that makes the camera non-static.  
\end{itemize}

{\bf Quality data requires effective oversight.}
Our human evaluations (\autoref{sec:prior_work_errors}) show that even recent benchmarks such as TUNA-Bench from Kling\cite{kong2025tuna}, which claims to have a specification (though unreleased), still suffer from issues caused by limited oversight. In our case, oversight is even more critical because following our caption policy to write an extremely detailed caption (200–400 words) from scratch is cognitively demanding even for trained experts. To efficiently collect high-quality captions, we design an oversight framework that combines (a) {\it human–AI collaboration} to reduce writing and hallucination errors, and (b) {\it screening, training, and incentives} to ensure annotators are skilled and motivated to keep details accurate. 

{\bf (a) Human–AI collaboration for data curation.}
Inspired by scalable oversight in NLP~\cite{bowman2022measuring, saunders2022self}, we design a simple but effective workflow that divides tasks between humans and models so each focuses on what it does best. Since large language models are already more efficient writers than most humans~\cite{chatterji2025people}, our workflow lets models write fluent text that closely follows human instruction, while humans focus on visual fact-checking of model outputs. \autoref{fig:oversight} shows the steps:
(1) Humans first label all visual and motion {\bf primitives}~\footnote{To reduce cognitive load, primitive labeling is done on a separate platform with its own quality-controlled process, allowing annotators to focus purely on captioning during the main workflow.} that are important but easy to miss when writing captions from scratch (e.g., camera shake, focal-plane shifts, point of view, shot size, and overlays such as framing or subtitles);
(2) a video–language model drafts a {\bf pre-caption} from these labels, covering as many relevant details as possible according to our captioning guideline; (3) humans review the pre-caption and write a {\bf correctional critique} explaining what is wrong or missing and how to fix it; (4) the model incorporates this critique to produce a refined {\bf post-caption}; and (5) humans revise the critique if needed until the caption is fully accurate. \autoref{sec:oversight_details} provides more details on this process and includes screenshots of the annotation interface.
Our preliminary study shows that annotators find this division of labor faster and less mentally demanding, producing captions that are (1) more accurate, as humans focus on {\it verification} rather than {\it generation}; (2) more complete, as models better follow comprehensive guidelines and make fuller use of all labeled primitives; and (3) more fluent, since all text is polished by the model.
To further improve accuracy, annotators first complete {\it subject} and {\it scene} post-captions, which then prompt the model to generate more accurate {\it motion} and {\it spatial} pre-captions. %

{\bf (b) Screening, training, and incentives.} While this human–AI workflow is effective, both humans and models can still miss fine spatial or motion details; for example, for a camera-facing person, ``moving to his left'' appears on the frame's right, which is easy to confuse with ``moving to the frame’s left.'' To ensure precise descriptions, we focus on selecting, training, and motivating the most capable experts. We recruit annotators only with prior experience in content creation, such as filmmaking, motion design, or game capture. All applicants must complete six rounds of multiple-choice exams based on our primitives, covering over 150 videos that test understanding of camera motion, setup, and video cinematography. Only the top 3\% of applicants (from over 600) who rank in the top 20\% after six exams are selected. They then complete a month of (paid) training on our labeling platform, practicing primitive-level tasks to build the discrimination skills needed for professional-quality captioning. We also encourage annotators to raise questions during training if they disagree with any ground-truth answer or are unsure how to handle edge cases. High performers are promoted to captioning roles and begin by shadowing expert critiques on 100 gold-standard captions authored and reviewed by the paper’s authors (some of whom are also content creators). To further improve accuracy, we introduce a reviewer role (promoted from top annotators with a strong track record) that checks every critique and post-caption and corrects errors using the same workflow. Annotators can see how their work is corrected and may appeal to the managers (the paper’s authors) if they disagree. This peer-review process is reinforced by an accuracy-based bonus: annotators earn rewards for error-free annotations, and reviewers for correcting mistakes, keeping both roles motivated to maintain precise captions. See \autoref{sec:oversight_details} for details.

\begin{figure*}[t!]
\centering
\includegraphics[width=\textwidth]{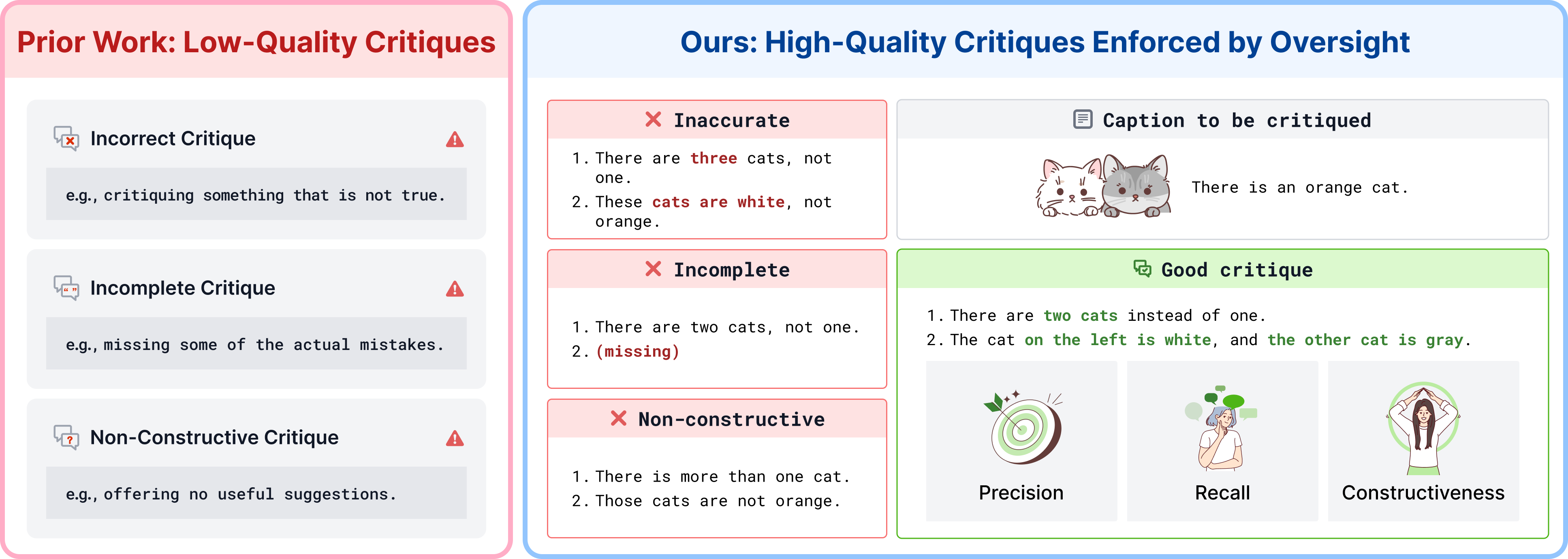}
\caption{\small {\bf Critique quality matters for post-training.}
Prior work~\cite{saunders2022self, zhang2025mm} often collects critiques that are inaccurate (e.g., hallucinating information), incomplete (e.g., skipping mistakes), or non-constructive (e.g., noting errors without explaining how to fix them). Our curation framework instead requires each critique to directly guide the model in producing the final post-caption, forcing annotators to write critiques that are accurate, complete, and constructive. \autoref{sec:critique_quality} shows that removing any of these qualities leads to substantially worse results.
}
\label{fig:critique_quality}
\end{figure*}

\begin{table*}[t]
\centering
\renewcommand{\arraystretch}{1.25}
\caption{\small \textbf{Model performance across captioning, reward modeling, and critique generation.} We note that: (1) Off-the-shelf models perform well on {\it subject} and {\it scene} but struggle with {\it motion} and {\it camera} aspects. (2) Our fine-tuned model achieves state-of-the-art performance across all tasks, outperforming the close-sourced Gemini-2.5/3.1. (3) Adding explicit {\it preference} and {\it critique} signals (full-data post-training) improves both SFT and RL. Note that Gemini-3.1 does not support logprobs for reward modeling; therefore, we directly use the generated text as a binary reward score (0 or 1), indicated with an asterisk (*). \autoref{sec:post_training} provides experimental details.
}
\scalebox{0.60}{
\begin{NiceTabular}{lcccccccccccccccccc}
\toprule[1.5pt]
\multirow{2}{*}{\textbf{Method}} 
& \multicolumn{6}{c}{\textbf{Caption Generation}} 
& \multicolumn{6}{c}{\textbf{Reward Modeling}} 
& \multicolumn{6}{c}{\textbf{Critique Generation}} \\
\cmidrule(l){2-7} \cmidrule(l){8-13} \cmidrule(l){14-19}
& {\scriptsize Subject} & {\scriptsize Scene} & {\scriptsize Motion} & {\scriptsize Spatial} & {\scriptsize Camera} & {\scriptsize Avg}
& {\scriptsize Subject} & {\scriptsize Scene} & {\scriptsize Motion} & {\scriptsize Spatial} & {\scriptsize Camera} & {\scriptsize Avg}
& {\scriptsize Subject} & {\scriptsize Scene} & {\scriptsize Motion} & {\scriptsize Spatial} & {\scriptsize Camera} & {\scriptsize Avg} \\
\midrule

\multicolumn{19}{l}{\it Open-source models} \\

PerceptionLM~\cite{cho2025perceptionlm} 
  & 8.2 & 4.8 & 5.0 & 7.0 & 7.5 & 6.5 & 38.2 & 32.4 & 29.9 & 34.9 & 39.9 & 35.1 & 2.5 & 1.5 & 2.0 & 1.8 & 2.2 & 2.0 \\

OmniVinci~\cite{ye2025omnivinci}
  & 2.8 & 5.2 & 3.5 & 5.5 & 3.0 & 4.0 & 35.9 & 42.9 & 37.5 & 32.8 & 34.7 & 36.8 & 1.2 & 2.2 & 1.8 & 2.5 & 1.3 & 1.8 \\

VideoChat-R1.5~\cite{yan2025videochat}
  & 6.5 & 5.8 & 4.2 & 2.8 & 5.0 & 4.9 & 42.5 & 44.3 & 41.0 & 43.3 & 49.3 & 44.1 & 1.0 & 2.5 & 1.8 & 2.0 & 2.2 & 1.9 \\

SkyReels-V2~\cite{chen2025skyreels}
  & 1.8 & 4.0 & 2.5 & 4.5 & 3.2 & 3.2 & 52.7 & 58.0 & 55.2 & 51.0 & 59.9 & 55.4 & 2.2 & 1.0 & 1.5 & 2.8 & 1.5 & 1.8 \\

OwlCap~\cite{zhong2025owlcap}
  & 4.8 & 5.5 & 3.8 & 5.2 & 2.5 & 4.4 & 48.4 & 51.3 & 49.7 & 47.4 & 55.2 & 50.4 & 1.5 & 2.5 & 2.0 & 1.2 & 1.8 & 1.8 \\

video-SALMONN-2~\cite{tang2025video}
  & 2.5 & 1.5 & 2.0 & 3.5 & 3.8 & 2.7 & 53.1 & 61.9 & 57.8 & 56.1 & 61.2 & 58.0 & 1.8 & 1.0 & 2.5 & 1.5 & 2.2 & 1.8 \\

ShareGPT4Video~\cite{chen2025sharegpt4video}
  & 5.5 & 6.8 & 3.5 & 6.0 & 4.5 & 5.3 & 49.5 & 42.5 & 46.7 & 44.2 & 51.4 & 46.9 & 2.0 & 1.5 & 1.8 & 2.5 & 1.0 & 1.8 \\

MotionSight~\cite{du2025motionsight}
  & 4.0 & 3.5 & 5.8 & 4.2 & 2.2 & 3.9 & 45.3 & 48.2 & 53.1 & 46.7 & 49.3 & 48.5 & 2.8 & 1.5 & 1.0 & 2.0 & 2.2 & 1.9 \\

MMR1~\cite{leng2025mmr1}
  & 2.0 & 4.2 & 2.8 & 5.0 & 3.5 & 3.5 & 55.2 & 60.2 & 59.5 & 57.8 & 58.2 & 58.2 & 1.5 & 2.0 & 2.5 & 1.0 & 1.5 & 1.7 \\

SynPO~\cite{dang2025synpo}
  & 6.8 & 7.2 & 5.0 & 3.8 & 6.5 & 5.9 & 46.0 & 43.9 & 44.6 & 45.0 & 51.8 & 46.3 & 1.2 & 2.5 & 1.8 & 2.2 & 1.8 & 1.9 \\

VideoPASTA~\cite{kulkarni2025videopasta}
  & 3.0 & 2.8 & 1.8 & 3.8 & 1.5 & 2.6 & 56.9 & 56.7 & 54.0 & 51.0 & 49.5 & 53.6 & 2.0 & 1.0 & 1.5 & 2.5 & 2.2 & 1.8 \\

Omni-Captioner~\cite{ma2025omni}
  & 8.0 & 7.8 & 4.5 & 7.5 & 6.5 & 6.9 & 41.0 & 43.9 & 42.5 & 42.5 & 50.1 & 44.0 & 1.5 & 2.2 & 2.5 & 1.0 & 1.8 & 1.8 \\

InternVL-3.5~\cite{wang2025internvl3}
  & 3.8 & 2.4 & 1.8 & 3.2 & 1.8 & 2.6 & 46.0 & 63.7 & 61.6 & 59.5 & 63.1 & 58.8 & 2.8 & 2.0 & 2.5 & 2.2 & 1.5 & 2.2 \\

Granite 3.3~\cite{team2025granite}
  & 2.0 & 1.8 & 0.8 & 2.2 & 0.8 & 1.5 & 41.2 & 44.3 & 38.9 & 42.5 & 46.7 & 42.7 & 1.0 & 1.5 & 0.8 & 1.2 & 2.0 & 1.3 \\

Qwen2.5-VL-7B~\cite{qwen25vl}
  & 5.1 & 1.4 & 4.3 & 4.6 & 4.6 & 4.0 & 31.0 & 27.2 & 35.0 & 26.1 & 64.9 & 36.8 & 2.2 & 2.8 & 1.8 & 1.5 & 2.0 & 2.1 \\

Qwen2.5-VL-72B~\cite{qwen25vl}
  & 7.0 & 6.5 & 3.0 & 6.5 & 6.0 & 5.8 & 33.3 & 28.8 & 36.1 & 27.8 & 63.9 & 38.0 & 3.0 & 2.5 & 2.8 & 2.2 & 1.8 & 2.5 \\

Qwen3-VL-8B-Instruct~\cite{qwen3}
  & 4.4 & 4.0 & 3.7 & 5.2 & 1.3 & 3.7 & 35.7 & 37.4 & 33.9 & 31.3 & 53.5 & 38.4 & 1.0 & 0.8 & 2.1 & 2.0 & 0.8 & 1.3 \\
  
Qwen3-VL-32B-Instruct~\cite{qwen3}
  & 4.1 & 4.8 & 3.5 & 5.9 & 1.6 & 4.0 & 37.2 & 36.1 & 35.4 & 29.8 & 54.8 & 38.7 & 1.2 & 0.6 & 2.3 & 1.8 & 1.1 & 1.4 \\
\midrule

\multicolumn{19}{l}{\it Closed-source models} \\

GPT-4o~\cite{gpt4}
  & 4.7 & 3.0 & 3.8 & 4.5 & 4.9 & 4.2 & 49.2 & 53.7 & 51.1 & 51.9 & 56.6 & 52.5 & 1.5 & 2.0 & 2.5 & 1.2 & 1.8 & 1.8 \\
GPT-5
  & 5.9 & 6.3 & 5.3 & 5.8 & 5.1 & 5.7 & 55.7 & 61.6 & 58.2 & 59.0 & 62.9 & 59.5 & 3.0 & 2.5 & 3.5 & 2.8 & 2.2 & 2.8 \\
Gemini-2.5-Pro~\cite{comanici2025gemini}
  & 6.3 & 6.8 & 6.1 & 7.2 & 5.5 & 6.2 & 58.2 & 64.9 & 57.3 & 62.9 & 66.5 & 62.0 & 3.5 & 2.8 & 2.5 & 3.2 & 3.0 & 3.0 \\
Gemini-3.1-Pro
  & 6.1 & 6.0 & 1.4 & 7.2 & 4.6 & 5.1 & 48.3* & 51.6* & 49.6* & 52.3* & 47.8* & 49.9* & 3.8 & 2.8 & 3.5 & 3.4 & 2.8 & 3.3 \\
\midrule

\multicolumn{19}{l}{\it Caption-only post-training (Qwen3-VL-8B-Instruct)} \\
RLHF-V (Caption)
  & 8.8 & 8.1 & 5.4 & 12.1 & 9.4 & 8.8 & 56.3 & 65.5 & 44.3 & 42.7 & 41.0 & 50.0 & 1.8 & 1.4 & 1.6 & 3.1 & 1.8 & 1.9 \\
DPO (Caption)
  & 9.4 & 8.6 & 5.8 & 11.6 & 9.8 & 9.0 & 56.7 & 65.0 & 44.7 & 42.3 & 41.5 & 50.0 & 2.1 & 1.1 & 1.4 & 2.8 & 2.1 & 1.9 \\
SFT (Caption)
  & 11.4 & 11.1 & 7.3 & 15.4 & 14.9 & 12.0 & 56.6 & 53.4 & 55.6 & 44.7 & 44.1 & 50.9 & 4.5 & 4.0 & 5.0 & 9.2 & 4.7 & 5.5 \\
\midrule
\multicolumn{19}{l}{\it \textbf{Full data post-training (Qwen3-VL-8B-Instruct)}} \\
\textbf{RLHF-V (All)}
  & 15.8 & 16.0 & 13.8 & 17.5 & 15.4 & 15.7 & 83.5 & 86.2 & 76.8 & 80.4 & 78.1 & 81.0 & 23.5 & 24.0 & 25.3 & 26.7 & 29.0 & 25.7 \\
\textbf{DPO (All)}
  & 15.5 & 16.0 & 14.0 & 17.7 & 15.8 & 15.8 & 82.8 & 87.4 & 75.5 & 79.6 & 78.5 & 80.8 & 23.0 & 24.5 & 25.0 & 26.7 & 28.3 & 25.5 \\
\textbf{SFT (All)}
  & \textbf{18.0} & \textbf{18.2 }& \textbf{17.3} & \textbf{19.3} & \textbf{18.3} & \textbf{18.2} & \textbf{89.8} & \textbf{91.0} & \textbf{89.1} & \textbf{91.2} & \textbf{88.1} & \textbf{89.8} & \textbf{36.0} & \textbf{41.6} & \textbf{42.9} & \textbf{43.1} & \textbf{46.2} & \textbf{41.7} \\
\bottomrule[1.5pt]
\end{NiceTabular}
}
\vspace{-2mm}
\label{tab:caption_reward_critique}
\end{table*}

\section{Evaluation and Post-Training}
Our oversight framework naturally yields triplets of {\it (pre-caption, critique, post-caption)} that support evaluation and post-training for video–language models.

{\bf Benchmarks and tasks.}
We curate $~\sim$20k triplets in total, allocating 5k as a held-out benchmark and using the rest for training. This scale far exceeds recent public benchmarks (typically $\sim$1k), such as DREAM-1K~\cite{tarsier} and TUNA-Bench~\cite{kong2025tuna}, and uniquely covers five aspects per video: {\it subject}, {\it scene}, {\it motion}, {\it spatial}, and {\it camera}. \autoref{tab:dataset_comparison_policy} provides additional statistics such as video duration, FPS, and word count. To our knowledge, ours is the first unified benchmark that evaluates not only {\it video captioning} but also {\it reward modeling} and {\it critique generation}, which have become increasingly important for post-training~\cite{mcaleese2024llm, gordon2025unblocking}. We expect all three tasks to advance video understanding; for example, \autoref{sec:inference_scaling} shows that strong reward/critique models enable inference-time scaling for video captioning. The tasks are:

\begin{enumerate}
\item {\bf Caption generation} (Video $\rightarrow$ Caption): generate a comprehensive caption that follows our policy for the five aspects. We report the BLEU-4 against the reference caption in \autoref{tab:caption_reward_critique} and defer other metrics such as LLM-as-judge to \autoref{tab:caption_critique_metrics}.

\item {\bf Reward modeling} ($[$Video, Caption$]$ $\rightarrow$ Score): predict which caption better matches the video by checking whether the post-caption receives a higher score than the pre-caption. We report binary accuracy (chance = 0.5).

\item {\bf Critique generation} ($[$Video, Caption$]$ $\rightarrow$ Critique): produce a correctional critique that identifies errors or omissions and explains how to fix them. We report BLEU-4 against the reference critique in the main paper, and include ROUGE and a ``critique-guided revision'' proxy metric~\cite{gordon2025unblocking} in \autoref{tab:caption_critique_metrics}.
\end{enumerate}

{\bf Preliminaries: offline post-training.} %
We first consider standard supervised fine-tuning ({\bf SFT}), which directly trains the model to generate the preferred outputs (post-captions). We also explore offline RL methods like {\bf DPO}~\cite{rafailov2023direct}, which adds a contrastive objective that rewards preferred outputs and penalizes rejected ones with a KL regularization term. {\bf RLHF-V}~\cite{yu2024rlhf} further extends DPO by increasing gradients on the edited text segments between rejected and preferred outputs. Our data naturally supports these methods, as samples that include both a preferred (post-caption) and rejected (pre-caption) output form ready-made preference pairs.

{\bf Ours: training with explicit preferences and critiques.}
While DPO and RLHF-V only compare preferred and rejected outputs implicitly, we further consider explicitly training models to generate (1) critiques and (2) preference labels.
\begin{enumerate}
    \item {\bf Critiques:} We train the model to generate a $\text{critique}$ for each $(\text{video}, \text{caption})$ pair using reference critiques when available. If a caption is already correct, the target critique is ``{\it The caption is accurate and requires no edits.}'' This training enables the model to perform more effective self-critique during inference.
    \item {\bf Preference labels:} We train the model to classify each $(\text{video}, \text{caption})$ pair into $\{{\tt Yes}, {\tt No}\}$, where ``${\tt Yes}$'' denotes the preferred (post-)caption and ``${\tt No}$'' the rejected (pre-)caption. At inference time, following VQAScore~\cite{lin2024evaluating}, the model's probability of ``${\tt Yes}$'' is used as a reward score, which we find more reliable than prompting for Likert-scale ratings in \autoref{fig:scoring_ablation}.
\end{enumerate}

{\bf Findings.}
\autoref{tab:caption_reward_critique} presents our main results. (1) Current models perform well on {\it subject} and {\it scene} but struggle with {\it motion} and {\it camera} aspects, which are likely underrepresented in training data. (2) Our fine-tuned model achieves state-of-the-art performance, surpassing both open- and closed-source baselines. (3) Adding explicit {\it preference} and {\it critique} signals consistently improves both SFT and RL methods across all tasks. Additional results in \autoref{sec:inference_scaling} show that (4) inference-time scaling brings further gains without extra human supervision on reward modeling and caption generation tasks.

{\bf Critique quality is the key.}
Effective critiques must be {\it accurate}, {\it complete}, and {\it constructive}, explaining not only what is wrong but also how to fix it (see \autoref{fig:critique_quality} for an example).
In our workflow, annotators must write critiques that directly guide the model to produce the final post-caption, which naturally enforces all three properties.
To show why they matter, we use Gemini-2.5-Pro to take our ground-truth critiques and inject controlled errors:
(1) making critiques {\it inaccurate} by replacing a correct point with an incorrect one (e.g., changing ``{\it the man is wearing black, not white}'' to ``{\it the man is wearing blue, not white}'') or inserting a spurious point when the pre-caption is already correct;
(2) making critiques {\it incomplete} by removing a necessary correction (unless the critique is already ``{\it The caption is accurate and requires no edit}'');
and (3) making critiques {\it non-constructive} by removing the fix (e.g., reducing ``{\it the man is wearing black, not white}'' to ``{\it the man is not wearing white}'').
We also compare against critiques from Gemini-2.5-Pro given both the video and the pre-caption, from a blind Gemini-2.5-Pro that hallucinates a critique using only the pre-caption, and from annotators before the second-stage quality control.
\autoref{tab:critique_comparison} confirms that weakening any of the three properties leads to a substantial drop in performance.
Conversely, adding a second stage of quality control noticeably improves results, presumably by strengthening all three dimensions.
Critiques generated by Gemini-2.5-Pro perform poorly even when the model has access to the video, suggesting that current models still struggle to produce accurate, complete, and constructive critiques.
As discussed further in \autoref{sec:critique_quality}, these qualities are often overlooked; prior work such as MM-RLHF~\cite{zhang2025mm} and OpenAI's GDC~\cite{saunders2022self} often collects feedback that identifies errors but offers no corrections. This underscores that critique quality, not just quantity, is essential for scalable oversight.

\begin{table*}[tb!]
\centering
\caption{\small \textbf{Critique quality determines post-training success.}
We analyze three critique failure types ({\it Inaccurate}, {\it Incomplete}, and {\it Unhelpful}), corresponding to violations of precision, recall, and constructiveness.
\autoref{sec:critique_quality} details how we generate these negative critiques from ground-truth human critiques using LLMs.
We also compare against critiques from Gemini-2.5-Pro given both the video and the pre-caption, from a blind Gemini-2.5-Pro that hallucinates critiques from the pre-caption alone, and from annotators before the second-stage quality control.
We see that weakening any of the three properties leads to a substantial drop in post-training performance, while adding a second stage of quality control noticeably improves results.}
\resizebox{0.8\textwidth}{!}{
\begin{tabular}{@{}lcccccc@{}}
\toprule[1.2pt]
\multirow{2}{*}{\textbf{Critique Type}} &
\multicolumn{3}{c}{\textbf{Quality Metrics}} &
\multicolumn{3}{c}{\textbf{Task Performance}} \\
\cmidrule(lr){2-4} \cmidrule(lr){5-7}
 & \textbf{Precision} & \textbf{Recall} & \textbf{Constructiveness} & \textbf{Caption} & \textbf{Reward} & \textbf{Critique} \\
\midrule
Blind Gemini-2.5 & -- & -- & -- & 10.9 & 44.5 & 21.1 \\
Gemini-2.5 & -- & -- & -- & 12.7 & 62.0 & 26.2 \\
Inaccurate critique & \xmark & \cmark & \cmark & 12.1 & 47.1 & 21.9 \\
Incomplete critique & \cmark & \xmark & \cmark & 12.5 & 56.6 & 28.7 \\
Unhelpful critique & \cmark & \cmark & \xmark & 13.4 & 67.2 & 32.9 \\
\midrule
Our critique (w/o quality check) & -- & -- & -- & 14.8 & 73.1 & 35.7 \\
Our critique (w/ quality check) & \cmark & \cmark & \cmark & {\bf 18.2} & {\bf 89.8} & {\bf 41.7} \\
\bottomrule[1.2pt]
\end{tabular}
}
\label{tab:critique_comparison}
\end{table*}

\begin{figure}[t!]
\centering
    \includegraphics[width=0.48\textwidth]{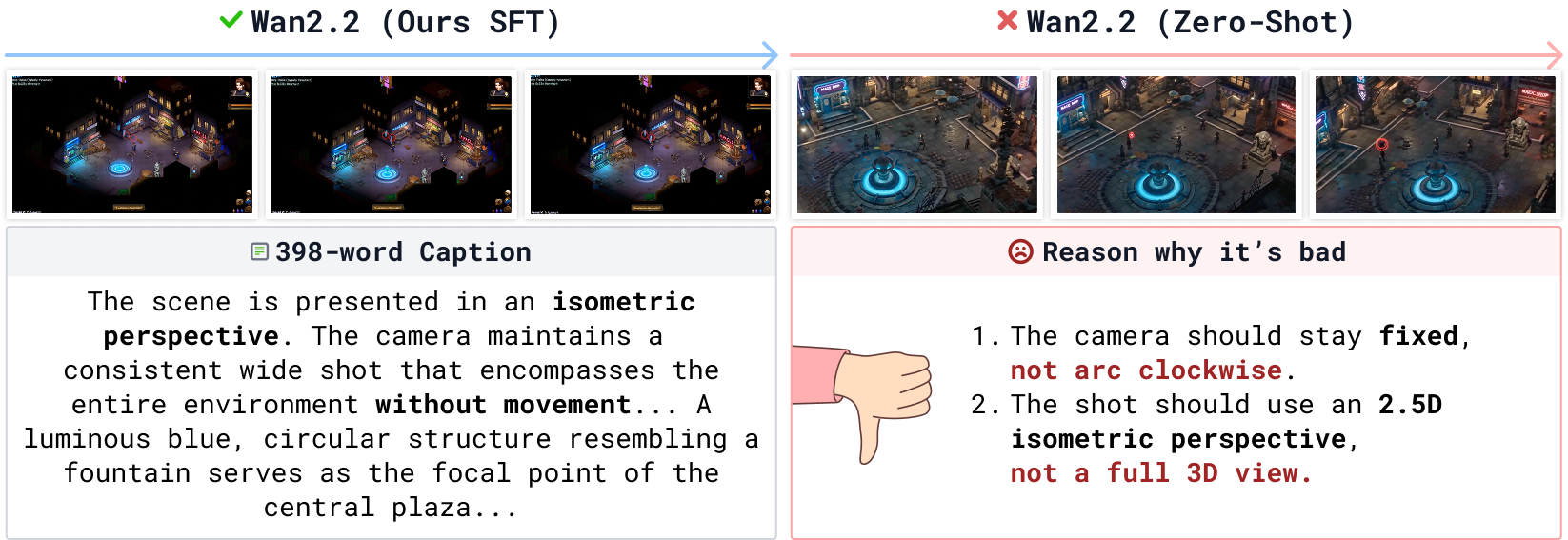}
    \vspace{6pt}
    \includegraphics[width=0.48\textwidth]{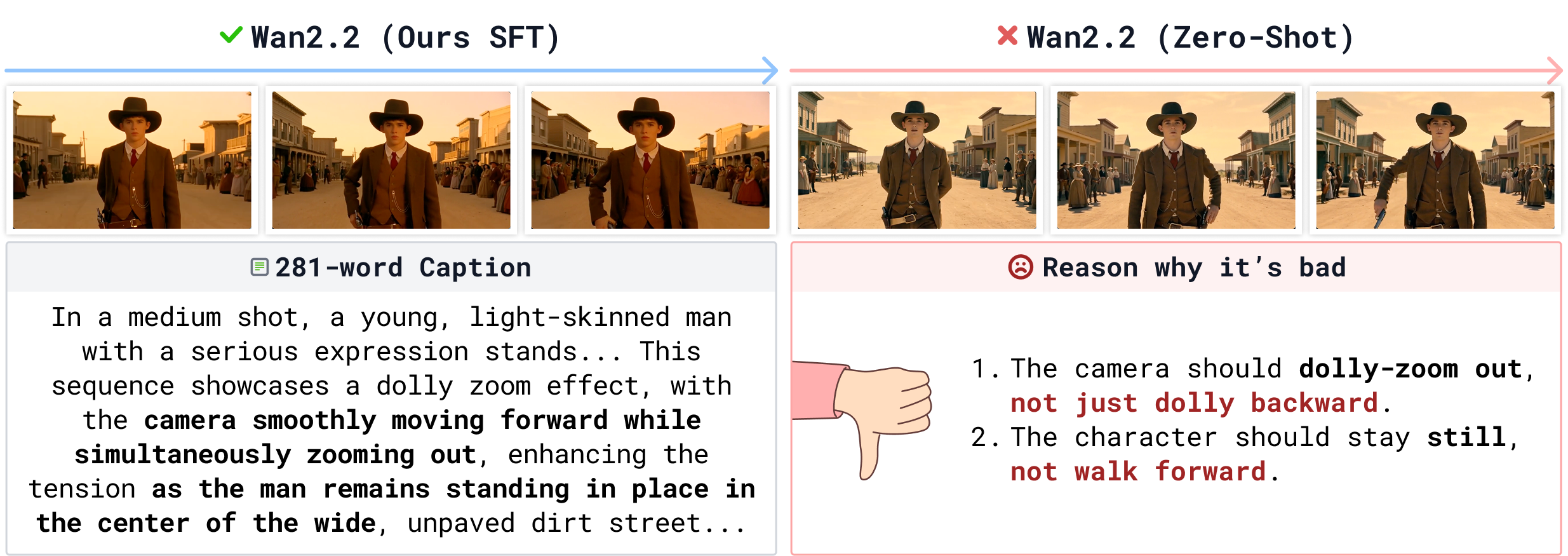}
    \caption{{\bf Re-captioning improves video generation.} After fine-tuning on large-scale videos re-captioned by our post-trained Qwen3-VL, Wan2.2 can better follow detailed prompts (up to 400 words) more faithfully and achieve finer control over camera motion and video cinematography, such as dolly zoom movements and isometric (2.5D) views. Evaluations in \autoref{sec:video_generation_details} confirm that this pipeline outperforms both zero-shot Wan and Wan fine-tuned using zero-shot Qwen3-VL captions.}\label{fig:videogeneration}
\end{figure}

\section{Improving Professional Video Generation}
\label{sec:video_generation}
High-quality captions are crucial for improving visual generation~\cite{dalle3, chatterjee2025stable}. We use our post-trained model to re-caption large-scale professional videos and fine-tune Wan2.2~\cite{wan2025} for better prompt following. We manually collect about $\sim$150K videos from YouTube channels covering films, commercials, music videos, and game footage under standard (non-commercial) licenses.

{\bf Re-captioning improves text-to-video.} \autoref{fig:videogeneration} shows that our fine-tuned Wan model better follows detailed prompts and supports finer control such as {\it dolly zoom} and {\it isometric} (2.5D) perspectives. Additional examples in \autoref{sec:video_generation_details} demonstrate control over camera movements ({\it rolling} in \autoref{fig:rolling_cw}), camera angles ({\it Dutch angle} in \autoref{fig:dutch_angle}),  focus shifts ({\it rack focus} in \autoref{fig:rack_focus}), playback speed ({\it speed ramp} in \autoref{fig:speed_ramp}), game perspective ({\it side-view} in \autoref{fig:side_view}), overlays ({\it watermark} in \autoref{fig:watermark}), height transitions ({\it underwater} to {\it above-water} in \autoref{fig:waterlevels}), shot size ({\it medium shot} to {\it close-up} in \autoref{fig:shot_size}), and subject transition ({\it revealing shot} in \autoref{fig:revealing_shot}). Evaluations on 200 samples show that fine-tuning on our re-captioned data significantly improves prompt following compared to Wan2.2 fine-tuned with zero-shot Qwen3-VL captions (\autoref{sec:video_generation_details}).

\section{Conclusion}
{\bf Limitations and future work.}
Our CHAI framework shows that high-quality data can be curated efficiently through human–AI collaboration, even with modest academic resources.
Future work can further scale this process by using stronger critique models to assist human annotators~\cite{saunders2022self, bowman2022measuring}.
While this work focuses on video understanding, building video generation benchmarks is a natural next step.

{\bf Summary.}
We present an open recipe, dataset, and benchmarks for precise video captioning, combining clear specification, scalable oversight, and effective post-training.
We hope this work advances scalable human–AI collaboration for high-quality data curation in video understanding and generation.

\section*{Acknowledgements}
We are grateful to Xue Bai, Rushikesh Zawar, Qunlin Jin, Yu Huang, Jackie Li, Daniel Jiang, Gautam Gare, Sally Chen, Nikhil Keetha, Jay Karhade, Jean de Dieu Nyandwi, and Graham Neubig for their generous feedback, thoughtful discussions, and encouragement throughout this work. This material is based upon work supported by the National Science Foundation Graduate Research Fellowship Program under Grant No(s) (NSF grant number: DGE2140739 ). Any opinions, findings, and conclusions or recommendations expressed in this material are those of the author(s) and do not necessarily reflect the views of the National Science Foundation.

\clearpage
{
    \small
    \bibliographystyle{ieeenat_fullname}
    \bibliography{main}
}

\newpage
\clearpage
{
    \centering
    \Large
    \textbf{Building a Precise Video Language with Human–AI Oversight} \\ \vspace{0.5em} {Supplementary Material}\\
    \large
}

\appendix

\vspace{-0.1in}

\section*{}
\begin{center}
    \emph{\bf \em \large Outline}
\end{center}

{Below is the outline of the supplement:
\begin{itemize}

\item {\bf Section \ref{sec:prior_work_errors}} provides a detailed error analysis of prior datasets and crowdsourced annotations.

\item {\bf Section \ref{sec:specification_details}} presents our full specification, including creator demographics and captioning policy.

\item {\bf Section \ref{sec:oversight_details}} details our complete oversight pipeline, including annotator screening, training, platform design, and promotion criteria.

\item {\bf Section \ref{sec:more_examples}} shows additional examples of (pre-caption, critique, post-caption) triplets from our dataset.

\item {\bf Section \ref{sec:critique_quality}} details how we generate negative critiques that violate precision, recall, and constructiveness using LLMs, and examines the issue of non-constructive critiques in prior work.

\item {\bf Section \ref{sec:post_training}} provides additional ablations of evaluation metrics and reward scoring formats, details SFT dataset preparation, explains adversarial negative critique generation, and reports hyperparameters.

\item {\bf Section \ref{sec:inference_scaling}} shows inference-time scaling results.

\item {\bf Section \ref{sec:video_generation_details}} presents details of the web-scraped video set and the fine-tuning experiments for text-to-video generative models.

\item {\bf Section \ref{sec:captioning_policy}} provides the complete captioning guidelines for human annotators.

\item {\bf Section \ref{sec:pseudocode_precaption}} presents Python-style pseudocode for converting labeled primitives into model instructions for generating pre-captions.

\end{itemize}
}

\section{Error Analysis of Prior Datasets}
\label{sec:prior_work_errors}

We worked with our top-performing annotators to manually evaluate eight widely used video–text datasets and benchmarks released between 2016 and 2025: MSR-VTT~\cite{xu2016msr}, ActivityNet~\cite{activitynet}, Dream1K~\cite{tarsier}, VDC~\cite{chai2024auroracap}, UltraVideo~\cite{xue2025ultravideo}, ShareGPT4Video~\cite{chen2025sharegpt4video}, TUNA-Bench~\cite{kong2025tuna}, and PerceptionLM~\cite{cho2025perceptionlm}.
This evaluation reveals systematic issues in caption accuracy, completeness, and consistency, which we support with quantitative results and visual examples.
We first summarize the main findings, then provide detailed per-dataset analyses and recommendations for future dataset curation.

{\bf Issues in prior datasets.}
Without clear specification, annotators lack guidance on what to describe and how much detail to include.
Without proper oversight, errors remain uncorrected and propagate through the dataset.
Together, these gaps produce captions that are inconsistent, incomplete, or incorrect.
We illustrate both problems below with examples from prior datasets.

{\bf Issue: lack of specification.} Without clear rules on what and how to describe, captions often become inconsistent and inaccurate:

\begin{itemize}
    \item {\bf (1) Imprecise terminology}. For example, the caption may misuse terms for visual effects.
    \item {\bf (2) Missing information}. For example, the caption may omit motion or spatial details.
    \item {\bf (3) Subjective descriptions}. For example, the caption may introduce emotional interpretations.
\end{itemize}

{\bf Issue: lack of oversight.} Curating long, detailed video descriptions is inherently demanding and error-prone. Without proper oversight, both humans and models can produce captions with errors that go unchecked:

\begin{itemize}
    \item {\bf (4) Poor writing.} For example, there may be typos, grammatical mistakes, or awkward phrasing. More subtle issues include describing multiple temporal events in the wrong order or mentioning multiple subjects without giving clear references based on visual attributes.
    \item {\bf (5) Visual hallucinations.} For example, the caption may hallucinate non-existent objects or motions.
    \item {\bf (6) Inaccurate details.} For example, the caption may confuse subtle visual details.
\end{itemize}

\begin{figure*}[t!]
\centering
\includegraphics[width=\textwidth]{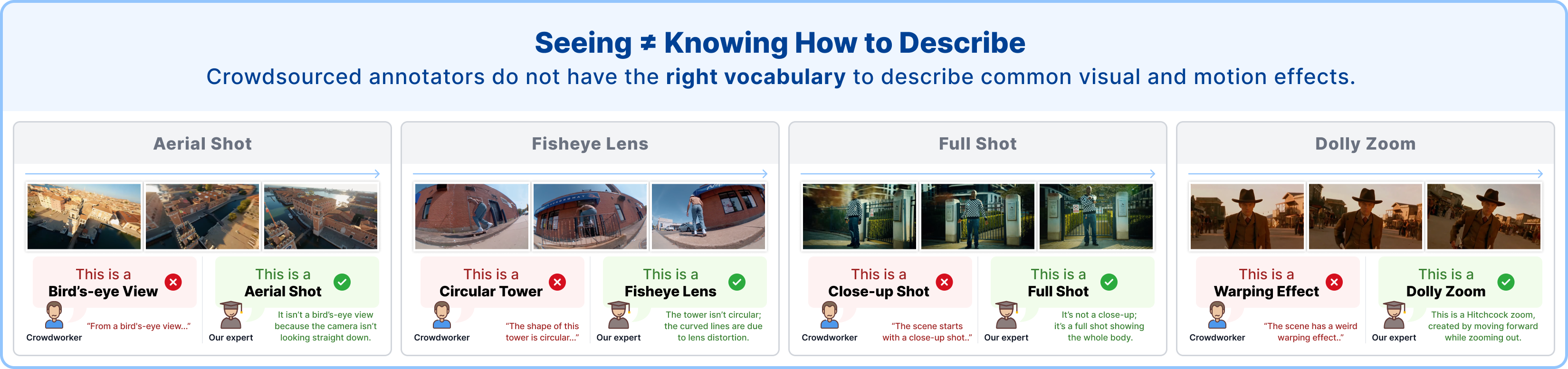}
    \caption{\small {\bf Errors in crowdsourced video captions.} Crowdsourced captions often omit key details, producing overly brief descriptions, and lack the visual vocabulary needed to describe common cinematic or motion effects. For example, crowdworkers may confuse a high vantage viewpoint with a ``{\it bird's-eye view}'' (which refers to a top-down angle), describe a regular building as a ``{\it circular tower}'' because a fisheye lens bends straight lines outward, mistake a full shot of a human body for a ``{\it close-up},'' or describe a dolly zoom (where the camera moves while zooming in the opposite direction) as a ``{\it weird warping effect}.'' These errors motivate our decision to work with professional video creators to define the specification and annotate captions.
    }\label{fig:crowdsourced}
\end{figure*}

\begin{table*}[tb!]
\centering
\caption{\small \textbf{Comparison of video--text datasets on video statistics, caption source, task rating, and known issues.} We omit the Overall score for MSR-VTT and ActivityNet Captions because they primarily describe human actions without covering other aspects.
We mark only issues that occur frequently (in more than 15\% of the samples we manually evaluated).
To ensure robustness, we evaluate over 20 random samples per dataset and provide their original dataset IDs.
Caption collection methods are reported from the original papers, except VDC, which claims manual evaluation but appears AI-generated.
Scores are not directly comparable across datasets, as video content, annotation specifications, and caption detail levels vary significantly; for example, ShareGPT4Video achieves a similar score to UltraVideo despite using an older image-based GPT-4V model without temporal reasoning, because its videos are generally less dynamic.
Among existing datasets, TUNA-Bench (1k videos) achieves the highest quality across most aspects, especially subject, scene, and motion.
Detailed qualitative examples and findings are presented in \autoref{sec:prior_work_errors}.
We emphasize that these scores serve only as illustrative references and encourage readers to review the error cases on our website for deeper insights.}
\resizebox{\textwidth}{!}{
\begin{tabular}{@{}lcccccccccccccccccc@{}}
\toprule[1.5pt]
\multirow{2}{*}{\textbf{Dataset}} &
\multicolumn{3}{c}{\textbf{Video Statistics}} &
\multicolumn{2}{c}{\textbf{Caption}} &
\multicolumn{6}{c}{\textbf{Task Rating}} &
\multicolumn{3}{c}{\textbf{Specification}} &
\multicolumn{3}{c}{\textbf{Oversight}} \\
\cmidrule(lr){2-4} \cmidrule(lr){5-6} \cmidrule(lr){7-12} \cmidrule(lr){13-15} \cmidrule(lr){16-18}
 & \textbf{Size} & \textbf{Duration} & \textbf{FPS} & \textbf{Annotation} & \textbf{\# Word} & \textbf{Overall} & \textbf{Subject} & \textbf{Scene} & \textbf{Motion} & \textbf{Spatial} & \textbf{Camera} & \begin{tabular}[c]{@{}c@{}}Precise\\ Terminology\end{tabular} & \begin{tabular}[c]{@{}c@{}}Complete\\ Information\end{tabular} & \begin{tabular}[c]{@{}c@{}}Objective\\ Description\end{tabular} & \begin{tabular}[c]{@{}c@{}}Clear\\ Writing\end{tabular} & \begin{tabular}[c]{@{}c@{}}No\\ Hallucination\end{tabular} & \begin{tabular}[c]{@{}c@{}}Accurate\\ Details\end{tabular} \\
\midrule
MSR-VTT~\cite{xu2016msr}                 & 10k & $\sim$15s & 3 & Human       & $\sim$10 & -- & 1.8 & 1.2 & 3.3 & 1.0 & 1.0 & \xmark & \xmark & \cmark & \xmark & \cmark & \xmark \\
ActivityNet Captions~\cite{activitynet}           & 20k & $\sim$120s & 3 & Human       & $\sim$50 & -- & 1.5 & 1.5 & 3.7 & 1.3 & 1.0 & \xmark & \xmark & \cmark & \xmark & \xmark & \xmark \\
ShareGPT4Video~\cite{chen2025sharegpt4video} & 40k & $\sim$30s & $\sim$30 & AI          & $\sim$280 & 3.7 & 3.8 & 4.7 & 3.8 & 1.5 & 2.1 & \xmark & \xmark & \xmark & \cmark & \xmark & \xmark \\
UltraVideo~\cite{xue2025ultravideo}   & 59k  & $\sim$13s & $\sim$37  & AI          & $\sim$824 & 3.7 & 4.7 & 4.8 & 4.5 & 2.6 & 3.2 & \xmark & \xmark & \xmark & \cmark & \xmark & \xmark \\
VDC~\cite{chai2024auroracap}             & 1k & $\sim$30s & $\sim$30 & AI\qmark    & $\sim$500 & 3.6 & 4.4 & 4.6 & 4.4 & 2.4 & 2.7 & \xmark & \xmark & \xmark & \cmark & \xmark & \xmark \\
Dream1K~\cite{tarsier}                   & 1k & $\sim$9s & $\sim$30 & Human       & $\sim$60 & 3.3 & 2.3 & 1.7 & 4.7 & 1.6 & 2.0 & \xmark & \xmark & \cmark & \xmark & \cmark & \xmark \\
PerceptionLM~\cite{cho2025perceptionlm}  & 120k & $\sim$15s & $\sim$28 & AI $\rightarrow$ Human       & $\sim$50 & 4.0 & 4.0 & 3.2 & 4.7 & 2.1 & 1.4 & \xmark & \xmark & \cmark & \xmark & \cmark & \xmark \\
TUNA-Bench~\cite{kong2025tuna}           & 1k & $\sim$15s & $\sim$20 & Expert       & $\sim$200 & 4.1 & 4.9 & 4.7 & 4.9 & 3.5 & 2.2 & \xmark & \xmark & \cmark & \cmark & \cmark & \xmark \\
\midrule
Crowdsourced        & 1k & $\sim$6s & $\sim$30 & Human   & $\sim$110 & 3.2 & 3.4 & 3.5 & 3.5 & 2.4 & 2.5 & \xmark & \xmark & \xmark & \xmark & \xmark & \xmark \\
Ours (pre-caption w/o human)             & 4k & $\sim$6s & $\sim$30 & Expert $\leftrightarrow$ AI  & $\sim$395 & 3.9 & 3.9 & 3.9 & 3.9 & 3.4 & 4.5 & \cmark & \cmark & \cmark & \cmark & \xmark & \xmark \\
Ours                               &  4k     & $\sim$6s & $\sim$30  & Expert $\leftrightarrow$ AI  & $\sim$405 & 5.0 & 5.0 & 5.0 & 5.0 & 5.0 & 5.0 & \cmark & \cmark & \cmark & \cmark & \cmark & \cmark \\
\bottomrule[1.5pt]
\end{tabular}
}
\vspace{-3mm}
\label{tab:dataset_comparison_policy}
\end{table*}

\begin{figure*}[t!]
\centering
\setlength{\tabcolsep}{1pt} %
\renewcommand{\arraystretch}{0.5} %
\begin{tabular}{cc}
    \includegraphics[width=0.5\textwidth]{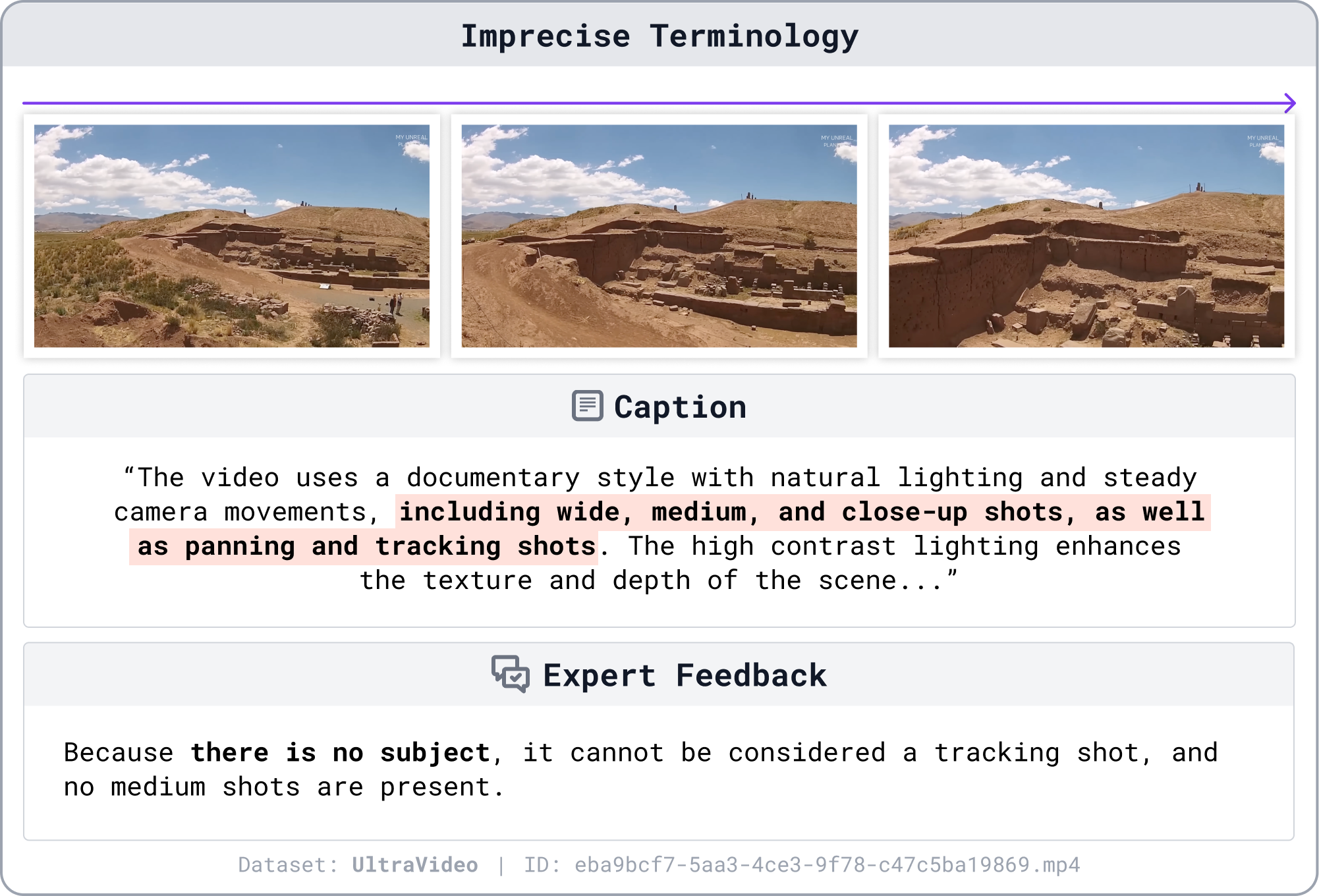} &
    \includegraphics[width=0.5\textwidth]{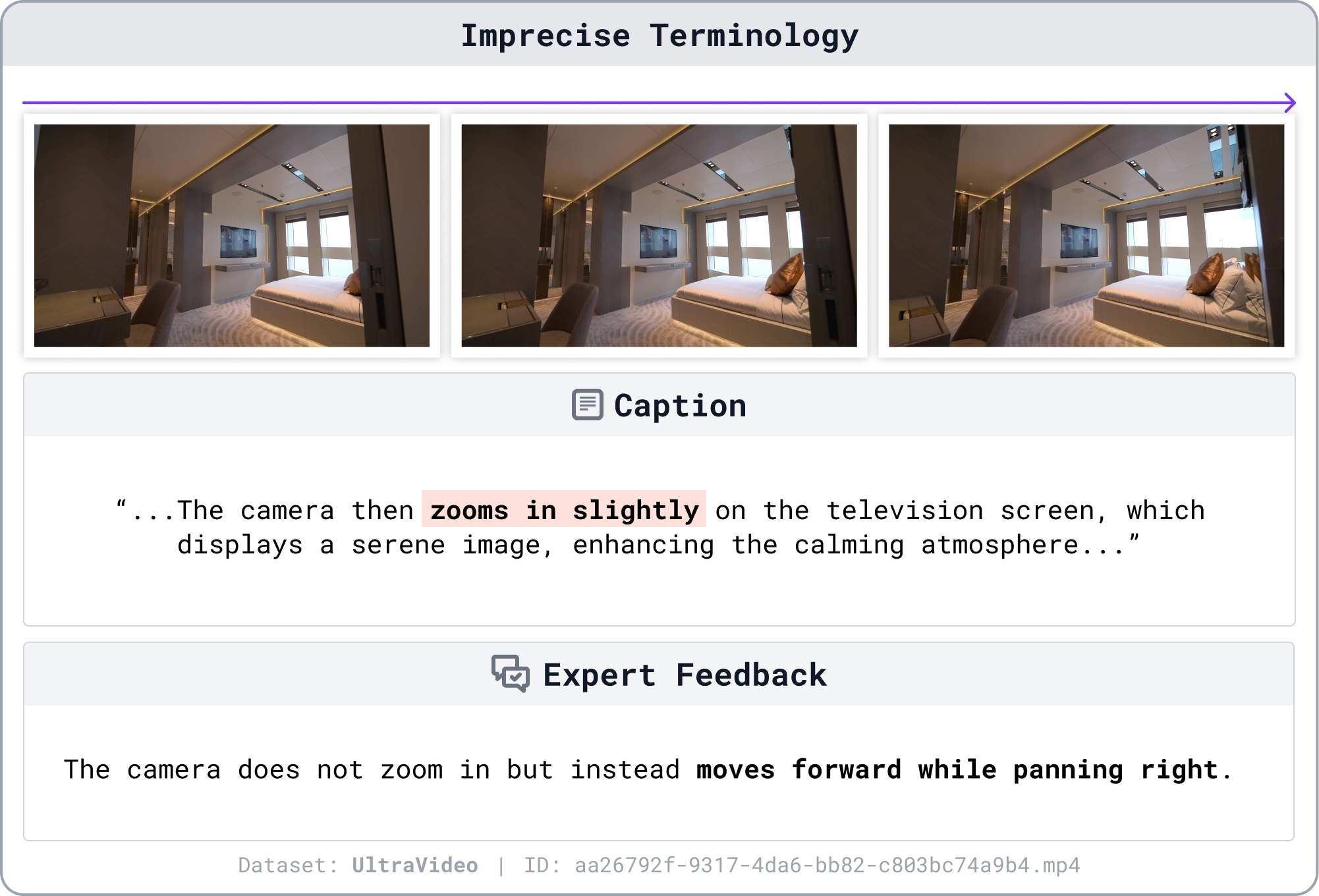} \\
    \multicolumn{2}{c}{\small \textbf{(a) Imprecise Terminology}} \\
    \noalign{\vspace{4mm}} %

    \includegraphics[width=0.5\textwidth]{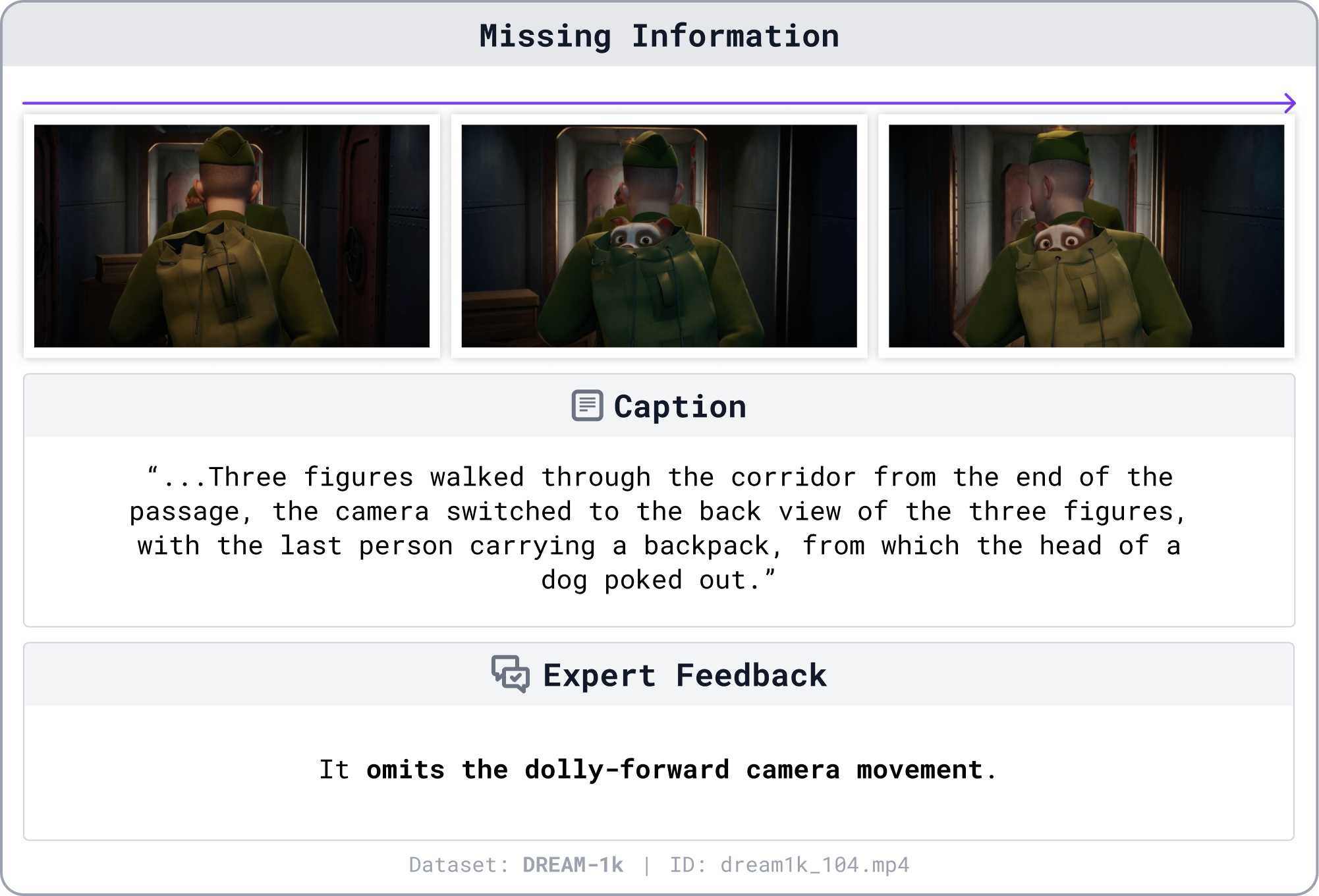} &
    \includegraphics[width=0.5\textwidth]{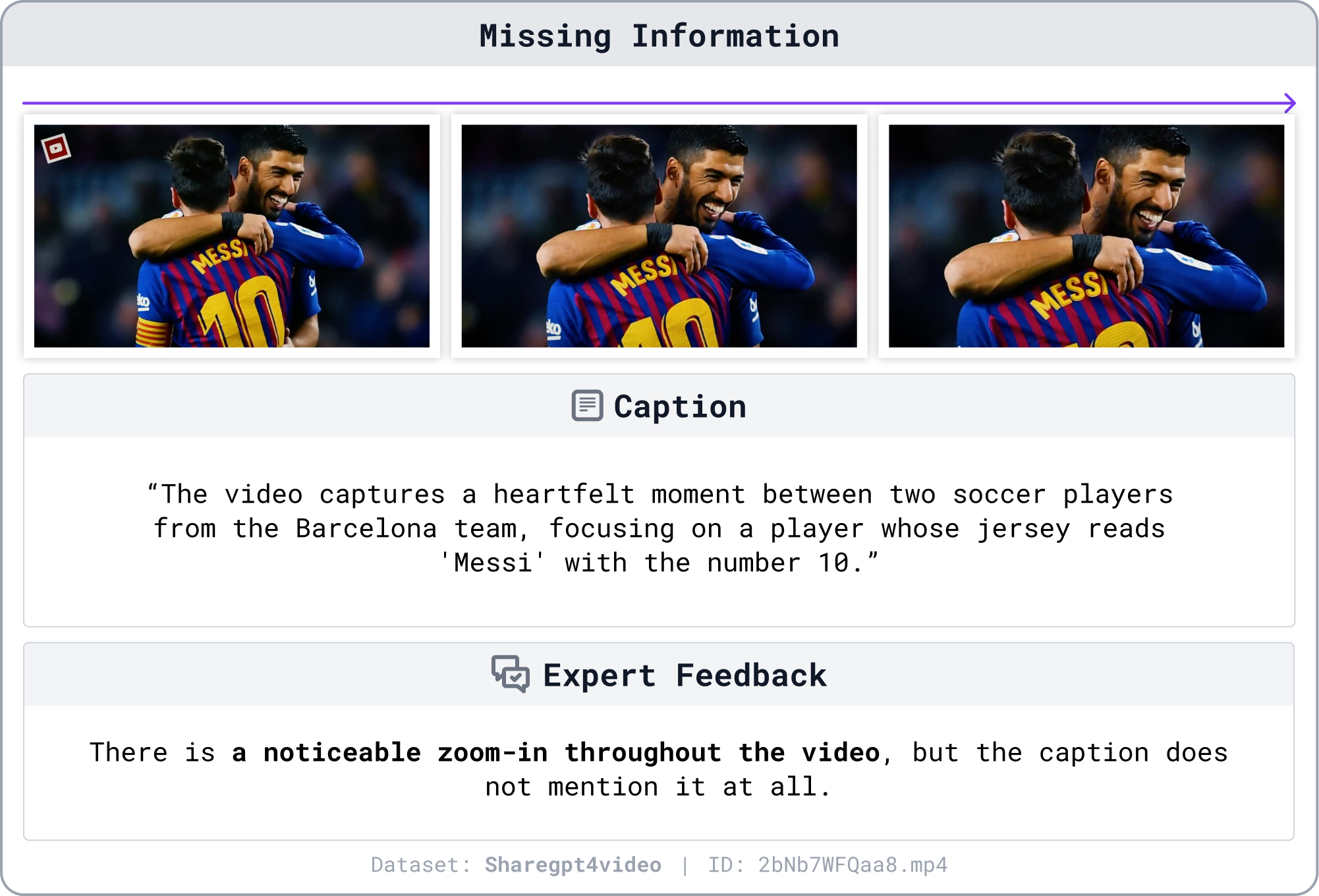} \\
    \multicolumn{2}{c}{\small \textbf{(b) Missing Information}} \\
    \noalign{\vspace{4mm}}

    \includegraphics[width=0.5\textwidth]{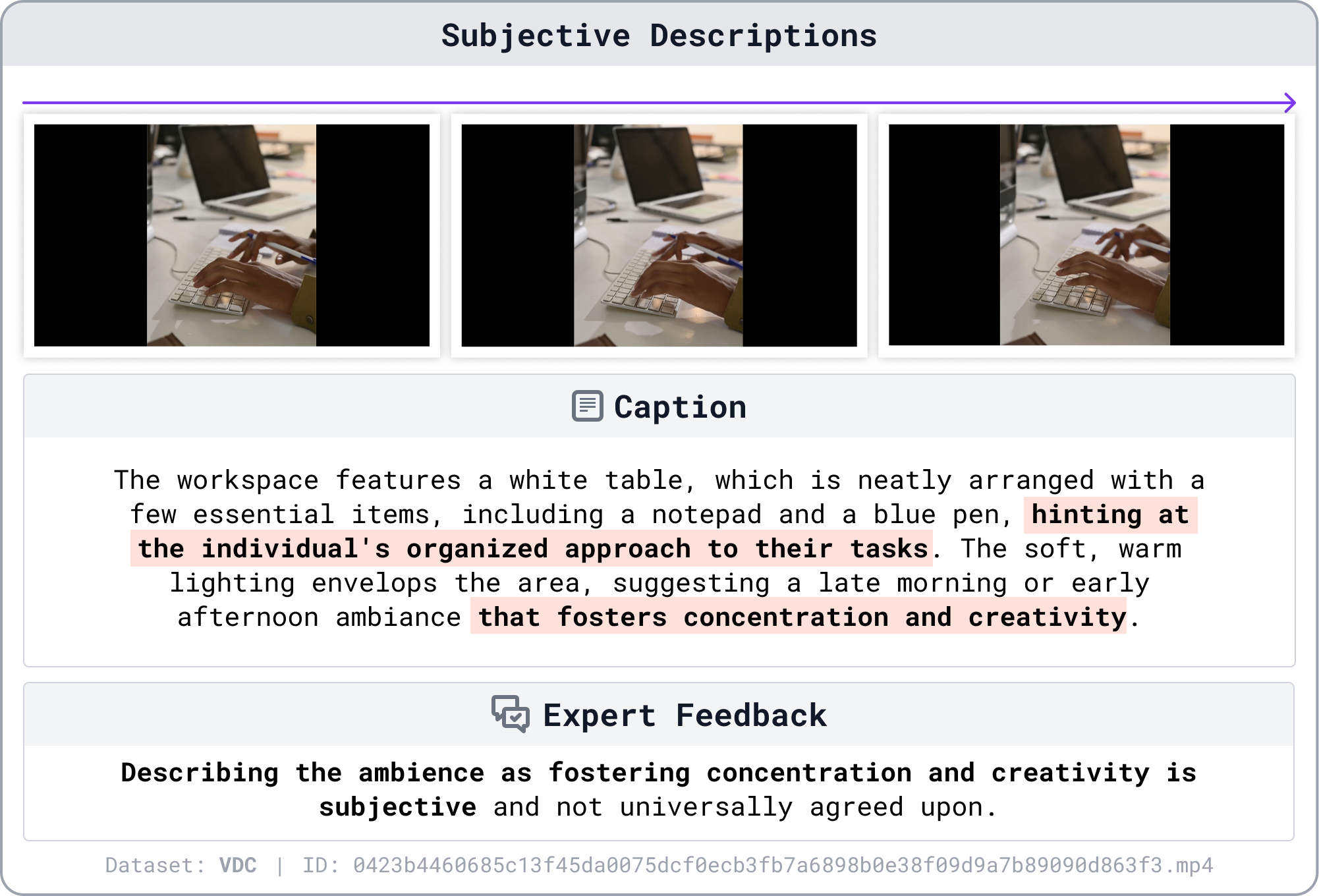} &
    \includegraphics[width=0.5\textwidth]{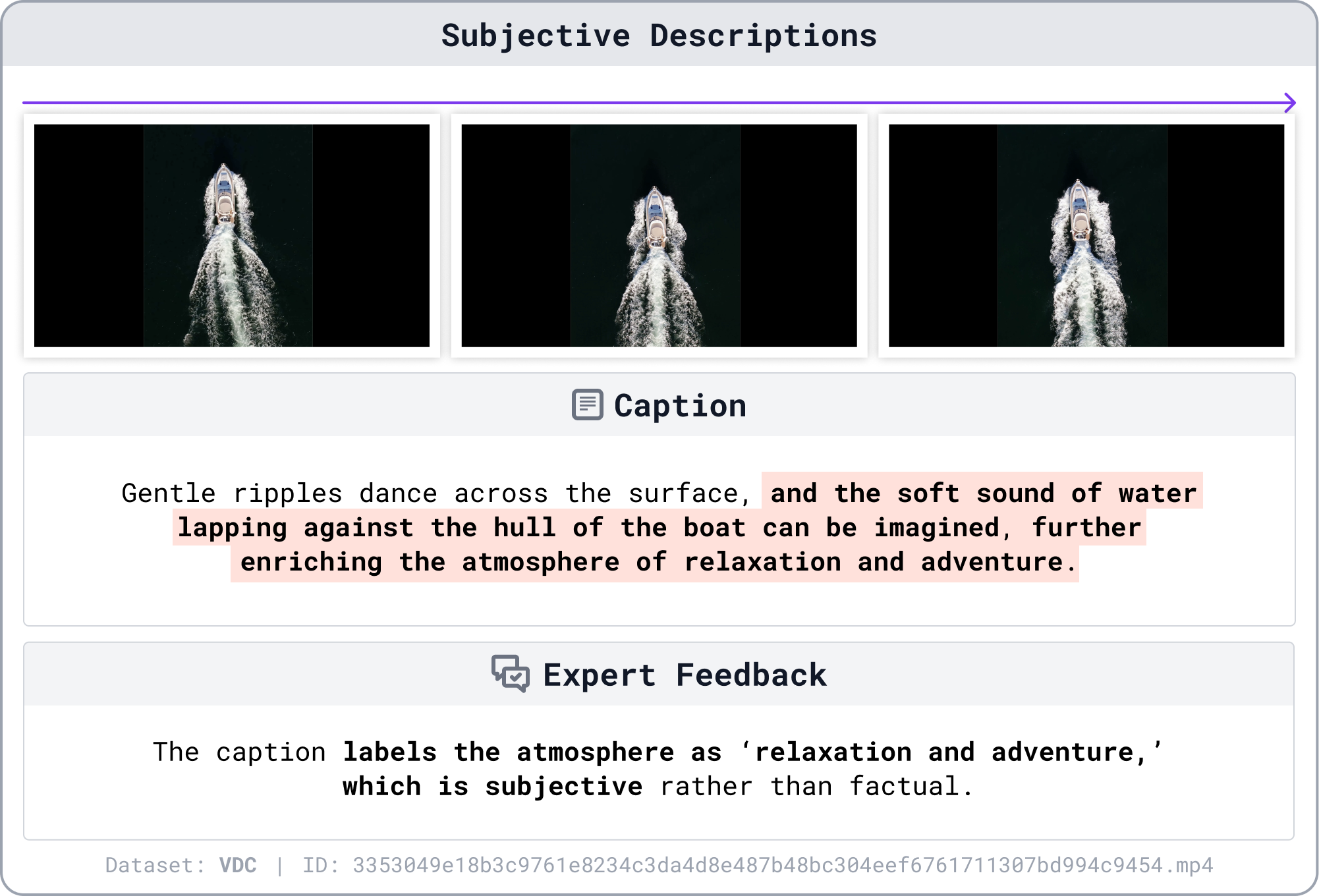} \\
    \multicolumn{2}{c}{\small \textbf{(c) Subjective Descriptions}} \\
\end{tabular}
\vspace{6mm}
\caption{\small \textbf{Examples of errors caused by lack of specification.} Without clear rules, captions often suffer from (a) misuse of terminology, (b) omission of details, or (c) inclusion of subjective interpretations.}
\label{fig:errors_specification}
\end{figure*}

\begin{figure*}[t!]
\centering
\setlength{\tabcolsep}{1pt}
\renewcommand{\arraystretch}{0.5}
\begin{tabular}{cc}
    \includegraphics[width=0.5\textwidth]{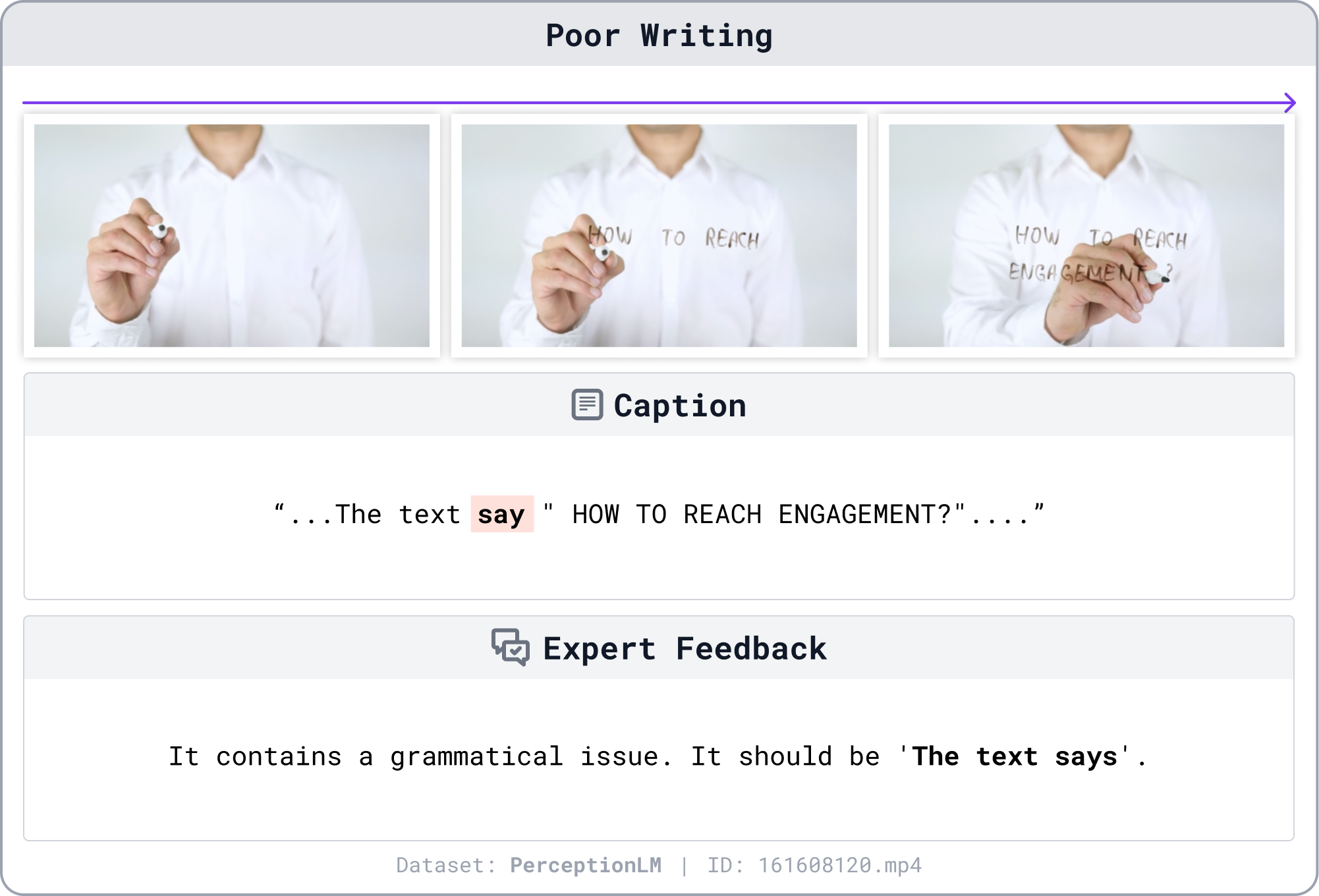} &
    \includegraphics[width=0.5\textwidth]{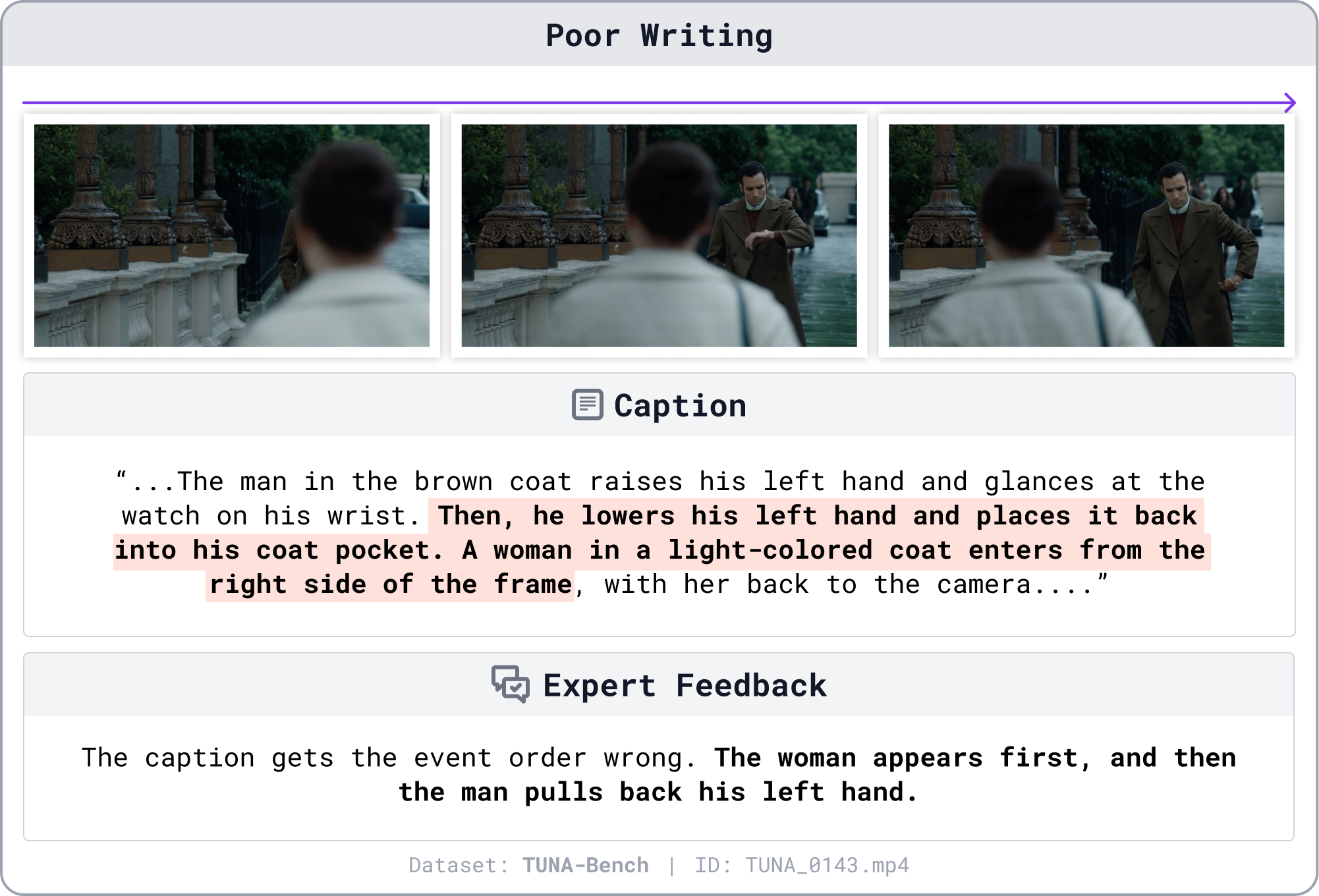} \\
    \multicolumn{2}{c}{\small \textbf{(a) Poor Writing}} \\
    \noalign{\vspace{4mm}}

    \includegraphics[width=0.5\textwidth]{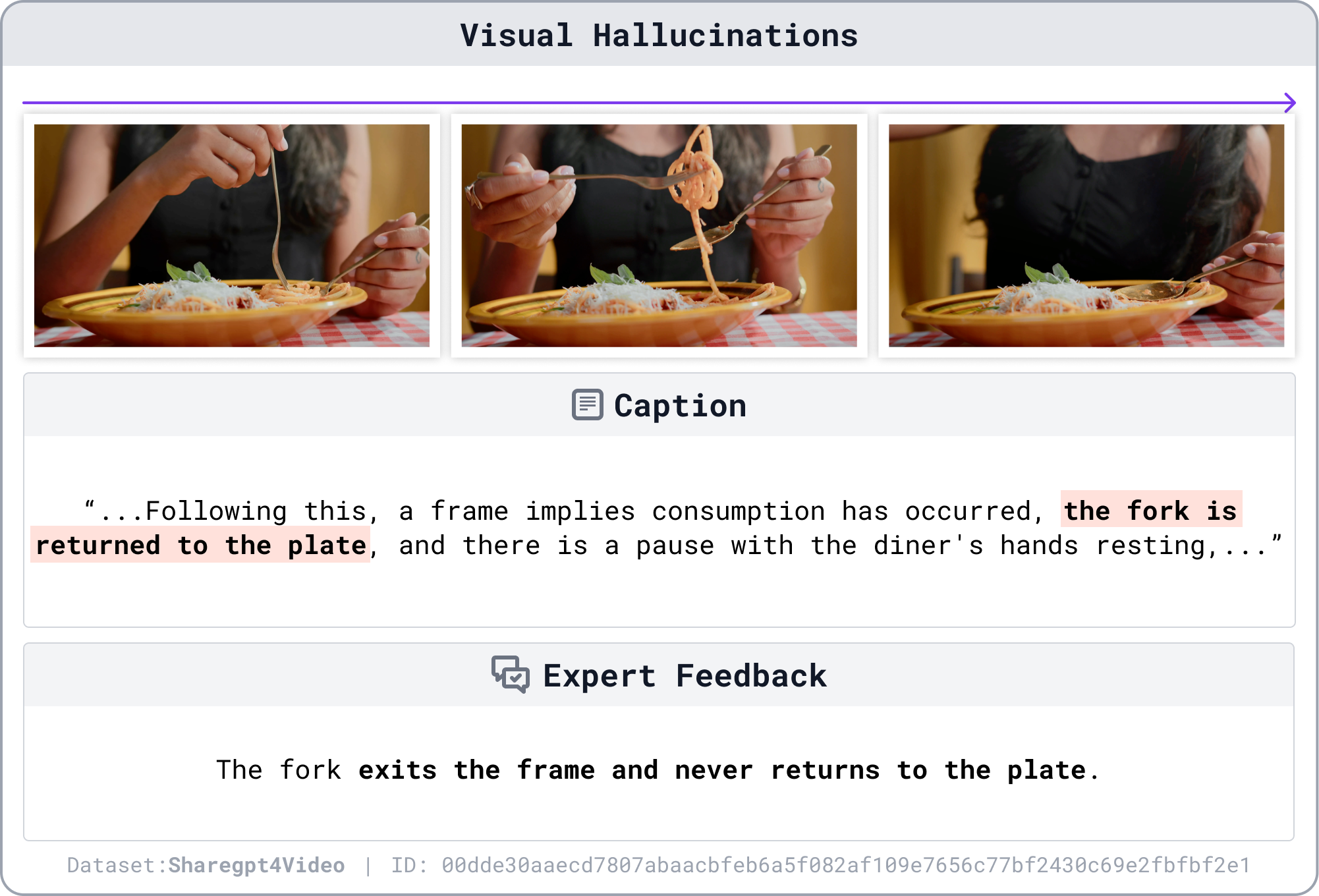} &
    \includegraphics[width=0.5\textwidth]{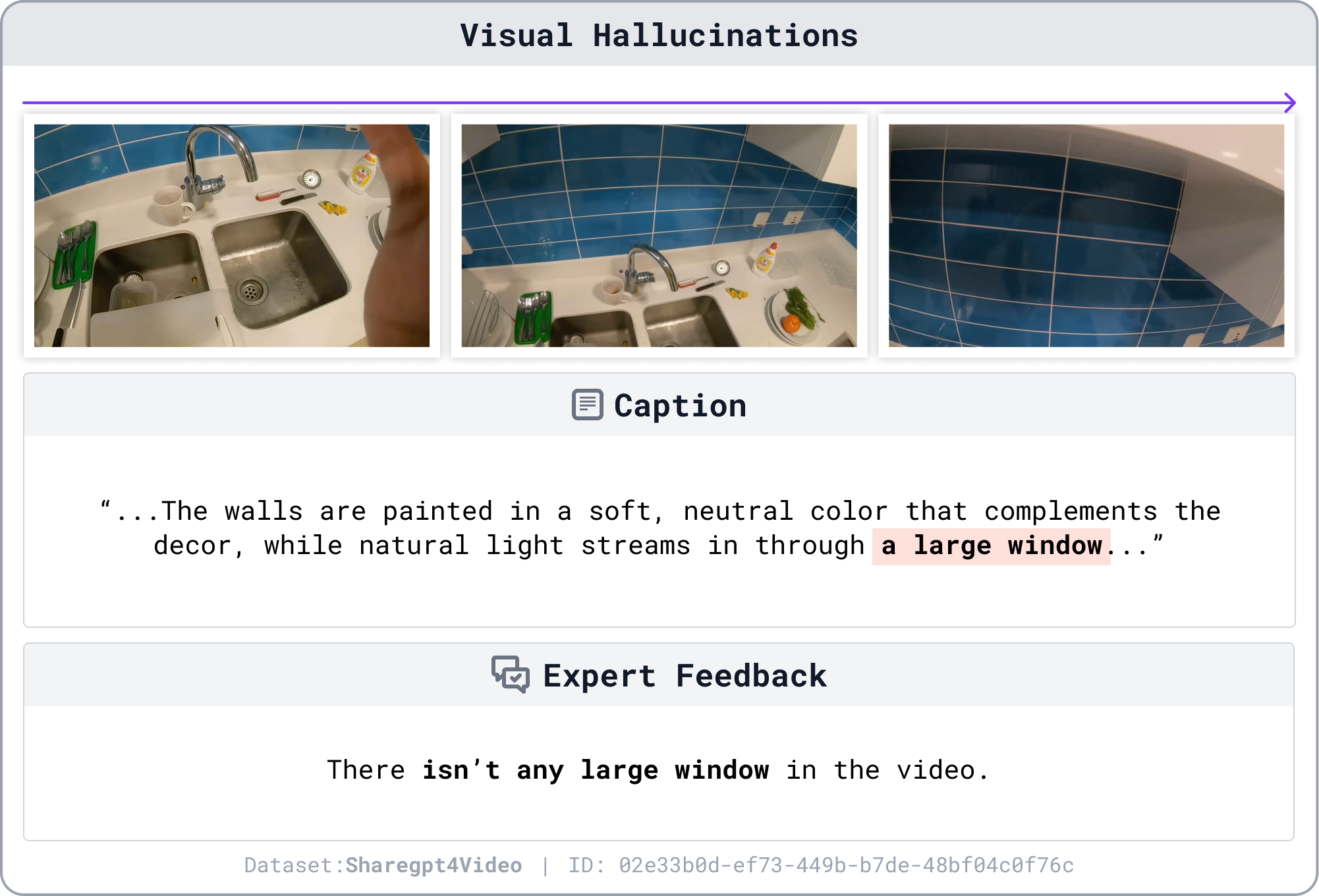} \\
    \multicolumn{2}{c}{\small \textbf{(b) Visual Hallucinations}} \\
    \noalign{\vspace{4mm}}

    \includegraphics[width=0.5\textwidth]{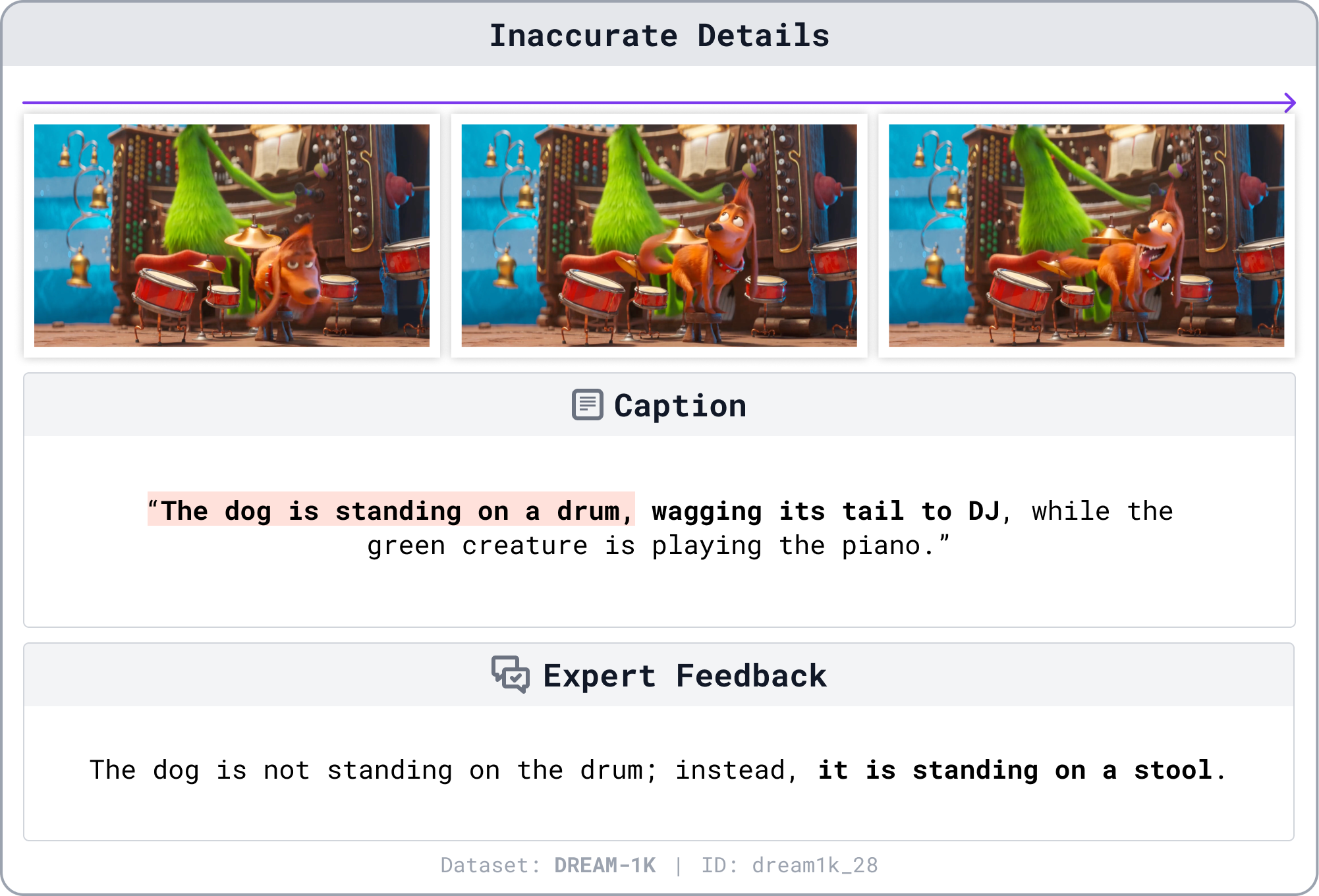} &
    \includegraphics[width=0.5\textwidth]{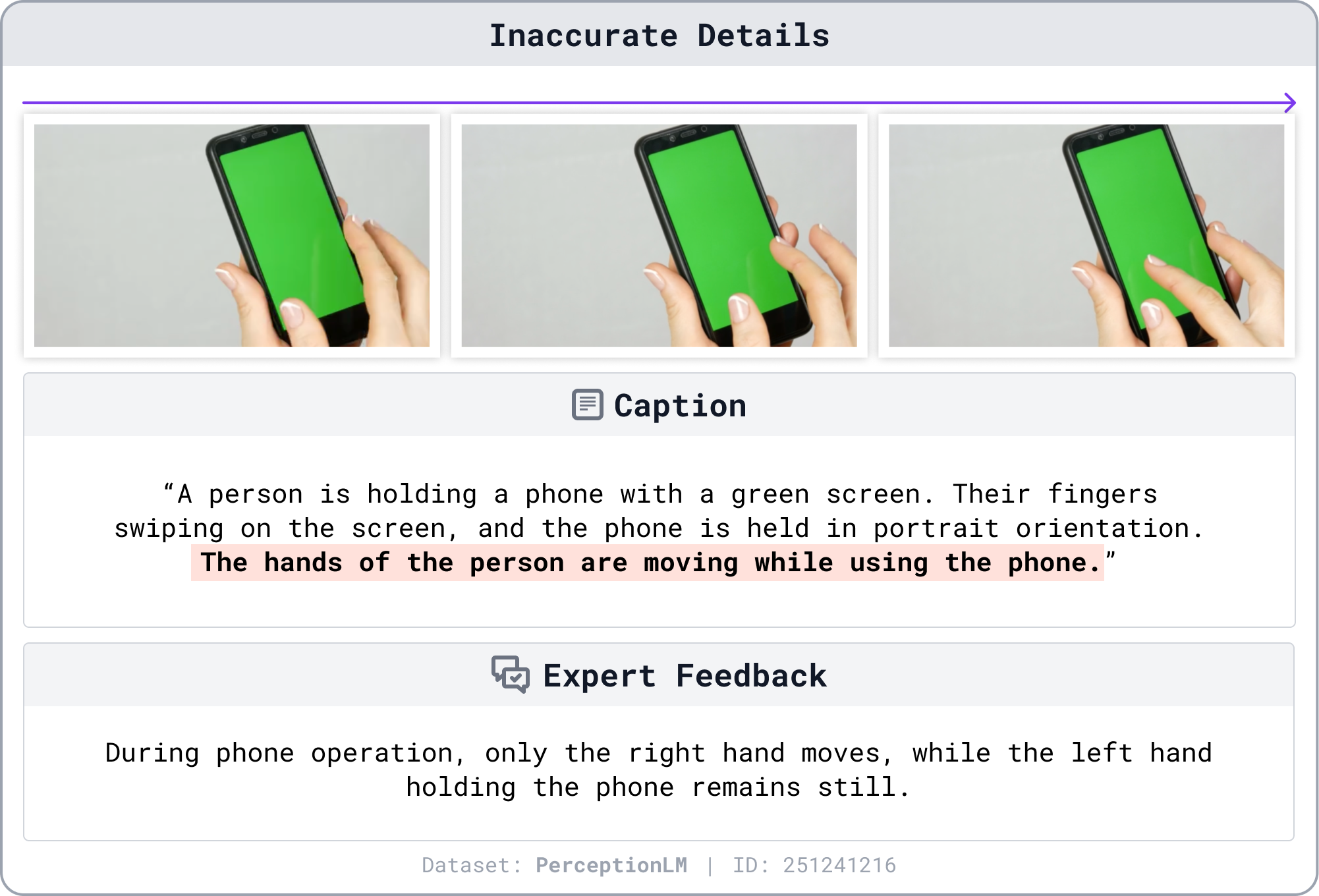} \\
    \multicolumn{2}{c}{\small \textbf{(c) Inaccurate Details}} \\
\end{tabular}
\vspace{6mm}
\caption{\small \textbf{Examples of errors caused by lack of oversight.} Even with detailed prompts, unchecked captions may contain (a) typos, grammar mistakes, or more serious issues like describing events out of temporal order, (b) hallucinations of non-existent objects or events, or (c) confusion of subtle visual details.}
\label{fig:errors_oversight}
\end{figure*}

\begin{figure*}[t!]
\centering
    \includegraphics[width=\textwidth]{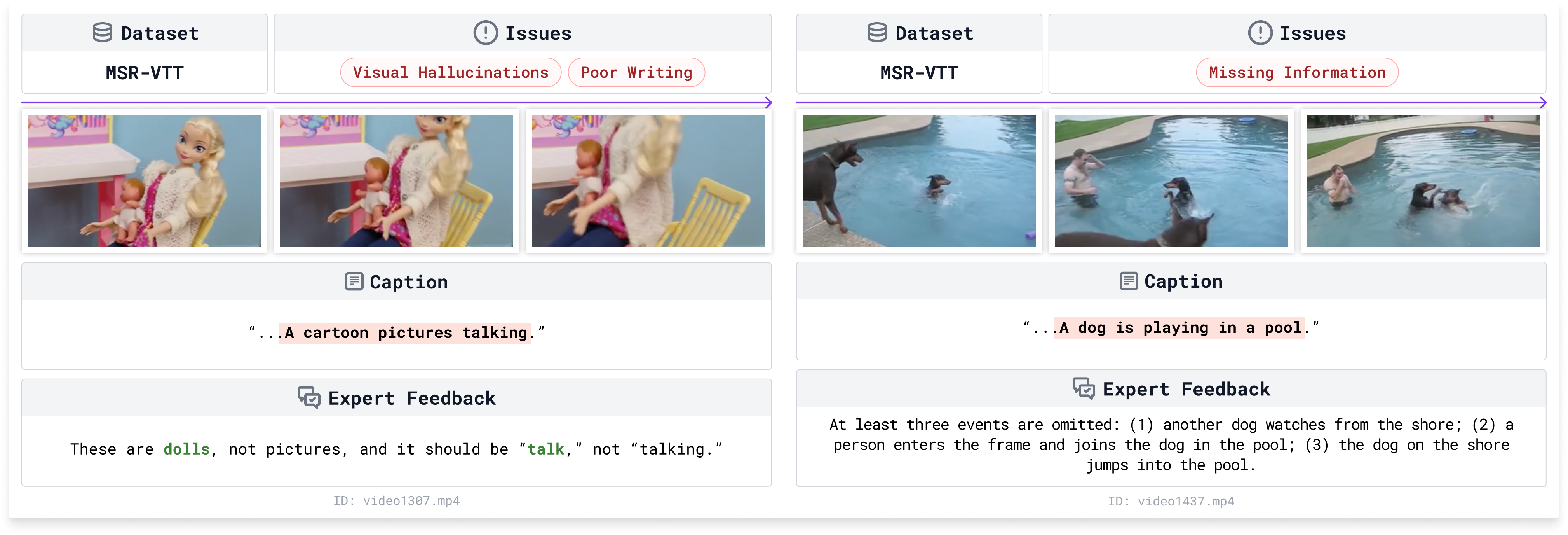}
    \vspace{1mm}
    \caption{\small {\bf Examples of error cases in MSR-VTT~\cite{xu2016msr}.} 
    }\label{fig:errors_msrvtt}
\end{figure*}

\begin{figure*}[t!]
\centering
    \includegraphics[width=\textwidth]{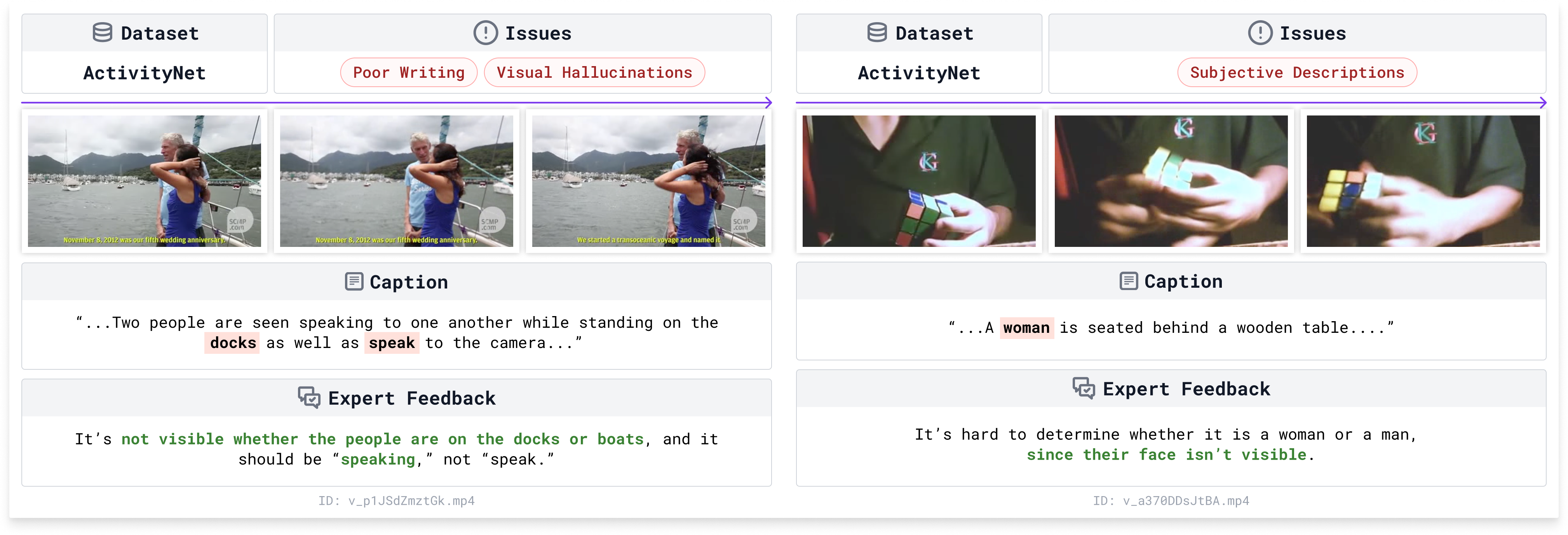}
    \caption{\small {\bf Examples of error cases in ActivityNet Captions~\cite{activitynet}.} 
    }\label{fig:errors_activitynet}
\end{figure*}

\begin{figure*}[t!]
\centering
    \includegraphics[width=\textwidth]{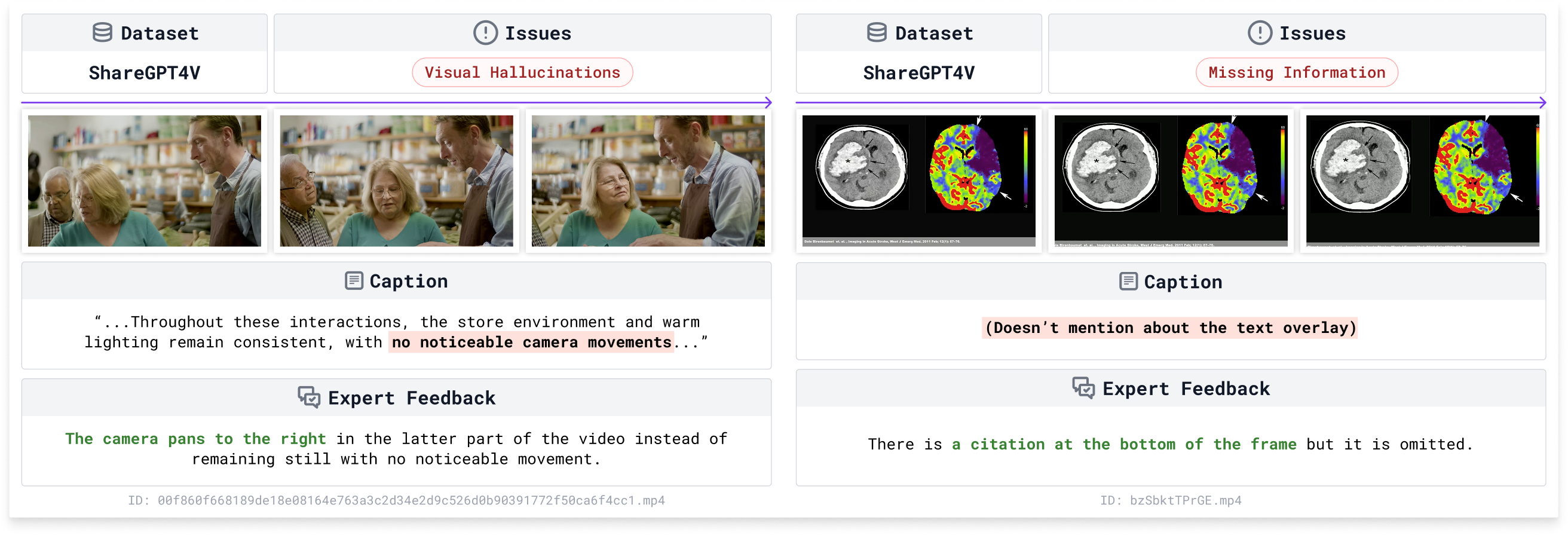}
    \caption{\small {\bf Examples of error cases in ShareGPT4Video~\cite{chen2025sharegpt4video}.} 
    }\label{fig:errors_sharegpt4video}
\end{figure*}

\begin{figure*}[t!]
\centering
    \includegraphics[width=\textwidth]{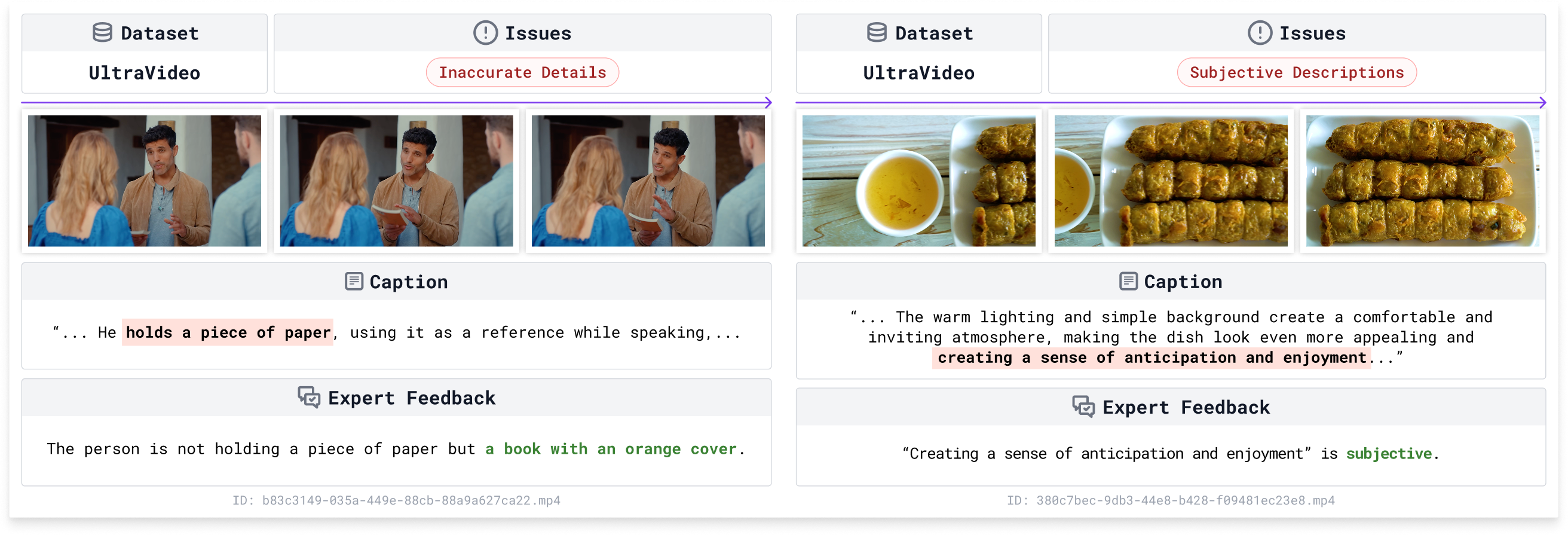}
    \vspace{1mm}
    \caption{\small {\bf Examples of error cases in UltraVideo~\cite{xue2025ultravideo}.} 
    }\label{fig:errors_ultravideo}
\end{figure*}

\begin{figure*}[t!]
\centering
    \includegraphics[width=\textwidth]{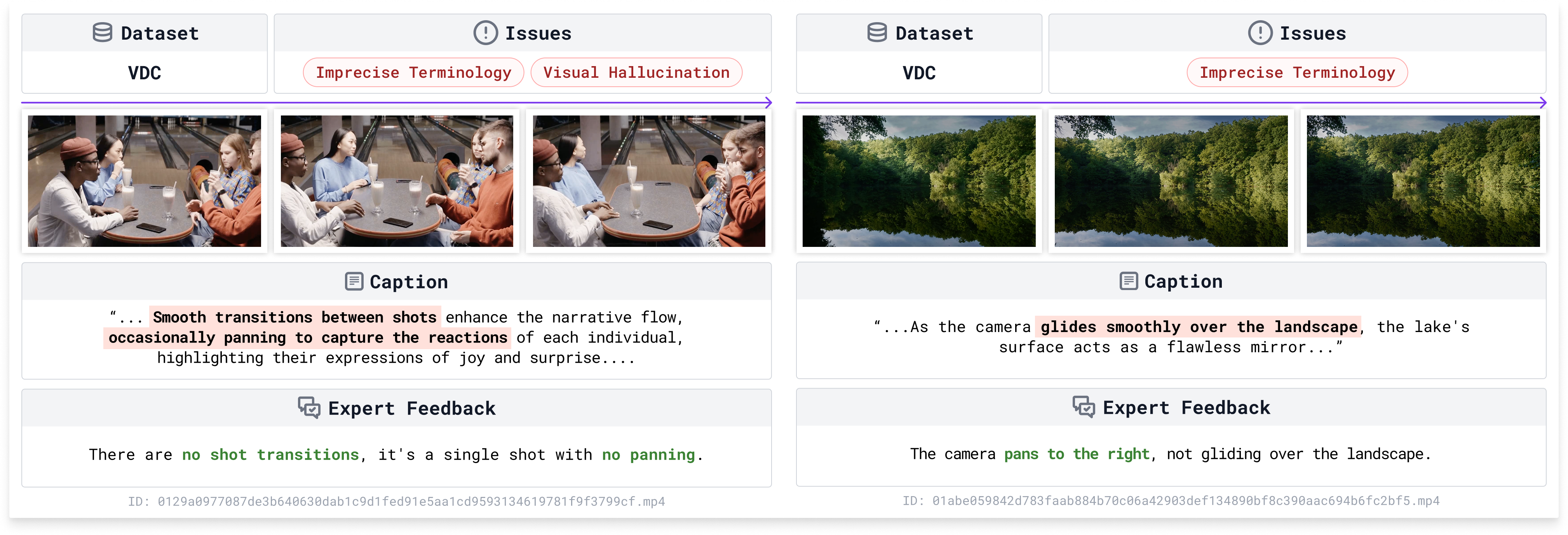}
    \caption{\small {\bf Examples of error cases in VDC~\cite{chai2024auroracap}.} 
    }\label{fig:errors_vdc}
\end{figure*}

\begin{figure*}[t!]
\centering
    \includegraphics[width=\textwidth]{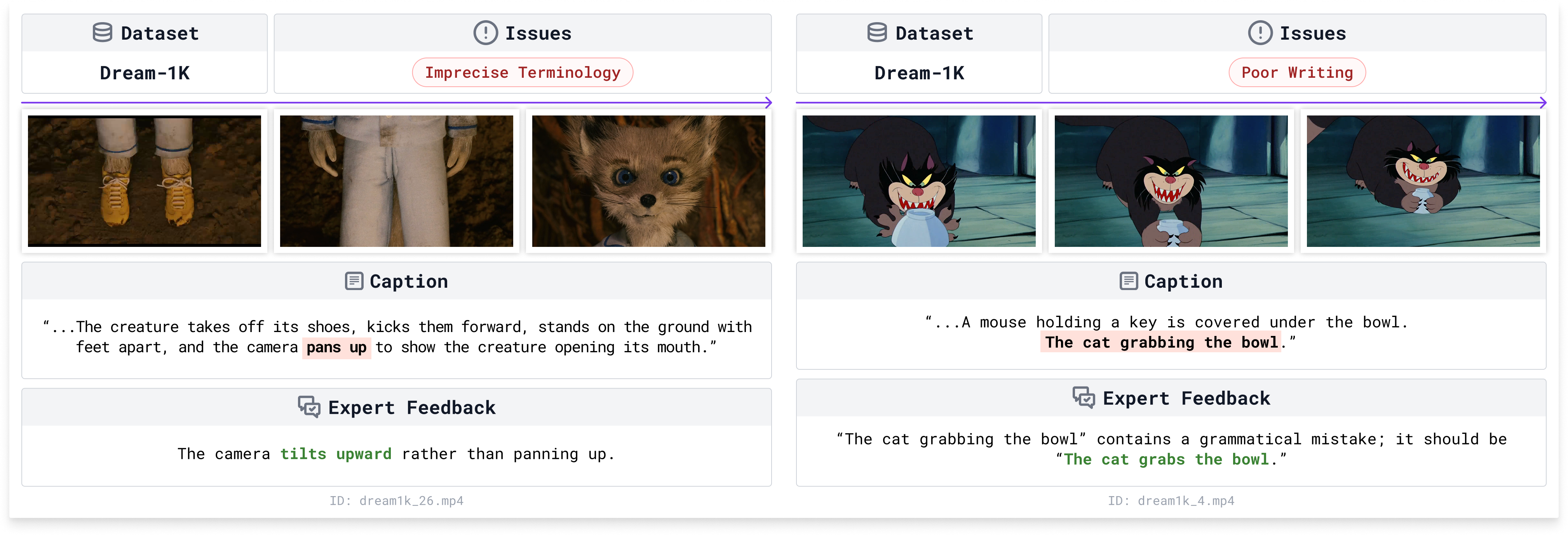}
    \caption{\small {\bf Examples of error cases in Dream1K~\cite{tarsier}.} 
    }\label{fig:errors_dream1k}
\end{figure*}

\begin{figure*}[t!]
\centering
    \includegraphics[width=\textwidth]{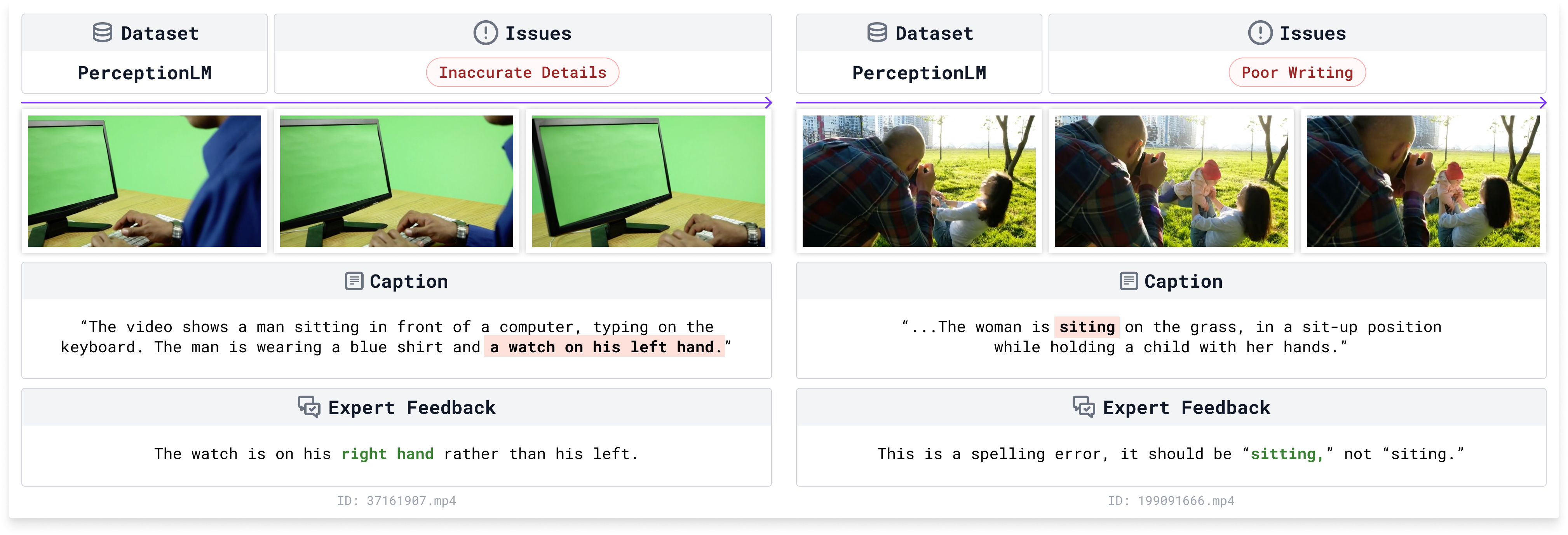}
    \caption{\small {\bf Examples of error cases in PerceptionLM (PE-Video)~\cite{cho2025perceptionlm}.} 
    }\label{fig:errors_perceptionlm}
\end{figure*}

\begin{figure*}[t!]
\centering
    \includegraphics[width=\textwidth]{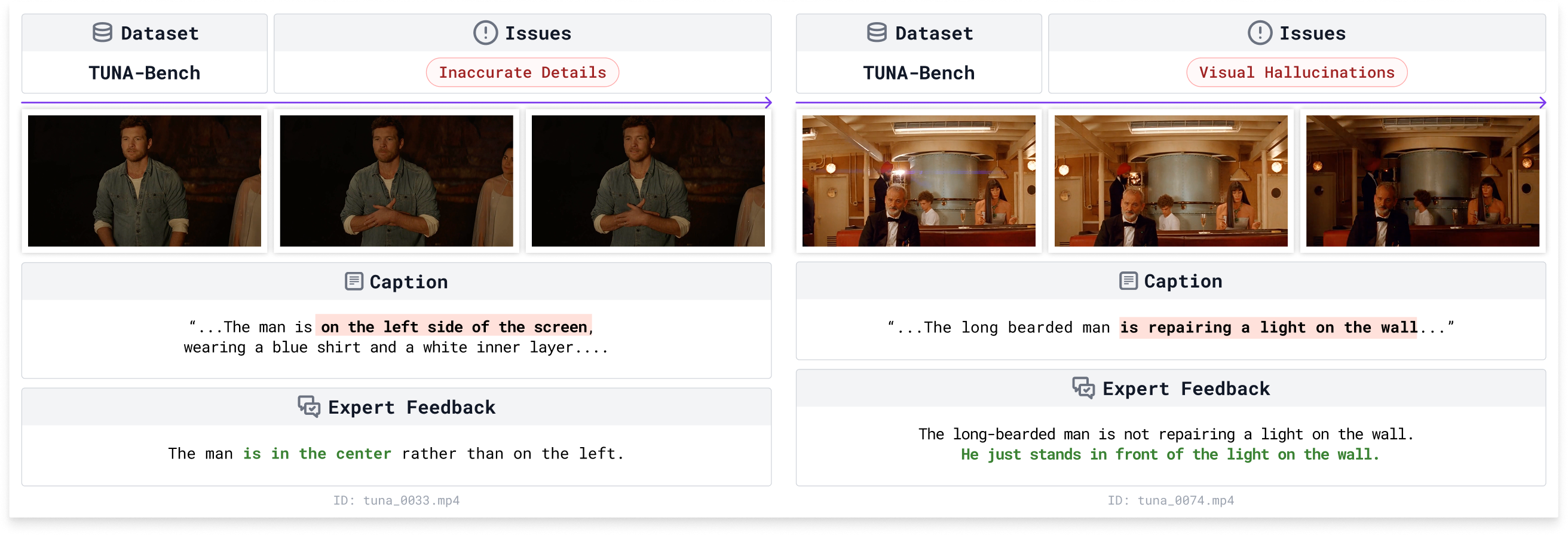}
    \caption{\small {\bf Examples of error cases in TUNA-Bench~\cite{kong2025tuna}.} 
    }\label{fig:errors_tunabench}
\end{figure*}

{\bf Error analysis.}
We present representative error cases from prior datasets in \autoref{fig:errors_specification} (missing specification) and \autoref{fig:errors_oversight} (missing oversight).
Additional dataset-specific examples are shown in \autoref{fig:errors_msrvtt} (MSR-VTT), \autoref{fig:errors_activitynet} (ActivityNet), \autoref{fig:errors_sharegpt4video} (ShareGPT4Video), \autoref{fig:errors_ultravideo} (UltraVideo), \autoref{fig:errors_vdc} (VDC), \autoref{fig:errors_dream1k} (Dream1K), \autoref{fig:errors_perceptionlm} (PerceptionLM), and \autoref{fig:errors_tunabench} (TUNA-Bench).

{\bf Detailed reports.}
For each dataset below, we provide an overview, describe its specification and oversight method, and highlight key issues.
Beyond prior datasets, we report statistics for three versions of our own dataset:
(1)~{\it Crowdsourced}: captions from untrained workers given no specification beyond ``describe the subject, background, motion, and camera in detail'';
(2)~{\it Ours (without human)}: pre-captions generated by our label-then-caption approach, where expert-labeled primitives are provided to Gemini-2.5-Pro or Qwen-2.5-VL-72B to produce more accurate captions;
and (3)~{\it Ours}: the final human-revised version after second-stage quality control, used for training and evaluation.

{\bf Human rating rubrics.}
We rate all datasets with the same Likert-scale rubric across our five caption aspects.
We also report an Overall rating summarizing coverage and accuracy across all aspects, except for MSR-VTT and ActivityNet Captions, which cover only human activities.
Note that scores across datasets are not directly comparable, as the video distributions, caption detail levels, and covered aspects differ:
\begin{enumerate}
    \item[5.] Completely correct; no changes needed.
    \item[4.] Excellent, with only minor adjustments required (e.g., a few inaccurate, missing, or hallucinated words). Roughly one sentence may need addition, modification, or deletion.
    \item[3.] Mostly correct but with notable omissions, hallucinations, or errors (e.g., more than two sentences require rewriting, deletion, or addition).
    \item[2.] Mostly incorrect, requiring substantial revisions (e.g., more than half of the caption is incorrect and must be redone).
    \item[1.] Almost entirely wrong, requiring a complete rewrite (e.g., the entire caption is irrelevant or fails to mention the specified aspect at all).
\end{enumerate}

{\bf Summary of human evaluations.}
\autoref{tab:dataset_comparison_policy} summarizes our human evaluation results.
Among prior work, TUNA-Bench scores highest, likely due to its more comprehensive internal specification and quality control, though its guidelines are not public and avoidable issues such as typos and grammatical errors remain.
On our videos, captions from crowdworkers are generally unreliable: they are much shorter than those produced under our structured specification and oversight framework (${\sim}$100 words vs.\ ${\sim}$400 words), and their errors are often severe, as crowdworkers typically lack the visual vocabulary needed to describe common cinematic or motion effects.
Illustrative examples are shown in \autoref{fig:crowdsourced}.

\begin{itemize}
  \item {\bf MSR-VTT}~\cite{xu2016msr} (2016)
    \begin{itemize}
      \item {\bf Overview:} As one of the earliest video–text datasets, MSR-VTT includes short clips (10–30 seconds each) annotated with a single short sentence.
      \item {\bf Specification:} No public annotation policy was released. Captions mainly describe subjects and their immediate actions, without mentioning scene context, camera motion, or spatial layout. 
      \item {\bf Oversight:} Captions were collected from thousands of crowdsourced AMT workers, with each clip annotated by several independent annotators. The only quality control mentioned is removing duplicate or very short sentences during post-processing. 
      \item {\bf Issues identified:} Crowdsourced captions suffer from severe quality issues. Many describe only part of the clip or focus on one action while ignoring others. They also often contain grammatical mistakes. Because the captions are short (typically under 10 words), we do not observe major issues with subjective descriptions or hallucinated content.
    \end{itemize}
    \item {\bf ActivityNet Captions}~\cite{activitynet} (2017)
    \begin{itemize}
      \item {\bf Overview:} ActivityNet Captions include temporally grounded human-annotated descriptions for $\sim$20k videos (around 100k sentences  in total).
      \item {\bf Specification:} Annotators first wrote a paragraph enumerating all major human actions; then each sentence was tied to a start and end time. Each sentence describes a single event.
      \item {\bf Oversight:} Annotators were trained through examples of good and bad annotations shown in the interface. The interface enforced minimum length ($\geq$3 sentences, $\geq$5 words each) and required all sentences to be timestamped before submission.
      \item {\bf Issues identified:} Like MSR-VTT, captions mainly focus on human actions. Despite better quality control, quality still varies. Many describe only part of the clip; overlapping events are merged into one sentence. Camera motion and visual style are never mentioned. Due to crowdsourced annotation, Grammar errors (e.g., missing ``to'') remain frequent. Because annotators attempt to include more details than in MSR-VTT, hallucinations-especially with subjects and spatial relations—become more common.
    \end{itemize}
    \item {\bf ShareGPT4Video}~\cite{chen2025sharegpt4video} (2024)
    \begin{itemize}
        \item {\bf Overview:} ShareGPT4Video annotates 40k videos automatically using GPT-4V without any human-in-the-loop, by sending static frames sampled every 2 seconds to the model.
        \item {\bf Specification:} The authors provide only a brief instruction to GPT-4V, e.g., ``conveying changes in actions, behaviors, environments, states and attributes of objects, and camera movements between adjacent frames.''
        \item {\bf Oversight:} No quality control was applied to the generated captions.
        \item {\bf Issues identified:} We manually evaluate only the summary captions from all individual frames. GPT-4V is good at describing general scene context but still hallucinates often, especially for spatial and camera details (e.g., confusing left vs. right). It also makes basic visual mistakes, such as treating overlay graphics as real objects, or saying an object or motion exists when it does not. There are no writing errors since all captions are model-generated, but the model frequently adds subjective phrases like ``reflective ambiance,'' ``warm atmosphere,'' or ``themes of speed and connectivity'' that humans may not agree with. Their videos are also relatively static, which partly explains the higher score compared to newer datasets. Because descriptions are based on static frames, camera motion is often wrong, e.g., calling a moving shot static, or the opposite. Although a short specification is provided, it is inconsistently followed (e.g., some captions mention shot size, others do not). We recommend caution when using this dataset for training video-language models.
    \end{itemize}
    \item {\bf UltraVideo}~\cite{xue2025ultravideo} (2025)
    \begin{itemize}
        \item {\bf Overview:} UltraVideo includes 59k  high-resolution (4K to 8K) videos and provide Qwen2.5-VL generated structured captions.
        \item {\bf Specification:} The authors instruct Qwen2.5-VL to generate nine structured captions—Brief Description, Detailed Description, Background, Theme Description, Style, Shot Type, Camera Movement, Lighting, and Video Atmosphere—plus one summarized caption by Qwen3 (average 824 words total). From manual inspection, we find heavy redundancy among the structured captions, so we focus only on the summarized caption for human evaluation.
        \item {\bf Oversight:} No quality control was applied to the generated captions.
        \item {\bf Issues identified:}We manually evaluate only the summarized captions. While they roughly follow our specification and include many relevant aspects, each caption focuses on different elements, making them inconsistent. Similar to GPT-4V in ShareGPT4Video, Qwen2.5-VL performs well at describing general scene context but still hallucinates frequently, especially for spatial and camera details (e.g., confusing left vs. right or misnaming camera motion). It rarely mentions camera height, angle, lens distortion, or spatial depth. Hallucinations are also common when describing subjects or actions that do not exist in the video, and key visual subjects are often omitted. There are no grammatical errors since captions are model-generated, but they frequently include subjective phrases like ``meditative viewing experience,'' ``engaging and informative tone,'' or ``inspire awe and wonder'' which humans may not agree with. 
    \end{itemize}
    \item {\bf VDC}~\cite{chai2024auroracap} (2024)
    \begin{itemize}
        \item {\bf Overview:} VDC (Video Detailed Captioning) includes 1k videos with long, structured captions developed as part of the AuroraCap benchmark.
  \item {\bf Specification:} We cannot find any human annotation policy released by the authors. The structured detailed captions includes camera, short, background, main object, and detailed captions. Each part only have a brief task description, e.g., ``Camera caption. Describe the camera work in detail, including shot types, angles, movements, transitions, and any special effects used to enhance the video.''
  \item {\bf Oversight:} Although the paper mentions ``manual quality inspection is employed to ensure the quality of the video captions,'' the captions appear to be automatically generated. They are too long, uniformly fluent, and free of writing errors, suggesting no human revision was applied.
  \item {\bf Issues identified:} The captions are very long, so we only focus on the detailed caption for evaluation. We refer readers to ~\cite{camerabench} for issues in their camera captions. We find frequent hallucinations and subjective language. Examples include describing a white object as ``colorful,'' calling a gliding shot ``panning,'' claiming the lighting changes when it does not, or saying the camera angle shifts when it stays fixed. Captions sometimes mention shot size but not consistently, and descriptions of camera motion are often wrong. While there are no grammar mistakes, many captions use subjective or emotional phrases such as ``harmonious escape from the hustle and bustle of everyday life,'' ``celebrates the dynamic and ever-changing beauty of nature,'' and ``beauty of concentration and the art of multitasking in a modern office setting.'' These patterns strongly suggest the captions are model-generated without human quality control. We do not recommend using this benchmark to evaluate video detailed captioning due to these major issues.
    \end{itemize}
 \item {\bf DREAM-1K}~\cite{tarsier} (2024)
   \begin{itemize}
    \item {\bf Overview:} Dream1K is a human-annotated test set released with the Tarsier paper, containing 1k challenging movie clips with multiple shots, subjects, and events.
    \item {\bf Specification:} The paper provides limited details about the human captioning policy. The authors claim that the human annotation focuses on describing complex videos at different levels of granularity and camera motions such as zooming, translating, panning, and rotating. No detailed public guideline or policy is released beyond these descriptions.
    \item {\bf Oversight:} The authors claim the dataset was ``carefully described in detail by human annotators,'' but no explicit quality control procedure is described. Captions show clear signs of human writing, including grammar issues and uneven phrasing, but also reveal inconsistent precision across samples.
    \item {\bf Issues identified:} Although claimed to be carefully annotated, we find occasional hallucinations, especially in fine-grained details such as spatial relations (e.g., confusing reference frames like  saying ``to her right''' instead of ``to the right of the frame``) and ambiguous object references (referring ``the object'' instead of ``the person''). Camera motion is frequently omitted even when clearly visible, and subject appearance and scene context are rarely mentioned or described comprehensively, e.g., saying multicolored hair as ``green''. We also observe frequent grammar mistakes that result in incomplete sentences, e.g., ``The cat grabbing the bowl.''
\end{itemize}
\item {\bf PerceptionLM (PE-Video)}~\cite{cho2025perceptionlm} (2025)
\begin{itemize}
    \item {\bf Overview:} PerceptionLM builds on the PE-Video dataset, containing about 1M videos, of which around 120k have human-refined captions. According to the paper, each caption is first generated by a model and then refined by human annotators.
    \item {\bf Specification:} The authors instruct human annotators to improve the synthetic captions by ``{\it removing any hallucinations, correcting words that describe the video inaccurately, eliminating repetitive or redundant words to make the caption more concise, and adding any missing actions being performed in the video.}'' In practice, the captions mostly focus on activities and actions but sometimes include subject appearance, scene context, or camera motion, leading to inconsistent coverage across samples.
    \item {\bf Oversight:} Similar to our dataset, they first use a video captioner to produce pre-captions and then hire 200 annotators to refine them. However, there is no explicit mention of any quality-check or review procedure. We find frequent hallucinations and writing issues, suggesting the absence of systematic quality control.
    \item {\bf Issues identified:} Similar to our pipeline, they first use models to draft a pre-caption for each video and then hire annotators to correct the mistakes. However, we suspect that no model was used to polish the final human-edited captions, as many contain writing errors such as ``The text say'' instead of ``The text says'' or ``siting'' instead of ``sitting.'' Descriptions of camera motion are often inaccurate, for example confusing camera rotation (``pan across'') with camera translation. Spatial details are rarely mentioned, and when they appear, they are often wrong; for instance, confusing a person's left hand as the right. Fine-grained visual details are also frequently misinterpreted, such as describing a green paper on a screen as a device with a green screen. When multiple subjects appear, captions use vague pronouns like ``it'' when there are multiple potential subjects, making references unclear. Overall, although human refinement was employed to improve accuracy, captions remain inconsistent and prone to fine-detail visual and writing errors.
    \end{itemize}
\item {\bf TUNA-Bench~\cite{kong2025tuna} (2025)}
\begin{itemize}
    \item {\bf Overview:} TUNA-Bench consists of 1k videos with detailed captions annotated by trained human experts.
    \item {\bf Specification:} The authors claim that ``{\it we prepare a detailed note document for instructing human annotators on annotation.}'' However, the actual instruction materials are not released publicly and only briefly mentioned in the paper as instructing annotators to ``{\it Strictly follow chronological ordering of events}'' and to provide ``{\it Objective descriptions without summarization and subjective feelings.}'' They also emphasize distinguishing similar objects ``{\it by unique attributes (e.g., age, dress, etc.).}'' The element rules require descriptions of ``{\it Camera states, including panning, rotating, zooming, following, shaking, transition, etc.},'' along with scene context, actions, and visual attributes such as ``{\it characters’ gender, age, and dress, objects’ color, shape, and number.}'' 
    \item {\bf Oversight:} The paper describes the annotator screening and training procedure only at a high level without releasing implementation details. The authors state that ``{\it all annotators have TEM-4 or TEM-8 English proficiency, and have experience in video captioning annotation (e.g., several annotators have previously annotated video-caption pairs for Kling project). Prior to formal annotation, they undergo our specialized training to guarantee the quality of their annotation.}'' They further claim to ``{\it ensure annotation quality and consistency, we implemented a rigorous annotator selection and training process. Initially, all potential annotators underwent a trial annotation phase using a shared subset of videos... Through careful evaluation of their trial annotations, we selected only those annotators who demonstrated high consistency, accuracy, and thorough understanding of the annotation guidelines.}'' Their captions occasionally shows writing problems such as typos and grammar mistakes, likely because they do not use LLMs to polish the human caption. Overall, the procedure appears rigorous, although no concrete examples or released materials are provided.  
    \item {\bf Issues identified:} Among all benchmarks we manually evaluated, TUNA-Bench provides the highest overall human caption quality and is recommended for evaluating subject motion description. Specification-wise, the authors claim to have a detailed annotation guideline, but it is not publicly released. The captions are strong in describing subject appearance, scene context, and general motion or activity, yet show common human issues in spatial and camera details. Many captions misinterpret camera movement (e.g., saying the camera is moving when it is actually rotating, or mentioning rotation without specifying whether it arcs around a subject or rotates along its own axis). Annotators often confuse ``shot transitions (hard cuts)'' with ``zooming'' and rarely describe spatial depth, sometimes labeling a ``midground car'' as ``background.'' Overlay elements such as on-screen subtitles are also frequently omitted. Minor hallucinations occur (e.g., ``a man on the left'' when centered, ``patterned'' when the object has plain texture, or ``one person'' when there are two), but these are relatively infrequent given the level of detail. Overall, this dataset highlights the value of clear specification and quality control.
\end{itemize}

\end{itemize}

\section{Specification Details}
\label{sec:specification_details}

In this section, we describe how we develop our specification in collaboration with video content creators.

\begin{figure*}[t!]
\centering
\setlength{\tabcolsep}{4pt} %
\begin{tabular}{cc}
    \includegraphics[width=0.45\textwidth]{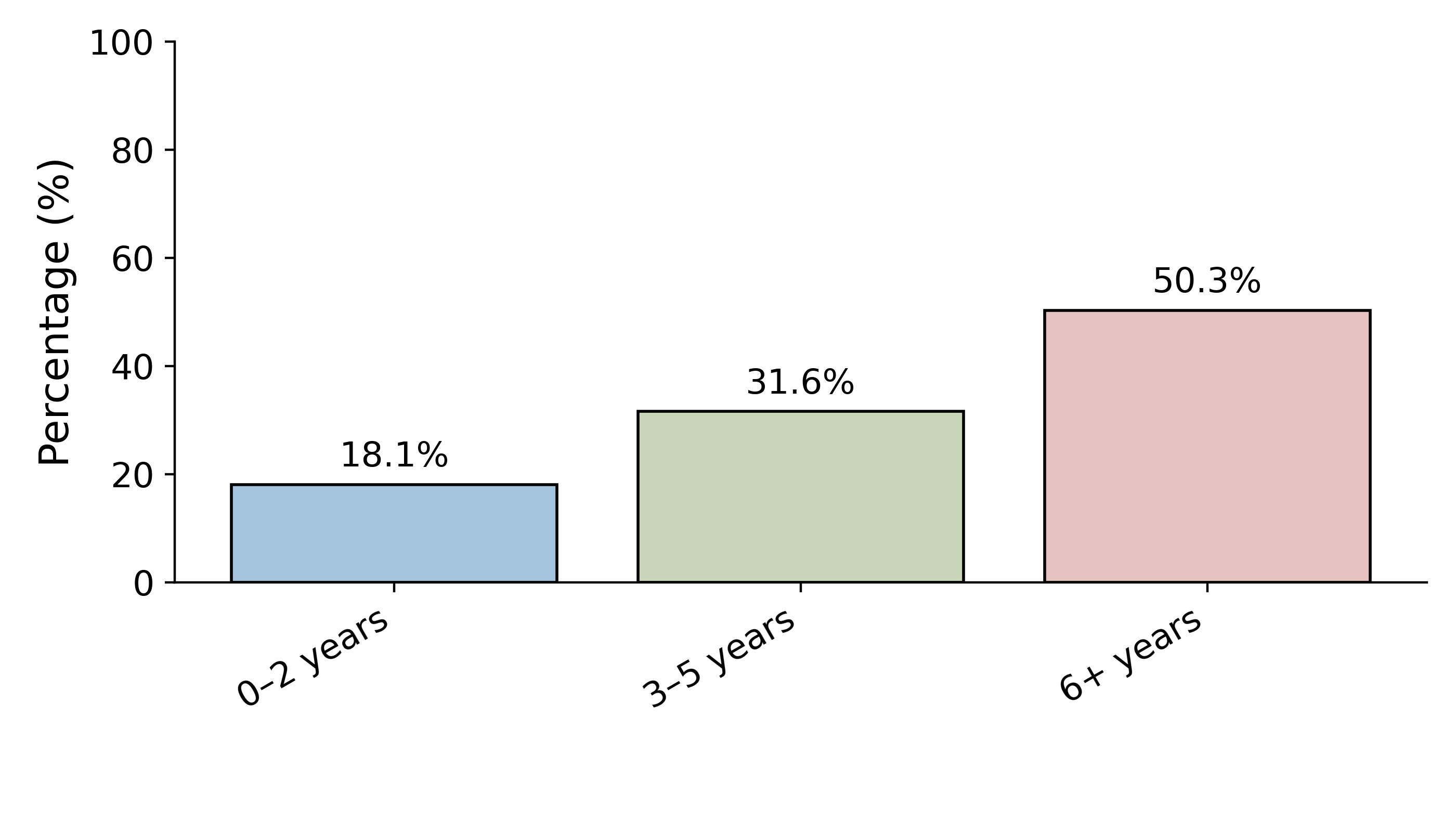} &
    \includegraphics[width=0.45\textwidth]{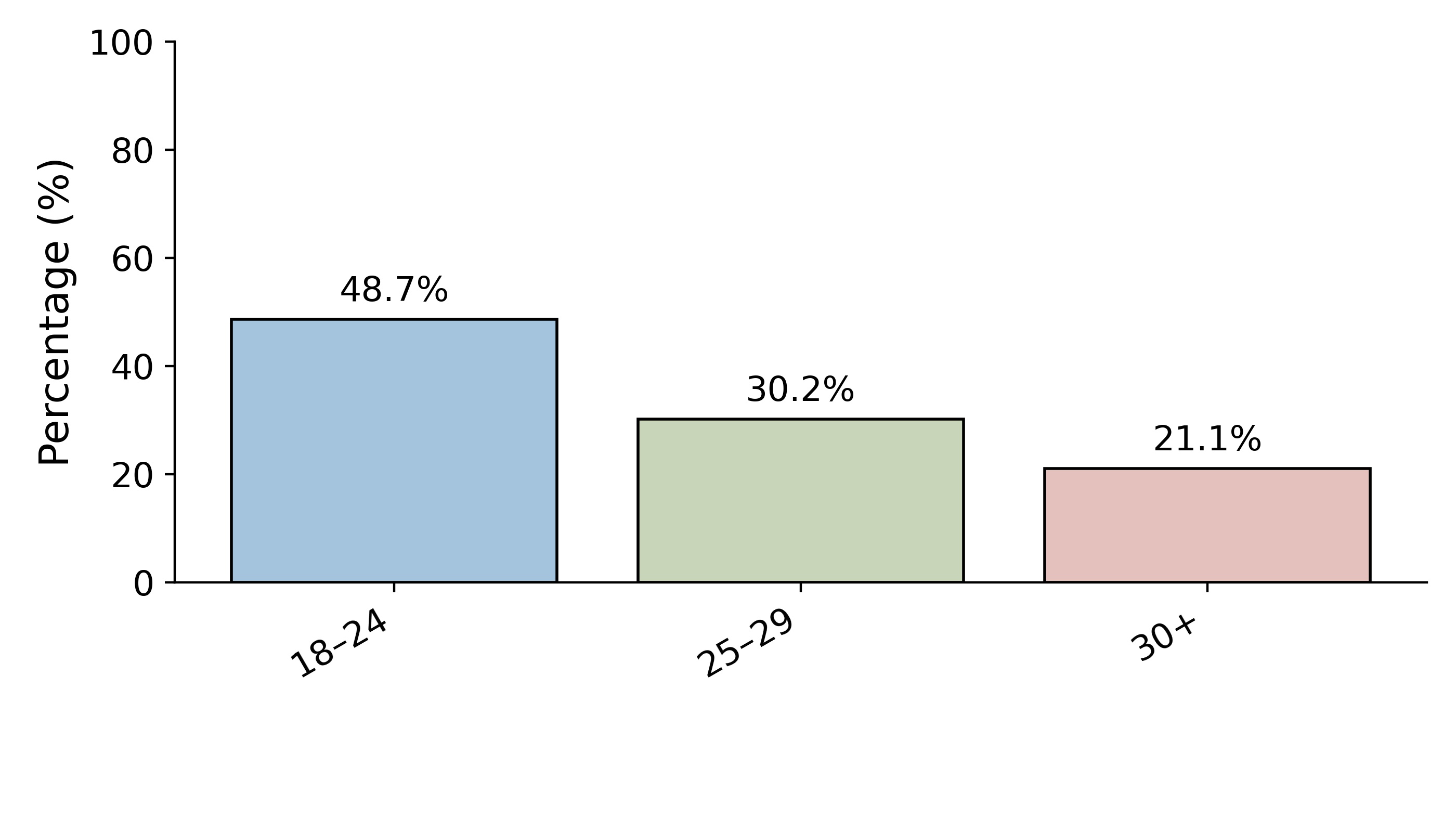} \\
    \small (a) Professional experience &
    \small (b) Age distribution \\[4pt]
    \includegraphics[width=0.45\textwidth]{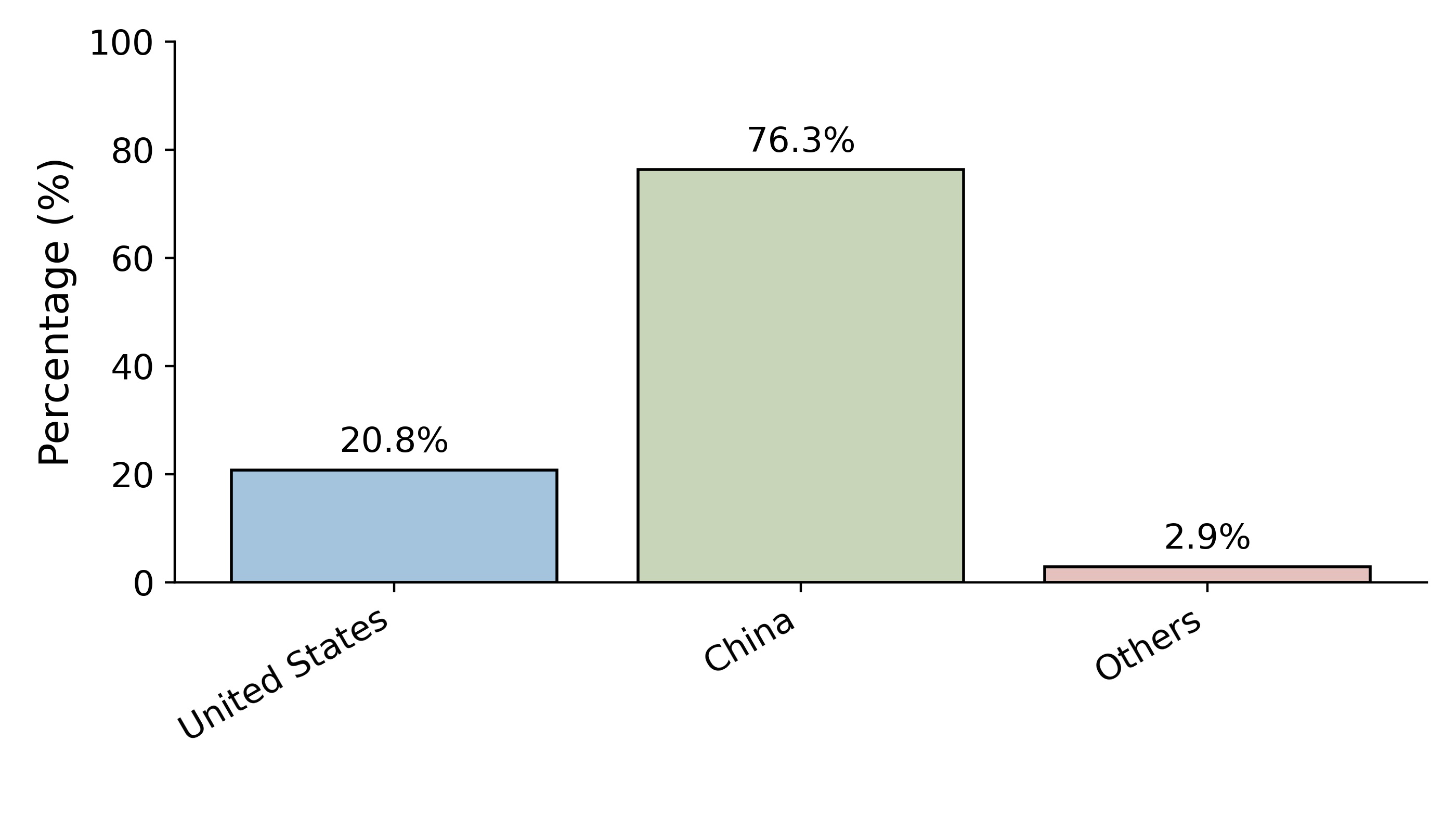} &
    \includegraphics[width=0.45\textwidth]{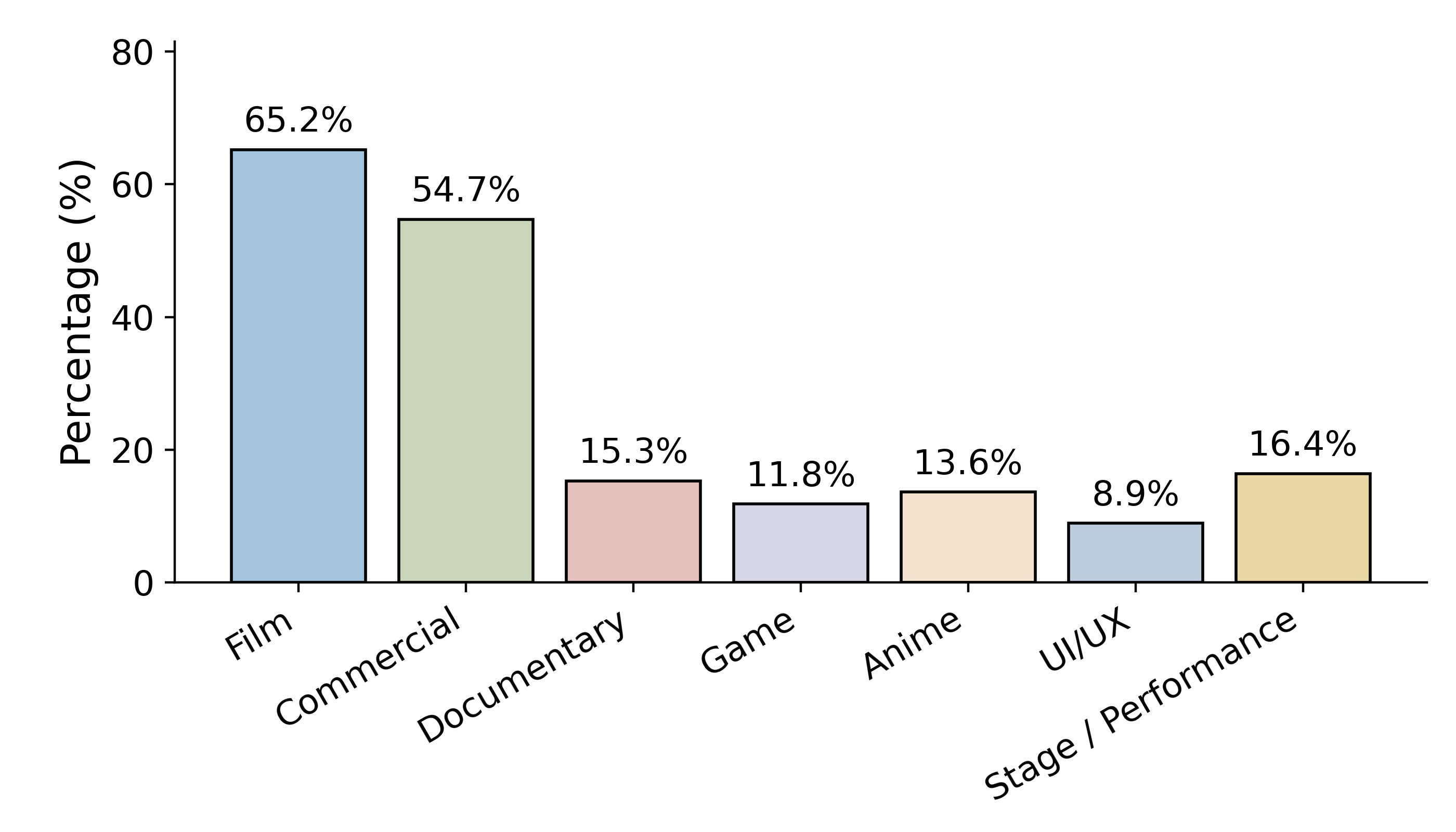} \\
    \small (c) Geographic region &
    \small (d) Professional domains \\
\end{tabular}
\caption{\small {\bf Demographics of video content creators.}}
\label{fig:demographics}
\end{figure*}

{\bf Demographics of content creators.}
Our specification draws on direct input from professional video creators experienced in camera movement and shot composition.
A core team of 30 creators (filmmakers, motion designers, cinematographers, and game artists) meets weekly with paper authors to refine the taxonomy, resolve ambiguous cases, and verify primitive definitions across hundreds of video samples.
We additionally collaborate with over 100 creators selected from a pool of more than 600 applicants, who provide feedback, raise clarification questions during training, and help refine definitions and decision rules.
Annotators are also drawn from this trained pool, ensuring strong visual literacy and familiarity with the specification.
\autoref{fig:demographics} summarizes the distribution of creator experience, age, geographic region, and professional domain.

\begin{figure*}[t!]
\centering
\includegraphics[width=\textwidth]{images/overall.png}
    \caption{\small {\bf Overview of our specification.} As the arrows indicate, motion and spatial captions depend on the completed subject and scene captions; \autoref{sec:pseudocode_precaption} provides the full prompts detailing this dependency.
    }\label{fig:all_aspects}
\end{figure*}

{\bf Specification overview.} \autoref{fig:all_aspects} provides a high-level overview of our specification, organized into five major aspects and their corresponding subaspects.

{\bf From primitives to captioning.} We convert labeled $\sim$200 {\bf primitives} (from CameraBench-Pro~\cite{rao2026moodio}) into {\bf pre-captions} by prompting the model with dynamically generated instructions that depend on each video’s labels. Unlike Socratic models~\cite{zeng2022socratic}, which use a fixed template, our prompts change according to the specific primitives. For example, if a video contains multiple people with one clear main subject, the prompt asks the model to describe that subject’s appearance, clothing, and relationships to others. If the video is a scenery shot, the prompt instead tells the model to describe the environment and note the absence of a main subject. \autoref{sec:pseudocode_precaption} includes Python-style pseudocode showing how raw primitives are converted into {\it subject}, {\it scene}, {\it motion}, {\it spatial}, and {\it camera} pre-captions.

\section{Oversight Details}
\label{sec:oversight_details}

In this section, we detail our oversight framework covering annotator screening, training, captioning, and human–AI team data curation and incentive structures.

{\bf Who is in the loop.} Our oversight framework involves three roles drawn from a unified pool of professional video creators. First, a {\bf core team of 30} creators (filmmakers, motion designers, cinematographers, game artists) meets weekly with the paper authors to develop the specification, resolve ambiguous cases, and author gold-standard reference captions. Second, out of more than {\bf 600 applicants}, roughly {\bf 100 pass our multi-stage screening} and form the part-time annotator pool; of these, $\sim$50 label primitives and $\sim$20 are promoted to write captions. Third, the top-performing $\sim$8 captioners are further promoted to {\bf reviewer} roles, who perform stage-2 quality checks. All roles are compensated, and detailed pipeline and promotion criteria follow below.

{\bf Annotator screening and training.}
We recruit annotators primarily from our pool of professional video content creators. Similar to CameraBench~\cite{camerabench}, we find that professionals learn our training guideline quickly and make fewer mistakes once they begin real tasks. We also include a small number of highly motivated non-professionals who spend over three months training with our guideline and platform (including paper authors who originally came from computer vision backgrounds but became fluent in creative visual vocabulary through this process), confirming that the guideline is sufficient to train non-professionals to a professional level with enough effort. Screening begins with studying the training materials and completing primitive labeling for two projects: camera motion and video cinematography (including camera setup). Each project requires passing two to six multiple-choice exams. We only accept annotators who score above 90\% across all primitives. In practice, most professionals reach 80–90\% accuracy within two or three exams; annotators who cannot reach 90\% after three attempts in any section are removed. Out of 600 applicants, roughly 50 pass all exams, and about 30 exceed 90\%, forming our main annotator pool. These 30 begin with primitive labeling; only after completing more than 300 videos with a strong track record are they promoted to captioning, yielding about 20 captioners. Among them, only around 8 are eventually promoted to reviewer roles for final-stage quality checks.

\begin{figure*}[t!]
\centering
\setlength{\tabcolsep}{4pt}
\begin{tabular}{cc}
    \includegraphics[width=0.45\textwidth]{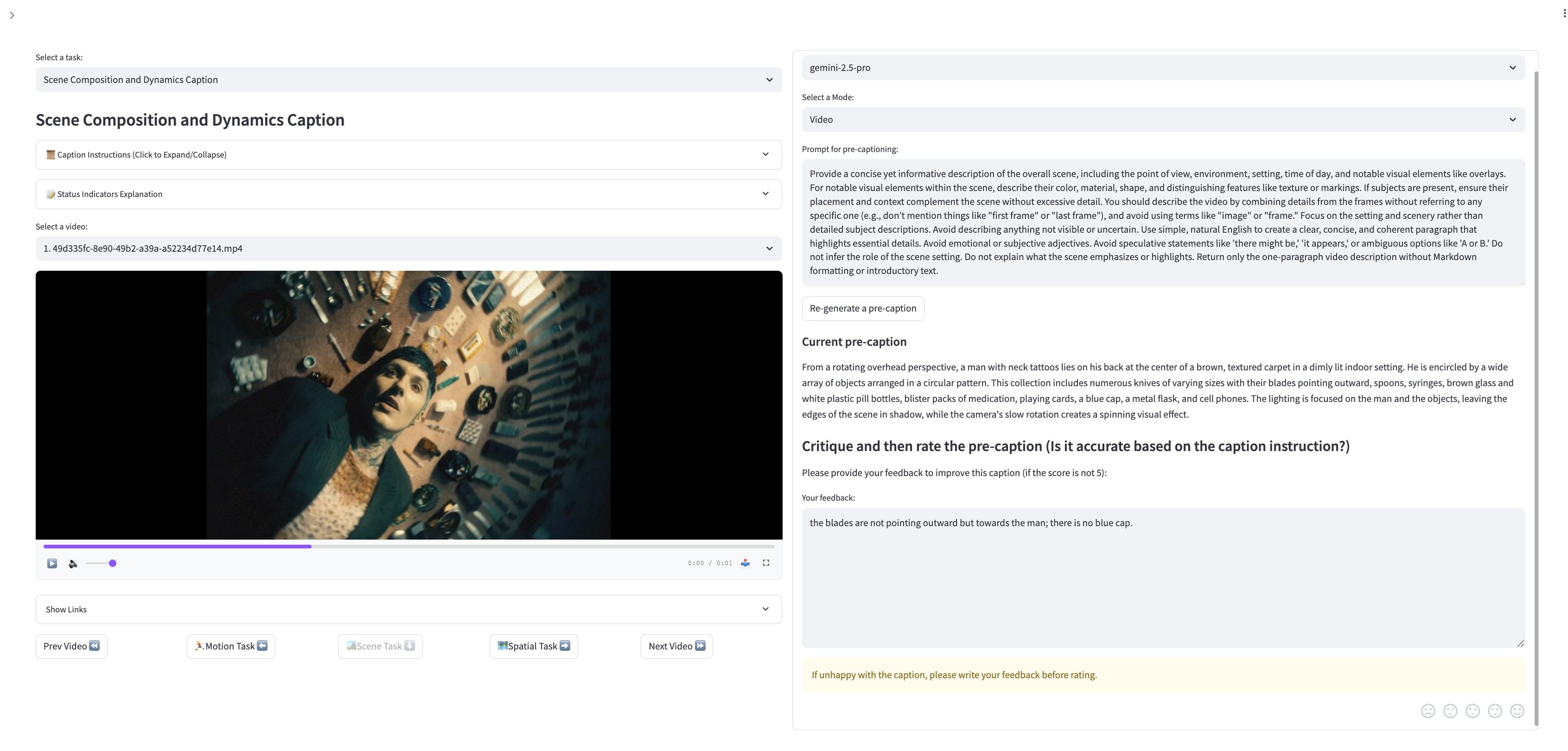} &
    \includegraphics[width=0.45\textwidth]{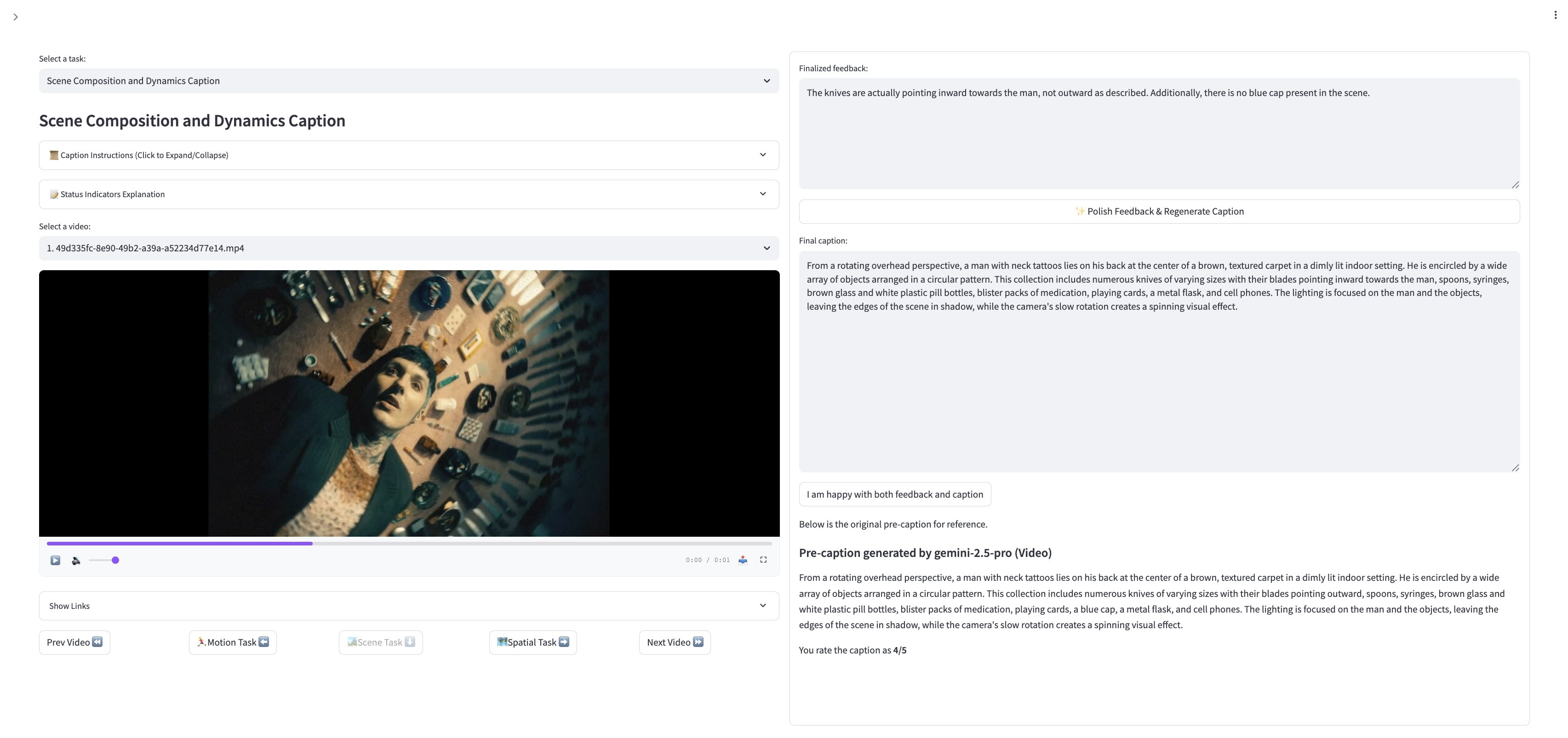} \\
    \small (a) Critiquing model pre-caption &
    \small (b) Finalizing critique and post-caption \\[4pt]

    \includegraphics[width=0.45\textwidth]{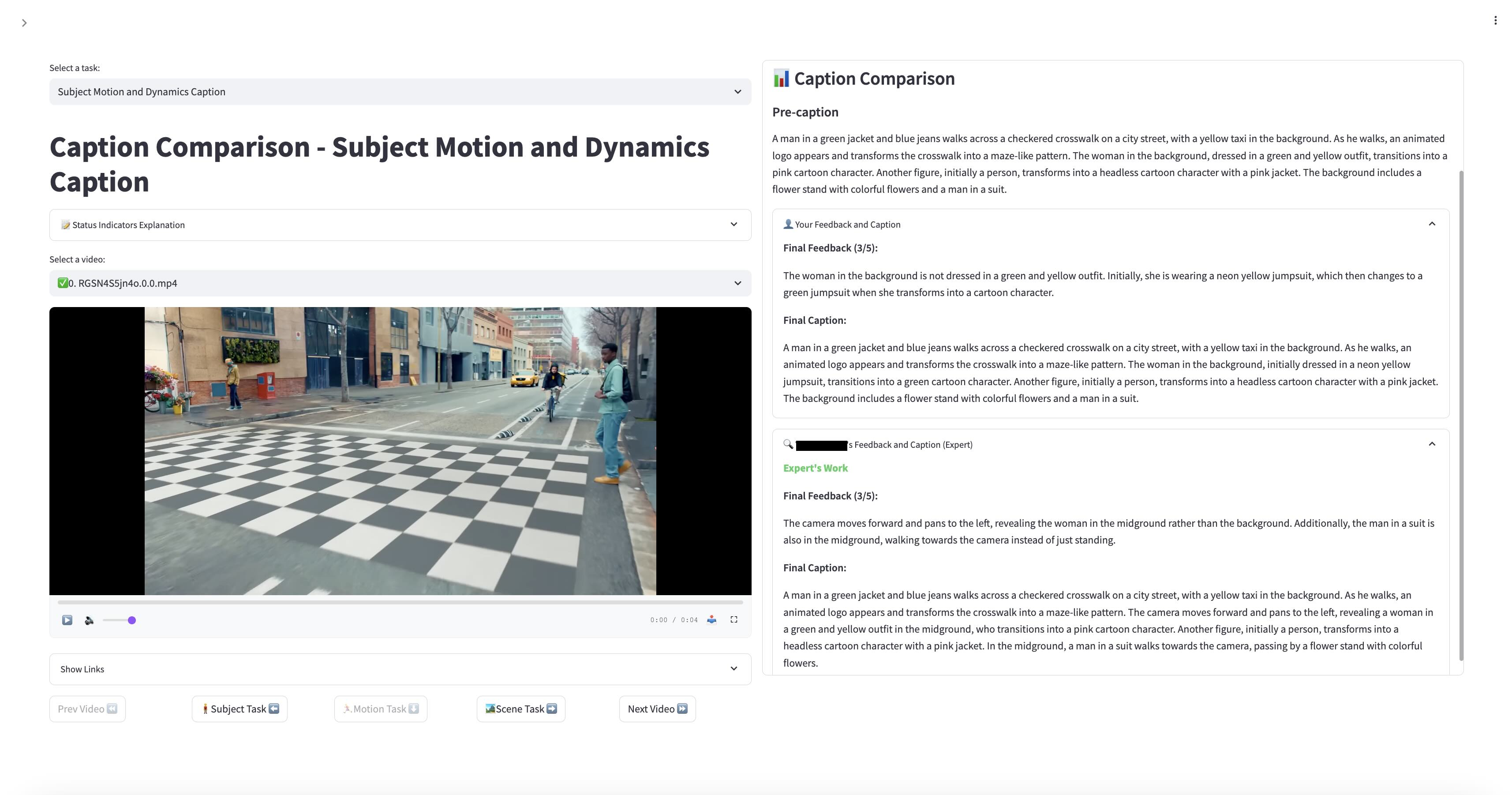} &
    \includegraphics[width=0.45\textwidth]{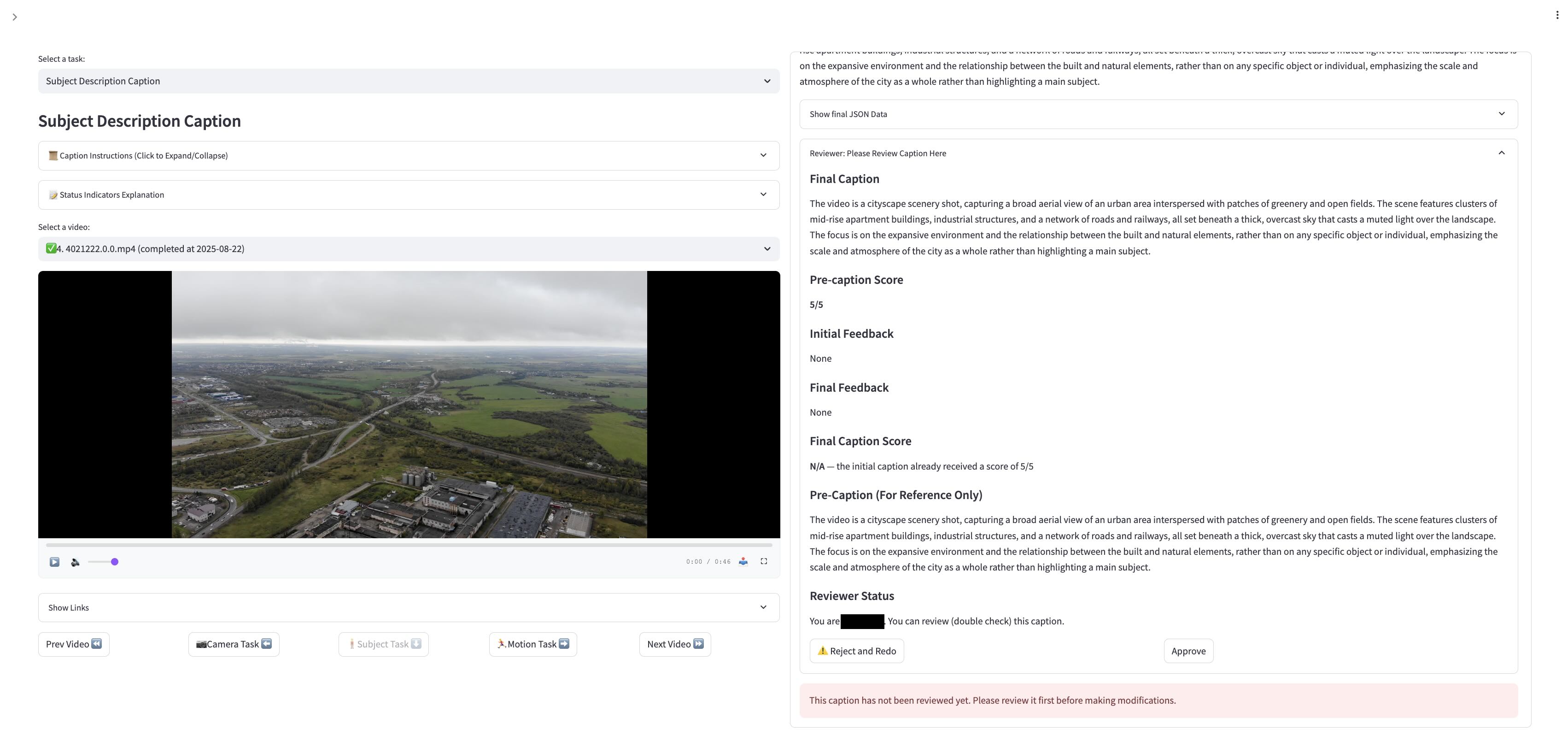} \\
    \small (c) Incoming annotator shadowing expert critique &
    \small (d) Reviewer reviewing annotator's work \\[4pt]

    \includegraphics[width=0.45\textwidth]{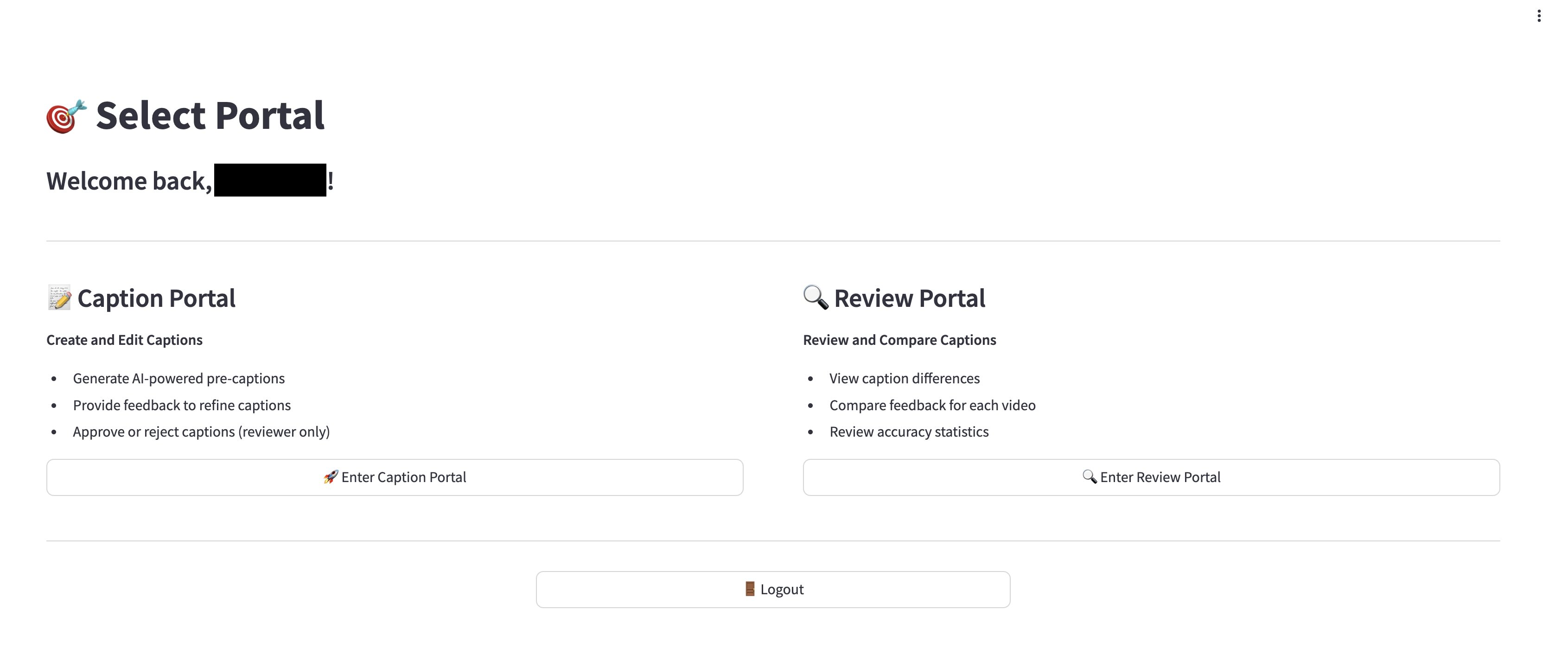} &
    \includegraphics[width=0.45\textwidth]{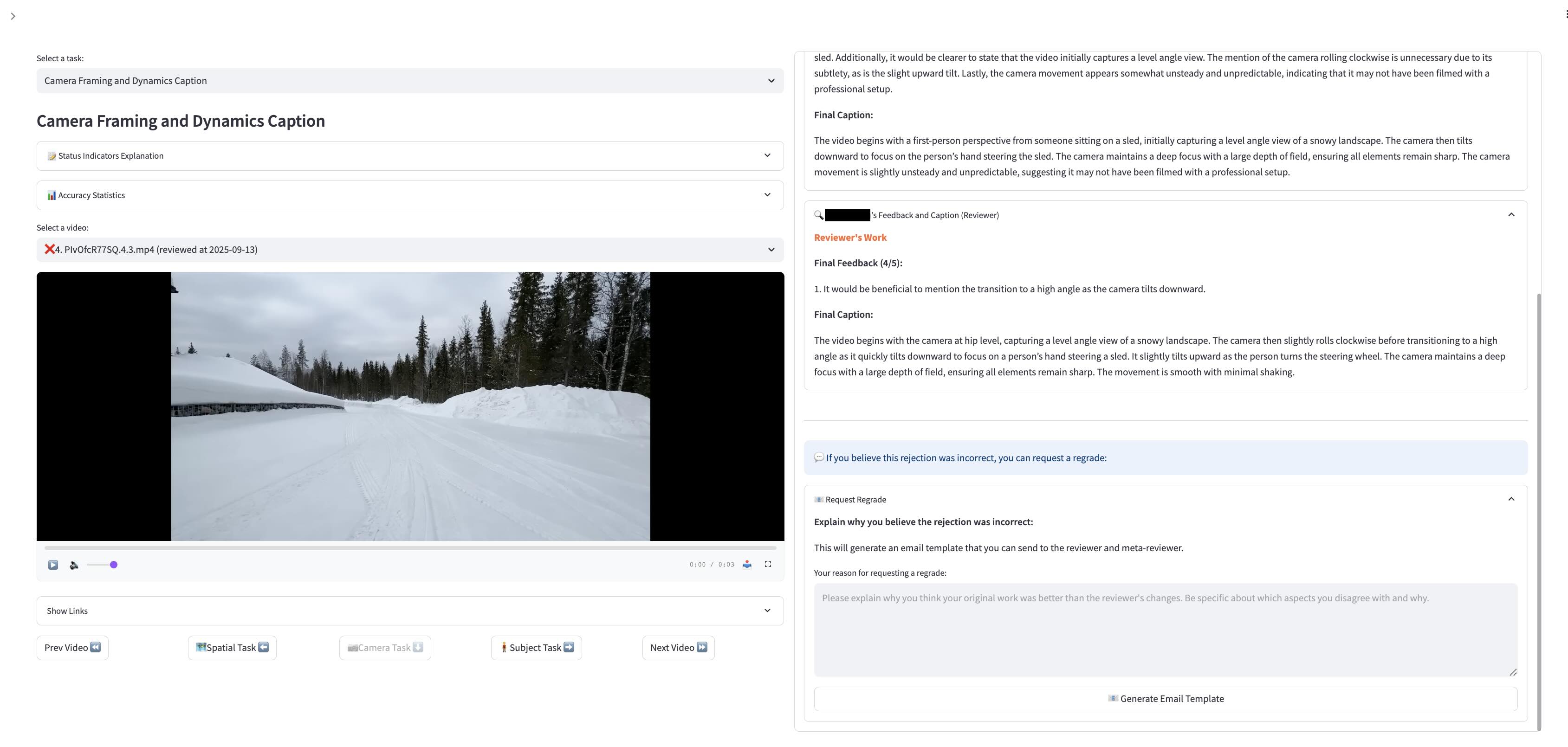} \\
    \small (e) Choosing caption or review portal &
    \small (f) Review portal allows annotator appealing rejection \\
\end{tabular}

\caption{\small {\bf Example screenshots from our captioning platform.}
(a) shows how annotators (or reviewers) use our platform to generate and critique model pre-captions, and
(b) shows how they continue polishing the critique and post-caption until satisfied.
(c) shows how new annotators can perform training by first completing the above steps, then comparing their critique with expert versions verified by multiple paper authors.
(d) reviewers can review annotator work by examining their critique and post-caption, approving good work or rejecting and redoing incorrect submissions.
(e) annotators can choose the caption or review portal, and
(f) shows how they can appeal via regrade request by sending email to the reviewer.}
\label{fig:caption_platform}
\end{figure*}

\begin{figure*}[t!]
\centering
\includegraphics[width=1.0\textwidth]{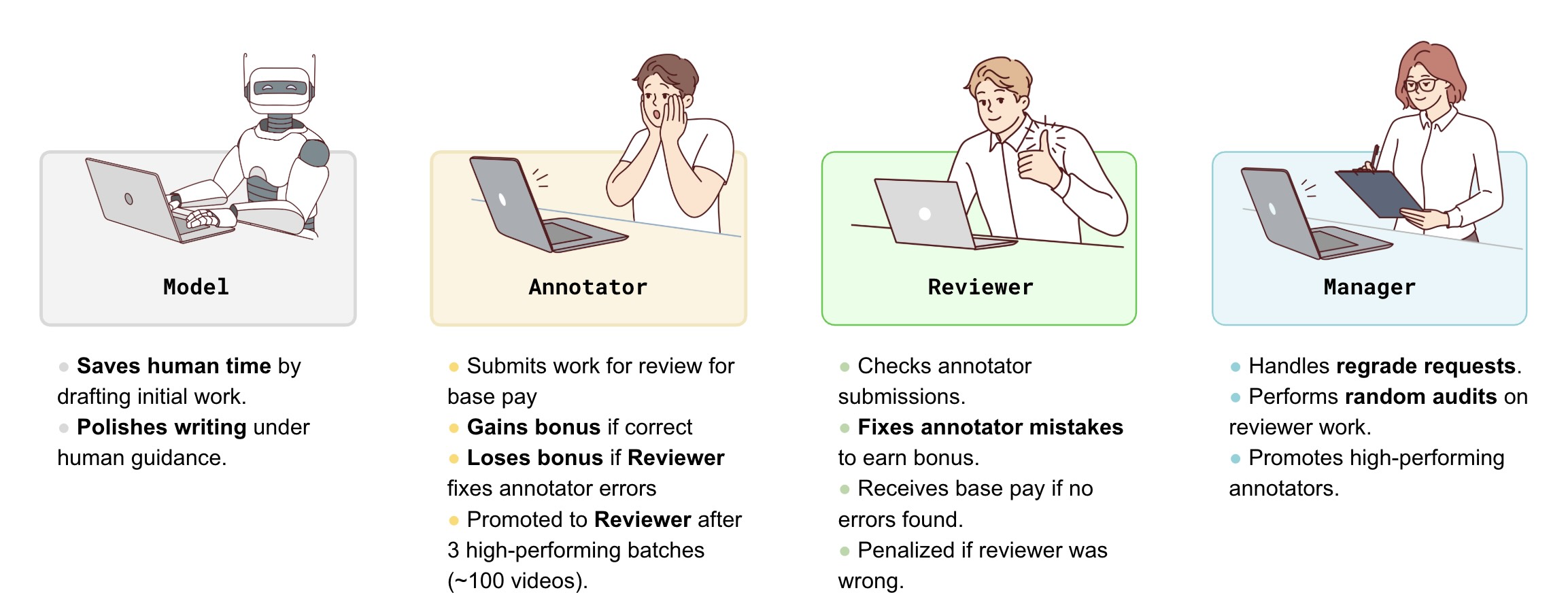}
\caption{\small {\bf Human-AI team structure.} }
\label{fig:human_ai_team}
\end{figure*}

{\bf Platform for human-AI caption curation.}
\autoref{fig:caption_platform} shows screenshots of our in-house captioning platform, which supports efficient human–AI collaboration. The interface includes tools for annotators to refine critiques and captions, allows new annotators to shadow expert work, allows reviewers to approve or reject submitted captions, and provides an appeal workflow where annotators can request a regrade by emailing the reviewer and manager. We show the human-AI team structure in \autoref{fig:human_ai_team}.

{\bf CHAI annotation statistics.} To characterize the cost and behavior of our oversight pipeline in practice, we log per-aspect statistics across all annotation sessions. \autoref{tab:chai_stats} reports the average number of critique-revision iterations per aspect, the Likert score of the initial Gemini-2.5-Pro pre-caption, the word counts of the final post-caption and the critique, and the wall-clock time spent per video. Three observations stand out. First, {\em most videos converge in one iteration}: across all five aspects, 95\% of samples reach an accurate post-caption after a single critique-revision cycle, and only the remaining 5\% require a second reviewer pass. Second, {\em spatial is the hardest aspect}: spatial captions have the lowest pre-caption score (3.3/5) and the highest iteration count (1.4), consistent with the general observation that current VLMs struggle with reference-frame disambiguation (e.g., subject's left vs.\ frame's left). Third, {\em effort shifts from writing to verification, not away}: total time per video under CHAI (15 min) is comparable to crowdsourced write-from-scratch captioning (15 min) in our pilot study, yet CHAI captions are $\sim$3.7$\times$ longer (405 vs.\ 110 words; \autoref{sec:prior_work_errors}) and substantially more accurate. In other words, our oversight framework re-allocates human effort rather than eliminating it, which is precisely what enables the caption-quality improvements reported in the main paper.

\begin{table}[h!]
\centering
\renewcommand{\arraystretch}{1.15}
\caption{\small \textbf{Per-aspect CHAI annotation statistics.} Averages computed across our full training set ($\sim$4k videos $\times$ 5 aspects). Pre-caption scores are on a 1--5 human Likert scale.}
\label{tab:chai_stats}
\scalebox{0.7}{
\begin{tabular}{lccccc}
\toprule[1.2pt]
 & Subject & Scene & Motion & Spatial & Camera \\
\midrule
\#Critique-revision iterations & 1.1 & 1.1 & 1.2 & 1.4 & 1.0 \\
Pre-caption score (out of 5)   & 3.7 & 3.8 & 3.8 & 3.3 & 4.4 \\
Post-caption word count        & 93  & 97  & 54  & 117 & 45  \\
Critique word count            & 48  & 46  & 38  & 63  & 27  \\
Time per video (minutes)       & 3   & 3   & 2   & 5   & 2   \\
\bottomrule[1.2pt]
\end{tabular}
}
\end{table}

{\bf Promotion and salary.}
Our annotation pipeline uses a quality-based compensation system that rewards accuracy and encourages consistent performance. For captioning, annotators earn a base of \$30 per set (10 videos, 50 captions). Accuracy-based deductions of \$5, \$10, or \$15 apply if any task falls to 70\%, 50\%, or 30\% accuracy; a \$5 bonus is awarded if all tasks reach 90\% or above. Annotators with consistently low accuracy may be stopped early and asked to self-correct or complete additional onboarding tasks. Those who achieve 80--90\% accuracy for three consecutive sets and complete at least one set per day may be promoted to caption reviewer. Caption reviewers earn a base of \$15 per set, with dynamic increases to \$20, \$25, \$30, or \$35 depending on annotator accuracy (all tasks $\geq$ 90\%, $\geq$ 70\%, $\leq$ 70\%, $\leq$ 50\%, or $\leq$ 30\% respectively). Manager spot-checks are conducted at random; if major issues are found, the reviewer must redo the batch, and three such incidents result in suspension.

{\bf Cost of building annotator expertise.} The single largest cost in our pipeline is not the annotation itself but the upfront training of annotators. A new annotator typically requires {\bf 1.5--3 months of paid training} on our primitive-labeling platform before reaching $\geq$90\% accuracy on held-out exams. This overhead is the primary bottleneck for scaling oversight beyond academic-budget dataset sizes, and it is the main reason we cap the initial release at $\sim$4k videos. Reducing this overhead—for instance, by using stronger critique models to accelerate the training loop, or by transferring annotator skills across related tasks—is a natural direction for follow-up work.

\section{Examples of Pre-captions, Critiques, and Post-Captions}
\label{sec:more_examples}

\begin{figure*}[t!]
\centering
\setlength{\tabcolsep}{6pt}
\begin{tabular}{cc}
    \includegraphics[width=0.45\textwidth]{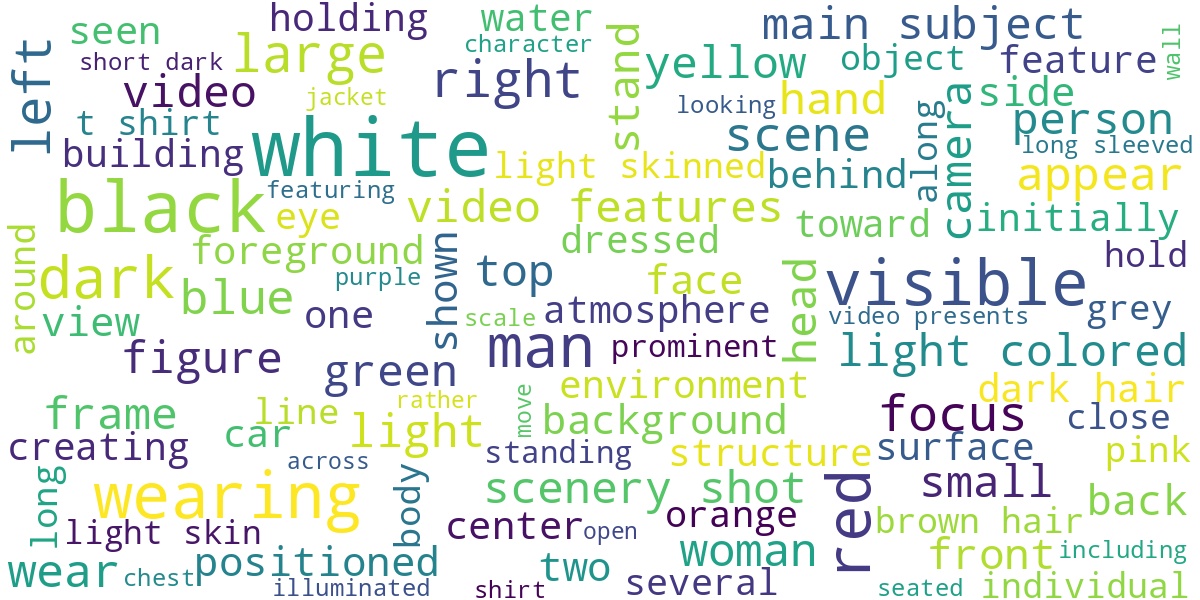} &
    \includegraphics[width=0.45\textwidth]{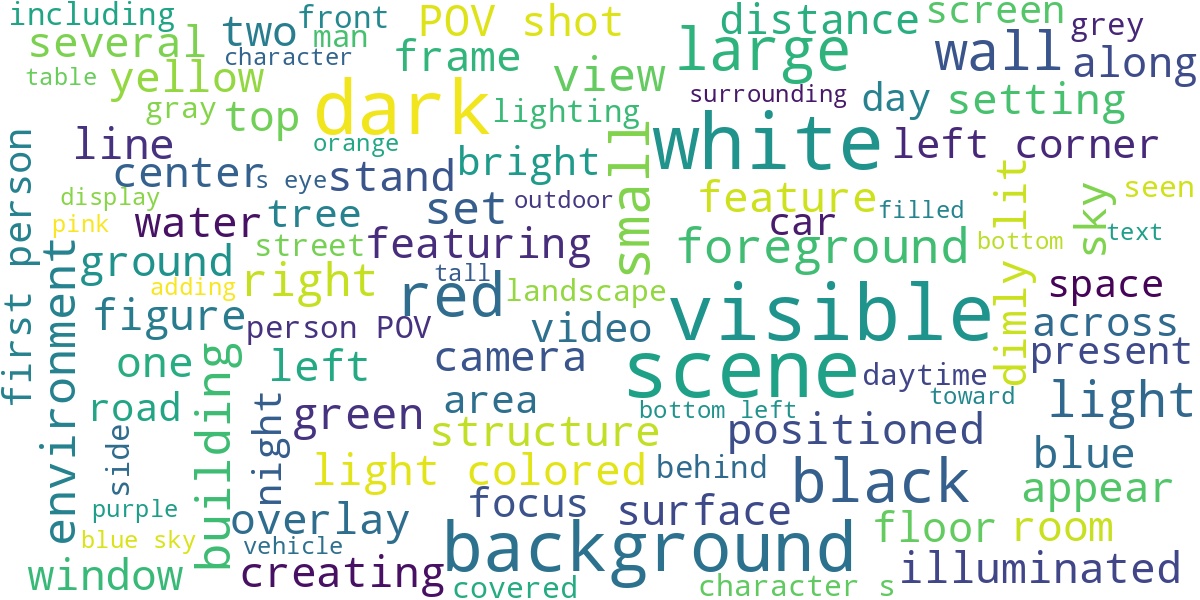} \\
    \small (1) Subject captions &
    \small (2) Scene captions \\[8pt]

    \includegraphics[width=0.45\textwidth]{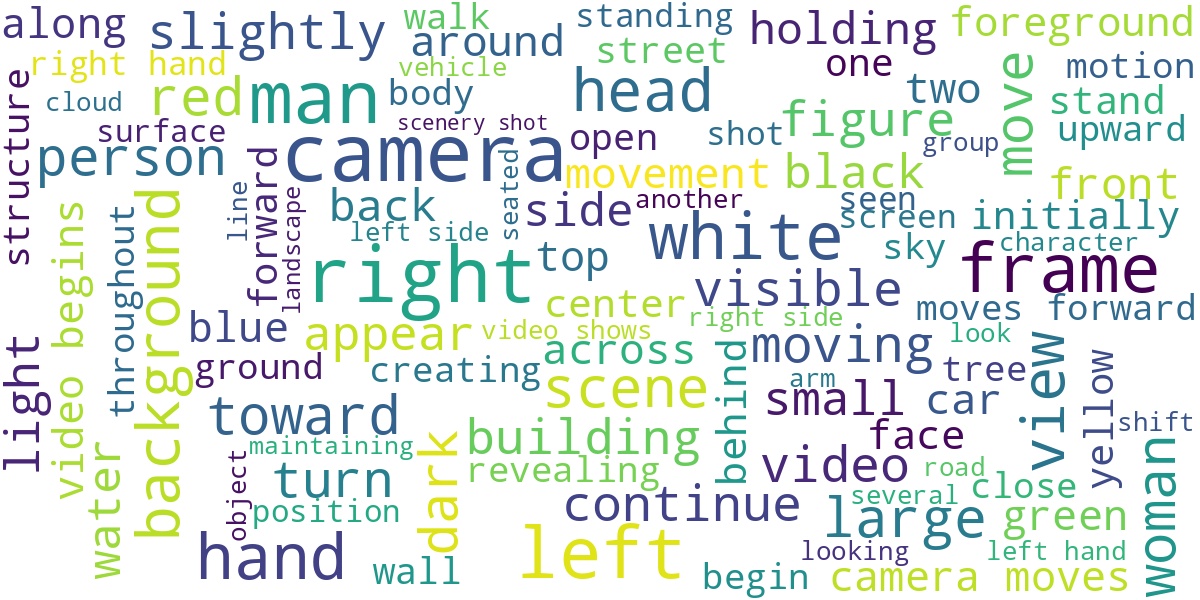} &
    \includegraphics[width=0.45\textwidth]{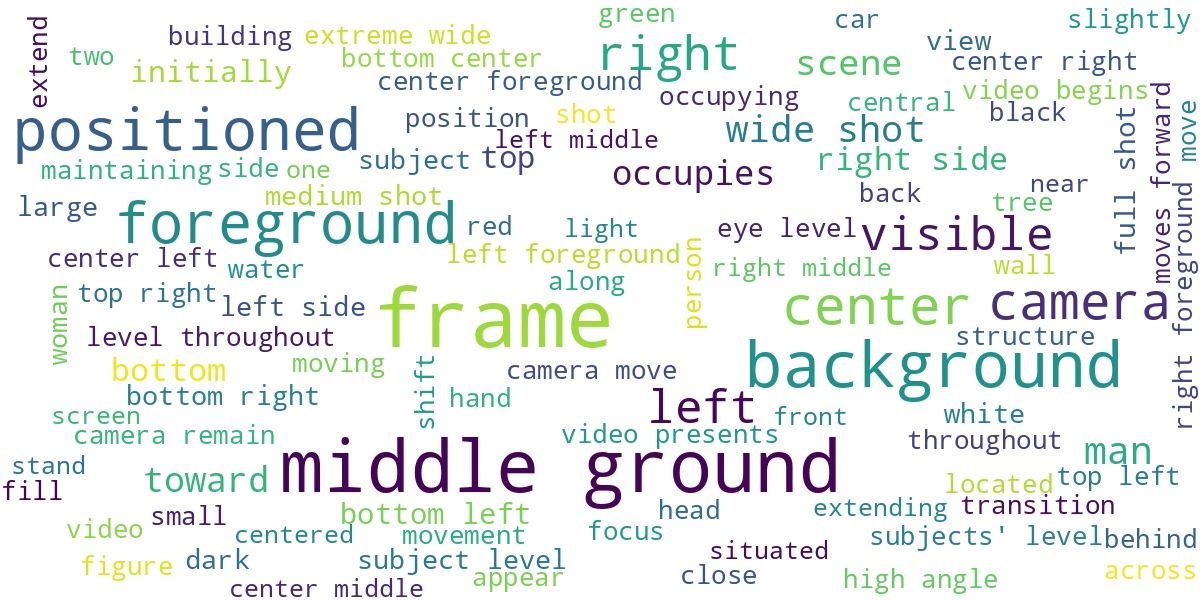} \\
    \small (3) Motion captions &
    \small (4) Spatial captions \\[8pt]

    \multicolumn{2}{c}{
        \includegraphics[width=0.45\textwidth]{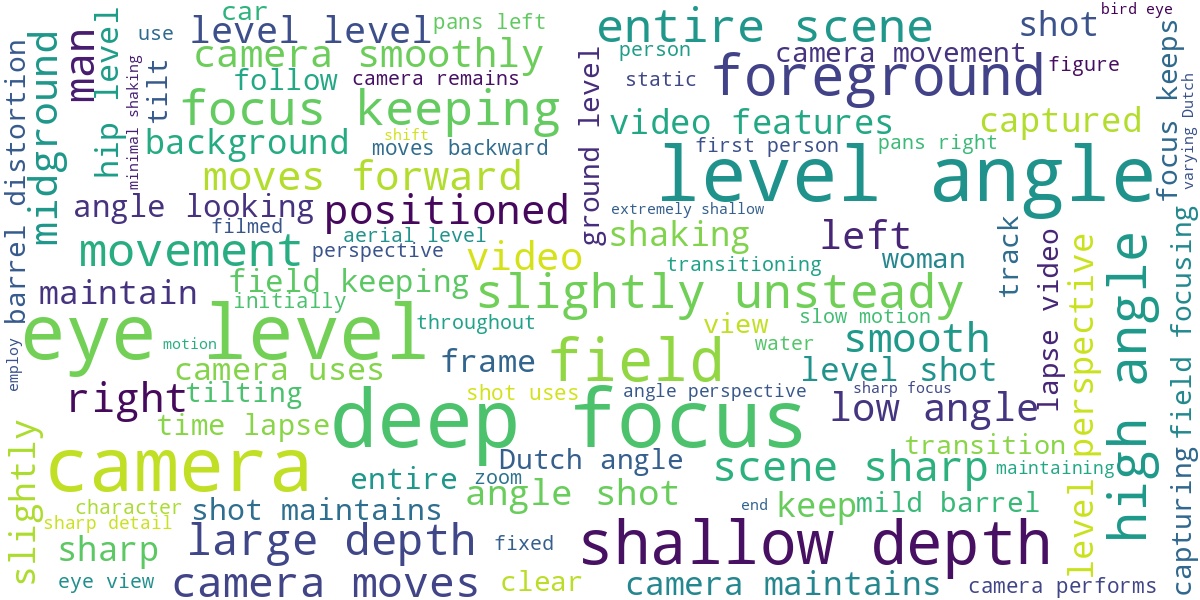}
    } \\
    \multicolumn{2}{c}{\small (5) Camera captions} \\
\end{tabular}

\caption{\small {\bf Word clouds of captions.}}
\label{fig:caption_wordclouds}
\end{figure*}

\begin{figure*}[t!]
\centering
\begin{tabular}{c}
    \includegraphics[width=0.95\textwidth]{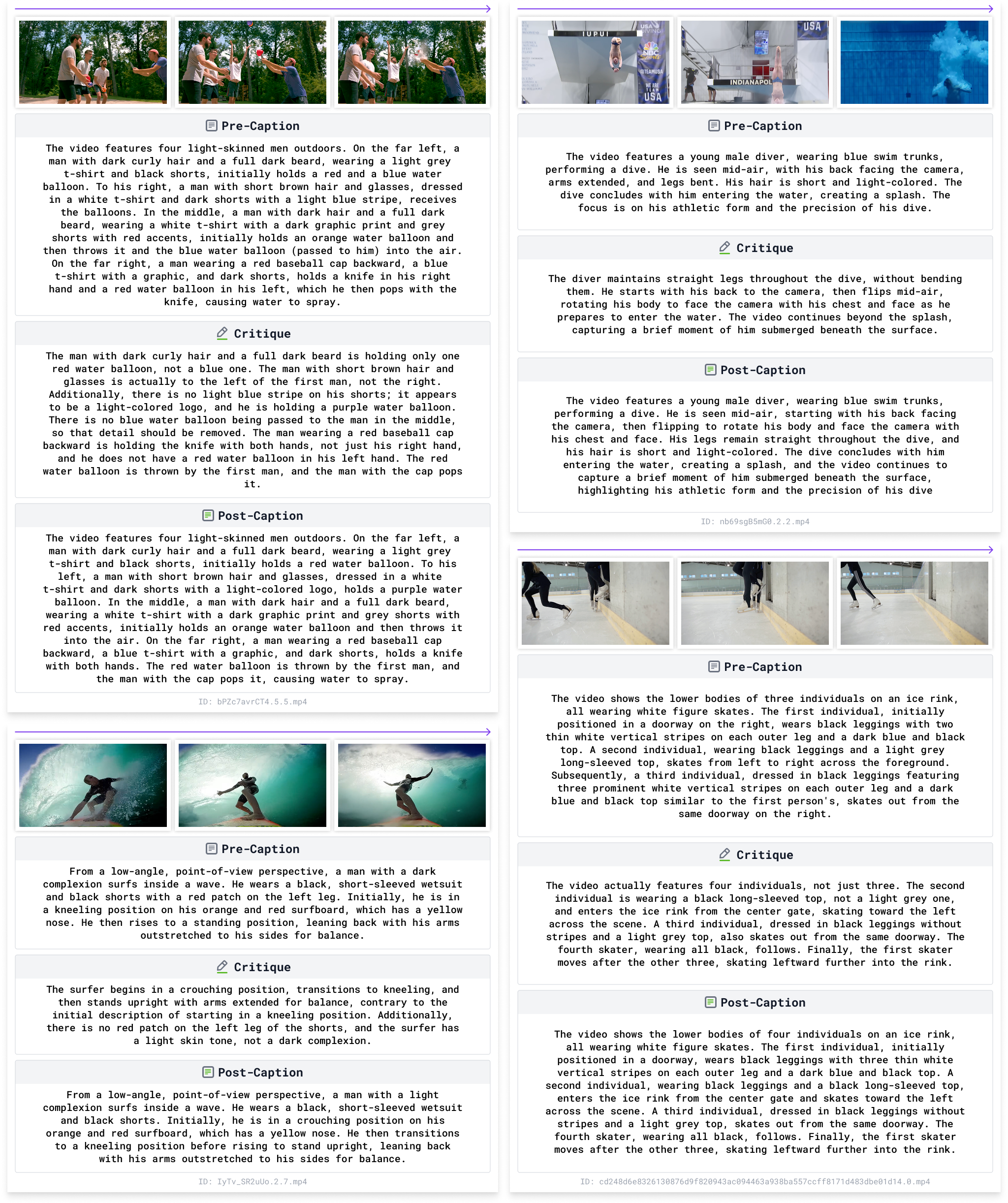} \\
\end{tabular}
\caption{\small {\bf Subject caption examples.} }
\label{fig:subject_captions}
\end{figure*}

\begin{figure*}[t!]
\centering
\begin{tabular}{c}
    \includegraphics[width=0.95\textwidth]{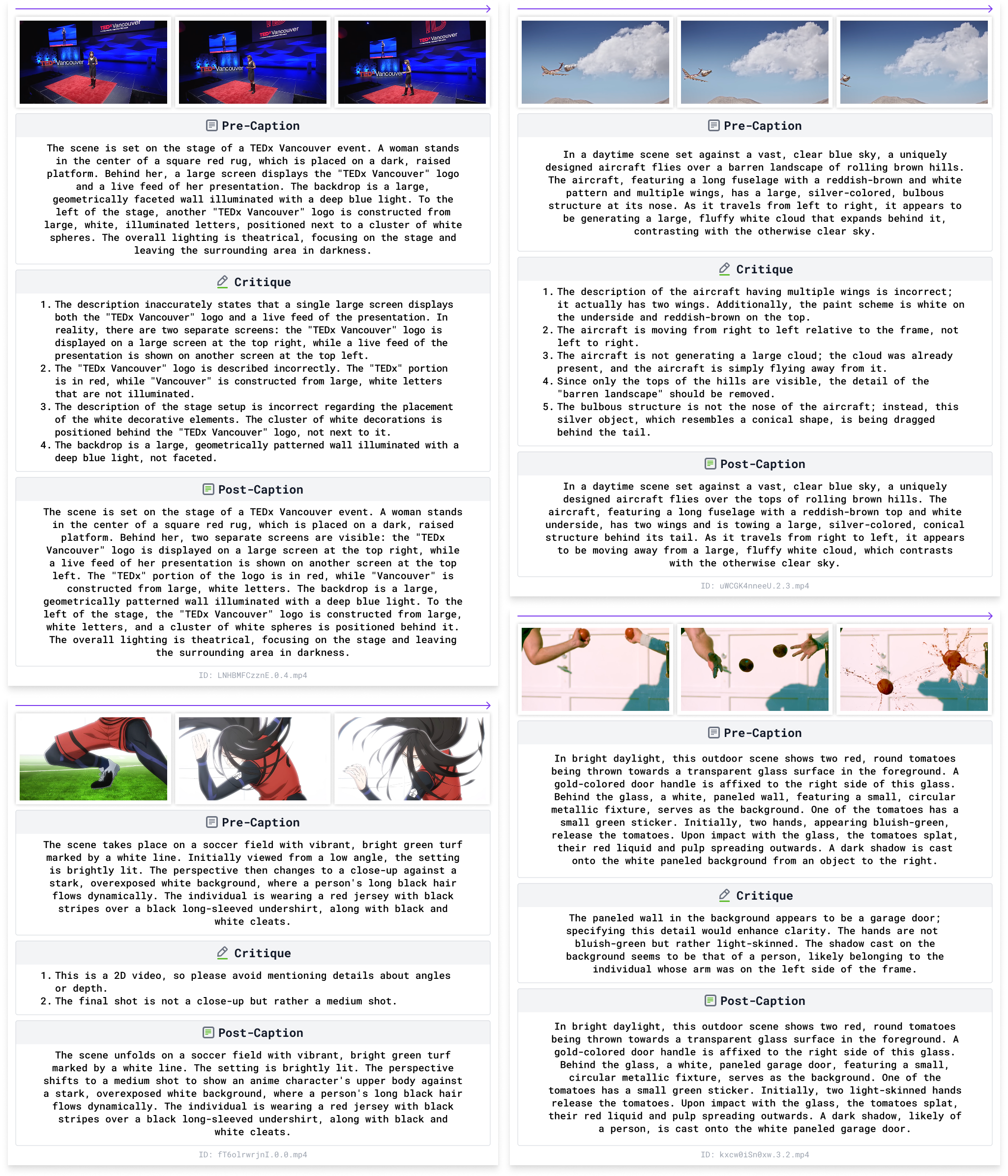} \\
\end{tabular}
\caption{\small {\bf Scene caption examples.} }
\label{fig:scene_captions}
\end{figure*}

\begin{figure*}[t!]
\centering
\begin{tabular}{c}
    \includegraphics[width=0.95\textwidth]{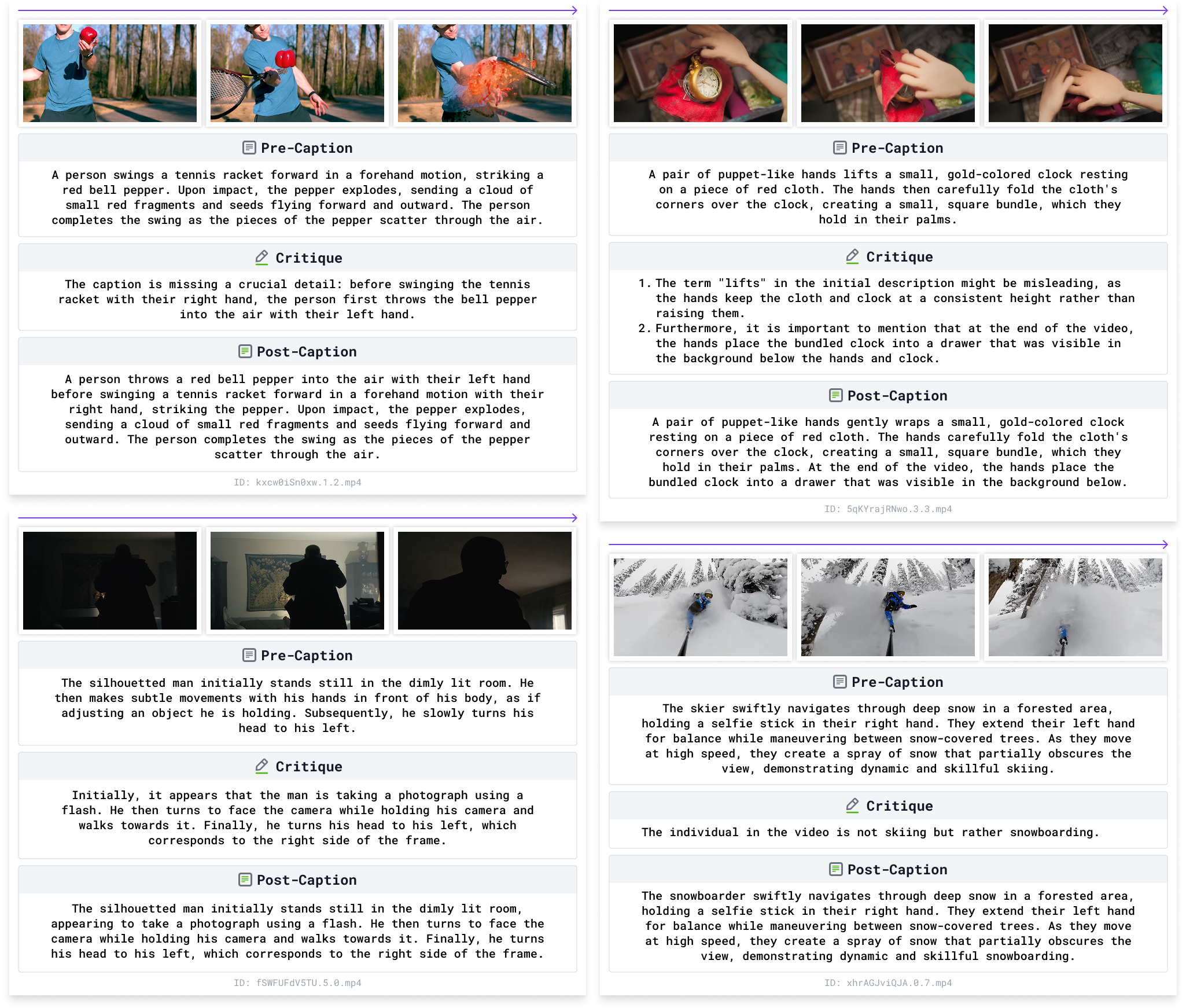} \\
\end{tabular}
\caption{\small {\bf Motion caption examples.} }
\label{fig:motion_captions}
\end{figure*}

\begin{figure*}[t!]
\centering
\begin{tabular}{c}
    \includegraphics[width=0.95\textwidth]{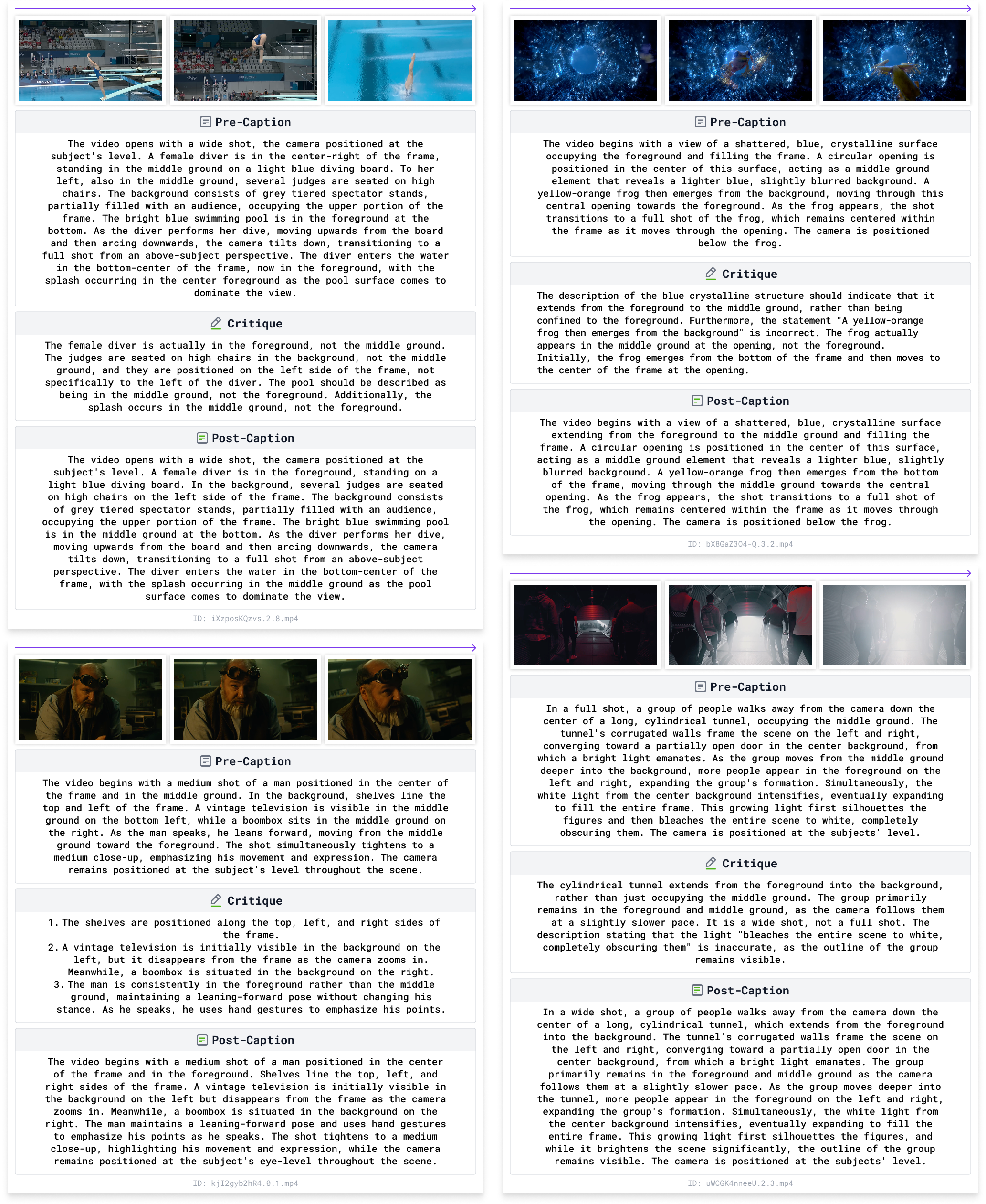} \\
\end{tabular}
\caption{\small {\bf Spatial caption examples.} }
\label{fig:spatial_captions}
\end{figure*}

\begin{figure*}[t!]
\centering
\begin{tabular}{c}
    \includegraphics[width=0.95\textwidth]{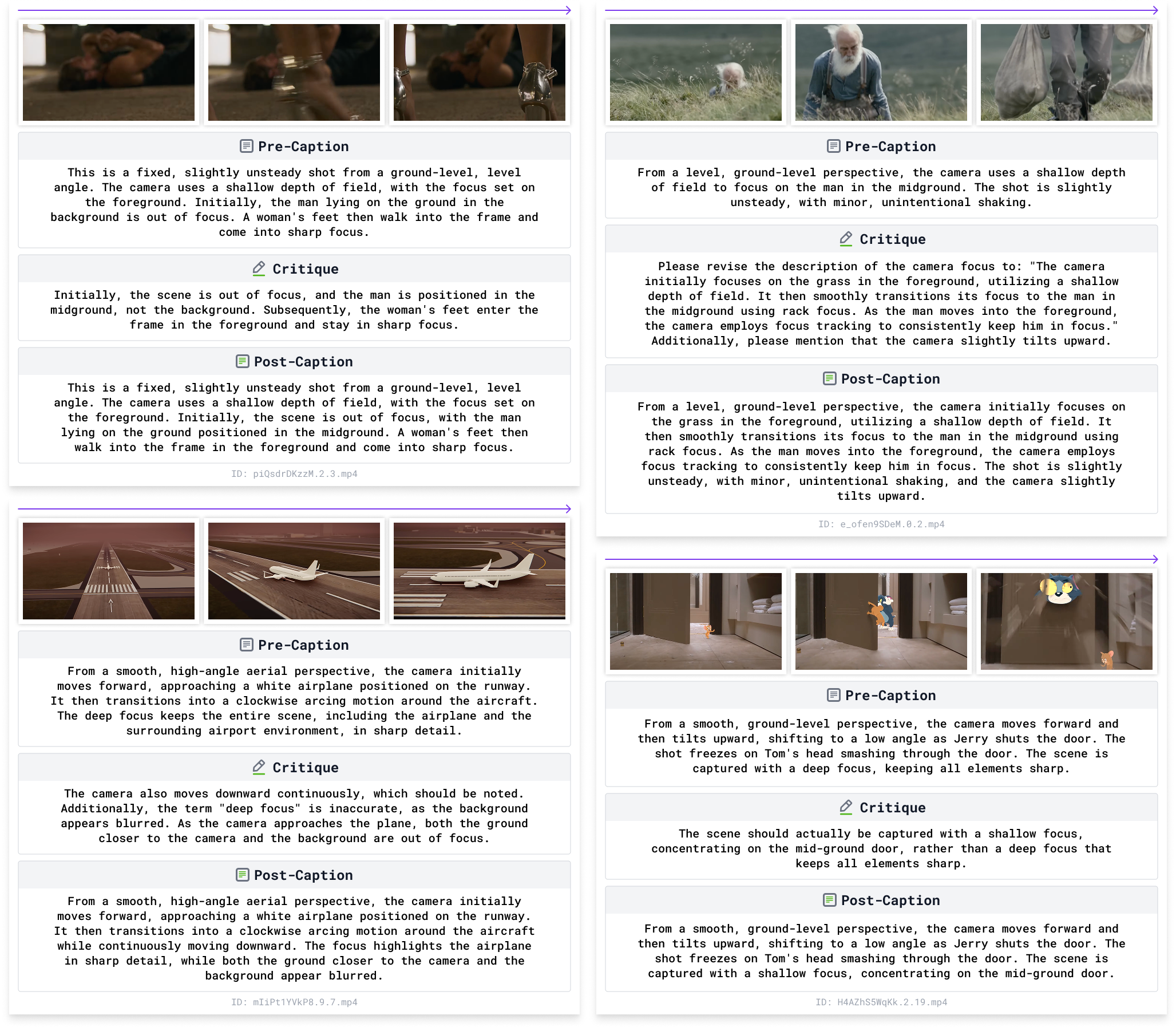} \\
\end{tabular}
\caption{\small {\bf Camera caption examples.}}
\label{fig:camera_captions}
\end{figure*}

In this section, we show more examples of video pre-captions, critiques, and post-captions:
\begin{itemize}
\item Word clouds: \autoref{fig:caption_wordclouds}
\item Subject captions: \autoref{fig:subject_captions}
\item Scene captions: \autoref{fig:scene_captions}
\item Motion captions: \autoref{fig:motion_captions}
\item Spatial captions: \autoref{fig:spatial_captions}
\item Camera captions: \autoref{fig:camera_captions}
\end{itemize}

\clearpage
\onecolumn
\section{Critique Quality Matters}
\label{sec:critique_quality}

In this section, we review the core properties of good critiques, describe the prompts used to inject controlled errors for our ablations in \autoref{tab:critique_comparison}, and analyze why critiques in prior work are often low quality, particularly in constructiveness.

{\bf Our oversight framework enforces quality critiques.}
In our workflow, annotators must write critiques that directly guide the model to produce the final post-caption.
This requirement naturally enforces three properties: critiques must be {\bf accurate} (no hallucinated information), {\bf complete} (every error identified), and {\bf constructive} (explaining not only what is wrong but how to fix it).

\subsection{Prompts for Controlled Critique Error Injection}
\label{sec:critique_prompts}

To isolate the effect of each property, we use LLMs to take our ground-truth human critiques and inject one type of error at a time.
All prompts receive three inputs: the \texttt{caption\_instruction} (the annotation guideline for that caption type), the \texttt{caption} (the pre-caption), and, where applicable, the \texttt{feedback} (the ground-truth human critique).
We produce three degraded variants, detailed below with their full prompt templates and generation models.

\subsubsection{Inaccurate Critiques}

We generate two types of inaccurate critiques: \emph{insertion errors}, which add a spurious point to the original critique (e.g., inserting a fabricated detail when the pre-caption is already correct), and \emph{replacement errors}, which swap a correct point for a plausible but incorrect one (e.g., changing ``{\it the man is wearing black, not white}'' to ``{\it the man is wearing blue, not white}'').

\begin{tcolorbox}[title={\textbf{Insertion Error Critique Prompt} \hfill \texttt{GPT-4.1}}, colback=blue!3, colframe=blue!40, fonttitle=\small\sffamily, breakable]
\small
Please modify the following feedback by adding one extra irrelevant or incorrect detail that was not present in the original critique.

\medskip
\textbf{Caption Instruction:} \texttt{\{caption\_instruction\}}

\textbf{Original Caption:} \texttt{\{caption\}}

\textbf{Original Feedback:} \texttt{\{feedback\}}

\medskip
\textbf{Instructions:}
\begin{enumerate}[leftmargin=*, nosep]
\item Insert one additional detail at a random position in the critique that is either irrelevant to the caption task or factually incorrect about the caption.
\item Make the addition feel natural and integrated with the original feedback. Inserted feedback should provide concrete details related to the caption instruction, not unrelated visual content or vague suggestions.
\item If the original feedback is empty, add one detail that is not included in the original caption and is incorrect about what is shown.
\item Return only the modified feedback paragraph without any additional text or explanations.
\item Avoid repeating any content from the original caption.
\item Do not include non-visual elements (e.g., background music, narration).
\item Provide explicit information, not ambiguous details.
\item Please try to use affirmative sentences rather than negative or interrogative ones.
\item Please do not delete the original feedback content, only insert.
\item Do not use negative statements (e.g., ``there is no \ldots'' or ``avoid mentioning \ldots'') in your inserted feedback.
\end{enumerate}
\end{tcolorbox}

\begin{tcolorbox}[title={\textbf{Replacement Error Critique Prompt} \hfill \texttt{GPT-4.1}}, colback=blue!3, colframe=blue!40, fonttitle=\small\sffamily, breakable]
\small
Please modify the following feedback by replacing one correct detail with wrong or misleading information.

\medskip
\textbf{Caption Instruction:} \texttt{\{caption\_instruction\}}

\textbf{Original Caption:} \texttt{\{caption\}}

\textbf{Original Feedback:} \texttt{\{feedback\}}

\medskip
\textbf{Instructions:}
\begin{enumerate}[leftmargin=*, nosep]
\item Identify one factual detail or suggestion in the original feedback.
\item Replace this detail with an incorrect alternative that sounds plausible but is wrong.
\item Keep the overall structure and tone of the original feedback.
\item Return only the modified feedback paragraph without any additional text or explanations.
\item Avoid repeating any content from the original caption.
\item Do not include non-visual elements (e.g., background music, narration).
\item Provide explicit information, not ambiguous details.
\item If the feedback includes phrases such as ``not xxx'', please keep them, as they indicate errors in the original caption.
\end{enumerate}
\end{tcolorbox}

\subsubsection{Incomplete Critiques}

Incomplete critiques are created by removing a necessary correction from the ground-truth critique (unless the critique is already ``{\it The caption is accurate and requires no edit}'').

\begin{tcolorbox}[title={\textbf{Deletion Error Critique Prompt} \hfill \texttt{GPT-4.1}}, colback=orange!3, colframe=orange!40, fonttitle=\small\sffamily, breakable]
\small
Please modify the following feedback by removing one important detail to make it incomplete.

\medskip
\textbf{Caption Instruction:} \texttt{\{caption\_instruction\}}

\textbf{Original Caption:} \texttt{\{caption\}}

\textbf{Original Feedback:} \texttt{\{feedback\}}

\medskip
\textbf{Instructions:}
\begin{enumerate}[leftmargin=*, nosep]
\item Remove one key detail, suggestion, or explanation from the original feedback only if it is sufficiently long.
\item If the original feedback consists of only a single sentence or item, do not simply shorten it, but replace it with ``The caption is accurate and requires no edits, so it should remain exactly the same.''
\item Return only the modified feedback paragraph without any additional text or explanations.
\item If the feedback is presented as a numbered list, when deleting, remove one item at random rather than automatically deleting the last entry.
\item Identify the portions of the feedback that conflict with the caption. These conflicting elements are relatively significant and can be prioritized for deletion, but delete only one full element from the original feedback.
\end{enumerate}
\end{tcolorbox}

\subsubsection{Non-Constructive Critiques}

Non-constructive critiques retain only the error identification while stripping the fix (e.g., reducing ``{\it the man is wearing black, not white}'' to ``{\it the man is not wearing white}'').

\begin{tcolorbox}[title={\textbf{Non-Constructive Critique Prompt} \hfill \texttt{Gemini-2.5-Pro}}, colback=red!3, colframe=red!40, fonttitle=\small\sffamily, breakable]
\small
Please modify the following feedback to only point out problems without providing any constructive suggestions or solutions.

\medskip
\textbf{Caption Instruction:} \texttt{\{caption\_instruction\}}

\textbf{Original Caption:} \texttt{\{caption\}}

\textbf{Original Feedback:} \texttt{\{feedback\}}

\medskip
\textbf{Instructions:}
\begin{enumerate}[leftmargin=*, nosep]
\item Convert all constructive suggestions in the feedback into criticism only: state only what is wrong in the caption that conflicts with the feedback, without mentioning what is correct.
\item Remove all helpful guidance or improvement suggestions. If the feedback is only guidance or suggestions, replace it with: ``The caption is accurate and requires no edits, so it should remain exactly the same.''
\item Return only the non-constructive feedback paragraph, with no extra text or explanation.
\item When feedback suggests adding something (not changing one thing to another), rephrase it to say the caption is missing that thing, stated generally without details.
\item If the feedback only provides the corrected version without explaining the issues, identify the problematic parts in the caption and state which parts are wrong.
\end{enumerate}
\end{tcolorbox}

\subsection{Model-Generated Critiques}

We also compare against critiques generated by Gemini-2.5-Pro in two settings: a \emph{video model} that receives the full video alongside the pre-caption, and a \emph{blind model} that receives only the pre-caption and must hallucinate its critique.
Note that unlike the error-injection prompts above, these prompts do not receive ground-truth feedback.

\begin{tcolorbox}[title={\textbf{Video Model Critique Prompt} \hfill \texttt{Gemini-2.5-Pro} + Video}, colback=green!3, colframe=green!40, fonttitle=\small\sffamily, breakable]
\small
Please provide detailed feedback on how well this caption follows the given instruction. Carefully watch the video and analyze the caption against the instruction requirements.

\medskip
\textbf{Caption Instruction:} \texttt{\{caption\_instruction\}}

\textbf{Caption:} \texttt{\{caption\}}

\medskip
\textbf{Instructions:}
\begin{enumerate}[leftmargin=*, nosep]
\item Carefully watch the video and review the caption against the specific requirements in the caption instruction.
\item Identify any missing elements, inaccuracies, or areas for improvement based on what you observe in the video.
\item Provide specific, actionable suggestions for how to improve the caption.
\item Be thorough and constructive in your analysis.
\item If the caption is already excellent, simply state ``The caption is accurate and requires no edits, so it should remain exactly the same.''
\item Return only your feedback paragraph without any additional text or explanations.
\item If you discover any missing elements in the caption---details present in the video but omitted---point out which element has been left out.
\item If you find any factual errors in the caption that conflict with the actual video, identify where the error occurs and explain how it should be corrected.
\item If the caption is overly long and contains information unrelated to the Caption Instruction or is significantly redundant, point out those parts and explain that they need to be deleted.
\item Do not offer feedback on things not specified in the Caption Instruction. Do not be wordy; keep suggestions concise, direct, and constructive.
\end{enumerate}
\end{tcolorbox}

\begin{tcolorbox}[title={\textbf{Blind Model Critique Prompt} \hfill \texttt{Gemini-2.5-Pro} (no video)}, colback=gray!5, colframe=gray!50, fonttitle=\small\sffamily, breakable]
\small
Please provide feedback on this caption by imagining you have watched the video. Generate a critique by assuming you have visual access to the content (you can imagine anything in the video).

\medskip
\textbf{Caption Instruction:} \texttt{\{caption\_instruction\}}

\textbf{Caption:} \texttt{\{caption\}}

\medskip
\textbf{Instructions:}
\begin{enumerate}[leftmargin=*, nosep]
\item Pretend you have watched the video and generate feedback based on your imagined visual content.
\item Create specific critique points about what you imagine might be missing or incorrect in the caption.
\item Provide suggestions for improvement based on your imagined video content.
\item Make the feedback substantial and detailed.
\item You can imagine any visual elements that seem plausible for this type of content.
\item Return only your feedback paragraph without any additional text or explanations.
\end{enumerate}
\end{tcolorbox}

\subsection{Critiques in Prior Work Are Often Non-Constructive}

We find that critiques collected in prior work are frequently non-constructive, which may partly explain why they are not consistently helpful for downstream training or evaluation.
We study OpenAI's GDC~\cite{saunders2022self}, which collects critiques for topic-based summarization and text question answering, and MM-RLHF~\cite{zhang2025mm}, which collects critiques for video question answering.
Using ChatGPT to classify critiques as constructive or non-constructive, combined with manual verification over ${\sim}200$ samples per dataset (ensuring ${\geq}{\sim}95\%$ accuracy), we estimate a lower bound on the fraction of critiques that are non-constructive.
As shown in \autoref{tab:mmrlhf_gdc_ablation}, more than half of the critiques in these datasets (and over 90\% in MM-RLHF for responses requiring critique) are non-constructive.
Random qualitative examples are shown in \autoref{tab:gdc_summarization} (GDC summarization), \autoref{tab:gdc_qa} (GDC text QA), and \autoref{tab:mmrlhf} (MM-RLHF video QA).
This contrasts sharply with our approach: in GDC and MM-RLHF, annotators are asked only to critique model outputs, but not to use that critique to correct them.
In our oversight framework, critiques are written expressly to {\it fix} the model output, which naturally enforces precision, recall, and constructiveness.

\begin{table*}[t]
\centering
\renewcommand{\arraystretch}{1.25}
\caption{\small \textbf{Prior work collects non-constructive critiques.}
We report the distribution of constructive versus non-constructive critiques for each benchmark task. Surprisingly, these datasets contain a high proportion of unhelpful critiques that only point out what is wrong without explaining what is correct or how to fix it.}
\scalebox{0.90}{
\begin{NiceTabular}{llccc}
\toprule[1.5pt]
\multirow{2}{*}{\textbf{Dataset / Benchmark}} 
& \multirow{2}{*}{\textbf{Task}} 
& \multirow{2}{*}{\textbf{Size}} 
& \multicolumn{2}{c}{\textbf{Critique Distribution}} \\
\cmidrule(l){4-5}
& & & \textbf{Constructive} & \textbf{Non-Constructive} \\
\midrule

\multirow{2}{*}{OpenAI GDC~\cite{saunders2022self}} 
  & Topic-based Summarization 
  & 1728 
  & 53\% & 47\% \\

  & Text Question Answering 
  & 1851 
  & 51\% & 49\% \\
\midrule
MM-RLHF~\cite{zhang2025mm}
  & Video Question Answering 
  & 8751 
  & 10\% & 90\% \\

\bottomrule[1.5pt]
\end{NiceTabular}
}
\vspace{-2mm}
\label{tab:mmrlhf_gdc_ablation}
\end{table*}

\begin{table*}[h]
\centering
\small
\renewcommand{\arraystretch}{1.2}
\scalebox{0.8}{
\begin{NiceTabular}{M{0.12\linewidth} M{0.22\linewidth} M{0.15\linewidth} M{0.45\linewidth} c}
\toprule[1.5pt]
\textbf{Question} & \textbf{Answer} & \textbf{Critique} & \textbf{Is the Critique Constructive?} & \textbf{Label} \\ \midrule

What is the purpose of the U.S red flag list? & 
To require that U.S companies treat customers named on the list with caution. US suppliers are to no longer use existing licenses to sell these customers products or servIce installed EquItment. & 
Replace capital ``I'' in ``servIce''. Use the correct spelling for ``EquItment''. & 
The critique provides specific instructions on how to improve the answer by correcting the capitalization and spelling errors. It does not require additional information from the source material to make these changes. The editor can directly implement these corrections. & 
\textcolor{green!60!black}{\textbf{Yes}} \\ \midrule

What are some related fields to circle packing? & 
Sphere packing is a related field. & 
The passage also mentions ring lemma and Doyle spirals as related topics. & 
The critique provides specific additional related topics, ``ring lemma'' and ``Doyle spirals,'' that can be directly added to the answer. This allows for a concrete improvement to the answer without needing to refer back to the source material. & 
\textcolor{green!60!black}{\textbf{Yes}} \\ \midrule

Summarize the global reaction to Trump's remarks. & 
Botswana summoned the US ambassador to clarify if they thought of Botswana as a shithole country. Rupert Colville, UN Human Rights spokesperson, said the comments were "shocking and shameful." Haiti's US Ambassador Paul Altidor believes that the remarks Trump made about immigrants who come to the US are wrong. & 
The answer should include the response of the African Union. & 
The critique specifies that the answer should include the response of the African Union. However, \textcolor{red}{it does not provide any details about what the African Union's response was.} Without this information, I cannot improve the answer without referring back to the source material. & 
\textcolor{red}{\textbf{No}} \\ \midrule

Why is the protagonist behaving in such a self-destructive way? & 
This information can't be found. & 
The information is not in the story but it can be implied. & 
The critique suggests inferring behavior, but without access to the source material, \textcolor{red}{it is impossible to determine what those implications might be.} The critique does not provide specific guidance or details for a concrete improvement. & 
\textcolor{red}{\textbf{No}} \\ \midrule

Summarize all the information relating to the mountainous Austrian Tyrol. & 
Averil Demuth's first story was set in the Austrian Tyrol. The girl Trudi, the boy Hansel, the cow Lotti, the dog Berni, the hen Griselda and the raven Kraak all go up the mountain to see Riese the giant, and then all come down again. Eleanor Farjeon gave the book a positive review. The House in the Mountains was also set in the Austrian Tyrol. & 
The sentence is unnecessary. & 
The critique states that ``The sentence is unnecessary,'' but \textcolor{red}{it does not specify which sentence is unnecessary.} Since there are multiple sentences, it is unclear which one should be removed without referring back to the source material. & 
\textcolor{red}{\textbf{No}} \\ \midrule

Who is the protagonist of this story? & 
The protagonist of this story is Stephen. & 
Describe him more. & 
The critique asks for a description of the protagonist, Stephen, but \textcolor{red}{does not provide any specific details or characteristics about him}. Without additional information from the source material, it is not possible to expand on the description of Stephen. Therefore, the critique does not provide enough information to improve the answer without referring back to the source material. & 
\textcolor{red}{\textbf{No}} \\ 

\bottomrule[1.5pt]
\end{NiceTabular}
}
\caption{\small \textbf{Critique analysis on OpenAI GDC~\cite{saunders2022self} (Topic-based Summarization).} We show random examples from the test set (1728 samples) and find that more than 47\% of critiques in this task are non-constructive.}
\label{tab:gdc_summarization}
\end{table*}

\begin{table*}[h]
\centering
\small
\renewcommand{\arraystretch}{1.2}
\scalebox{0.8}{
\begin{NiceTabular}{M{0.12\linewidth} M{0.22\linewidth} M{0.15\linewidth} M{0.45\linewidth} c}
\toprule[1.5pt]
\textbf{Question} & \textbf{Answer} & \textbf{Critique} & \textbf{Is the Critique Constructive?} & \textbf{Label} \\ \midrule

Why does the narrator suggest that they must terminate the patient's pregnancy? & 
The narrator suggests that they must terminate the patient's pregnancy because the current sector has too many male children. This is due to the fact that the State has mandated that all pregnancies must be terminated if the fetus is male. & 
The second sentence ``mandated that all pregnancies must be terminated if the fetus is male'' should be changed to ``has regulates the gender distribution''. & 
The critique provides specific guidance on how to improve the answer by suggesting a change to the wording. It implies the issue is about regulating gender distribution, allowing for concrete improvement without source material. & 
\textcolor{green!60!black}{\textbf{Yes}} \\ \midrule

How fast is Jiranová? & 
Jiranová is a Czech runner and sprinter who has personal bests in the 400 metres, 100 metres hurdles, and 60 metres hurdles. Her 400 metres best is 53.77 seconds. Her 100 metres hurdles best is 13.37 seconds, and her 60 metres hurdles best is 8.31 seconds. & 
The first sentence is unnecessary for answering the question. & 
The critique specifies that the first sentence is unnecessary. This implies the answer should focus solely on her personal best times. The editor can improve the answer by removing the extraneous sentence. & 
\textcolor{green!60!black}{\textbf{Yes}} \\ \midrule

What is the voting process in Switzerland? & 
Information can't be found. & 
This information can be found in the article. & 
The critique states the information exists but \textcolor{red}{does not provide any specific details or corrections.} It does not imply what the correct information is, so the editor cannot make a concrete improvement without the source. & 
\textcolor{red}{\textbf{No}} \\ \midrule

How old was Sanuth when his cricket career took off? & 
Sanuth was 25 years old when his cricket career took off. & 
The answer is incorrect. & 
The critique states the answer is incorrect but \textcolor{red}{does not provide any information about the correct age.} Without additional details, the editor cannot determine the correct age to improve the answer. & 
\textcolor{red}{\textbf{No}} \\ \midrule

How is a consensus reached? & 
No answer. & 
No answer to the question. & 
The critique points out the lack of an answer but \textcolor{red}{does not provide any details or implications about what should be included.} The editor needs the source material to construct a response. & 
\textcolor{red}{\textbf{No}} \\ \midrule

Why did the Leicester mayor find the move by the health secretary disappointing? & 
The mayor found the move by the health secretary disappointing because it means the premier's planned speech will shift away from coronavirus and onto the lockdown, making COVID an even greater focus. & 
``...because it means the premier's planned speech will shift away from coronavirus and onto the lcokdown...'' This sentence lacks clarity and accuracy. & 
The critique claims the sentence lacks clarity/accuracy but \textcolor{red}{does not provide specific information on how to clarify or correct it.} It is impossible to improve without understanding the correct context from the source. & 
\textcolor{red}{\textbf{No}} \\ 

\bottomrule[1.5pt]
\end{NiceTabular}
}
\caption{\small \textbf{Critique analysis on OpenAI GDC (Text Question Answering).} We show random examples from the test set. Out of 1851 test samples, we find that 49\% of critiques in this task are non-constructive.}
\label{tab:gdc_qa}
\end{table*}

\begin{table*}[h]
\centering
\small
\renewcommand{\arraystretch}{1.2}
\scalebox{0.8}{
\begin{NiceTabular}{M{0.12\linewidth} M{0.22\linewidth} M{0.15\linewidth} M{0.45\linewidth} c}
\toprule[1.5pt]
\textbf{Question} & \textbf{Answer} & \textbf{Critique} & \textbf{Is the Critique Constructive?} & \textbf{Label} \\ \midrule

What is depicted in the frames throughout the video? & 
The frames depict a hand holding a small, toy car-shaped chocolate. & 
The description is incorrect; it is a car-shaped chocolate candy. & 
The critique specifies the object is a ``car-shaped chocolate candy'' rather than a ``toy car.'' This provides enough information to correct the description without referring back to the source video. & 
\textcolor{green!60!black}{\textbf{Yes}} \\ \midrule

What colors are showcased in the video? & 
The video showcases a variety of colors, including green grass, blue sky, white clouds, and snow-capped mountains. & 
Correct answer, but the description and analysis are minimal. & 
The critique points out minimal analysis but \textcolor{red}{does not specify what additional description is needed.} Without specific guidance, it is impossible to improve the answer without viewing the video. & 
\textcolor{red}{\textbf{No}} \\ \midrule

What kind of tool is the person holding in their hands? & 
The person is holding a tool, possibly a screwdriver or a similar implement. & 
Failure to identify the tool’s purpose. & 
The critique mentions a failure to identify the purpose but \textcolor{red}{does not provide information about what the purpose is.} It is impossible to specify the tool's use without this info. & 
\textcolor{red}{\textbf{No}} \\ \midrule

What are the two primary colors represented on the Christmas tree? & 
The two primary colors represented on the Christmas tree are green and purple. & 
Green description error. & 
The critique mentions a ``green description error'' but \textcolor{red}{does not provide specific information about what the error is or how to correct it.} Without details, the answer cannot be improved. & 
\textcolor{red}{\textbf{No}} \\ \midrule

Who is the main focus of the video and what action do they take? & 
I'm unable to view the video and identify the main focus or actions taken. Could you describe the scene for me? & 
No answer to the question. & 
The critique notes the lack of answer but \textcolor{red}{does not provide any information or guidance on what the correct answer should be.} The editor cannot improve the answer without the source content. & 
\textcolor{red}{\textbf{No}} \\ \midrule

What is the individual wearing in the video? & 
The individual is wearing a long-sleeved pajama top with a pattern, possibly featuring small illustrations, and matching pajama pants. & 
The description of the top is accurate, but the color of the pants is incorrect. & 
The critique indicates the pants color is incorrect but \textcolor{red}{does not provide the correct color.} The editor would need to view the video to determine the correct color. & 
\textcolor{red}{\textbf{No}} \\ 

\bottomrule[1.5pt]
\end{NiceTabular}
}
\caption{\small \textbf{Critique analysis on MM-RLHF~\cite{zhang2025mm} (Video Question Answering).} We show random examples from MM-RLHF that have an non-empty critique (58\% out of the 8751 test samples). Of those, we find that 90\% are non-constructive.}
\label{tab:mmrlhf}
\end{table*}

\clearpage
\clearpage
\section{Post-training details}
\label{sec:post_training}

In this section we provide additional details on post-training experiments. We ablate reference-based evaluation metrics to measure how well they align with human Likert scale ratings, report additional reference-based metrics for caption and critique generation, and include the task instructions used for the five captioning tasks. We also describe the formats used in the SFT dataset, including how we enforce a balanced dataset for training reward models for both captions and critiques, compare text-based versus logit-based reward scoring, and finally report all training hyperparameters for offline experiments (SFT, DPO, RLHF-V).

{\bf LLM-Judge-Direct.} This method directly sends the reference caption and the generated candidate caption to GPT-4o and takes the P(Yes)~\cite{lin2024evaluating} for the question below:
{
\ttfamily

Reference caption: \{reference\}

Candidate caption: \{candidate\}

Does the candidate caption accurately match the reference caption in terms of content and meaning? Answer only Yes or No.
}

{\bf LLM-Judge-Instruct.} We also try sending the task-specific instruction to GPT-4o for computing P(Yes):
{
\ttfamily

Task Instruction: \{instruction\}

Reference caption: \{reference\}

Candidate caption: \{candidate\}

Does the candidate caption accurately follow the task instruction and match the reference? Answer only Yes or No.
}

{\bf Task instructions.} The instruction for each captioning task is shown below:
\begin{itemize}
    \item {\bf Subject:} {\it Provide a concise yet informative description of the subjects in this video, including their types, appearances (e.g., clothing, facial expressions, gender, ethnicity, color, shape), and poses. When multiple subjects are present, clearly distinguish them using unique traits, position, actions, or relationships, and describe them in temporal or prominence-based order to ensure clarity.}
    \item {\bf Scene:} {\it Provide a concise yet informative description of the overall scene, including the point of view, environment, setting, time of day, overlays, and notable visual elements. If subjects are present, the scene description should complement their descriptions by establishing their location and possible context. Aim to give enough detail to convey the setting while avoiding unnecessary information.}
    \item {\bf Motion:} {\it Provide a concise yet informative description of the subject's motion in this video, including individual actions, subject–object or subject–subject interactions, and group activities when a crowd is present. Event order matters—if multiple actions occur, present them in chronological order.}
    \item {\bf Spatial:} {\it Provide a concise yet informative description of how subjects and elements are spatially framed within the scene, including the shot size of the subject (or the shot size of the scenery if there is no salient subject), their 2D position within the frame, spatial depth within the scene (foreground, middle ground, background), height relative to the camera, and any notable spatial movement.}
    \item {\bf Camera:} {\it Provide a concise yet informative description of the video and camera configuration, including playback speed, lens distortion (if present), camera angle, camera height relative to the ground plane, camera movements (translation, rotation, zooming, steadiness, speed, intensity, and complexity), and focus (depth, focus plane, and any changes in focus).}
\end{itemize}

\begin{figure*}[tb!]
\centering
    \includegraphics[width=0.5\textwidth]{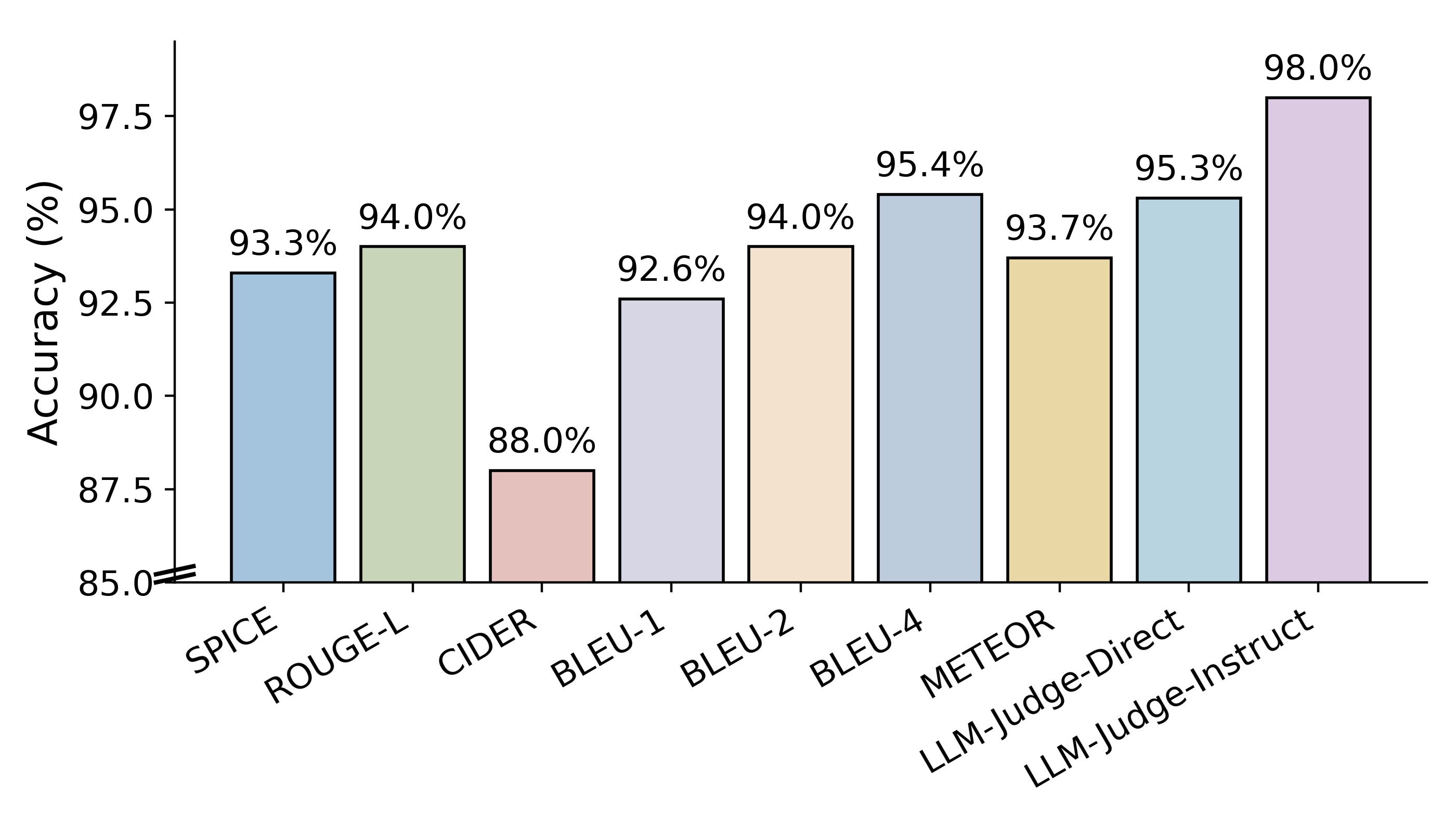}
    \caption{\small {\bf Comparing reference-based metrics.}
We evaluate each reference-based metric by measuring its pairwise accuracy (with tie optimization~\cite{deutsch2023ties}) when comparing scores between any two captions in the evaluation set, which includes model pre-captions rated with human 1-to-5 Likert scores and adversarial negative captions that automatically receive a score of 1. LLM-Judge-Instruct with GPT-4o achieves the best performance (98.0\%) by incorporating the caption instruction when comparing reference and candidate captions, followed by LLM-Judge-Direct with GPT-4o (95.3\%); BLEU-4 (95.4\%) remains the most reliable classic metric.
}
\label{fig:reference_based_metrics_ablation}
\end{figure*}

{\bf Ablating reference-based evaluation metrics.}
We compare a range of reference-based metrics to assess how well they align with human judgement. Since we collect 1-to-5 human Likert scores for all model pre-captions and also generate adversarial negative captions that automatically receive a score of 1 by hallucinating all details (explained later in this section), we can meta-evaluate a metric by computing its pairwise accuracy across all items: for every pair of captions in the evaluation set (consisting of both pre-captions and adversarial negative captions), the metric is correct if it assigns a higher score to the caption with the higher human Likert score. We follow Pairwise Accuracy with tie optimization~\cite{deutsch2023ties, li2024evaluating}, which is more robust for this setting because its tie-optimization procedure handles the many tied pairs that naturally arise from discrete 1-to-5 Likert scores. The metrics we evaluate include SPICE, ROUGE-L, METEOR, CIDER, BLEU-1/2/4, as well as the two versions of LLM-Judge described above. \autoref{fig:reference_based_metrics_ablation} shows that LLM-Judge-Instruct with GPT-4o achieves the highest alignment with human ratings (98.0\%), followed by LLM-Judge-Direct with GPT-4o (95.3\%); among classic metrics, BLEU-4 is the most reliable (95.4\%). Providing the task instruction consistently improves alignment, since it lets the judge weight task-relevant details (e.g., spatial framing vs.\ scene atmosphere) when scoring the candidate caption. For the main experiments, we continue to use BLEU-4 as the most reproducible classic metric, and additionally report LLM-Judge-Instruct in \autoref{tab:caption_critique_metrics} as a complementary signal.

{\bf Evaluating critique generation via caption-revision.} For critique generation, we introduce a more reliable metric, {\bf Critique-Revision}, which avoids comparing against the reference critique directly (since critiques vary widely in style, such as whether annotators use bullet points to point out mistakes). Instead, we use the generated critique to revise the caption with ChatGPT using the prompt shown below, and then compare the revised caption to the reference caption:
{
\ttfamily
Given a video caption and user feedback, please provide an improved version of the caption that addresses the feedback. Note that the user feedback could be poorly written, so please try your best to guess what it means.

Original caption: \{pre\_caption\}

User feedback: \{critique\}

Respond with the improved caption only, without quotation marks or JSON formatting.
}

{\bf Results with more reference-based metrics.} \autoref{tab:caption_critique_metrics} reports additional reference-based metrics for caption and critique generation, including ROUGE-L, LLM-Judge-Instruct scores for captioning, and Critique-Revision scores for critique generation. The ranking of methods is stable across all metrics: explicit preference and critique supervision improves standard SFT and RL methods, and our full-data post-training recipe outperforms all baselines.

{\bf Details for the offline SFT dataset.} Below, we describe the eight formats used during SFT training. We also apply text cleaning by stripping newlines from all collected captions and critiques.

\begin{figure*}[tb!]
\centering
    \includegraphics[width=0.5\textwidth]{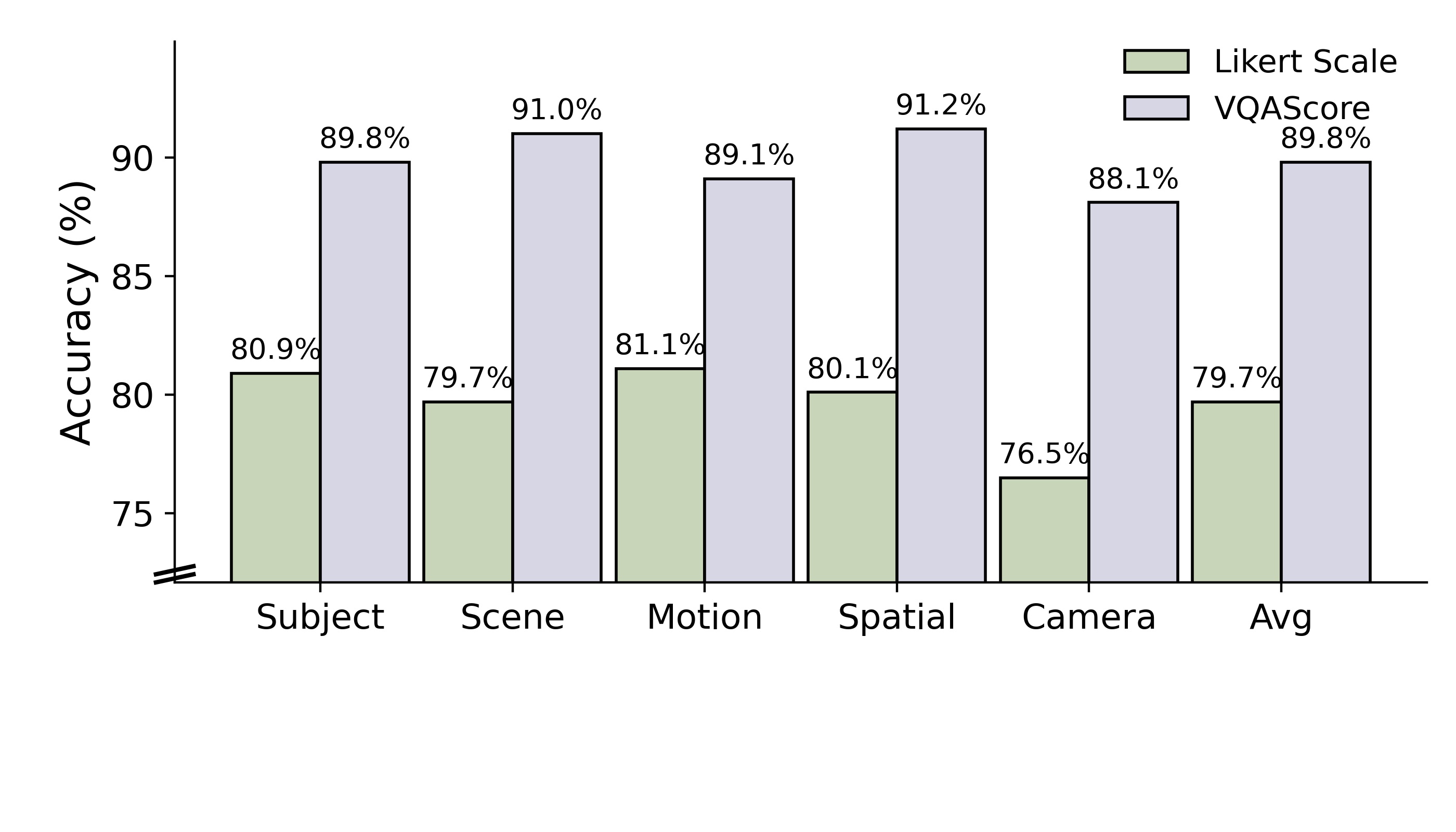}
    \caption{\small {\bf Comparing reward models using VQAScore versus text generation.} We find that VQAScore (a logit-based probability score) significantly outperforms asking the model to directly output a Likert-scale score from 1 to 5 after supervised fine-tuning (SFT). Therefore, we stick to VQAScore for reward modeling in this work.
    }\label{fig:scoring_ablation}
\end{figure*}

\begin{enumerate}[leftmargin=1.5em]

\item {\bf Caption Generation}

\begin{itemize}
    \item {\bf Input:} Video, Task Instruction
    \item {\bf Output:} Caption
\end{itemize}

\vspace{0.6em}

\item {\bf Critique Generation}

\begin{itemize}
    \item {\bf Input:}
    \begin{itemize}
        \item Video, Task Instruction, Caption
        \item {\tt Provide a critique of the video's caption based on how accurately it follows the task instruction. Point out what is wrong or missing and how to fix it. If the caption is already accurate, output: ``The caption is accurate and requires no edits, so it should remain exactly the same.''}
    \end{itemize}
    \item {\bf Output:} Critique
\end{itemize}

\vspace{0.6em}

\item {\bf Caption Reward (VQAScore)}

\begin{itemize}
    \item {\bf Input:}
    \begin{itemize}
        \item Video, Task Instruction, Caption
        \item {\tt Does the video's caption accurately follow the task instruction? Please answer Yes or No only.}
    \end{itemize}

    \item {\bf Output:} Yes (if human Likert scale is 5) or No (if human Likert scale is 1-4)

    \item {\bf Note:} To prevent the reward model from collapsing to a constant prediction, we balance Yes and No examples following~\cite{naturalbench}: for each post-caption labeled Yes, we treat the corresponding pre-caption as a No example if its human score is below 5, and exclude pairs where the pre-caption already scores a perfect 5. A variant that directly predicts the Likert score via text generation performs worse. See \autoref{fig:scoring_ablation} for results.
\end{itemize}

\vspace{0.6em}

\item {\bf Critique Reward (VQAScore)}

\begin{itemize}
    \item {\bf Input:}
    \begin{itemize}
        \item Video, Task Instruction, Caption, Critique
        \item {\tt Does this critique of the video's caption provide accurate and constructive feedback to help the caption better follow the task instruction? Please answer Yes or No only.}
    \end{itemize}

    \item {\bf Output:} Yes or No

    \item {\bf Note:} We construct balanced pairs only from (video, pre-caption) samples that require critique.
The reward model answers ``Yes'' for (video, post-caption, ``{\tt The caption is accurate and requires no edits, so it should remain exactly the same}'') and (video, pre-caption, human-critique), and ``No'' for (video, pre-caption, ``{\tt The caption is accurate and requires no edits, so it should remain exactly the same}'') and (video, post-caption, \{insertion or replacement error critique\}).
This balances Yes and No answers by pairing correct assessments against adversarial negative critiques.
We follow the procedure in \autoref{sec:critique_quality} to generate these critiques.
\end{itemize}

\vspace{0.6em}

\item {\bf Caption Revision}

\begin{itemize}
    \item {\bf Input:}
    \begin{itemize}
        \item Video, Task Instruction, Caption
        \item {\tt Provide an improved caption that better follows the task instruction. If the original caption is already accurate, keep it exactly the same with no edits.}
    \end{itemize}
    \item {\bf Output:} Revised Caption
\end{itemize}

\vspace{0.6em}

\item {\bf Caption Revision (with Given Critique)}

\begin{itemize}
    \item {\bf Input:}
    \begin{itemize}
        \item Video, Task Instruction, Caption, Critique
        \item {\tt Provide an improved caption that better follows the task instruction. You are provided a critique of the caption with respect to the task instruction as context. If the original caption is already accurate, keep it exactly the same with no edits.}
    \end{itemize}
    \item {\bf Output:} Revised Caption

    \item {\bf Note:} During training, critique is human-generated.
\end{itemize}

\vspace{0.6em}

\item {\bf Critique-based Caption Revision}

\begin{itemize}
    \item {\bf Input:}
    \begin{itemize}
        \item Video, Task Instruction, Caption
        \item {\tt Provide an improved caption that better follows the task instruction, after first writing a critique. If the caption is already accurate enough, the critique should be: `The caption is accurate and requires no edits, so it should remain exactly the same.' And keep the original caption exactly the same with no edits. Write your critique after `Critique:', then on a new line write the improved caption after `Improved Caption:'}
    \end{itemize}
    \item {\bf Output:} (Critique, Revised Caption)
\end{itemize}

\vspace{0.6em}

\item {\bf Caption Scoring}

\begin{itemize}
    \item {\bf Input:}
    \begin{itemize}
        \item Video, Task Instruction, Caption
        \item {\tt Score the video's caption based on how well it follows the task instruction. Rate 1–5 (1=poor, 5=excellent).}
    \end{itemize}

    \item {\bf Output:} Likert scale score (1-to-5)
    \item {\bf Note:} This format is used only for the reward-scoring ablation in \autoref{fig:scoring_ablation} and is not included in any other experiments. To train this format more effectively, because model pre-captions almost never receive a score of 1, we add adversarially generated negative captions labeled with score 1 to balance the score distribution in the training set.
\end{itemize}

\end{enumerate}

{\bf Constructing negative responses for DPO and RLHF-V.} Our offline dataset can easily pair each positive response with a negative one. For caption generation, we use the pre-caption when it has a score below 5, or skip it otherwise. For critique generation, we sample an adversarial critique that is inaccurate, incomplete, or non-constructive. For reward modeling, we flip the Yes/No answer. For revision, the pre-caption serves as the negative and the post-caption as the positive. This construction means every triplet in our dataset contributes training signal to all post-training methods under comparison, which is important for apples-to-apples evaluation.

{\bf Reward scoring format: VQAScore vs.\ text-generated Likert.} A natural question is whether the reward model should emit a scalar score directly (by generating a Likert digit 1–5 as text) or whether it should predict a binary Yes/No label whose $P(\text{Yes})$ is then used as a continuous score via VQAScore~\cite{lin2024evaluating}. We ablate both in \autoref{fig:scoring_ablation} and find that VQAScore substantially outperforms text-generated Likert after SFT, across all five aspects. We attribute this to two factors: (i) the Likert head must allocate probability mass across five discrete outputs, which dilutes the supervision signal; and (ii) human Likert labels are imbalanced (most pre-captions score 3–4, rarely 1 or 5), which hurts Likert-style training even after adversarial balancing. VQAScore avoids both issues by reducing to a binary classification with a continuous logit-based score. We therefore use VQAScore for all reward-modeling results in the main paper.

{\bf Training details for SFT, DPO, RLHF-V.}
We fine-tune Qwen3-VL-8B-Instruct under three paradigms.
All use a learning rate of $3 \times 10^{-5}$, a maximum sequence length of 2048 tokens, cosine scheduling, and process videos at 8 FPS with up to 128 frames.
For \textbf{SFT}, we perform full-parameter training with the vision tower and multi-modal projector frozen, batch size of 10 per device with 2 gradient accumulation steps, 5\% warmup, and train for 3 epochs with ZeRO Stage 3 and BF16 precision.
For \textbf{DPO} and \textbf{RLHF-V}, we apply LoRA (rank 16, targeting all modules) with batch size of 1 per device with 8 gradient accumulation steps, 10\% warmup, and train for 3 epochs.
DPO uses $\beta=0.1$ with the sigmoid loss.
For \textbf{RLHF-V}~\cite{yu2024rlhf}, we follow the original paper's defaults with $\beta=0.5$ and segment-weighting coefficient $\gamma=5$, using Python's \texttt{difflib} (\texttt{SequenceMatcher}) to label unchanged vs.\ corrected tokens.

\begin{table*}[tb!]
\centering
\renewcommand{\arraystretch}{1.25}
\caption{\small \textbf{More reference-based metrics for captioning and critique generation.}
We additionally report ROUGE-L for both captioning and critique generation, LLM-Judge-Instruct scores for captions, and Critique-based-Revision scores for critiques.}
\scalebox{0.75}{
\begin{NiceTabular}{lcccccc}
\toprule[1.5pt]
\multirow{2}{*}{\textbf{Method}}
& \multicolumn{3}{c}{\textbf{Caption Generation}}
& \multicolumn{3}{c}{\textbf{Critique Generation}} \\
\cmidrule(l){2-4} \cmidrule(l){5-7}
& {\scriptsize BLEU-4} & {\scriptsize ROUGE-L} & {\scriptsize LLM-Judge-Instruct}
& {\scriptsize BLEU-4} & {\scriptsize ROUGE-L} & {\scriptsize Caption-Revision} \\
\midrule
\multicolumn{7}{l}{\it Open-source models} \\
Qwen3-VL-8B-Instruct~\cite{qwen3}
  & 2.6 & 4.2 & 4.3 & 1.3 & 2.8 & 4.0 \\
Qwen3-VL-32B-Instruct~\cite{qwen3}
  & 5.3 & 8.4 & 8.8 & 1.4 & 3.0 & 7.1 \\
\midrule
\multicolumn{7}{l}{\it Closed-source models} \\
GPT-4o~\cite{gpt4}
  & 4.2 & 6.5 & 6.7 & 1.8 & 6.8 & 5.0 \\
GPT-5
  & 5.5 & 8.5 & 8.8 & 2.8 & 4.5 & 6.5 \\
Gemini-2.5-Pro~\cite{comanici2025gemini}
  & 6.0 & 9.3 & 9.6 & 3.0 & 5.7 & 7.0 \\
Gemini-3.1-Pro
  & 5.1 & 7.9 & 8.2 & 3.3 & 8.2 & 7.5 \\
\midrule
\multicolumn{7}{l}{\it Caption-only post-training (Qwen3-VL-8B-Instruct)} \\
RLHF-V (Caption)
  & 8.8 & 15.4 & 12.8 & 1.9 & 5.1 & 4.2 \\
DPO (Caption)
  & 9.0 & 15.7 & 11.2 & 1.9 & 5.2 & 4.3 \\
SFT (Caption)
  & 12.0 & 20.8 & 14.5 & 5.5 & 5.9 & 4.8 \\
\midrule
\multicolumn{7}{l}{\it \textbf{Full data post-training (Qwen3-VL-8B-Instruct)}} \\
\textbf{RLHF-V (All)}
  & 15.7 & 32.5 & 21.0 & 25.7 & 47.8 & 14.0 \\
\textbf{DPO (All)}
  & 15.8 & 32.8 & 20.9 & 25.5 & 47.4 & 13.7 \\
\textbf{SFT (All)}
  & 18.2 & 39.2 & 24.1 & 41.7 & 53.0 & 15.9 \\
\bottomrule[1.5pt]
\end{NiceTabular}
}
\label{tab:caption_critique_metrics}
\end{table*}

\clearpage
\section{Inference-Time Scaling}
\label{sec:inference_scaling}

\begin{table*}[t!]
\centering
\caption{\small \textbf{Test-time scaling strategies for reward modeling.}
We compare four reward scoring (VQAScore) modes: \textbf{Direct} computes $P(\text{Yes})$ in a single pass;
\textbf{Critique-First} generates a critique then computes $P(\text{Yes})$;
\textbf{Critique-Last} computes $P(\text{Yes})$ then generates a critique;
\textbf{Self-Critique-Conditioned} generates a self-critique in pass~1 and computes $P(\text{Yes} \mid \text{critique})$ in pass~2.
In addition, Self-Consistency~\cite{wang2022self} uses the average of P(Yes) over $N$ rollouts.
Inference passes count total forward passes per sample. {\bf Critique-Last} gives the best overall performance, showing that adding critique at inference time can yield further gains.
However, Critique-First and Self-Critique-Conditioned underperform, likely because the Qwen3-VL model was not sufficiently trained on these formats to follow them reliably; increasing the scale of critique-augmented training data may improve their performance.}
\resizebox{\textwidth}{!}{
\begin{tabular}{@{}l c ccccc@{}}
\toprule[1.2pt]
\multirow{2}{*}{\textbf{Mode}} &
\multirow{2}{*}{\textbf{Passes}} &
\multicolumn{5}{c}{\textbf{Reward Modeling Accuracy}} \\
\cmidrule(lr){3-7}
 & &
 \textbf{Subject} & \textbf{Scene} & \textbf{Motion} & \textbf{Spatial} & \textbf{Camera} \\
\midrule
\multicolumn{7}{@{}l}{\textit{VQAScore-based methods}} \\
Direct                     & $1$  & 89.8 & 91.0 & 89.1 & 91.2 & 88.1 \\
Critique-First             & $1$  & 79.1 & 72.9 & 80.7 & 90.7 & 74.6 \\
Critique-Last              & $1$  & 89.5 & 89.6 & 93.2 & 94.9 & 86.0 \\
Self-Critique-Conditioned  & $2$  & 80.4 & 78.1 & 83.3 & 90.0 & 81.7 \\
\midrule
\multicolumn{7}{@{}l}{\textit{Self-Consistency using 5 rollouts}} \\
Direct                     & $5$  & 89.8 & \textbf{91.0} & 89.1 & 91.2 & 88.1 \\
Critique-First             & $5$  & 81.2 & 75.6 & 83.0 & 92.0 & 74.7 \\
Critique-Last              & $5$  & {\bf 91.9} & {\bf 91.0} & {\bf 92.6} & {\bf 93.0} & {\bf 89.7} \\
Self-Critique-Conditioned  & $10$ & 82.5 & 81.8 & 85.4 & {\bf 96.3} & 83.9 \\
\bottomrule[1.2pt]
\end{tabular}
}
\label{tab:reward_scaling}
\end{table*}

\subsection{Test-time Scaling for Reward Modeling}
While directly generating Yes/No is already effective for reward modeling, we study whether generating a critique alongside the answer can further improve performance.
We evaluate both VQAScore (probability of Yes) and Self-Consistency (ratio of Yes to No over 5 rollouts).

{\bf Results.} \autoref{tab:reward_scaling} shows that outputting the critique after Yes/No (Critique-Last) marginally outperforms direct VQAScore, suggesting that inference-time critique can yield further gains.

\begin{table*}[t!]
\centering
\caption{\small \textbf{Test-time scaling strategies for caption generation.} We compare parallel Best-of-$N$ strategies and sequential iterative strategies. Generation cost counts caption, critique, and revision calls (each normalized to cost $1$, with critique-based revision counted as cost $2$), while reward cost counts calls to the reward model. The final five columns report caption performance at different test-time scales ($N \in \{1, 4,8,16\}$). Vanilla Best-of-N Caption yields the best performance given the same inference budget.
While critique-based approaches show marginal improvement, we believe increasing the scale of critique-augmented training data to improve critique generation quality can make these approaches perform even better in future work.}
\resizebox{\textwidth}{!}{
\begin{tabular}{@{}l l c c cccc@{}}
\toprule[1.2pt]
\multirow{2}{*}{\textbf{Mode}} &
\multirow{2}{*}{\textbf{Strategy}} &
\multicolumn{2}{c}{\textbf{Inference Cost}} &
\multicolumn{4}{c}{\textbf{Result (BLEU-4)}} \\
\cmidrule(lr){3-4} \cmidrule(lr){5-8}
 &  & \textbf{Generation} & \textbf{Reward} &
 \textbf{N=1} &
 \textbf{N=4} & \textbf{N=8} & \textbf{N=16}   \\
\midrule
\multirow{5}{*}{\textbf{Parallel (Best-of-$N$)}}
 & Best-of-$N$ Caption              & $N$     & $N$ & \textbf{17.1} & {\bf 20.4} & {\bf 22.8} & {\bf 25.1}  \\
 & Best-of-$N$ Revision             & $1+N$   & $N$ & 16.2 & 18.9 & 20.3 & 21.2  \\
 & Best-of-$N$ Critique-then-Rev.   & $1+2N$  & $N$ & 15.5 & 18.5 & 20.1 & 21.0  \\
 & Best-of-$N$ Critique-based Rev.  & $1+2N$  & $N$ & {\bf 17.1} & 19.2 & 20.6 & 21.4  \\
 & Best-of-$N$ Critique             & $2+N$   & $N$ & 15.2 & 17.8 & 19.2 & 20.1  \\
\midrule
\multirow{3}{*}{\textbf{Sequential (Iterative)}}
 & Iterative Revision               & $1+N$   & $0$ & 15.9 & 16.9 & 17.3 & 17.5  \\
 & Iterative Critique-then-Rev.     & $1+2N$  & $0$ & 16.0 & 17.6 & 18.1 & 18.4  \\
 & Iterative Critique-based Rev.    & $1+2N$  & $0$ & 16.4 & 17.5 & 17.8 & 18.0 \\
\bottomrule[1.2pt]
\end{tabular}
}
\label{tab:testtime_scaling}
\end{table*}

\subsection{Test-time Scaling for Caption Generation}

We ablate different approaches to scale caption generation performance using inference-time compute.
For cost estimation, we separate {\it generation cost} (caption, critique, or revision) from {\it reward cost} (scoring a completed sequence), which is significantly cheaper due to KV caching.
We treat caption and critique generation as equal cost, and count a critique-based revision as $2\times$ generation cost because it requires producing a self-critique followed by a revised caption.

{\bf Parallel (best-of-$N$) approaches.}
These methods generate $N$ candidates independently and select the best one using the reward model. All candidates can be produced in parallel, so total cost scales directly with $N$.

\begin{enumerate}
    \item \textbf{Best-of-$N$ Caption.}
    Sample $N$ captions in parallel and select the highest-scoring one using the reward model.
    \begin{itemize}
        \item {\bf Generation cost:} $N$
        \item {\bf Reward cost:} $N$
    \end{itemize}

    \item \textbf{Best-of-$N$ Revision.}
    Generate one caption and produce $N$ independent revisions. The reward model scores all $N$ revised captions, and we select the best revision.
    \begin{itemize}
        \item {\bf Generation cost:} $1 + N$
        \item {\bf Reward cost:} $N$
    \end{itemize}

    \item \textbf{Best-of-$N$ Critique-then-Revision.}
    Generate one caption, produce $N$ critiques, and revise once per critique. Each revised caption is scored with the reward model to select the best one.
    \begin{itemize}
        \item {\bf Generation cost:} $1 + 2N$
        \item {\bf Reward cost:} $N$
    \end{itemize}

    \item \textbf{Best-of-$N$ Critique-based Revision.}
    Generate one caption and run $N$ critique-based revision steps (each step internally generates a self-critique and a revised caption). All revised candidates are scored to select the best one.
    \begin{itemize}
        \item {\bf Generation cost:} $1 + 2N$
        \item {\bf Reward cost:} $N$
    \end{itemize}

    \item \textbf{Best-of-$N$ Critique.}
    Generate one caption and produce $N$ critiques. We use the \emph{critique reward model} (trained specifically to evaluate critique quality) to score the $N$ critiques, select the best critique, and perform a single critique-based revision based on that critique.
    \begin{itemize}
        \item {\bf Generation cost:} $2 + N$
        \item {\bf Reward cost:} $N$
    \end{itemize}
\end{enumerate}

{\bf Sequential approaches.}
These methods refine a single caption over multiple dependent steps. Each iteration uses the previous output, so computation accumulates linearly with $N$.

\begin{enumerate}
    \item \textbf{Iterative Revision.}
    Generate one caption and apply up to $N$ sequential revisions.
    \begin{itemize}
        \item {\bf Generation cost:} $1 + N$
        \item {\bf Reward cost:} $0$
    \end{itemize}

    \item \textbf{Iterative Critique-then-Revision.}
    At each step, generate a critique of the current caption and revise accordingly, repeated for $N$ steps.
    \begin{itemize}
        \item {\bf Generation cost:} $1 + 2N$
        \item {\bf Reward cost:} $0$
    \end{itemize}

    \item \textbf{Iterative Critique-based Revision.}
    Perform $N$ critique-based revision steps, each internally generating a critique and a revision.
    \begin{itemize}
        \item {\bf Generation cost:} $1 + 2N$
        \item {\bf Reward cost:} $0$
    \end{itemize}
\end{enumerate}

{\bf Results.} 
\autoref{tab:testtime_scaling} compares these strategies across computational budgets ($N \in \{1,4,8,16\}$).
Vanilla best-of-$N$ caption achieves the strongest performance at $N \geq 4$, reaching 25.1 BLEU-4 at $N=16$, despite being the simplest parallel strategy.
Parallel methods consistently outperform sequential iterative approaches, which show diminishing returns as $N$ increases (e.g., iterative critique-based revision improves only from 16.4 to 18.0 across $N=1$ to $16$).
This highlights that reward-based selection over parallel candidates is more effective than sequential refinement without discriminative filtering.

{\bf Limitations.} We focus on pure parallel and sequential strategies and do not explore hybrid search procedures. We also do not incorporate reward-guided early stopping or pruning mechanisms (e.g., discarding low-scoring candidates in a beam-search–like process). These extensions are left for future work.

\section{Video Generation}
\label{sec:video_generation_details}
In this section, we report statistics of the $\sim$150K professional videos we curated from YouTube, summarize their genre distribution, show additional qualitative comparisons with zero-shot Wan, and include a human study on prompt following using 200 prompts generated from our test set.

\begin{figure*}[t!]
\centering
    \includegraphics[width=0.6\textwidth]{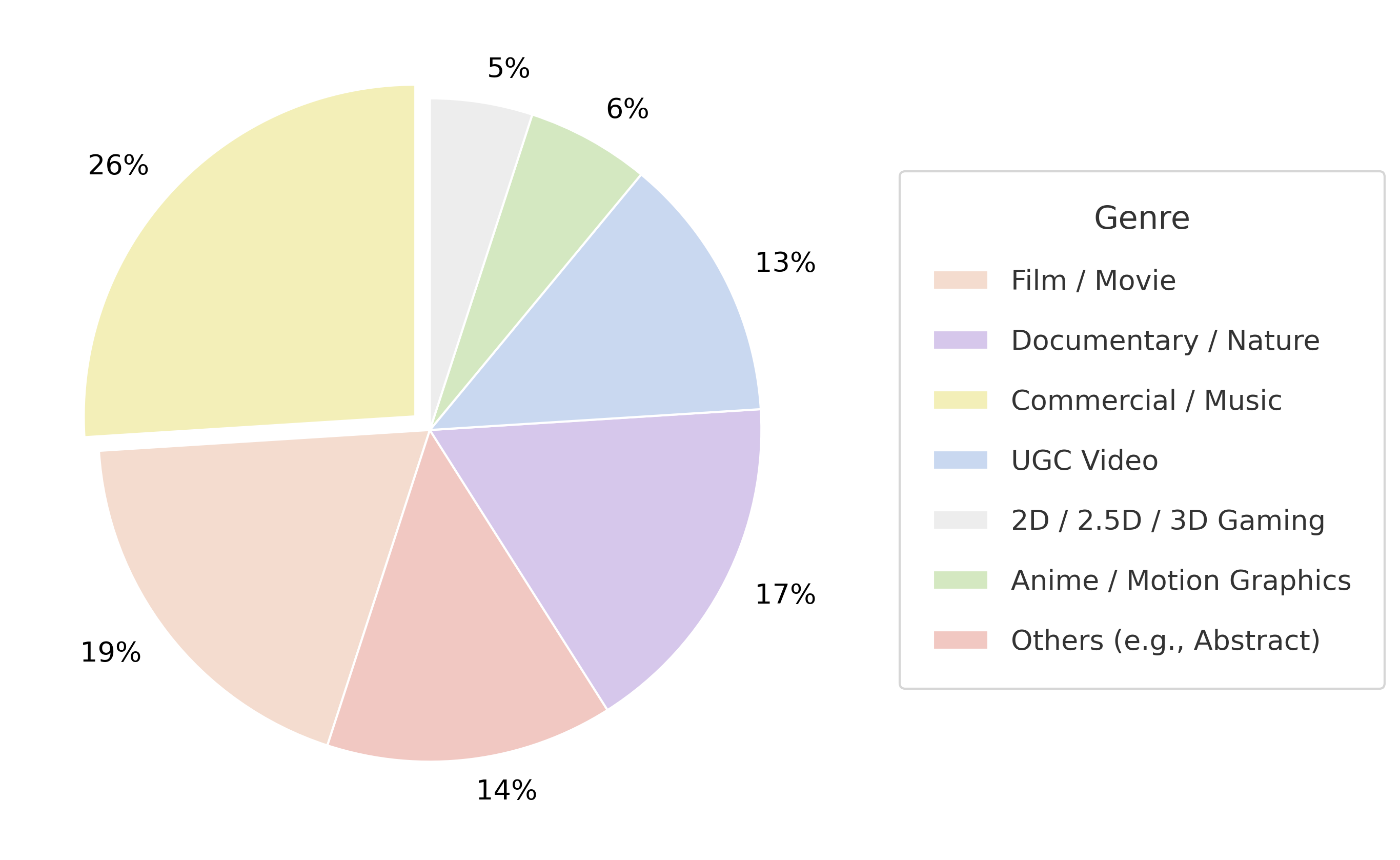}
    \caption{\small {\bf Genre distribution for $\sim$150k curated videos.}}
    \label{fig:video_genres}
\end{figure*}

{\bf Video statistics.} We hire our annotators to manually source high-quality and diverse YouTube channels spanning films, commercials, music videos, user-generated content (including vlogs, first-person GoPro footage, and etc.), documentary and nature videos (including drone shots, animals, and scenery), gaming footage (3D, 2.5D isometric, 2D side view, top-down, and etc.), and others (including dashcam, screen recordings, abstract visuals, and etc.). All videos follow the standard YouTube license for non-commercial use. We then apply TransNetV2~\cite{soucek2024transnet} to segment the videos and use a high threshold (0.15) to discard any clips with potential hard cuts or soft transitions, resulting in $\sim$150K videos with an average duration of $4.6$ seconds (we discard clips shorter than one second). The distribution of genres (measured by channel type) is shown in \autoref{fig:video_genres}.

{\bf Captioning.} We compare two approaches to caption these videos: (1) applying the zero-shot Qwen3-VL-8B-Instruct model (without any training on our dataset) to directly generate the five captions given our task instruction, and (2) using our own Qwen3-VL-8B-Instruct model fine-tuned via SFT. Because the five captions are too long to append naively (Wan2.2 accepts only 512 tokens or roughly up to 400 words), and because they contain substantial overlap (e.g., motion captions often also describe the subjects in detail), we merge them into a single summary caption that combines all subject, scene, motion, spatial, and camera information. This merging is done using ChatGPT with the prompt attached below.

\begin{lstlisting}[label={lst:merge_prompt},keywordstyle=\ttfamily]
Please merge the following five captions into a single, comprehensive caption that describes the video completely without any redundancy.

Caption Types:
1. Subject: Describes the subjects/people in the video
2. Scene: Describes the scene composition and environment
3. Motion: Describes the movement and dynamics of subjects
4. Spatial: Describes the spatial relationships and framing
5. Camera: Describes camera movements and framing choices

Input Captions:
{captions}

Instructions:
1. Use the SPATIAL caption as your BASE structure - it provides the core visual description and framing
2. Merge MOTION and CAMERA captions into the spatial description to create a temporally coherent narrative that describes how things change over time
3. Add information from SUBJECT and SCENE captions ONLY if they contain unique details not already covered in the Spatial caption
4. Eliminate ALL redundant information - if the same detail appears in multiple captions, mention it only ONCE
5. Preserve the EXACT wording from the original captions - do NOT paraphrase
6. When describing temporal changes, integrate motion and camera movements in chronological order to show how the scene evolves
7. CRITICAL: Every unique detail from all five captions must appear in the final merged caption - nothing should be omitted
8. Do NOT add any information not present in the original captions
9. Return only the merged caption without any additional text or formatting

Goal: A single, temporally coherent caption based on the Spatial description, with Motion and Camera information merged chronologically, and Subject/Scene details added only when they provide new information.
\end{lstlisting}

{\bf Finetuning text-to-video generation models.} We follow the default hyperparameters to finetune Wan2.2, as shown in \autoref{tab:wan_hyperparams}.

\begin{figure*}[t!]
\centering
    \includegraphics[width=0.85\textwidth]{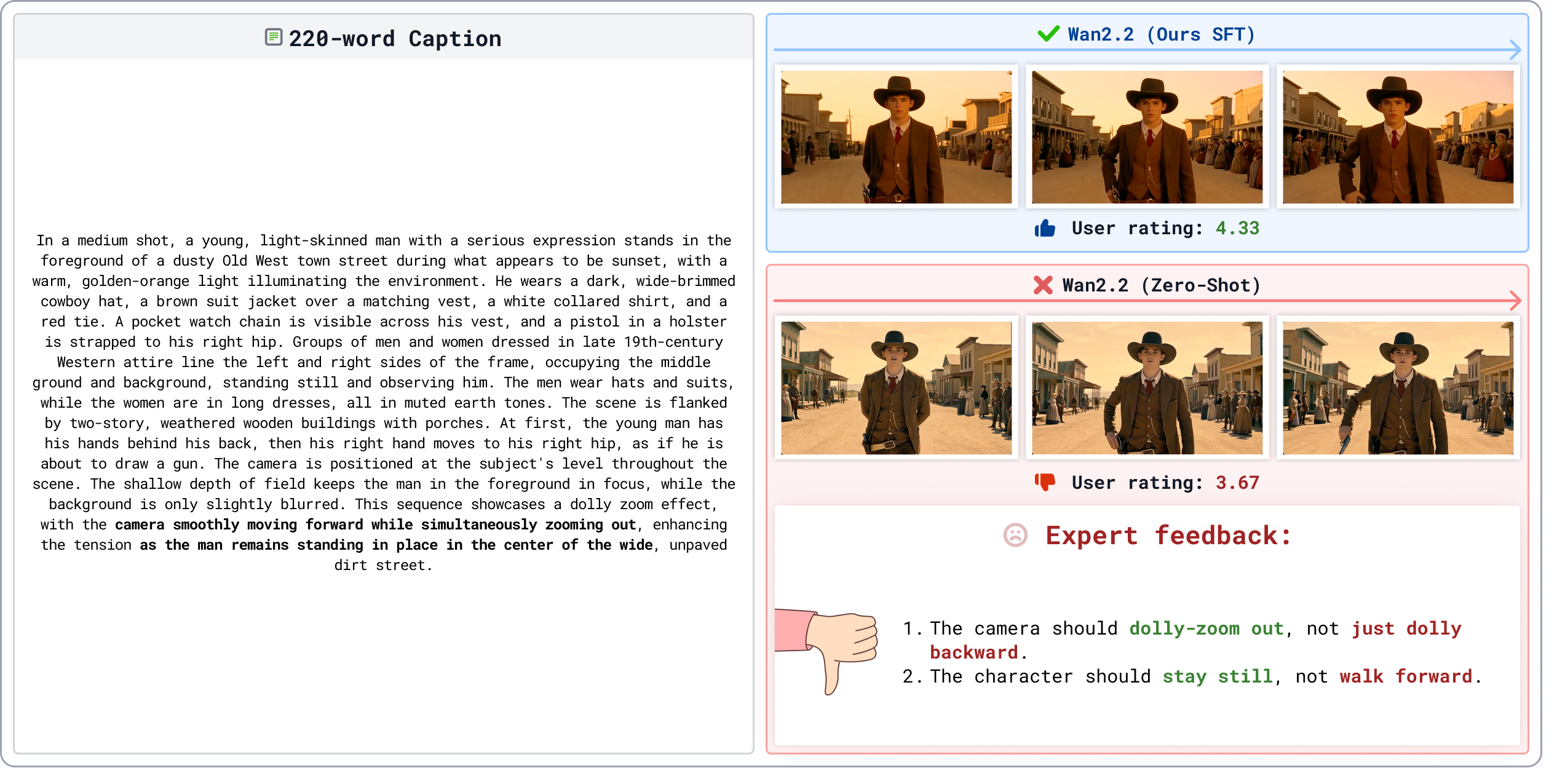}
    \caption{\bf Video generation example: dolly-zoom out.
    }\label{fig:dolly_zoom}
\end{figure*}

\begin{figure*}[t!]
\centering
    \includegraphics[width=0.85\textwidth]{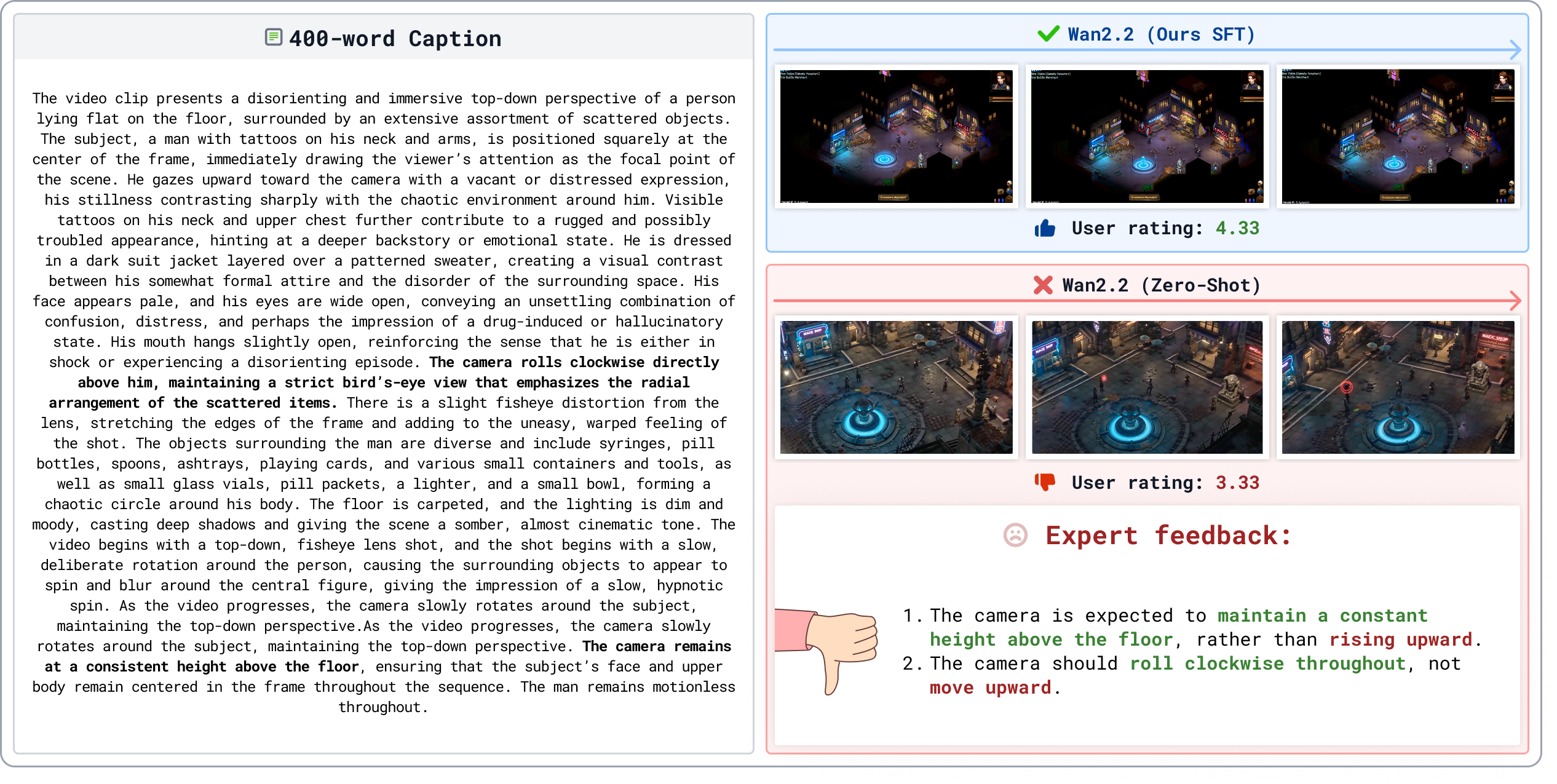}
    \caption{\bf Video generation example: isometric (2.5D) game perspective.
    }\label{fig:isometric}
\end{figure*}

\begin{figure*}[t!]
\centering
\includegraphics[width=0.85\textwidth]{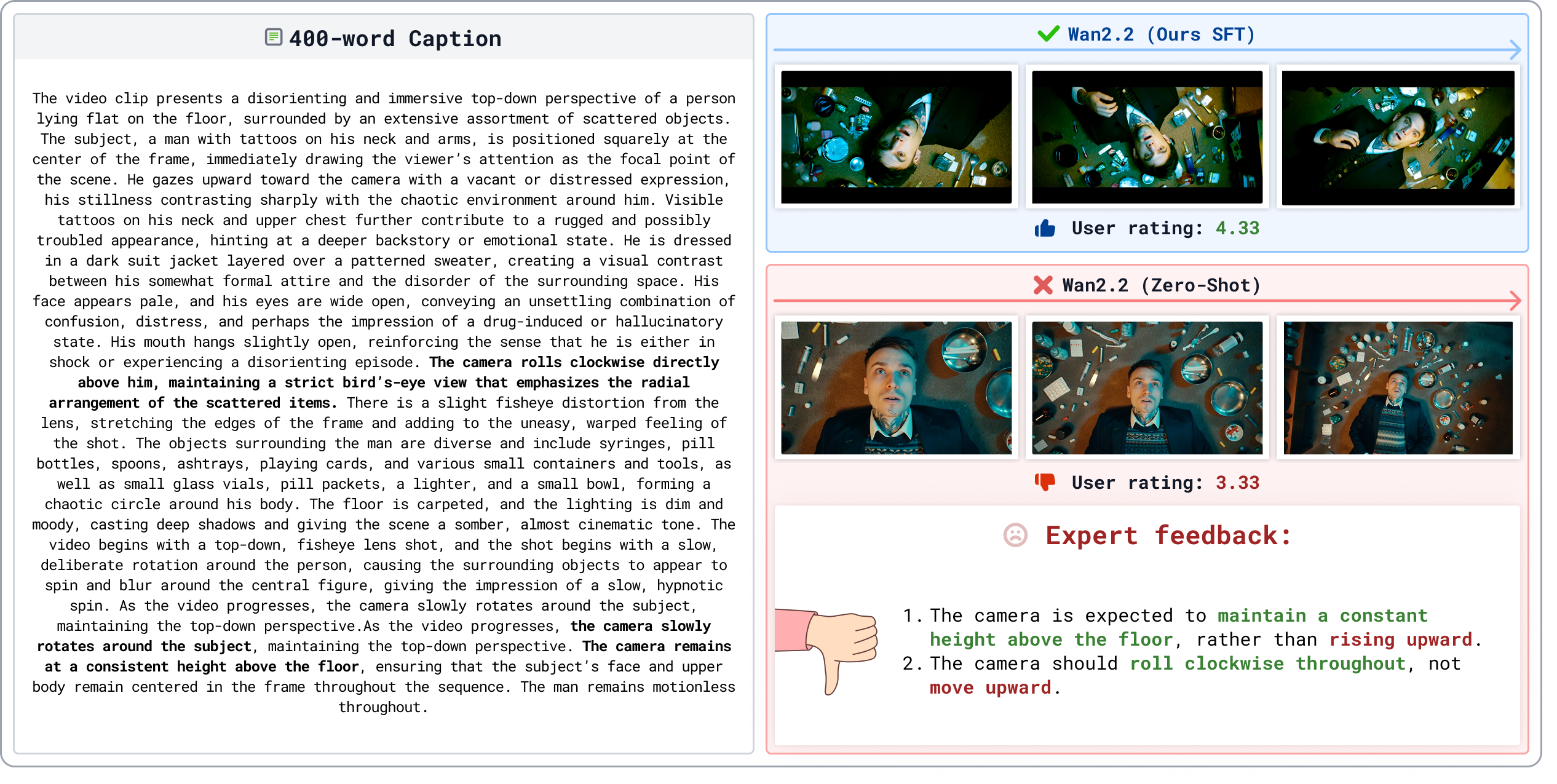}
    \caption{\bf Video generation example: clockwise roll.
    }\label{fig:rolling_cw}
\end{figure*}

\begin{figure*}[t!]
\centering
\includegraphics[width=0.85\textwidth]{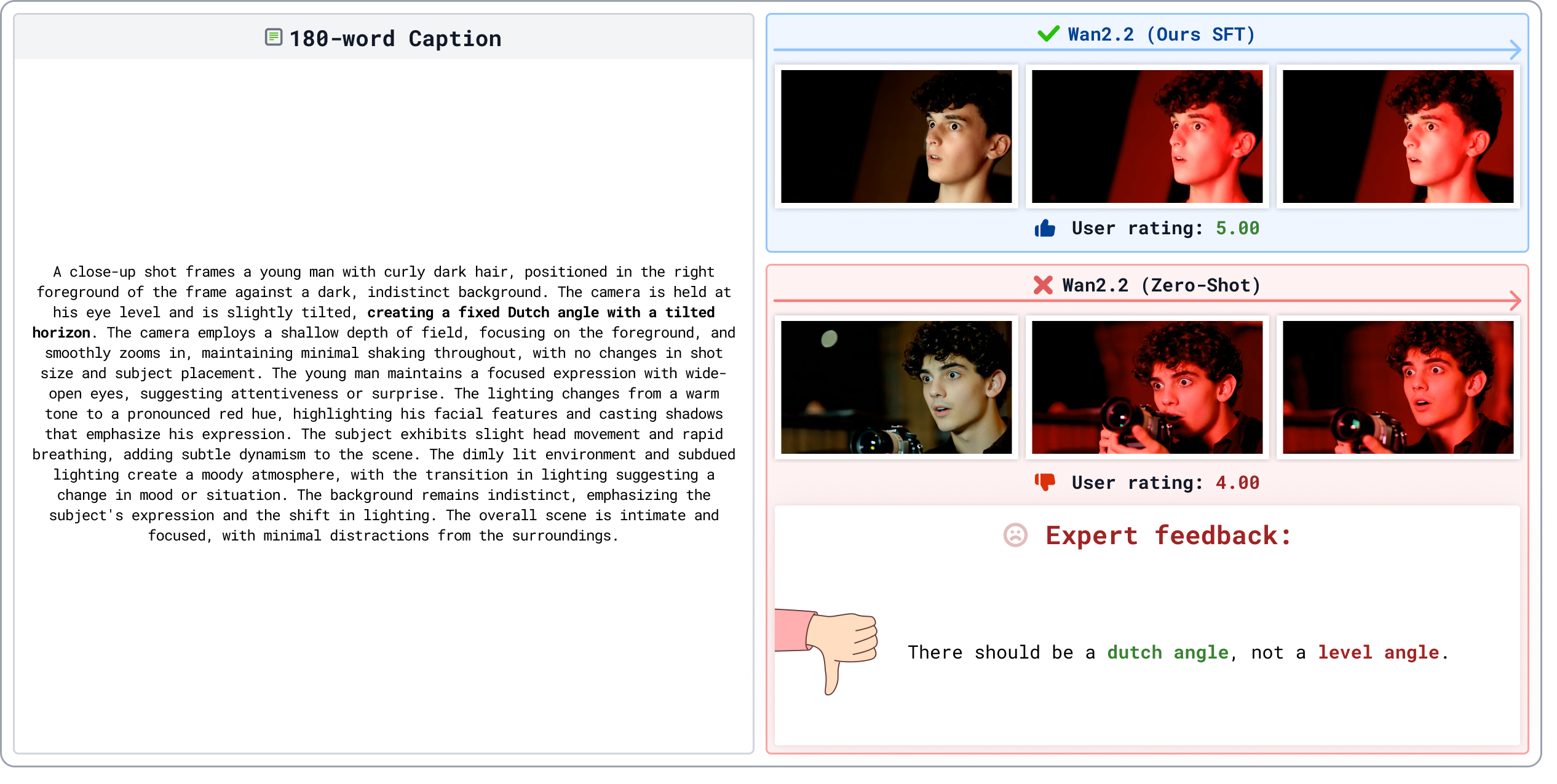}
    \caption{\bf Video generation example: Dutch angle.
    }\label{fig:dutch_angle}
\end{figure*}

\begin{figure*}[t!]
\centering
\includegraphics[width=0.85\textwidth]{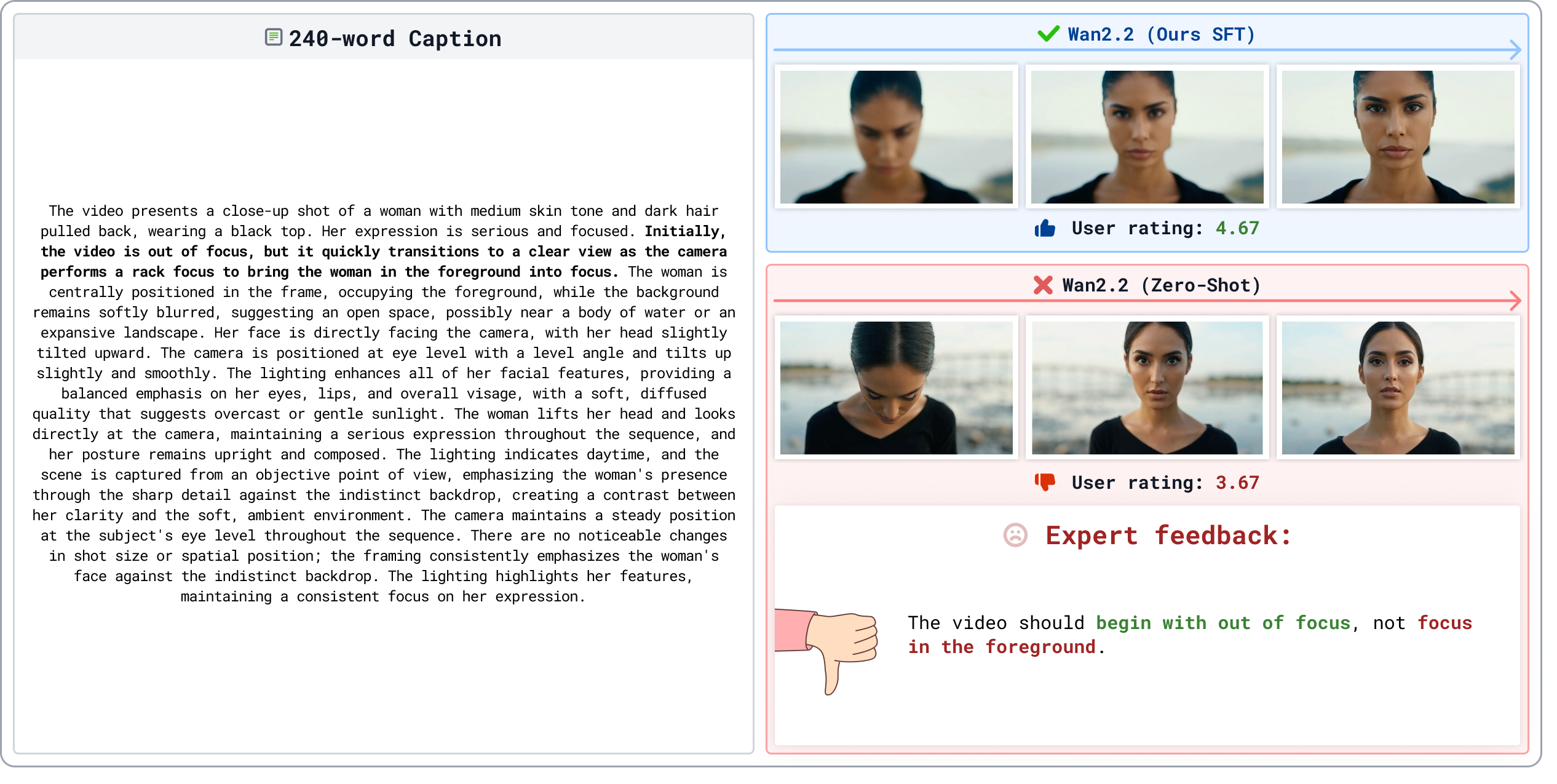}
    \caption{\bf Video generation example: rack focus.
    }\label{fig:rack_focus}
\end{figure*}

\begin{figure*}[t!]
\centering
\includegraphics[width=0.85\textwidth]{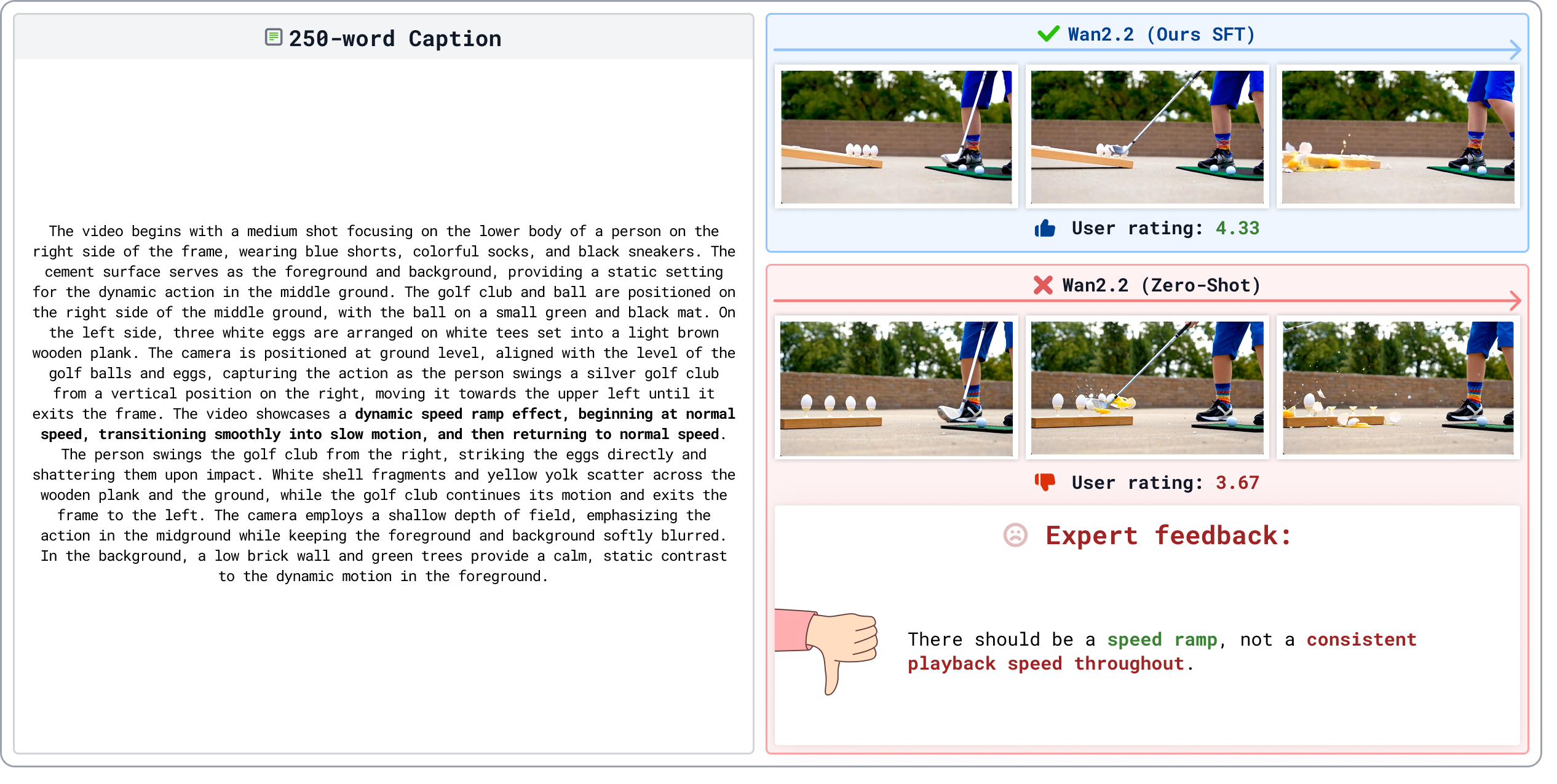}
    \caption{\bf Video generation example: speed ramp.
    }\label{fig:speed_ramp}
\end{figure*}

\begin{figure*}[t!]
\centering
\includegraphics[width=0.85\textwidth]{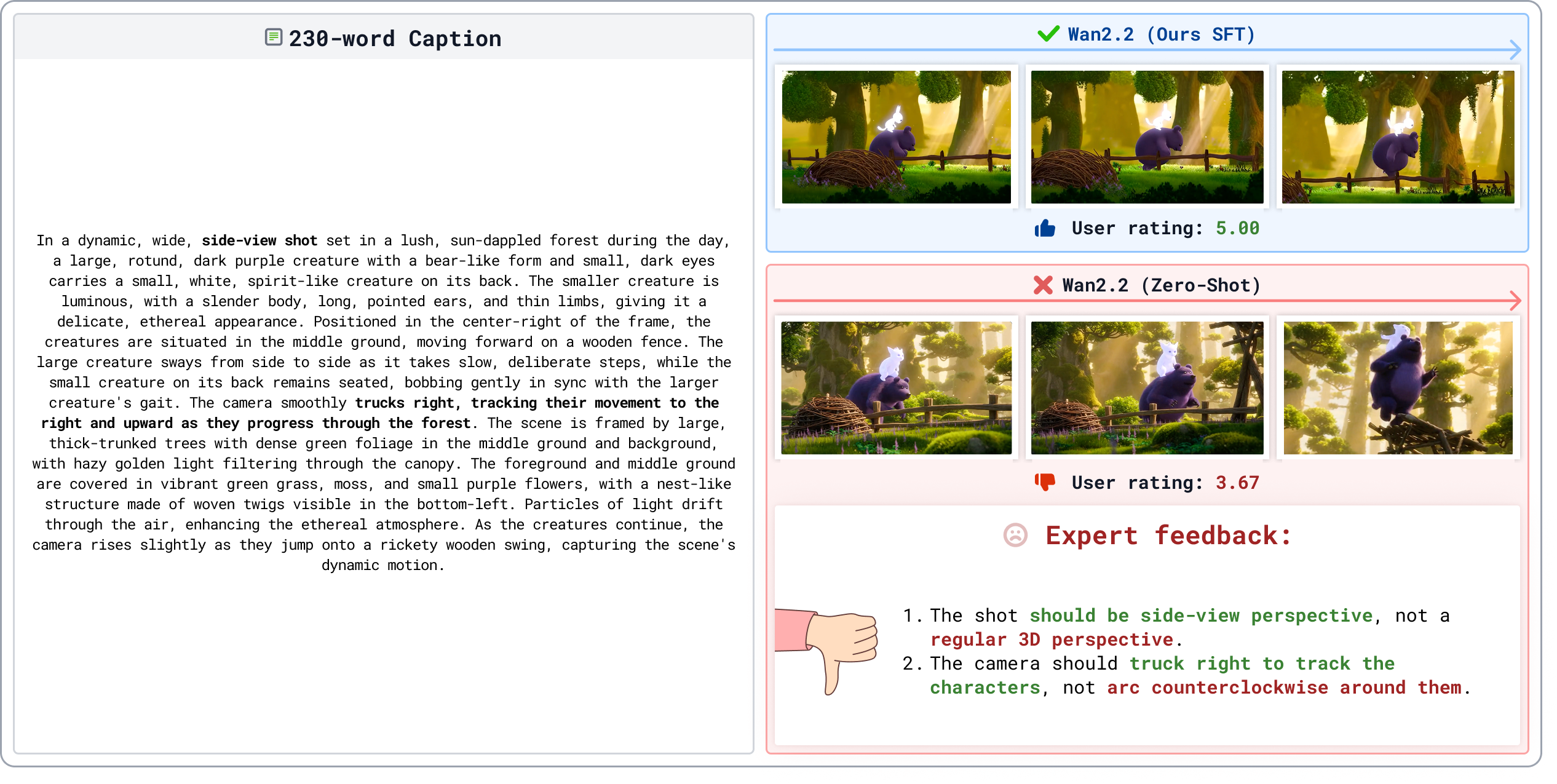}
    \caption{\bf Video generation example: side-view game perspective.
    }\label{fig:side_view}
\end{figure*}

\begin{figure*}[t!]
\centering
\includegraphics[width=0.85\textwidth]{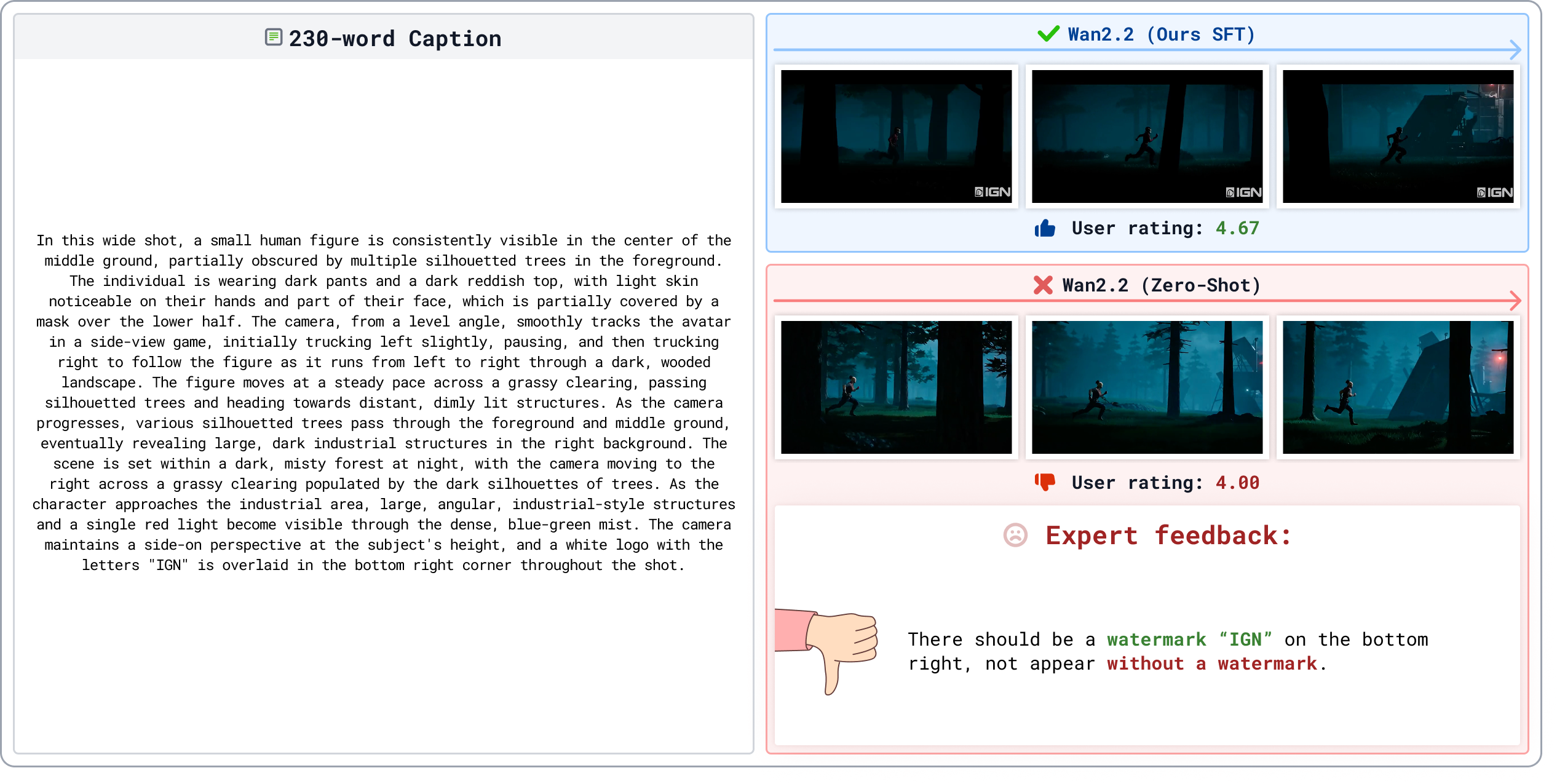}
    \caption{\bf Video generation example: watermark addition.
    }\label{fig:watermark}
\end{figure*}

\begin{figure*}[t!]
\centering
\includegraphics[width=0.85\textwidth]{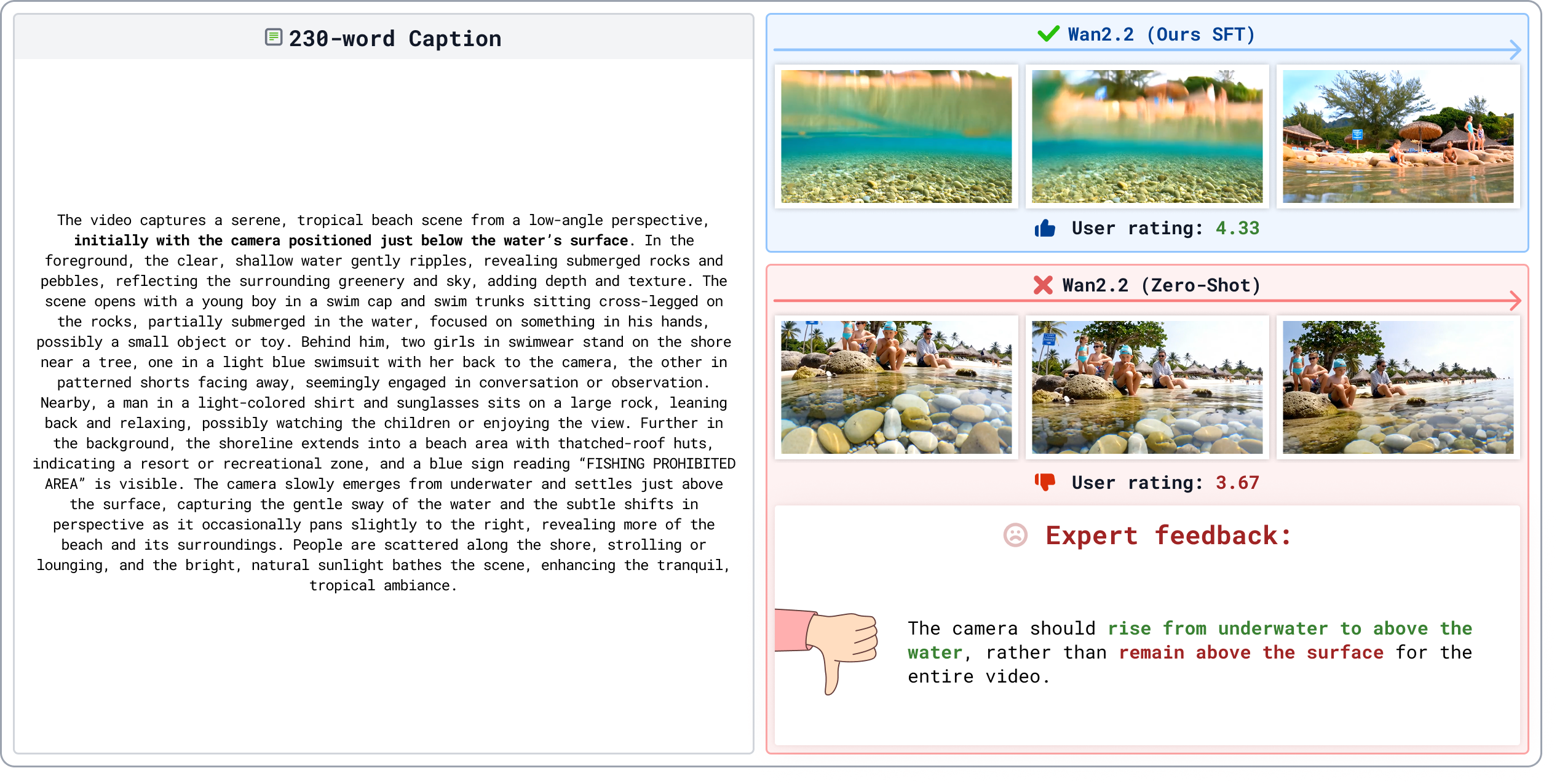}
    \caption{\bf Video generation example: height change from underwater to above water.
    }\label{fig:waterlevels}
\end{figure*}

\begin{figure*}[t!]
\centering
\includegraphics[width=0.85\textwidth]{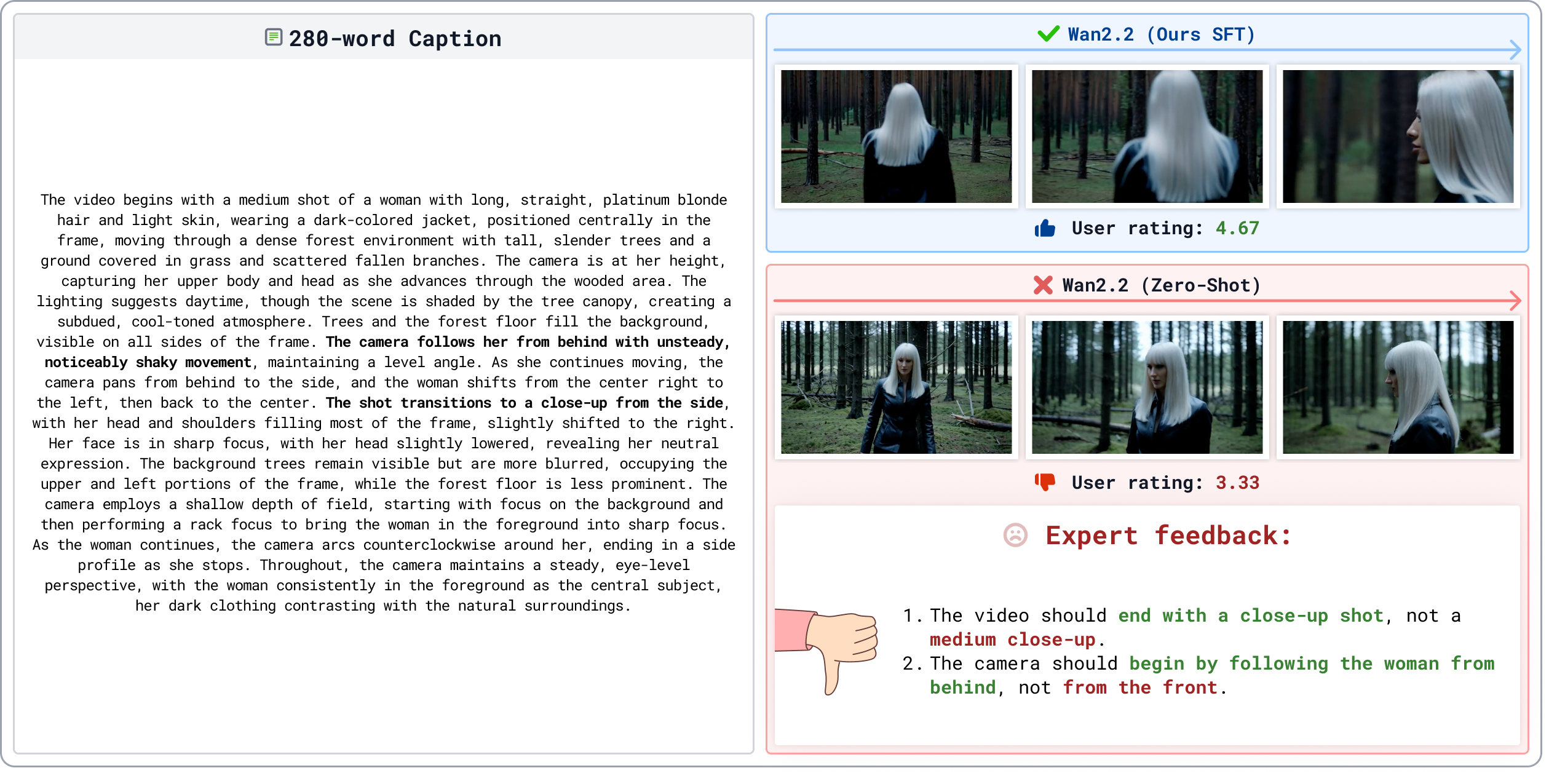}
    \caption{\bf Video generation example: shot size change from medium to close-up.
    }\label{fig:shot_size}
\end{figure*}

\begin{figure*}[t!]
\centering
\includegraphics[width=0.85\textwidth]{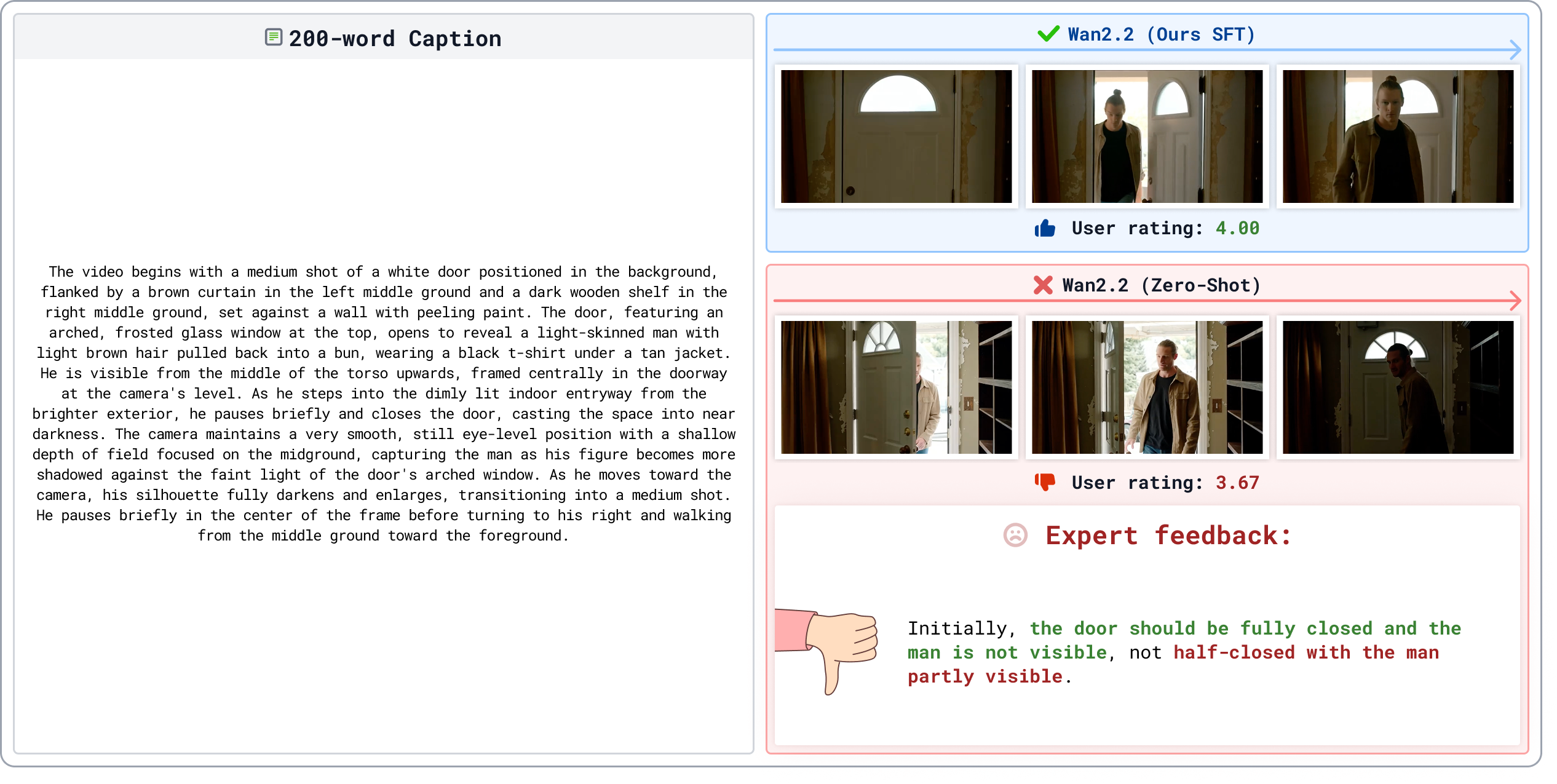}
    \caption{\bf Video generation example: revealing shot.
    }\label{fig:revealing_shot}
\end{figure*}

{\bf Human study.} We compare three Wan2.2-T2V-14B variants: {\bf Base}, {\bf ZS-Caption} (Wan2.2 finetuned using captions generated by zero-shot Qwen3-VL-8B-Instruct), and {\bf SFT-Caption} (Wan2.2 finetuned using captions from our SFT-trained Qwen3-VL-8B-Instruct). We evaluate all three models on 200 prompts summarized from the five captions using GPT-4o. As shown in \autoref{tab:human_preference}, the SFT-Caption variant achieves the best prompt following, measured using 1-to-5 human Likert scores averaged across three expert annotators. We attach qualitative samples in \autoref{fig:dolly_zoom} (dolly-zoom out), \autoref{fig:isometric} (isometric 2.5D gaming perspective), \autoref{fig:rolling_cw} (camera rolling), \autoref{fig:dutch_angle} (Dutch angle), \autoref{fig:rack_focus} (rack focus), \autoref{fig:speed_ramp} (speed ramping effect), \autoref{fig:side_view} (side-view gaming perspective), \autoref{fig:watermark} (watermark overlays), \autoref{fig:waterlevels} (camera rising from underwater to above water), \autoref{fig:shot_size} (changing shot size), and \autoref{fig:revealing_shot} (revealing shot).

\begin{table}[t]
\centering
\renewcommand{\arraystretch}{1.15}
\caption{\small \textbf{Human preference study.} Models fine-tuned on higher-quality captions receive higher average human Likert scores.}
\label{tab:human_preference}
\begin{tabular}{lc}
\toprule[1.5pt]
\textbf{Model} & \textbf{Avg. Human Likert Score} \\
\midrule
Base         & $3.23 \pm 0.78$ \\
ZS-Caption   & $3.64 \pm 0.54$ \\
SFT-Caption  & $3.95 \pm 0.45$ \\
\bottomrule[1.5pt]
\end{tabular}
\end{table}

\begin{table}[t]
\centering
\caption{{\bf Training hyperparameters} for Wan2.2 fine-tuning.}
\begin{tabular}{l|c c}
\toprule[1.5pt]
\textbf{Hyperparameter} & \textbf{High-Noise Stage} & \textbf{Low-Noise Stage} \\
\midrule
timestep range & 0 $\rightarrow$ 0.417 & 0.417 $\rightarrow$ 1 \\
resolution & \multicolumn{2}{c}{832$\times$480} \\
num frames & \multicolumn{2}{c}{49} \\
learning rate & \multicolumn{2}{c}{1e-5} \\
lr schedule & \multicolumn{2}{c}{cosine decay} \\
epochs & \multicolumn{2}{c}{2} \\
optimizer & \multicolumn{2}{c}{AdamW} \\
DeepSpeed stage & \multicolumn{2}{c}{ZeRO-2} \\
\bottomrule[1.5pt]
\end{tabular}
\label{tab:wan_hyperparams}
\end{table}

\clearpage
\section{Human Captioning Policy}
\label{sec:captioning_policy}

\subsection{Captioning Overview}
\begin{enumerate}
    \item \textbf{Subject Description}\\
    Includes: Subject Type, Subject Attributes, Relationships Between Multiple Subjects, Subject Transitions\\
    Example: ``The camera first shows a man in a black shirt with a brown beard standing to the right of a tree, facing the camera. The camera then moves left, revealing a young boy in red looking at the tree.''
    \item \textbf{Subject Motion \& Dynamics}\\
    Includes: Subject Action, Subject Dynamics, Subject-Object Interactions, Subject-Subject Interactions, Group Action\\
    Example: ``The video shows two teams playing soccer. A player from the red team sprints across the field and collides with an opponent in blue, causing both to fall.''
    \item \textbf{Scene Composition \& Dynamics}\\
    Includes: Point-of-View, Overlays, Scene Type, Scene Attributes, Scene Dynamics\\
    Example: ``A first-person perspective of walking on a sunny beach, with ocean waves crashing onto the shore.''
    \item \textbf{Spatial Framing \& Dynamics}\\
    Includes: Shot Size, Spatial Position, Spatial Motion\\
    Example: ``The video opens with a medium shot of a man positioned on the left side of the frame. As he walks forward toward the camera, moving from the midground to the foreground, the shot gradually transitions into a close-up of his face at the center of the frame.''
    \item \textbf{Camera Framing \& Dynamics}\\
    Includes: Video Playback Speed, Lens Distortion, Camera Height, Angle, Focus, Depth of Field, Camera Steadiness, Camera Movement\\
    Example: ``The video is in slow motion, capturing a low-angle, ground-level shot of a skateboarder performing a trick. A fisheye lens introduces barrel distortion, while deep focus ensures sharp clarity throughout the frame. The camera tracks the skateboarder from the side, trucking right, with an unsteady, handheld-like motion.''
\end{enumerate}

\subsection{High-Level Captioning Policy}
Subject Motion and Spatial Framing are dependent on the Subject/Scene descriptions.

\begin{itemize}
    \item \textbf{Focus on the Visual:} Ensure your captions accurately reflect the visual content by describing only what is visible in the frame. If speculation is necessary (e.g., about events outside the frame or the subject’s thoughts), use phrases like ``it could be that'' or ``this might suggest that'' to indicate uncertainty. Please ignore the audio information from the video.
    \item \textbf{Conciseness:} Keep descriptions clear and to the point, avoiding unnecessary details.
    \item \textbf{Objectivity:} Avoid personal opinions or interpretations in your descriptions.
    \item \textbf{Order Matters:} Describe subjects and events in temporal order when applicable, and prioritize the most important elements first for clarity.
    \item \textbf{Clarity:} Your caption should provide enough context so someone who hasn't seen the video understands what's happening.
\end{itemize}

\subsection{Subject Description Policy}
\textbf{Instructions for Subject Description}\\
Provide a concise yet informative description of the subjects in this video.

\begin{enumerate}
    \item \textbf{Subject Types:}
    \begin{itemize}
        \item Specify the subject's type precisely (e.g., ``man,'' ``woman,'' ``dog,'' ``car,'' ``tree''). Avoid vague terms like ``thing'' or ``item.''
        \item If the subject type is ambiguous, provide your best judgment and explain your reasoning.
    \end{itemize}

    \item \textbf{Visual Attributes:}\\
    Describe the subject's key visual characteristics using specific and descriptive language. Consider the following aspects where relevant:
    \begin{itemize}
        \item \textbf{Appearance:}\\
        People: Include details like clothing (including colors and style), hairstyle, facial hair, age (if discernible), gender, ethnicity (if relevant and clear), facial expression, and so on.\\
        Objects: Describe their color, material, shape, and any distinguishing marks (e.g., ``smooth,'' ``rough,'' ``furry,'' ``metallic'', ``black'', ``red'', etc.)
        \item \textbf{Pose/Orientation:}\\
        Describe the subject's posture and orientation within the frame (e.g., ``standing,'' ``sitting,'' ``lying down,'' ``walking,'' ``facing left,'' ``arms raised,'' ``facing the camera''). Pay particular attention to objects not in their usual state (e.g., a tilted lamp, a book lying open face down).
    \end{itemize}

    \item \textbf{How to Refer to Multiple Subjects:}\\
    Sometimes there's more than one important person or thing in a picture. When that happens, make sure it's clear which one you're talking about. Here are some ways to do that:
    \begin{itemize}
        \item \textbf{Type:} The simplest way to refer to a subject is by its category, e.g., ``the man,'' ``the dog,'' or ``the tree.''
        \item \textbf{Attributes:} If multiple subjects belong to the same category, use distinguishing features:
        \begin{itemize}
            \item \textbf{Unique Appearance:} Highlight distinct traits, such as ``the woman in the red dress,'' ``the man with the beard,'' ``the blue car,'' or ``the largest tree.''
            \item \textbf{Location:} Specify position within the scene, e.g., ``the man on the left,'' ``the dog in the background,'' ``the car in the midground,'' or ``the building in the middle.''
            \item \textbf{Action:} Describe their activity, e.g., ``the person walking,'' ``the child playing with a ball,'' ``the bird flying,'' or ``the cat sitting on the windowsill.''
            \item \textbf{Relationship to Each Other:} For example, ``the man next to the woman,'' (spatial relationship), ``the first man that enters the frame'' (temporal relationship), or ``the two cars parked side by side.''
        \end{itemize}
        \item \textbf{Combining Descriptions:} For maximum clarity, combine multiple attributes. Example:
        \begin{itemize}
            \item ``The woman in the red dress on the left, talking on her phone.''
            \item ``The dog in the background, running toward the ball.''
        \end{itemize}
        The key is to give enough information so that anyone reading your description can easily tell which person or object you are referring to. Don't assume we know! The more detail you give, the better.
    \end{itemize}

    \item \textbf{Order Matters When Describing Multiple Subjects:}\\
    When describing multiple subjects, the order in which you mention them matters. Prioritize elements based on their importance in the video, ensuring a natural and logical flow.
    \begin{itemize}
        \item \textbf{Temporal Order:} If the scene unfolds over time, describe subjects in the order they appear. For example, ``First, the car speeds past, then the cyclist enters the frame.''
        \item \textbf{Prominence-Based Order:} If temporal order isn’t relevant, start with the most visually striking or important subject before moving on to less prominent ones. For example, ``The video shows a bright red sports car in the foreground. In the midground, a blue sedan is right behind it.''
    \end{itemize}
\end{enumerate}

\textbf{Subject Description Policy for Annotators}\\
Use the following logic:

\begin{enumerate}
    \item Are there Shot Transitions (cuts or soft transitions)?\\
    If Yes $\rightarrow$ No need for description (ends here).\\
    If No $\rightarrow$ Proceed to the next step.

    \item Are there prominent subjects that strongly draw the viewer's attention?\\
    If No $\rightarrow$ Specify the type of shot and explain why there is no main subject (e.g, a scenery shot of landscape or cityscape; an establishing shot that sets the scene; a video featuring abstract visuals) (ends here).\\
    If Yes $\rightarrow$ Proceed to the next step.

    \item \textbf{Determining How to Describe Subjects in a Video:}\\
Before describing the subjects in a video, determine whether it features a single subject or a small group of subjects, or if it involves a complex subject scenario with multiple subjects or transitions. Identifying the scenario ensures a more fluent and logical description.
\begin{itemize}
    \item \textbf{Single or Small Group of Subjects:}\\
    If the video consistently features one subject or a small group of subjects, follow the ``Instructions for Subject Description.'' When describing a small group, clearly distinguish between them by following ``How to Refer to Multiple Subject.''

    \item \textbf{Complex Subject Scenarios}\\
    If the video involves multiple subjects or complex transitions between them, please point out if the below scenarios occur:
    \begin{itemize}
        \item \textbf{Determine the Primary Focus:}\\
        If there is a single clear main subject, describe the main subject in detail, explaining why this subject is the focus while others are not. Include relevant details such as appearance, actions, and positioning that make the subject stand out. Provide a less detailed overview of secondary subjects, mentioning only their general presence, or relationship to the main subject.\\
        If there are multiple subjects in focus, describe subjects in prominence-based order (e.g., humans before objects).\\
        If there is no clear main subject, give a brief overview of all subjects without excessive detail.
        \item \textbf{Identify Subject Transitions:}\\
        If the focus shifts between subjects, specify the type of transition:
        \begin{itemize}
            \item Subject Revealing (a new subject enters the frame)
            \item Subject Disappearing (a subject exits or is no longer visible)
            \item Subject Switching (the focus shifts from one subject to another through rack focus or other camera movements)
            \item Other Complex Changes (subjects alternate focus multiple times)
        \end{itemize}
        And explain how the transition occurs (through subject movement or camera movement).
    \end{itemize}
    Identifying these scenarios ensures a more fluent and coherent description. For instance, in cases of subject switching, describe subjects in temporal order by following ``Order Matters When Describing Multiple Subjects.''
\end{itemize}
\end{enumerate}

\subsection{Subject Motion and Dynamics Policy}
\textbf{Instructions for Subject Motion \& Dynamics Description}\\
Provide a concise yet informative description of the subject motion in this video. Please note that order matters. If multiple actions occur, present them in chronological order (e.g., ``The bird first takes flight, then soars in a circle, and finally lands on a branch'').

\textbf{Subject Motion Description Policy for Annotators}\\
Use the following logic:

\begin{enumerate}
    \item Are there Shot Transitions (cuts or soft transitions)?\\
    If Yes $\rightarrow$ No need for description (ends here).\\
    If No $\rightarrow$ Proceed to the next step.

    \item Are there prominent subjects that strongly draw the viewer's attention?\\
    If No $\rightarrow$ No need for description (ends here).\\
    If Yes $\rightarrow$ Proceed to the next step.

    \item Are the subjects performing any activities or undergoing dynamic changes?\\
    If Yes $\rightarrow$ 
    \begin{itemize}
        \item \textbf{Subject Actions and Dynamics:} Describe the actions and dynamic changes of individual subjects. Be specific about the manner of movement. Examples:
        \begin{itemize}
            \item ``A runner sprints across the finish line.'' (Instead of: ``A person is running.'')
            \item ``A hummingbird hovers delicately, wings beating rapidly as it sips nectar from a flower.'' (Instead of: ``A bird is flying.'')
            \item ``A caterpillar slowly inches its way along a leaf.'' (Instead of: ``An insect is moving.'')
            \item ``A time-lapse shows a sunflower turning its head to follow the sun across the sky.'' (Instead of: ``A plant is rotating.'')
            \item ``A seed sprouts, sending a root down and a sprout up.'' (Instead of ``A seed is growing'')
        \end{itemize}
    \end{itemize}

    \item Does the motion involve subjects interacting with objects?\\
    If Yes $\rightarrow$
    \begin{itemize}
        \item \textbf{Subject-Object Interactions:} Describe the interaction between a subject and an object. Be specific about the type of interaction and the object involved. Detail the effect of the interaction if relevant. Examples:
        \begin{itemize}
            \item ``A chef flips an omelet in a pan.'' (Instead of: ``A person is using a pan.'')
            \item ``A dog fetches a tennis ball thrown by its owner.'' (Instead of: ``A dog is playing.'')
            \item ``A construction worker operates a jackhammer, breaking up the pavement.'' (Instead of: ``A person is working.'')
            \item ``A car collides with a traffic sign, bending it at a sharp angle.'' (Instead of: ``A car crashed.'')
        \end{itemize}
    \end{itemize}

    \item Does the motion involve subjects interacting with other subjects?\\
    If Yes $\rightarrow$
    \begin{itemize}
        \item \textbf{Subject-Subject Interactions:} Highlight interactions between different subjects. Describe the nature of the interaction and the relative movements of the subjects. Examples:
        \begin{itemize}
            \item ``Two boxers exchange blows in the ring, circling each other cautiously.'' (Instead of: ``People are fighting.'')
            \item ``A mother bird feeds worms to her chicks in the nest.'' (Instead of: ``Birds are together.'')
            \item ``Dancers perform a complex tango, their movements synchronized and graceful.'' (Instead of: ``People are dancing.'')
            \item ``A pride of lions hunts a zebra, surrounding it and closing in for the kill.'' (Instead of: ``Animals are interacting.'')
        \end{itemize}
    \end{itemize}

    \item Is there a group engaging in collective behavior or action?\\
    If Yes $\rightarrow$
    \begin{itemize}
        \item \textbf{Group Activity:} Summarize collective behaviors or actions of a group. Describe the overall movement and any coordinated actions within the group. Specify the type of group if relevant. Examples:
        \begin{itemize}
            \item ``A flock of geese flies in a V-formation across the horizon.'' (Instead of: ``Birds are flying.'')
            \item ``A crowd of protesters marches down the street, carrying signs and banners.'' (Instead of: ``People are walking.'')
            \item ``A swarm of bees buzzes around a hive.'' (Instead of: ``Insects are moving.'')
            \item ``A school of fish swims in unison, changing direction as one unit.'' (Instead of: ``Fish are swimming.'')
        \end{itemize}
    \end{itemize}
\end{enumerate}

\subsection{Scene Composition and Dynamics Policy}
\textbf{Instructions for Scene Composition \& Dynamics Description}\\
Provide a concise yet informative description of the overall scene, including environment, setting, time of day, and notable visual elements. If there are subjects in this video, this scene description will also complement the subject descriptions, establishing where the subjects are and why they might be there. Your goal is to give enough context for understanding the setting while avoiding excessive detail.

\begin{enumerate}
    \item \textbf{Describe the Overlay Elements (If any exists)}\\
    Overlays (if applicable): Identify and describe any overlay elements present in the shot that are not part of the scene. These may include text or visuals such as titles, subtitles, captions, icons, watermarks, heads-up displays (HUD), or framing elements. Clearly specify that these are overlays (not part of the scene) and describe their content and placement.

    \item \textbf{Describe the Point of View (What Is the Context of the Shot?)}\\
    Point of View (POV) (if discernible): Describe how the scene is framed or captured when relevant. Examples:
    \begin{itemize}
        \item First-person: ``The camera follows a person's viewpoint as they walk through a dimly lit hallway.''
        \item Drone: ``Aerial footage shows a city skyline stretching toward the horizon.''
        \item Over-the-shoulder: ``A character is seen from behind, looking at a laptop screen in a dimly lit room.''
        \item Top-down: ``A top-down oblique view of the game, with pieces arranged mid-game.''
        \item Dashcam: ``A vehicle’s dashboard and windshield frame the road ahead, with headlights illuminating the wet pavement.''
        \item Objective/Neutral: ``The camera provides a neutral, detached view of the scene.'' (Use this when no specific POV is evident.)
    \end{itemize}

    \item \textbf{Describe the Setting (Where Does It Happen?)}\\
    Scene Type: Specify the general type of scene using precise and descriptive terms. Indicate whether it is indoors (e.g., ``a cozy living room,'' ``a modern office'') or outdoors (e.g., ``a bustling city street,'' ``a dense forest''). Avoid vague terms like ``place'' or ``area''—be as specific as possible while ensuring clarity.\\
    Good: ``A sunlit café with large windows and wooden tables.''\\
    Avoid: ``An indoor place.''

    \begin{itemize}
        \item \textbf{Visual Attributes:} Considering the following aspects where relevant:
        \item Location: If the setting is a well-known place, state it explicitly (e.g., ``Times Square,'' ``Grand Canyon,'' ``Tokyo subway station''). If the exact location is unclear, describe using its defining visual elements (e.g., ``a narrow alley with graffiti-covered walls,'' ``a vast desert with rolling dunes,'' ``a dimly lit space with metal walls,'' ``an open area with sand and sparse vegetation'').
        \item Time of Day (if discernible): Specify whether the scene occurs during the day, night, or a transitional period like sunset or dawn, if relevant.
        \item Architectural and Natural Features: Mention buildings, roads, vegetation, water bodies, or other landscape elements that structure the scene (e.g., ``a winding mountain path surrounded by tall pines,'' ``a bustling marketplace with food stalls and colorful banners'').
        \item Weather Conditions: If outdoors, describe weather effects (e.g., ``a rainy street with wet pavement reflecting city lights,'' ``a snowy mountain pass covered in thick fog'').
        \item Furniture and Props (for indoor scenes): Identify relevant furnishings that establish the setting (e.g., ``a wooden desk cluttered with books and a vintage lamp,'' ``a hospital room with a bed, medical monitors, and IV stands'').
        \item Style: If relevant, describe notable color schemes or stylistic choices (e.g., ``a monochromatic, grayscale environment,'' ``a vibrant and colorful carnival scene with neon lights'').
    \end{itemize}

    \item \textbf{Describe Any Movement or Changes in the Environment (Any Changes or Motion in The Scene?)}\\
    Describe any natural or human-made movement happening at the scene level:
    \begin{itemize}
        \item Natural Motion: e.g., ``Leaves sway in the wind,'' ``Waves crash against the shore,'' ``As the sun sets, it casts long shadows on the trees.''
        \item Man-Made Motion: e.g., ``Traffic moves steadily on the highway,'' ``A train passes in the distance,'' ``Factory workers operate machinery in the background.''
        \item Crowd \& Background Activity: e.g., ``Pedestrians walk along a busy street,'' ``A crowd cheers and waves hands,'' ``The office starts empty, but employees gradually arrive and take their seats.''
    \end{itemize}
    If the scene changes, describe how it happens in the order it appears:
    \begin{itemize}
        \item Time-Based Transitions: e.g., ``The shot begins during the day but transitions to nighttime.''
        \item Movement-Based Transitions: e.g., ``The shot begins with a view of a quiet street. Then, the camera pans to reveal a hidden alley behind the main street.''
    \end{itemize}
\end{enumerate}

\textbf{How to Refer to Multiple Scene Elements}\\
Use precise and concise language to refer to different elements within the scene. For example:
\begin{itemize}
    \item ``In the background, a mountain range is visible.''
    \item ``On the left side of the frame, there is a large tree.''
    \item ``A wide river runs through the center, with a bridge arching over it.''
\end{itemize}
Prioritize the most prominent and important aspects of the scene. Start with the overall setting, then move on to more specific details.

\textbf{Scene Composition \& Dynamics Description Policy for Annotators}\\
Use the following logic:

\begin{enumerate}
    \item Are there Shot Transitions (cuts or soft transitions)?\\
    If Yes $\rightarrow$ No need for description (ends here).\\
    If No $\rightarrow$ Proceed to the next step.

    \item Are there any Overlay Elements in this shot?\\
    If No $\rightarrow$ No need for description.\\
    If Yes $\rightarrow$ Describe the ``Overlay Elements'' following the instructions.

    \item Is the Point of View obvious in this shot?\\
    If No $\rightarrow$ An objective/neutral POV.\\
    If Yes $\rightarrow$ Describe the ``Point of View (Why and How Is the Shot Taken?)'' following the instructions.

    \item Are there any notable elements in the scene that should be described?\\
    If No $\rightarrow$ For example, an extreme-close-up shot with no scene details. (ends here)\\
    If Yes $\rightarrow$
    \begin{itemize}
        \item Describe the ``Setting (Where Does It Happen?)'' following the instructions.
        \item For complex scenes (e.g., those with many elements), clearly distinguish between multiple scene elements by following ``How to Refer to Multiple Scene Elements.''
    \end{itemize}

    \item Are there any changes or motion happening at the scene level?\\
    If No $\rightarrow$ No need for further description (ends here).\\
    If Yes $\rightarrow$
    \begin{itemize}
        \item Describe the ``Movement or Changes in the Environment'' following the instructions.
    \end{itemize}
\end{enumerate}

\subsection{Spatial Framing and Dynamics Policy}
\textbf{Instructions for Spatial Framing \& Dynamics Description}\\
Provide a concise yet informative description of how subjects and elements are framed within the scene, including their shot size, position, and movement within the scene or frame. Your goal is to describe the spatial composition and motion within the shot.

\textbf{Framing of Subjects (How Are They Positioned in the Shot?)}\\
\begin{enumerate}
    \item \textbf{Describe the Subject’s Shot Size}\\
    Specify the shot size based on how much of the subject is visible:
    \begin{itemize}
        \item Close-Up: ``A close-up shot of a human face.''
        \item Medium Shot: ``A medium shot of a man’s lower body.''
        \item Wide Shot: ``A wide shot of a man standing near the ocean.''
    \end{itemize}
    If the shot size does not follow a typical framing pattern or changes erratically throughout the shot, describe the overall framing instead of forcing a specific shot size. For example: ``The camera tracks a skateboarder in an unsteady manner, and it mostly captures the skateboarder’s lower body.''

    \item \textbf{Describe the Subject’s Position in the Scene}\\
    \textbf{Position within the Frame:} Describe the subject's approximate location within the frame (e.g., ``top-left corner,'' ``center,'' ``right of the frame'').\\
    Example: ``The person is in the bottom-left corner of the frame.''\\
    Example: ``The person is on the right of the frame.''
    \begin{itemize}
        \item \textbf{Depth within the Scene:} Describe the subject's placement in relation to the foreground, midground, or background.\\
        Example: ``In the foreground, a person is sitting in front of a computer.''
        \item \textbf{Position within the Scene:} Describe the subject’s physical placement in the scene.\\
        Example: ``The woman in the midground stands near a window, looking outside.''
        \item \textbf{Height Relative to the Camera:} Describe the subject’s vertical positioning relative to the camera:\\
        Example: ``The man is framed at eye level.''\\
        Example: ``A low-angle shot captures the person from below.''
    \end{itemize}
\end{enumerate}

\textbf{Framing of Scenery (How Is the Environment Captured?)}\\
For scenery-focused shots without a primary subject, describe how the scene is framed:

\begin{enumerate}
    \item \textbf{Describe the Scene’s Shot Size}\\
    Specify the type of shot capturing the environment, for example:
    \begin{itemize}
        \item Wide Shot: ``A wide shot of a mountain range stretching across the horizon.''
        \item Close-Up: ``A close-up of raindrops hitting a window.''
    \end{itemize}

    \item \textbf{Describe the Scene’s Spatial Composition}\\
    Spatial Positioning: specify where key elements appear within the frame (e.g., left, right, center, corners):
    \begin{itemize}
        \item ``A symmetrical shot of a hallway positioned at the center of the frame, leading toward a vanishing point.''
        \item ``A large tree stands in the left-bottom corner of the frame.''
        \item ``A streetlamp is visible on the right side of the frame.''
    \end{itemize}
    \textbf{Depth (Foreground, Midground, and Background Elements):}
    \begin{itemize}
        \item ``In the foreground, a bicycle is parked to the right against a fence, while in the background, skyscrapers rise against the sky.''
        \item ``The midground features a river cutting through the landscape.''
    \end{itemize}
\end{enumerate}

\textbf{Spatial Motion Within the Frame (How Do Subjects or Scene Elements Move?)}\\
If shot size or spatial position changes within the frame, describe how these transitions happen clearly, specifying both the initial and final state.

\begin{enumerate}
    \item \textbf{Changes in shot size and spatial position for Subjects}
    \begin{itemize}
        \item ``A medium shot of a man’s upper body near a doorway transitions into a close-up of his face as he walks toward the camera.''
        \item ``A woman walking from the background to the foreground transitions from a wide shot capturing both her and the street scenery to a medium shot focusing on her lower body.''
        \item ``A cyclist moves from the left to the right side of the frame, maintaining a full shot throughout.''
        \item ``A full-body shot of a child at eye level shifts as the camera tilts upward, reframing them from a low angle looking up.''
        \item ``A wide shot captures a person near a park bench, who then walks diagonally from the bottom-left to the top-right corner of the frame.''
    \end{itemize}

    \item \textbf{Changes in shot size and spatial composition for Scenery shots}
    \begin{itemize}
        \item ``The shot begins with an aerial view of a city skyline, then tilts downward to focus on a busy intersection.''
        \item ``The camera moves forward, transitioning from a wide view of a dense forest to a close-up of a single tree trunk covered in moss.''
    \end{itemize}
\end{enumerate}

\textbf{Spatial Framing \& Dynamics Description Policy for Annotators}\\
Use the following logic:

\begin{enumerate}
    \item Are there Shot Transitions (cuts or soft transitions)?\\
    If Yes $\rightarrow$ No need for description (ends here).\\
    If No $\rightarrow$ Proceed to the next step.

    \item Are there prominent subjects that strongly draw the viewer's attention?\\
    If No $\rightarrow$ Follow ``Framing of Scenery'', and note any changes in the spatial composition of scene elements according to ``Spatial Motion Within the Frame''. (ends here)\\
    If Yes $\rightarrow$ Proceed to the next step.

    \item \textbf{Determining How to Describe Spatial Framing \& Motion for Subjects}\\
Before describing spatial framing and motion for subjects, determine whether the video features a single subject or small group of subjects or involves a complex subject scenario with multiple subjects or transitions.
\begin{itemize}
    \item \textbf{Single or Small Group of Subjects:}\\
    If the video consistently features one subject or a small group of subjects, follow the ``Framing of Subjects'' to describe:
    \begin{itemize}
        \item Framing size (or shot size)
        \item Frame location (position within the frame)
        \item Scene Depth (foreground, midground, background)
        \item Height relative to the camera (above, below, or at the height of the camera)
    \end{itemize}
    Then follow ``Spatial Motion Within the Frame'' to describe any movement affecting any of the above aspects. (Ends here)

    \item \textbf{Complex Subject Scenarios (Multiple Subjects or Subject Transitions)}\\
    If the video contains multiple subjects or complex subject transitions, follow these steps:
    \begin{itemize}
        \item \textbf{Determine the Primary Focus:}\\
        If there is a single clear main subject: follow the ``Framing of Subjects'' to describe this subject in detail, including shot size and spatial position. Then follow ``Spatial Motion Within the Frame'' to describe any spatial motion and changes. Provide a less detailed overview of secondary subjects.\\
        If the main subject is unclear: describe subjects’ spatial position and movement in prominence-based order (e.g., humans before objects). Instead of determining the shot size based on a random subject, specify it based on the most prominent subject (e.g., a human) if one is clearly dominant. Otherwise, if the subjects are relatively similar in size, use the average shot size. If the shot is even more complex, just directly say which (part of) subjects are visible and which are not.
        \item If there is no clear main subject: provide a general overview of subjects' spatial positions without excessive detail. Do not specify shot size, as it is not meaningful in this case. You may optionally describe the shot size following ``Framing of Scenery'' instead.
        \item \textbf{Identify Subject Transitions:}\\
        If subjects reveal, disappear, switch focus, or undergo other complex changes, describe their shot size (if relevant), spatial position, and movement accordingly. Ensure that the description follows the temporal order in which subjects appear.
    \end{itemize}
\end{itemize}
\end{enumerate}

\subsection{Camera Framing and Dynamics Policy}
\textbf{Instructions for Camera Framing \& Dynamics}\\
Provide a concise yet informative description of the video’s and camera's configuration (video speed, lens distortion, camera angle, camera height), movements (translation, rotation, steadiness, intensity, complexity), and focus (depth, focus plane, focus changes) in this video.

\textbf{Camera Framing \& Dynamics Policy for Annotators}\\
Use the following logic:

\begin{enumerate}
    \item Are there Shot Transitions (cuts or soft transitions)?\\
    If Yes $\rightarrow$ No need for description (ends here).\\
    If No $\rightarrow$ Proceed to the next step.

    \item Is the video played at a different speed than real-time?\\
    If Yes $\rightarrow$ If the video speed is altered, specify:
    \begin{itemize}
        \item Time-lapse: Events unfold significantly faster (e.g., ``Clouds move rapidly across the sky.'')
        \item Fast-Motion: Slightly faster than real-time (1x-3x speed).
        \item Slow-Motion: Slower playback than real-time.
        \item Stop-Motion: Frame-by-frame animation with discrete movements.
        \item Speed-Ramp: A mix of fast and slow speeds within the same video.
        \item Time-Reversed: The video plays in reverse.
    \end{itemize}

    \item Is there any noticeable lens distortion?\\
    If Yes $\rightarrow$ Describe the distortion type and degree:
    \begin{itemize}
        \item Fisheye: Extreme distortion with strong curvature, making the edges appear bent outward.
        \item Barrel: Mild distortion causing straight lines near the edges to bow outward.
    \end{itemize}

    \item Can we tell the camera height (relative to the ground)?\\
    If Yes $\rightarrow$
    \begin{itemize}
        \item Specify the height (Aerial-level, Overhead-level, Eye-level, Hip-level, Ground-level, Water-level, Underwater)
        \item Mention any camera movement that causes height changes.
    \end{itemize}

    \item Can we tell the camera angle (relative to the ground)?\\
    If Yes $\rightarrow$
    \begin{itemize}
        \item Specify the angle (Bird’s Eye, High Angle, Level Angle, Low Angle, Worm’s Eye)
        \item Mention any camera movement that changes the camera angle.
    \end{itemize}

    \item Is there a Dutch/Canted angle (relative to the ground)?\\
    If Yes $\rightarrow$ Describe how the Dutch Angle behaves:
    \begin{itemize}
        \item The Dutch angle remains fixed throughout the shot.
        \item The Dutch angle varies, changing due to camera rolling.
    \end{itemize}

    \item Can we tell the type of camera focus (or depth of field)?\\
    If No $\rightarrow$ Explain if the video doesn’t seem to be filmed with a real camera, lacks a realistic depth of field effect, or is too blurry or dark to determine.\\
    If Yes $\rightarrow$ Proceed to next step.

    \item Is the camera using a shallow depth of field?\\
    If No $\rightarrow$ Deep Focus.\\
    If Yes $\rightarrow$
    \begin{itemize}
        \item Describe the depth of field (shallow or extremely shallow).
        \item Specify which part of the frame is in focus (Foreground/Midground/Background/Out-of-Focus).
        \item If the focus changes, describe the transition (Rack/Pull Focus, Focus tracking).
    \end{itemize}

    \item Is the camera moving (including shaking)?\\
    If No $\rightarrow$ A static, fixed camera. (ends here).\\
    If Yes $\rightarrow$ Proceed to the next step.

    \item Is the camera shaking or wobbling?\\
    If Yes $\rightarrow$ Describe the degree of shaking or wobbling (minimal, moderate, severe shaking).

    \item Does the camera follow or move with any objects?\\
    If Yes $\rightarrow$ Describe how the camera moves with the subject (e.g., Tracking Shot, Arcing, Craning).

    \item How is the camera moving?\\
    Describe why the camera is moving (e.g., tracking a subject, revealing a scene, creating emphasis). Describe the motion using precise movement terms:
    \begin{itemize}
        \item Dolly In/Out: Moving forward or backward toward or away from the subject.
        \item Zoom In/Out: Changing focal length to create the illusion of moving closer or farther.
        \item Pan Left/Right: Rotating the camera horizontally.
        \item Truck Left/Right: Moving the camera laterally left or right.
        \item Tilt Up/Down: Angling the camera up or down.
        \item Pedestal Up/Down: Lifting or lowering the camera while keeping it level.
        \item Rolling Clockwise/Counterclockwise: Rotating the camera around its lens axis.
    \end{itemize}
    Mention the speed of movement if noticeably slow or fast. In addition, if different movements occur at different speeds, clearly distinguish them. For example: ``The camera slowly dollies forward while trucking quickly to the right.''\\
    Describe motion in temporal order if multiple movements occur. For example: ``The camera first pans right, then tilts upward to follow the subject.''\\
    However, if the movement appears too fragmented or random, avoid excessive detail.

    \item \textbf{Example (First-Person Video Game Scenario):}\\
    Excessive Detail (Too much description):\\
    ``As the player explores, the camera moves left, then quickly tilts up, followed by a rapid pan right. The player hesitates, looking down, then abruptly swings the camera left again before slightly tilting upward and making another quick turn to the right.''\\
    Better Description (Concise \& clear):\\
    ``The first-person camera moves randomly as the player looks around, frequently changing direction without a clear pattern.''
\end{enumerate}

\subsection{Policy for Grading the Caption and Feedback}
\begin{itemize}
    \item 5 points: Completely correct; no changes needed.
    \item 4 points: Excellent, with only minor adjustments required (e.g., (1) a few inaccurate words need correction, (2) a few missing words need addition, or (3) a few hallucinated words need deletion). Overall, about one sentence's worth of content may need to be added, changed, or deleted.
    \item 3 points: Mostly correct, but with some important omissions, hallucinations, or errors (e.g., more than two sentences need rewriting, deleting, or adding).
    \item 2 points: Mostly incorrect, requiring significant revisions (e.g., more than half of the caption is incorrect and needs to be redone).
    \item 1 point: Almost entirely wrong, requiring a complete rewrite (e.g., the entire caption is irrelevant).
\end{itemize}

\subsection{Special Cases}
\textbf{Special Case 1: Missing Descriptive Details}\\
If a caption is too brief, we can improve its readability by adding more descriptive details. As a general rule, a strong caption should be self-contained—if you read it to a friend who hasn’t seen the video, they should still be able to picture the subjects, scenery, motion, and camera work. Please aim to write captions that achieve this level of clarity. In particular, some camera captions might be too brief, such as saying the camera is focusing on the midground. Instead, we can improve readability by saying, ``the camera is focusing on the woman wearing a white cap in the midground.'' We should enhance the readability of these captions as much as possible. For readability issues, follow the same policy to deduct points.

\textbf{Special Case 2: Formatting Errors}\\
If the caption contains formatting errors (e.g., mentions of ``first frame'' or ``second frame''), do not rate or correct it. Instead, regenerate a new pre-caption by retrying and modify the prompt to explicitly instruct the model not to mention ``frame.'' In other words, only correct factual mistakes in the video caption.

\textbf{Special Case 3: Labelling Errors}\\
If you discover that the original labels provided by humans are incorrect, please report these cases here and correct those labels in the Labelling projects.

\textbf{Special Case 4: When the Polished Caption Isn't Perfect}\\
If the post-caption isn't perfect, the mistakes usually fall into two categories:
\begin{itemize}
    \item Type-A (Ignoring Feedback): ChatGPT doesn’t fully address all the points from our feedback.
    \item Type-B (Missed Feedback): We accidentally overlook some issues in our original feedback.
\end{itemize}
If you identify a Type-B issue, restart the feedback process. Only correct Type-A issues. In other words, ensure your original feedback is thorough and addresses all mistakes.

\subsection{Common Mistakes to Avoid}
\begin{itemize}
    \item Do not say things like ``this is a scene caption, so no need to mention the subject.'' For pre-captions, simply correct what’s wrong and keep what’s right—no need to add unnecessary explanations.
    \item Focus on describing the visual content. Avoid adding extra commentary like ``adds a sense of vibrancy to the scene.'' However, if the pre-caption includes such descriptions and they are also accurate, it’s fine to keep them.
    \item Ensure consistency across the five captioning tasks. Camera motion information may be mentioned in subject, scene, or spatial captions—make sure it aligns with what’s described in the camera caption.
    \item If a 2D animation clearly conveys depth through layering, you may describe elements like the foreground, middle ground, and background—but be sure to specify that it is a 2D animation attempting to depict depth. If the scene appears flat and lacks depth cues, it's best not to mention depth at all.
    \item A static shot is one where the camera remains entirely still—without any movement, focus change, or zoom. If any of these occur, the shot should not be described as static.
    \item Do not describe overlays as being in the foreground.
\end{itemize}

\section{Pseudocode for Generating Pre-Captions from Labeled Primitives}
\label{sec:pseudocode_precaption}
This section presents the Python implementations of caption generation policies for video annotation. Each policy dynamically generates prompts based on video cinematography and camera motion primitives.

\subsection{Subject Description Policy}

\begin{lstlisting}[language=Python,caption={Subject Description Policy},label={lst:subject-policy}]
class SubjectPolicy(PromptGenerator):
    """Generate prompts for subject description captions based on shot composition primitives."""
    
    def __init__(self):
        name = "Subject Description"
        info = "A policy that uses existing labels to prompt a human or model to provide structured captions for Subject."
        caption_fields = ["subject_description"]
        super().__init__(name, info, caption_fields)
    
    def get_prompt(self, data: VideoData) -> str:
        """Generate annotation prompt based on video's shot composition primitives."""
        
        # Base policy and format instructions
        POLICY_BASE = """Provide a concise yet informative description of the subjects in this video. Keep the description concise and clear, focusing on subject types and visual attributes. You should describe the video by combining details from the frames without referring to any specific one (e.g., don't mention things like "first frame" or "last frame"), and avoid using terms like "image" or "frame." Don't mention the background or motion unless it's necessary to distinguish subjects by location, action, or relationships. You must avoid describing what is not visible or what you are unsure about. You must use simple, natural English and ensure the description is a clear, concise, and coherent paragraph that highlights the most essential details. You must avoid subjective adjectives that convey emotions. Whenever you mention a subject, please describe its key visual attributes in detail. Return only the one-paragraph video description without Markdown formatting or introductory text.

Clearly identify each subject's type, using precise terms such as "man," "woman," "dog," "car," or "tree," rather than vague words like "thing" or "item." If the subject type is ambiguous, use your best judgment and briefly explain your reasoning.

Describe key visual attributes with specific and descriptive language. For people, include details such as clothing color and style, skin tone, hairstyle, facial hair, age (if discernible), gender, ethnicity (if relevant and clear), and facial expression. For objects, describe their color, material, shape, and distinguishing features like texture or markings. Additionally, note the subject's pose and orientation within the frame, such as standing, sitting, walking, or facing a certain direction. Pay attention to any objects that are not in their usual state, like a tilted lamp or an open book lying face down.

If there are multiple subjects to describe, ensure clarity in referring to each. The simplest way is by type, such as "the man," "the dog," or "the tree." If multiple subjects belong to the same category, distinguish them using unique appearance traits (e.g., "the woman in the red dress," "the man with the beard"), location within the scene (e.g., "the man on the left," "the car in the midground"), actions (e.g., "the child playing with a ball," "the bird flying"), or relationships to each other (e.g., "the man next to the woman," "the first man that enters the frame"). Also, when describing multiple subjects, the order in which they are mentioned matters. Prioritize based on relevance, starting with the largest or most centered subject. If the scene unfolds over time, describe subjects in the order they appear. If temporal order isn't relevant, begin with the most visually striking or important subject before moving to less prominent ones. The goal is to provide enough detail so that anyone reading the description can easily identify each subject."""

        FORMAT_INST = """Please avoid using phrases like "the first frame" or "the last frame" in your description. Instead, refer to the entire sequence simply as "the video." Your description should integrate observations from all frames into a cohesive, temporally and logically consistent narrative, rather than describing frames in isolation. Whenever you mention a subject, be sure to include detailed descriptions of its key visual attributes. The final output should be a single, fluent paragraph describing the video, with no Markdown formatting or introductory text. Don't mention the surroundings or the subject's motion unless necessary to distinguish subjects by location, action, or relationship."""

        # Special case: Has Shot Transition
        if data.Has_Shot_Transition:
            policy = POLICY_BASE
            policy += "This video contains one or more shot transitions. Please describe the subject of each segment in a single fluent paragraph."
            return policy

        # Special case: Scenery Shot (not Framing Subject)
        if data.Framing_Subject == False:
            policy = "The video is a scenery shot. You do not need to describe the subject. Please concisely specify the type of scenery shot (e.g., a landscape or cityscape scenery shot) in a single fluent paragraph. Also explain why there is no main subject, such as the focus being on the environment, atmosphere, or scale rather than a specific object. Just note that briefly in one to three sentences."
            policy += "" + FORMAT_INST
            return policy
        
        # Initialize with base policy for subject-containing videos
        policy = POLICY_BASE
        
        # Handle ambiguous framing (Framing_Subject == None)
        if data.Framing_Subject == None:
            if data.Many_Subject_with_No_Clear_Focus:
                policy += "Please note that this video contains **multiple subjects with no clear main focus**. Because it does not emphasize any specific subject, please briefly describe the types of subjects without going into too much detail. You may also describe the subjects collectively as a group."
                policy += "" + FORMAT_INST
                return policy
        
        # Add subject-type specific instructions
        if data.Human_Shot:
            policy += "Please note that the video features salient **human** subjects, so the description should focus on them."
        
        elif data.Non_Human_Shot:
            policy += "Please note that the video features salient **non-human** subjects, so the description should focus on them."
        
        # Handle subject transition cases
        elif data.Change_of_Subject_Shot:
            if data.Subject_Revealing:
                policy += "Please note that the video is a **revealing shot of the subject**, so the description should reflect this by explaining how the subject is revealed through either subject movement or camera movement."
            elif data.Subject_Disappearing:
                policy += "Please note that the video features the main subjects **disappearing from the frame**, so the description should reflect this by explaining how they exit, whether through subject movement or camera movement."
            elif data.Subject_Switching:
                policy += "Please note that the video features the main subjects **switching from one to another**, so the description should reflect this by explaining how the transition occurs, whether through subject movement or camera movement."
        
        # Handle dynamic size variations
        elif data.Clear_Subject_with_Dynamic_Shot_Size:
            policy += "Please note that the video has a **main subject with dynamic size**, so the description should focus on them. Don't mention the background scene and other motion."
        
        # Handle atypical appearance
        elif data.Clear_Subject_with_Atypical_Shot_Size:
            policy += "Focus on describing the **atypical appearance** of the main subjects in the video. Avoid mentioning the background or subject movements."
        
        # Handle multiple subjects with one focus
        elif data.Many_Subject_with_One_Clear_Focus:
            policy += "Please note that the video features **multiple subjects with one clear main focus**, so you need to clarify who the main subject is. The description should focus on the details of the main subject while concisely summarizing secondary subjects and describing their relationship to the main subject if clear."
        
        # Handle multiple different subjects in focus
        elif data.Different_Subjects_in_Focus:
            policy += "Please note that the video features **multiple different subjects in focus**, so the description should clearly distinguish their types and relationships."
        
        # Handle complex/abstract scenarios
        elif data.complex_shot_type == "unknown":
            policy += "Please note that the video features a **complex scenario** with ambiguous subjects or it is an abstract shot. Please try your best to describe the main subjects or objects in the video."
        
        # Use existing shot size description if available
        else:
            HAS_SHOT_SIZE = """In addition, the human-written caption below already describes the subjects (if any) in this video, including framing information like shot size. Use this caption as a reference to draft the subject description, but do not rely on it completely. Expand or refine it to fully capture the subject's type, attributes, appearance, unique features, pose, orientation, relationships between subjects, or any changes in the main focused subject, such as revealing, disappearing, or shifting focus. However, if the description below does not mention any subjects, please do not describe subjects and only specify the type of shot (e.g., a landscape or cityscape scenery shot or a FPV shot) and explain why there is no main subject.

Human-Written Caption: **{shot_size_description}**"""
            policy += "" + HAS_SHOT_SIZE.format(shot_size_description=data.shot_size_description)
        
        # Append format instructions and return
        policy += "" + FORMAT_INST
        return policy
\end{lstlisting}

\noindent\textbf{Primitives used:} Has Shot Transition, Framing Subject, Many Subject with No Clear Focus, Human Shot, Non-Human Shot, Change of Subject Shot, Subject Revealing, Subject Disappearing, Subject Switching, Clear Subject with Dynamic Shot Size, Clear Subject with Atypical Shot Size, Many Subject with One Clear Focus, Different Subjects in Focus.

\subsection{Scene Description Policy}

\begin{lstlisting}[language=Python,caption={Scene Description Policy},label={lst:scene-policy}]
class ScenePolicy(PromptGenerator):
    """Generate prompts for scene description captions based on point of view and overlay primitives."""
    
    def __init__(self):
        name = "Scene Description"
        info = "A policy that uses existing labels to prompt a human or model to provide structured captions for Scene."
        caption_fields = ["scene_composition_dynamics"]
        super().__init__(name, info, caption_fields)
    
    def get_prompt(self, data: VideoData) -> str:
        """Generate annotation prompt based on point of view and overlay information."""
        
        POLICY_BASE = """Provide a concise yet informative description of the overall scene, including the point of view, environment, setting, time of day, and notable visual elements like overlays. For notable visual elements within the scene, describe their color, material, shape, and distinguishing features like texture or markings. If subjects are present, ensure their placement and context complement the scene without excessive detail. You should describe the video by combining details from the frames without referring to any specific one (e.g., don't mention things like "first frame" or "last frame"), and avoid using terms like "image" or "frame." Focus on the setting and scenery rather than detailed subject descriptions. Avoid describing anything not visible or uncertain. Use simple, natural English to create a clear, concise, and coherent paragraph that highlights essential details. Avoid emotional or subjective adjectives. Avoid speculative statements like 'there might be,' 'it appears,' or ambiguous options like 'A or B.' Do not infer the role of the scene setting. Do not explain what the scene emphasizes or highlights. Return only the one-paragraph video description without Markdown formatting or introductory text.

If relevant, indicate the **point of view**, such as first-person, drone shot, or dashcam, and describe how it influences the viewer's perception. Specify the **setting** by clearly identifying whether it is indoors or outdoors, using precise language. If the location is known, state it explicitly (e.g., "Times Square" or "Tokyo subway station"). Otherwise, describe defining features such as "a narrow alley with graffiti-covered walls" or "a vast desert with rolling dunes." Mention the **time of day** and any notable **architectural or natural features**, such as buildings, roads, forests, or bodies of water. Include relevant **weather conditions** if applicable, like "a rainy street with wet pavement reflecting city lights" or "a snowy mountain pass covered in thick fog." For indoor settings, describe key **furniture or props** that establish the environment, such as "a wooden desk cluttered with books and a vintage lamp." If notable, mention the **style** of the scene, such as a monochromatic color scheme or a vibrant carnival with neon lights. If the video contains **overlay elements** such as text, titles, subtitles, captions, icons, watermarks, heads-up displays (HUD), or framing elements, specify that they are overlays (not part of the scene) and describe their content and placement.

If the scene involves **motion or changes**, describe natural elements like wind blowing through trees or waves crashing against the shore, as well as human-made movements such as traffic flowing on a highway or pedestrians walking along a busy street."""

        FORMAT_INST = """Please avoid using phrases like "the first frame" or "the last frame" in your description. Instead, refer to the entire sequence simply as "the video." Your description should integrate observations from all frames into a cohesive, temporally and logically consistent narrative, rather than describing frames in isolation. Do not infer the role of the scene setting. Do not explain what the scene emphasizes or highlights. The final output should be a single, fluent paragraph. Focus on the setting and scenery, not on detailed descriptions of the subject."""

        policy = POLICY_BASE
        policy += "" + FORMAT_INST
        
        # Add point of view information
        
        pov_description = f"[Point of view information for {data.true_pov_attribute}]"
        if data.true_pov_attribute == "objective_pov":
            pov_description += " (no need to mention)."
        
        # Add overlay instruction if present
        if data.Overlays:
            policy += "Please note that the video includes overlay elements, such as text or visuals like titles, subtitles, captions, icons, watermarks, heads-up displays (HUD), or framing elements. In your description, specify that these are overlays (not part of the scene) and describe their content and placement."
        
        # Add point of view context
        policy += f"In addition, you do not need to infer the camera's point of view, as this information is already provided. Please integrate the following point of view information into your caption:Point of View Information: **{pov_description}**"
        
        return policy
\end{lstlisting}

\noindent\textbf{Primitives used:} Overlays, Point of View (Objective Pov, First Person Pov, Selfie Pov, Overhead Pov, Locked On Pov, Dashcam Pov, Drone Pov, Broadcast Pov, Screen Recording Pov, Third Person Over Shoulder Pov, Third Person Over Hip Pov, Third Person Full Body Game Pov, Third Person Top Down Game Pov, Third Person Side View Game Pov, Third Person Isometric Game Pov).

\subsection{Subject Motion \& Dynamics Policy}

\begin{lstlisting}[language=Python,caption={Subject Motion \& Dynamics Policy},label={lst:subject-motion-policy}]
class SubjectMotionPolicy(PromptGenerator):
    """Generate prompts for subject motion description based on shot type primitives."""
    
    def __init__(self):
        name = "Subject Motion & Dynamics Description"
        info = "A policy that uses existing labels to prompt a human or model to provide structured captions for Subject Motion and Dynamics."
        caption_fields = ["subject_motion_dynamics"]
        super().__init__(name, info, caption_fields)
    
    def get_prompt(self, data: VideoData) -> str:
        """Generate annotation prompt based on subject motion characteristics."""
        
        POLICY_BASE = """Provide a concise yet informative description of the subject's motion in this video, ensuring actions are presented in **chronological order** if multiple movements occur (e.g., "The bird first takes flight, then soars in a circle, and finally lands on a branch"). Focus on the subject's motion rather than repeating details already included in the human-written subject descriptions. Avoid describing anything not visible or uncertain. Use simple, natural English to create a clear, concise, and coherent paragraph that highlights essential details. Avoid emotional or subjective adjectives. Avoid speculative statements like 'there might be,' 'it appears,' or ambiguous options like 'A or B.' Return only the one-paragraph video description without Markdown formatting or introductory text.

If the subject in the video has no movement, please briefly mention that without going into too much detail.

Please only describe the content of the video. Don't mention the details of the subject's appearance unless you need to differentiate between multiple subjects by their appearance. Clearly describe the subject's motion.

Avoid abstract descriptions, such as "The car maintains a low, sleek profile as it maneuvers the bend, emphasizing its speed and agility" and "emphasizing its speed and agility as it maneuvers through the turn."

Below are detailed instructions:

Describe **individual subject actions** with clarity, specifying how they move rather than using generic descriptions. For example, instead of "a person is running," say "a runner sprints across the finish line."

If the subject interacts with an **object**, specify the type of interaction and its effect. Instead of "a person is working," say "a construction worker operates a jackhammer, breaking up the pavement."

If there are **interactions between subjects**, describe the nature of their relationship and movements relative to each other. Instead of "people are fighting," say "two boxers exchange blows in the ring, circling each other cautiously."

If there are collective behaviors for a group of subjects, describe **group activities** with specificity. Instead of "birds are flying," say "a flock of geese flies in a V-formation across the horizon." Instead of "people are walking," say "a crowd of protesters marches down the street, carrying signs and banners." Clearly convey the type of group, their coordinated actions, and any notable patterns in their movement."""

        # Special case: Scenery Shot (no subjects)
        if data.Framing_Subject == False:
            return "The video is a scenery shot. You do not need to describe the subject motion. Just note that briefly in one to three sentences."
        
        policy = POLICY_BASE
        
        # Handle ambiguous framing
        if data.Framing_Subject == None:
            if data.Many_Subject_with_No_Clear_Focus:
                policy += "Please note that this video contains **multiple subjects without a clear main focus**. Briefly describe the salient motions and dynamics of the primary subjects while providing a concise overview of secondary movements, or describe all subjects' collective motion if that is more appropriate."
                return policy
        
        # Add subject-type specific instructions
        if data.Human_Shot:
            policy += "Please note that the video features salient **human** subjects, so the description should focus on their motion and dynamics."
        
        elif data.Non_Human_Shot:
            policy += "Please note that the video features salient **non-human** subjects, so the description should focus on their motion and dynamics."
        
        # Handle subject transition cases
        elif data.Change_of_Subject_Shot:
            if data.Subject_Revealing:
                policy += "Please note that the video is a **revealing shot of the subject**, so the description should reflect this by explaining how the subject is revealed through either subject movement or camera movement."
            elif data.Subject_Disappearing:
                policy += "Please note that the video features the main subjects **disappearing from the frame**, so the description should reflect this by explaining how they exit, whether through subject movement or camera movement."
            elif data.Subject_Switching:
                policy += "Please note that the video features the main subjects **switching from one to another**, so the description should first describe the first subject's motion and dynamics, followed by the second's."
        
        # Handle dynamic size and atypical motion
        elif data.Clear_Subject_with_Dynamic_Shot_Size:
            policy += "Please note that the **main subject's framing is not stable** throughout the video, so the description should reflect how their motion and dynamics contribute to this instability."
        
        elif data.Clear_Subject_with_Atypical_Shot_Size:
            policy += "Please note that the main subjects in this video exhibit **atypical motion, posture, or anatomy**, so the description should reflect this."
        
        # Handle multiple subject configurations
        elif data.Many_Subject_with_One_Clear_Focus:
            policy += "Please note that the video features **multiple subjects with a clear main focus**, so the description should focus on the motion and dynamics of the main subject while providing a concise overview of secondary subjects' movements."
        
        elif data.Different_Subjects_in_Focus:
            policy += "Please note that the video features **multiple different subjects in focus**, so the description should clearly distinguish their types, movement patterns, and interactions."
        
        # Add reference to existing subject description
        HAS_SUBJECT_DESC = """In addition, the human-written caption below already describes the subjects (if any) in this video but does not comprehensively capture their motion. Use this caption as a reference or starting point to draft the description, but do not rely on it completely. Expand or refine it to fully capture the subjects' motion and dynamics. However, if the caption does not mention any subjects, do not add any description--simply note this briefly in the description. Please note that this caption is only to help you clarify which subject's motion needs to be analyzed, rather than adding more content based on this description.

Human-Written Caption: **{subject_description}**"""
        policy += "" + HAS_SUBJECT_DESC.format(subject_description=data.subject_description)
        
        return policy
\end{lstlisting}

\noindent\textbf{Primitives used:} Framing Subject, Many Subject with No Clear Focus, Human Shot, Non-Human Shot, Change of Subject Shot, Subject Revealing, Subject Disappearing, Subject Switching, Clear Subject with Dynamic Shot Size, Clear Subject with Atypical Shot Size, Many Subject with One Clear Focus, Different Subjects in Focus.\\
\noindent\textbf{Non-primitive attribute:} \texttt{subject\_description} (human-written subject caption).

\subsection{Spatial Framing \& Dynamics Policy}

\begin{lstlisting}[language=Python,caption={Spatial Framing \& Dynamics Policy},label={lst:spatial-policy}]
class SpatialPolicy(PromptGenerator):
    """Generate prompts for spatial framing description with shot size and camera height information."""
    
    def __init__(self):
        name = "Spatial Framing and Dynamics Description"
        info = "A policy that uses existing labels to prompt a human or model to provide structured captions for Spatial Framing and Dynamics."
        caption_fields = ["spatial_framing_dynamics"]
        super().__init__(name, info, caption_fields)
    
    def format_shot_size(self, shot_size: str) -> str:
        """Convert shot size enum to natural language description."""
        shot_size_map = {
            "unknown": "unknown",
            "extreme_wide": "an extreme wide shot",
            "wide": "a wide shot",
            "full": "a full shot",
            "medium_full": "a medium full shot",
            "medium": "a medium shot",
            "medium_close_up": "a medium close-up shot",
            "close_up": "a close-up shot",
            "extreme_close_up": "an extreme close-up shot"
        }
        return shot_size_map[shot_size]
    
    def format_height_wrt_subject(self, height: str) -> str:
        """Convert height relative to subject enum to natural language."""
        height_map = {
            "unknown": "unknown",
            "above_subject": "above the subject",
            "at_subject": "at the subject's level",
            "below_subject": "below the subject"
        }
        return height_map[height]
    
    def get_prompt(self, data: VideoData) -> str:
        """Generate annotation prompt with shot size and height information."""
        
        POLICY_BASE = """Analyze the subjects and elements in this video and provide a concise yet informative description of how they are spatially framed within the scene, including **shot size, position, depth, height relative to the camera, and any changes**. Your goal is to describe the **spatial framing and dynamics** of the subjects and elements within the shot, considering both their placement within the frame and their relative positions in the scene. Ensure the description covers any notable spatial movements. Avoid describing anything not visible or uncertain. Use simple, natural English to create a clear, concise, and coherent paragraph that highlights essential details. Avoid emotional or subjective adjectives. Avoid speculative statements like 'there might be,' 'it appears,' or ambiguous options like 'A or B'. Return only the one-paragraph video description without Markdown formatting or introductory text.

First, specify the **shot size** based on the subject's size in the frame if major subjects are present. If the shot size is unclear, describe how much of the subject is visible. If no major subject exists (e.g., a scenery shot), describe the shot size in relation to the scenery.

Next, describe the **spatial position of subjects and elements in the video**, if relevant. Indicate their approximate **2D position** within the frame using terms like **left, right, bottom left, bottom right, top right, top left, bottom, top, or center**. Additionally, describe their **3D position** within the scene as **foreground, middle ground, or background**. Analyze as many elements as possible, and for each element mentioned, provide both its **2D and 3D position**.

Finally, describe the **camera's height relative to the subject**, if relevant. Indicate whether the camera is positioned at the subject's height, above them, or below them. We already have this information provided at the end. If it's not provided, try to describe it by yourself.

If **shot size or spatial position** changes, describe how these transitions occur clearly, specifying both the **initial and final states**."""

        HAS_SUBJECT_SCENE = """In addition, the human-written captions below already describe the subjects (if any) and the scenery in this video but do not capture their spatial composition and movements. Use these captions as a reference, but do not rely on them completely. Your goal is to fully capture the spatial framing and movements in this video. Don't write too much about the subject's or scenery's appearance.

Human-Written Description for Subjects: **{subject_description}**
Human-Written Description for Scenery: **{scene_description}**"""

        policy = POLICY_BASE
        policy += "" + HAS_SUBJECT_SCENE.format(
            subject_description=data.subject_description,
            scene_description=data.scene_description
        )
        
        # Determine subject status for shot size information
        subject_status = None  # 'has_subject', 'no_subject', 'change_of_subject', 'has_description'
        
        # Add subject-type specific instructions
        if data.Human_Shot:
            policy += "Please note that the video features **salient human subjects**, so you should focus on describing the spatial framing and movements of them."
            subject_status = "has_subject"
        
        elif data.Non_Human_Shot:
            policy += "Please note that the video features **salient non-human subjects**, so you should focus on describing the spatial framing and movements of them."
            subject_status = "has_subject"
        
        # Handle subject transition cases with detailed shot size info
        elif data.Change_of_Subject_Shot:
            subject_status = "change_of_subject"
            
            if data.Subject_Revealing:
                policy += "Please note that the video is a **revealing shot of the subject**."
                policy += f"Shot Size Information: The video begins with no subject. It then becomes {self.format_shot_size(data.shot_size_info['end'])} of the subject."
                if data.Subject_Height_Applicable:
                    policy += f"When the subject is revealed, the camera is positioned {self.format_height_wrt_subject(data.height_wrt_subject_info['end'])}."
            
            elif data.Subject_Disappearing:
                policy += "Please note that the video features **main subjects disappearing from the frame**."
                policy += f"Shot Size Information: The video begins with {self.format_shot_size(data.shot_size_info['start'])} of the subject. Then the subject disappears."
                if data.Subject_Height_Applicable:
                    policy += f"Before the subject disappears, the camera is positioned {self.format_height_wrt_subject(data.height_wrt_subject_info['start'])}."
            
            elif data.Subject_Switching:
                policy += "Please note that the video features **main subjects switching from one to another**."
                policy += f"Shot Size Information: The video begins with {self.format_shot_size(data.shot_size_info['start'])} of the first subject. Then it becomes {self.format_shot_size(data.shot_size_info['end'])} of the second subject."
                if data.Subject_Height_Applicable:
                    policy += f"The camera is positioned {self.format_height_wrt_subject(data.height_wrt_subject_info['start'])} when the first subject is in focus, and {self.format_height_wrt_subject(data.height_wrt_subject_info['end'])} when the second subject is in focus."
        
        # Note: This policy also uses non-primitive attributes: Shot_Size_Description, Subject_Height_Description
        
        # Handle other shot configurations
        elif data.Clear_Subject_with_Dynamic_Shot_Size:
            policy += "Please note that the **main subject's framing (shot size) is not stable** throughout the video, so the description should emphasize this."
            subject_status = "has_subject"
        
        elif data.Clear_Subject_with_Atypical_Shot_Size:
            policy += "Please note that the **main subjects exhibit atypical posture or anatomy**, so the description should reflect this."
            subject_status = "has_subject"
        
        elif data.Many_Subject_with_One_Clear_Focus:
            policy += "Please note that the video features **multiple subjects with a clear main focus**, so the description should focus on the main subject."
            subject_status = "has_subject"
        
        elif data.Different_Subjects_in_Focus:
            policy += "Please note that the video features **multiple different subjects in focus**, so the description should clearly distinguish their types and relationships."
            subject_status = "has_subject"
        
        elif data.Many_Subject_with_No_Clear_Focus:
            policy += "Please note that this video contains **multiple subjects without a clear main focus**. Briefly describe the spatial positions and movements of salient subjects while providing a concise overview of secondary subjects, or describe all the spatial composition of all subjects collectively as a group if that is more appropriate."
            subject_status = "has_subject"
        
        elif data.Scenery_Shot:
            policy += "Please note that the video is a **scenery shot**. You do not need to describe the subjects. Just note that briefly in one to three sentences."
            subject_status = "no_subject"
        
        elif data.complex_shot_type == "unknown":
            policy += "Please note that the video features a **complex scenario** with ambiguous subjects or it is an abstract shot. Please try your best to describe the spatial positions and movements of the main subjects or objects in the video."
            subject_status = None
        
        else:
            # Has existing shot size description
            subject_status = "has_description"
            policy += "The description below already mentions the spatial framing information about the subjects or scenery in this video. Use this caption as a reference to draft the spatial framing and dynamics description. Simply expand on it to fully capture other spatial positions and movements. Do not infer the any spatial framing information already mentioned below."
            policy += f"Shot Size Information: {data.Shot_Size_Description}"
            
            if data.Subject_Height_Applicable:
                if data.Height_Wrt_Subject_Changes:
                    policy += f"Camera Height Relative to Subjects: The camera is initially positioned {self.format_height_wrt_subject(data.height_wrt_subject_info['start'])} and then changes to {self.format_height_wrt_subject(data.height_wrt_subject_info['end'])}."
                else:
                    policy += f"Camera Height Relative to Subjects: The camera is positioned {self.format_height_wrt_subject(data.height_wrt_subject_info['start'])}."
            elif data.Subject_Height_Description != "":
                policy += f"Camera Height Relative to Subjects: {data.Subject_Height_Description}"
        
        # Add shot size information based on subject status
        if subject_status == "has_subject":
            if data.Shot_Size_Changes:
                policy += f"Shot Size Information: The video begins with {self.format_shot_size(data.shot_size_info['start'])} of the subjects. It then changes to {self.format_shot_size(data.shot_size_info['end'])}."
            else:
                policy += f"Shot Size Information: The video shows {self.format_shot_size(data.shot_size_info['start'])} of the subjects."
            
            if data.Subject_Height_Applicable:
                if data.Height_Wrt_Subject_Changes:
                    policy += f"Camera Height Relative to Subjects: The camera is initially positioned {self.format_height_wrt_subject(data.height_wrt_subject_info['start'])}. It then changes to {self.format_height_wrt_subject(data.height_wrt_subject_info['end'])}."
                else:
                    policy += f"Camera Height Relative to Subjects: The camera is positioned {self.format_height_wrt_subject(data.height_wrt_subject_info['start'])}."
            elif data.Subject_Height_Description != "":
                policy += f"Camera Height Relative to Subjects: {data.Subject_Height_Description}"
        
        elif subject_status == "no_subject":
            if data.Shot_Size_Changes:
                policy += f"Shot Size Information: The video begins with {self.format_shot_size(data.shot_size_info['start'])} of the scenery. It then changes to {self.format_shot_size(data.shot_size_info['end'])}."
            else:
                policy += f"Shot Size Information: The video shows {self.format_shot_size(data.shot_size_info['start'])} of the scenery."
        
        elif subject_status == None:
            policy += "Shot Size Information: The video features a complex scenario with ambiguous subjects or it is an abstract shot. Please try your best to describe the spatial positions and movements of the main subjects or objects in the video. Do not use shot size to describe the spatial framing."
        
        return policy
\end{lstlisting}

\noindent\textbf{Primitives used:} Human Shot, Non-Human Shot, Change of Subject Shot, Subject Revealing, Subject Disappearing, Subject Switching, Clear Subject with Dynamic Shot Size, Clear Subject with Atypical Shot Size, Many Subject with One Clear Focus, Different Subjects in Focus, Many Subject with No Clear Focus, Scenery Shot, Shot Size Is [Extreme Close Up, Close Up, Medium Close Up, Medium, Medium Full, Full, Wide, Extreme Wide], Shot Size Changes, Shot Size Changes From Small to Large, Shot Size Changes From Large to Small, Height Is Always Above Subject, Height Is Always At Subject, Height Is Always Below Subject, Height Wrt Subject Changes.\\
\noindent\textbf{Non-primitive attributes:} \texttt{subject\_description}, \texttt{scene\_description}, \texttt{Shot\_Size\_Description}, \texttt{Subject\_Height\_Description}.

\subsection{Camera Framing \& Dynamics Policy}

\begin{lstlisting}[language=Python,caption={Camera Framing \& Dynamics Policy},label={lst:camera-policy}]
class CameraPolicy(PromptGenerator):
    """Generate prompts for camera description with comprehensive camera setup and motion information."""
    
    def __init__(self):
        name = "Camera Framing and Dynamics Description"
        info = "A policy that uses existing labels to prompt a human or model to provide structured captions for Camera Framing and Dynamics."
        caption_fields = ["camera_framing_dynamics"]
        super().__init__(name, info, caption_fields)
    
    def format_playback_speed(self, speed: str) -> str:
        """Convert playback speed enum to natural language (reads from JSON in actual implementation)."""
        # In actual implementation, reads from labels/cam_setup/video_speed/*.json
        speed_descriptions = {
            "time_lapse": "The video is a time-lapse",
            "fast_motion": "The video is in fast motion",
            "regular": "The video is at regular playback speed (no need to mention).",
            "slow_motion": "The video is in slow motion",
            "stop_motion": "The video uses stop motion animation",
            "speed_ramp": "The video uses speed ramping (changing between fast and slow motion)",
            "time_reversed": "The video plays in reverse"
        }
        return speed_descriptions.get(speed, "Unknown playback speed")
    
    def format_lens_distortion(self, distortion: str) -> str:
        """Convert lens distortion type to natural language."""
        if distortion == "regular":
            return "No lens distortion (no need to mention)."
        elif distortion == "barrel":
            return "The video features mild barrel distortion where lines near the frame edges bow slightly outward."
        elif distortion == "fisheye":
            return "The video shows extreme fisheye distortion where most lines curve strongly outward."
        return "Unknown lens distortion"
    
    def format_camera_height_start(self, height: str) -> str:
        """Format camera height at start of video."""
        height_map = {
            "aerial_level": "at an aerial level",
            "overhead_level": "at an overhead level (around second floor height)",
            "eye_level": "at an eye level (above the waist)",
            "hip_level": "at a hip level (below the waist and above the knees)",
            "ground_level": "at a ground level",
            "water_level": "above water",
            "underwater_level": "underwater"
        }
        return height_map.get(height, "at an unknown height")
    
    def format_camera_height_end(self, height: str) -> str:
        """Format camera height at end of video (for transitions)."""
        height_map = {
            "aerial_level": "to an aerial level",
            "overhead_level": "to an overhead level (around second floor height)",
            "eye_level": "to an eye level (above the waist)",
            "hip_level": "to a hip level (below the waist and above the knees)",
            "ground_level": "to a ground level",
            "water_level": "above water",
            "underwater_level": "underwater"
        }
        return height_map.get(height, "to an unknown height")
    
    def format_camera_angle(self, angle: str) -> str:
        """Convert camera angle enum to natural language."""
        angle_map = {
            "bird_eye_angle": "a bird's-eye view angle (looking down directly at the ground)",
            "high_angle": "a high angle (looking down from above)",
            "level_angle": "a level angle (looking straight ahead)",
            "low_angle": "a low angle (looking up from below)",
            "worm_eye_angle": "a worm's-eye view angle (looking directly up)"
        }
        return angle_map.get(angle, "an unknown angle")
    
    def format_focus_plane(self, plane: str) -> str:
        """Convert focus plane enum to natural language."""
        plane_map = {
            "foreground": "focused on the foreground",
            "middle_ground": "focused on the midground",
            "background": "focused on the background",
            "out_of_focus": "out of focus"
        }
        return plane_map.get(plane, "focus unknown")
    
    def format_camera_steadiness(self, steadiness: str) -> str:
        """Convert camera steadiness enum to natural language."""
        steadiness_map = {
            "static": "The camera is stationary",
            "very_smooth": "The camera movement is very smooth with no shaking",
            "smooth": "The camera movement is smooth with minimal shaking",
            "unsteady": "The camera movement is slightly unsteady with some shaking",
            "very_unsteady": "The camera movement is unsteady with noticeable shaking"
        }
        return steadiness_map.get(steadiness, "Unknown steadiness")
    
    def format_camera_motion_speed(self, speed: str) -> str:
        """Convert camera motion speed enum to natural language."""
        speed_map = {
            "slow": "moving slowly.",
            "regular": "moving at a regular speed (no need to mention).",
            "fast": "moving quickly."
        }
        return speed_map.get(speed, "moving at unknown speed.")
    
    def get_movement_description_simple(self, data: VideoData) -> str:
        """Generate description for simple camera movements."""
        movement_map = {
            "Roll_Clockwise": "rolling clockwise",
            "Roll_Counterclockwise": "rolling counterclockwise",
            "Forward": "moving forward",
            "Backward": "moving backward",
            "Zoom_In": "zooming in",
            "Zoom_Out": "zooming out",
            "Upward": "moving up",
            "Downward": "moving down",
            "Tilt_Up": "tilting up",
            "Tilt_Down": "tilting down",
            "Pan_Right": "panning right",
            "Pan_Left": "panning left",
            "Leftward": "moving left",
            "Rightward": "moving right",
            "Crane_Up": "craning up in an arc",
            "Crane_Down": "craning down in an arc",
            "Arc_Clockwise": "arcing clockwise",
            "Arc_Counterclockwise": "arcing counterclockwise"
        }
        
        # Get list of active movements
        true_movements = [name for name, desc in movement_map.items() if getattr(data, name, False)]
        
        if len(true_movements) == 0:
            return "The camera shows no clear or intentional movement."
        elif len(true_movements) == 1:
            return f"The camera is {movement_map[true_movements[0]]}."
        elif len(true_movements) == 2:
            return f"The camera is {movement_map[true_movements[0]]} and {movement_map[true_movements[1]]}."
        else:
            movements_str = ", ".join([movement_map[m] for m in true_movements[:-1]])
            return f"The camera is {movements_str}, and {movement_map[true_movements[-1]]}."
    
    def get_tracking_description(self, data: VideoData) -> str:
        """Generate description for tracking shots."""
        # This is a complex method that combines tracking types
        # Simplified version shown here
        tracking_types = data.tracking_shot_types  # e.g., ["side", "pan"]
        
        if "side" in tracking_types:
            base = "The camera is tracking the subject from the side"
        elif "tail" in tracking_types:
            base = "The camera is following the subject from behind"
        elif "lead" in tracking_types:
            base = "The camera is leading the subject from the front"
        elif "aerial" in tracking_types:
            base = "The camera is tracking the subject from an aerial view"
        else:
            base = "The camera is tracking the subject"
        
        # Add size change information if applicable
        if data.Tracking_Subject_Larger_Size:
            base += ". During the tracking shot, the subject becomes larger in the frame."
        elif data.Tracking_Subject_Smaller_Size:
            base += ". During the tracking shot, the subject becomes smaller in the frame."
        
        return base
    
    def get_prompt(self, data: VideoData) -> str:
        """Generate comprehensive camera annotation prompt."""
        
        POLICY_BASE = """Provide a concise yet informative description of the **video's and camera's configuration**, covering **video speed, lens distortion, camera angle, camera height, movements (translation, rotation, zooming, steadiness, arcing, craning, tracking, speed, complexity, and purpose), and focus (depth, focus plane, focus changes).**

If **video speed** is altered, specify the type, such as *time-lapse* ("Clouds move rapidly across the sky"), *slow-motion*, *fast-motion*, or *speed ramp* (changing between fast and slow motion). If the video is *time-reversed* or *stop-motion*, note this as well.

If **lens distortion** is present, describe the type and degree. For example, *fisheye distortion* creates extreme curvature, while *barrel distortion* causes mild outward bowing of straight lines near the edges.

Describe the **camera height** in relation to the ground, such as *eye-level, hip-level, ground-level, overhead-level, aerial-level, above water, or underwater*. If height changes due to movement, mention how it transitions. Similarly, specify the **camera angle**, such as *bird's eye, high angle, level angle, low angle, or worm's eye*, noting any shifts within the video. If a **Dutch angle** (tilted horizon) is present, indicate whether it remains fixed or varies due to camera rolling.

If discernible, describe the **camera focus and depth of field**. For example, *deep focus* keeps all elements sharp, while *shallow or ultra-shallow depth of field* blurs the background or foreground. If focus changes dynamically, note whether it's a *rack focus* (shifting focus between subjects) or *focus tracking* (following a subject's depth movement), and state the focus plane at each stage (foreground, midground, background, or out-of-focus). If the video lacks realistic depth of field, describe whether it appears artificial (without a physical camera) or overly blurry.

If the **camera is static**, simply state that the shot is static. If it moves, describe the **type, direction, steadiness, and speed** of movement. Specify movements such as *tracking (following a subject), arcing clockwise/counterclockwise (circling around a subject or frame center horizontally), craning up/down (circling around a subject or frame center vertically), zooming (changing focal length), dollying (moving forward/backward), trucking (moving left/right), pedestaling (moving up/down), panning (rotating the camera horizontally), tilting (rotating the camera up/down), or rolling (rotating around the lens axis).* If the camera is shaking or wobbling, indicate the degree (e.g., minimal, moderate, or severe). If different movements occur at different speeds, clearly distinguish them. If the camera performs multiple movements, describe them in temporal order (e.g., *"The camera first pans right, then tilts upward to follow the subject"*)."""

        policy = POLICY_BASE
        
        policy += "Crucially, instead of inferring these attributes from the video, we have already provided human-labeled ground truth for some of the elements specified above. You should directly use this information in your description and should not infer any details that are not already provided. Your description should be brief, and if anything is normal or unremarkable, you do not need to include it (e.g., if the video is at regular playback speed, there is no need to mention it)."
        
        # Add playback speed information
        policy += f"**Playback Speed:** {self.format_playback_speed(data.playback_speed)}"
        
        # Add lens distortion information
        policy += f"**Lens Distortion:** {self.format_lens_distortion(data.lens_distortion)}"
        
        # Add camera height information
        if data.Height_Wrt_Ground_Applicable:
            if data.Height_Changes_From_Low_To_High or data.Height_Changes_From_High_To_Low:
                policy += f"**Camera Height:** The camera starts {self.format_camera_height_start(data.height_wrt_ground_info['start'])} and then moves {self.format_camera_height_end(data.height_wrt_ground_info['end'])}."
            else:
                policy += f"**Camera Height:** The camera is {self.format_camera_height_start(data.height_wrt_ground_info['start'])}."
        elif data.Overall_Height_Description != "":
            policy += f"**Camera Height:** {data.Overall_Height_Description}"
        else:
            policy += "**Camera Height:** The camera height is unclear or not significant enough to mention (no need to mention)."
        
        # Add camera angle information
        if data.Camera_Angle_Applicable:
            if data.Camera_Angle_Changes:
                policy += f"**Camera Angle:** The camera angle is initially at {self.format_camera_angle(data.camera_angle_info['start'])} and then changes to {self.format_camera_angle(data.camera_angle_info['end'])} due to camera motion."
            else:
                policy += f"**Camera Angle:** The camera angle is at {self.format_camera_angle(data.camera_angle_info['start'])}."
            
            # Check for Dutch angle
            if data.Dutch_Angle_Varying:
                policy += " The camera is also at a dutch angle that varies due to camera rolling."
            elif data.Dutch_Angle_Fixed:
                policy += " The camera is also at a fixed dutch angle during the video."
        elif data.Camera_Angle_Description != "":
            policy += f"**Camera Angle:** {data.Camera_Angle_Description}"
        else:
            policy += "**Camera Angle:** The camera angle is unclear or not significant enough to mention (no need to mention)."
        
        # Add focus information
        if data.Focus_Applicable:
            if data.Deep_Focus:
                policy += "**Camera Focus:** The camera uses a deep focus with a large depth of field."
            else:
                if data.Ultra_Shallow_Focus:
                    policy += "**Camera Focus:** The camera uses an extremely shallow depth of field, focusing on a very narrow plane."
                else:
                    policy += "**Camera Focus:** The camera uses a shallow depth of field, keeping a limited range in focus."
                
                if data.Focus_Changes:
                    policy += f" The camera starts {self.format_focus_plane(data.focus_info['start'])}, and later becomes {self.format_focus_plane(data.focus_info['end'])}."
                else:
                    policy += f" The camera is {self.format_focus_plane(data.focus_info['start'])}."
                
                if data.Rack_Pull_Focus:
                    policy += " The focus plane changes through a rack focus."
                elif data.Focus_Tracking:
                    policy += " The camera uses focus tracking to keep the subject in focus."
        elif data.Camera_Focus_Description != "":
            policy += f"**Camera Focus:** {data.Camera_Focus_Description}"
        else:
            policy += "**Camera Focus:** The camera focus is unclear or not significant enough to mention (no need to mention)."
        
        # Add camera motion information
        if data.Fixed_Camera:
            if data.Fixed_Camera_With_Shake:
                policy += "**Camera Motion:** The camera is fixed but slightly unsteady, with no intentional movement."
            else:
                policy += "**Camera Motion:** The camera is completely static, with no movement or shaking."
        else:
            if data.Complex_Motion:
                policy += f"**Camera Motion:** {data.complex_motion_description}"
            else:
                if data.Minor_Motion:
                    policy += "**Camera Motion:** The camera shows some minor movement."
                elif data.Simple_Motion:
                    policy += "**Camera Motion:** The camera shows a clear movement pattern."
                
                # Add specific movement description
                policy += " " + self.get_movement_description_simple(data)
            
            # Add tracking information
            if data.Tracking_Shot:
                policy += f"**Subject Tracking:** {self.get_tracking_description(data)}"
            
            # Add steadiness information
            policy += f"**Camera Steadiness:** {self.format_camera_steadiness(data.steadiness)}."
            
            # Add motion speed if not regular
            if data.camera_motion_speed != "regular":
                policy += f"**Camera Motion Speed**: The camera is {self.format_camera_motion_speed(data.camera_motion_speed)}"
        
        policy += "If possible, specify the subject that the camera focuses on when describing camera work. For instance, use 'focus on the man in the foreground' rather than 'focus on the foreground.' Likewise, if the camera follows a subject, avoid the generic phrase 'tracking the subject(s).' Instead, identify the subject and describe the specific type of tracking."
        
        return policy
\end{lstlisting}

\noindent\textbf{Primitives used (Camera Setup):} Regular Speed, Slow Motion, Fast Motion, Time Lapse, Stop Motion, Time Reversed, Speed Ramp, No Lens Distortion, Barrel Distortion, Fisheye Distortion, Height Wrt Ground Is [Ground Level, Hip Level, Eye Level, Overhead Level, Aerial Level, Water Level, Underwater Level], Height Wrt Ground Applicable, Height Changes From Low To High, Height Changes From High To Low, Camera Angle Is [Bird Eye Angle, High Angle, Level Angle, Low Angle, Worm Eye Angle], Camera Angle Applicable, Camera Angle Changes, Dutch Angle Fixed, Dutch Angle Varying, Deep Focus, Shallow Focus, Ultra Shallow Focus, Focus Applicable, Focus Changes, Focus Is [Foreground, Middle Ground, Background, Out Of Focus], Rack Pull Focus, Focus Tracking.

\noindent\textbf{Primitives used (Camera Motion):} Fixed Camera, Fixed Camera With Shake, Moving Camera, Clear Moving Camera, Simple Motion, Complex Motion, Minor Motion, Stable Camera Motion, Very Stable Camera Motion, Shaky Camera, Very Shaky Camera, Slow Moving Camera, Fast Moving Camera, Forward, Backward, Upward, Downward, Leftward, Rightward, Pan Left, Pan Right, Tilt Up, Tilt Down, Roll Clockwise, Roll Counterclockwise, Zoom In, Zoom Out, Arc Clockwise, Arc Counterclockwise, Crane Up, Crane Down, Tracking Shot, Aerial Tracking Shot, Arc Tracking Shot, Front Side Tracking Shot, Rear Side Tracking Shot, Lead Tracking Shot, Follow Tracking Shot, Tilt Tracking Shot, Pan Tracking Shot, Side Tracking Shot, Side Tracking Shot Leftward, Side Tracking Shot Rightward, Tracking Subject Larger Size, Tracking Subject Smaller Size.\\
\noindent\textbf{Non-primitive attributes:} \texttt{complex\_motion\_description}, \texttt{Overall\_Height\_Description}, \texttt{Camera\_Angle\_Description}, \texttt{Camera\_Focus\_Description}.

\twocolumn

\end{document}